%% file: 01_main_AISTATS.tex
\documentclass[twoside]{article}
\usepackage[preprint]{aistats2027}

\usepackage[round]{natbib}
\renewcommand{\bibname}{References}
\renewcommand{\bibsection}{\subsubsection*{\bibname}}

\input{00_packages}
\input{00_custom}

\input{ICLR/math_commands.tex}

\usepackage{pgfplots}
\pgfplotsset{compat=1.18}
\usepackage{hyperref}
\usepackage{url}
\usepackage{pgfplots}
\usepgfplotslibrary{groupplots}
\usetikzlibrary{arrows.meta}
\pgfplotsset{compat=1.18}

\definecolor{swgreen}{RGB}{44,160,44}
\definecolor{lonegray}{RGB}{77,77,77}
\definecolor{flipred}{RGB}{139,58,58}

\begin{document}

\runningtitle{Continual Classification Beyond the Kernel Regime}

\twocolumn[

\aistatstitle{Continual Classification Beyond the Kernel Regime: Harnessing the Sparsity Bias of L2 Regularization in Diagonal Linear Networks}

\aistatsauthor{Gilad Karpel \And Itay Evron \And Ron Meir \And Daniel Soudry}

\aistatsaddress{Technion \And NVIDIA \And  Technion \And  Technion
}]

\begin{abstract}
We study logistic regression for continual learning with linear predictors in a high-dimensional statistical setting.
Existing theoretical work has largely focused on continual linear regression with variants of L2 regularization, motivated by their analytical simplicity and connection to the kernel regime of neural networks.
We first derive the optimal L2 regularization strength for continual linear \emph{classification} and characterize its generalization performance, recovering behavior analogous to prior results for continual linear regression.
Next, we examine the effect of L2 regularization in deeper models using a minimal two-layer diagonal linear network.
In the single-task setting, L2 regularization on the weights of such models is known to promote sparsity in the resulting predictor, yielding a rich, sparse feature-learning regime beyond the kernel regime (a similar effect arises in two-layer ReLU networks).
However, in continual learning, where L2 regularization is typically centered on previous-task weights, this sparsity bias is no longer preserved.
We therefore propose a weighted variant of L2 regularization that extends this beneficial sparsity bias to continual learning.
We show theoretically and empirically that, for a sufficiently sparse teacher, our variant converges substantially faster than directly learning a linear predictor with conventional continual L2 or L1 regularization, trading some long-horizon bias for markedly improved performance over moderate task horizons.
\end{abstract}

\pagebreak

\section{Introduction}

In continual learning (CL), a learner sequentially learns from a stream of tasks. 
The key challenge is to accumulate expertise by learning new tasks without forgetting previously seen ones \citep{mccloskey1989catastrophic, lopez2017GEM,parisi2019continual,dohare2024loss}.
In modern large models, continual adaptation is especially important, as environments evolve over time (\eg in user-facing agents, e-commerce, and social networks), requiring models to incorporate new knowledge while retaining previously acquired capabilities, whereas retraining from scratch requires enormous computational resources.

We theoretically study continual classification in a statistical setting.
Specifically, our work provides a \emph{precise} high-dimensional characterization of continual logistic regression with linear predictors under Gaussian features.
Such statistical settings have recently facilitated theoretical advances in continual linear regression, shedding light on factors such as task similarity, model complexity, and regularization strength \citep[e.g.,][]{lin2023theory,zhao2025highdimensional,karpel2026optimal}.
The framework we develop allows us, for the first time, to theoretically study both L2 and other regularizers in a statistical setting for continual classification, focusing on the effects of the number of learned tasks and the amount of sparsity in the system.

Regularization methods have been a central focus of the continual learning community
(e.g., \citealt{kirkpatrick2017overcoming}; see survey in \citealt{shi2025continual}). 
Typically, these works incorporate an additive L2 regularizer in the objective of task $t$, 
\ie minimize  
$
\mathcal{L}_{t}(\vw) + \lambda \norm{\vw-\vw_{t-1}}^2_{\mM}
$
over $\vw$, with $\mathcal{L}_{t}$ being the per-task loss, $\mathbf{w}_{t-1}$ the solution to the previous task, and $\mM$ typically being diagonal \citep[e.g.,][]{benzing2022unifying} or even simply isotropic, $\mM=\I$ \citep[e.g.,][]{lubana2022quadratic}.

While empirical works have focused on regularizing changes in the weights of neural networks (NNs), 
the theoretical literature on regularization has mostly analyzed shallow, linear models \citep[e.g.,][]{li2023fixed,li2025memory,cai2025lastIterate,levinstein2026optimal,karpel2026optimal}.
Thus, it remains unclear whether even isotropic L2 regularization induces a similar effect or algorithmic bias in linear models as it does in NNs.
For example, for a single task, minimizing L2 regularization over the weights of a two-layer ReLU network is equivalent to minimizing L1 regularization on the second layer \citep{neyshabur2014search,neyshabur2015norm,savarese2019infinite}.
In the infinite-width limit, this further corresponds to sparse penalties on the predictor in function space \citep{savarese2019infinite,ongie2019function}.
Similar effects arise in other homogeneous models, where L2 regularization on the weights of depth-$h$ models typically translates into ``sparser'' penalties, such as the L$_p$ quasi-norm with $p=2/h$ for linear networks \citep{NEURIPS2021_e22cb9d6} and certain deep ReLU networks \citep{savarese2019infinite,zhang2022deep}.

Such explicit or implicit sparse penalties are often viewed as driving the rich feature-learning behavior of NNs, enabling them to outperform standard kernel methods \citep{woodworth2020kernel,parhi2022kinds}.
Even in simple linear settings, when a sparse solution exists, an L1 penalty yields sample complexity with only logarithmic dependence on the dimension $d$, compared with the linear dependence under L2 regularization \citep{ng2004feature}.
In other words, sparsity bias may be a key mechanism enabling overparameterized models (\eg NNs) to generalize.

So far, the effect of sparsity bias in continual learning has not been explored theoretically.
Intuitively, sparse solutions may interfere less across tasks and thereby reduce forgetting.
As our paper shows, this intuition can be effective, but only when the sparsity-inducing penalty is designed appropriately.
A natural candidate is the L1 regularizer $\norm{\mathbf{w}-\mathbf{w}_{t-1}}_1$, the direct analogue of the standard L2 penalty $\norm{\mathbf{w}-\mathbf{w}_{t-1}}_2^2$. 
We prove that this L1 regularizer can exploit a sparse ground truth, with excess risk decaying nearly linearly in the sparsity factor $s$.
However, both of these continual regularizers are nonhomogeneous in $\vw$, making it hard to align them with the natural sparsity bias we observe in a single task of deeper models.

We design a continual regularizer that avoids this nonhomogeneity pitfall, preserving a direct connection to the literature on sparsity bias.
Specifically, we introduce a homogeneous and analyzable ``sign weighted L1'' regularizer in \emph{predictor space}, defined as
$$f(\mathbf{w},\mathbf{w}_{t-1})
=
\sum_{i=1}^{d}\frac{|w_{i}|}{[\mathrm{sign}(w_{i})w_{t-1,i}]_{+}+\epsilon}\,.$$ 
Intuitively, this regularizer promotes sparsity in the predictor while encouraging preservation of the coordinate-wise signs of the previous iterate (see \cref{fig:sw-l1-penalty}).

\input{02_penalty_figure}

Importantly, our proposed regularizer arises \emph{naturally} at the predictor level when weighted L2 regularization is applied to the \emph{parameter space} of a two-layer diagonal linear network (DLN; \citealt{woodworth2020kernel}).
Its benefit goes beyond this connection to the sparsity-bias literature:
we show theoretically and empirically that this regularizer achieves a substantially better convergence rate and exhibits less forgetting than the direct L1 analogue.

Specifically, our contributions are:
\begin{itemize}[leftmargin=1.2em]

    \item We develop the first high-dimensional characterization of continual linear \emph{classification}, deriving an analytic expression for the asymptotic performance of general regularized continual logistic regression.

    \item In a 1-layer setting (where simply
    $\hat{y}=\mathbf{w}^\top\mathbf{x}$):

    \begin{itemize}[leftmargin=1.2em]
    \item We derive the optimal L2 regularization strength and excess risk in continual linear classification in high dimensions. 
    We show that the optimal strength scales with the task horizon $T$ as $T/\log T$ and the excess risk scales as $\log T/T$, mirroring the linear regression case \citep{karpel2026optimal}. 

    \item 
    For the direct L1 regularizer $\Vert\mathbf{w}-\mathbf{w}_{t-1}\Vert_1$, we establish global convergence guarantees with near-linear dependence on the sparsity factor $s$, namely $s\log\left(\frac{1}{s}\right)$, but only polylogarithmic dependence on the task horizon $T$.

    \end{itemize}

    \item In a two-layer DLN 
    (where 
    $
    \hat{y}=
    \sum_{j}\theta_{(1),j}\theta_{(2),j}x_j
    $):
    \begin{itemize}[leftmargin=1.2em]
        \item We design a weighting scheme for L2 regularization on the \emph{weight space} that induces a sign weighted L1 regularizer the \emph{predictor space}. 
    
        \item We prove that the sign weighted regularizer attains a tight linear dependence on $s$.
    \end{itemize}

    \item We empirically validate, using linear models and neural networks, that sign weighted L1 outperforms competing regularizers in the presence of sparsity.
\end{itemize}

\section{Setting: Regularized Continual Logistic Regression}
\label{sec:setting}
\paragraph{Continual classification tasks.}
In our setting, the learner recieves a sequence of $T$ tasks $\{(\X_t,\y_t)\}_{t=1}^T$, each having $n$ samples.
In each iteration $t$ a learner is exposed to the samples from the current task $(\X_t,\y_t)$, with no access to data from \emph{previous tasks}. 

\paragraph{Statistical data model: Gaussian features and a noisy linear teacher.}
For each task $t$, datapoints are sampled \iid from the Gaussian distribution, \ie $\{\mathbf {x}_{t,i}\}_{i=1}^{n}\iidsim \mathcal N(\mathbf 0_d,d^{-1}\mathbf I)$ independently of all other tasks.
Labels are drawn from a logistic model using a teacher vector $\teacher\in \mathbb R^d$ whose entries are sampled independently from a distribution $\Pi$, and a label-flip noise level of $\rho\in [0,\frac{1}{2})$. 
That is, for any task $t=1,\dots,T$ and sample $i=1,\dots,n$, the label is drawn as
\begin{equation}
\label{eq:logpol}
\begin{aligned}
\PP(y_{t,i}=1\mid \mathbf {x}_{t,i})
&
=
1-\PP(y_{t,i}=0\mid \mathbf {x}_{t,i})
\\
&=
\rho+(1-2\rho)\frac{e^{\mathbf x_{t,i}^\top\teacher}}{1+e^{\mathbf x_{t,i}^\top\teacher}}
\,.
\end{aligned}
\end{equation}

This setting has been extensively studied in the deep learning theory literature \citep{salehi2019impact,thrampoulidis2020theoretical,gao2026raptor}. 

\paragraph{Studied algorithm: Regularized continual logistic regression.}
To learn \emph{sequentially} without forgetting, we follow a large body of continual learning works that penalize deviations from the previous predictor \citep{kirkpatrick2017overcoming,schwarz2018progress,zhao2024statistical} using weighted L2 regularization.
Extending the theoretical analysis of continual learning with weighted L2 regularization on linear predictors (\ie depth-one neural networks) to deep neural networks is nontrivial.
We therefore consider a depth-$h$ diagonal linear network (DLN), which is a popular  \citep{woodworth2020kernel,nacson2022implicit,Azulay2021} minimal model for the effect of depth, of the form
\[
\mathbf{w}(\boldsymbol{\theta})=\boldsymbol{\theta}_{(1)}\odot\cdots\odot \boldsymbol{\theta}_{(h)},
\]
where $\boldsymbol{\theta}=\left[\boldsymbol{\theta}_{(1)},\cdots,\boldsymbol{\theta}_{(h)}\right]\in\mathbb{R}^{d\times h}$ are the parameters of all the layers, $\odot$ denotes component-wise product, and for an input $\mathbf x$ the predictor is $ \hat{y}=\mathbf w^{\top}(\boldsymbol{\theta})\mathbf x $.
In this paper, we focus on two cases: $h=1$ and $h=2$.

Specifically, the objective function incorporates an additive regularization term $f(\boldsymbol{\theta},\boldsymbol{\theta}_{t-1})$ to discourage deviations from the previously learned predictor.
{
\begin{algorithm}[H]
  \caption{Regularized Continual Logistic Regr. 
   \strut
  \label{schm:regularized}}
\hspace{-0.9em}%
\begin{minipage}{\dimexpr\linewidth+1em\relax}
\begin{algorithmic}
\vspace{-0.1em}
   \STATE {
   \textbf{Input:} 
   $\{(\X_t,\y_t)\}_{t=1}^T$,
   regularization strength $\lambda>0$
   } 
   \STATE { 
   \textbf{Initialize} $ \boldsymbol{\theta}_0=\mathbf 0_d$ 
   }
   \STATE { 
   \textbf{For each} iteration $t=1,\dots,T$:
   }
   \vspace{0.1em}
   \STATE {
   \hspace{0.18em}
   $\!\!\displaystyle\boldsymbol{\theta}_{t}\!=\!\argmin_{\boldsymbol{\theta}\in\mathbb{R}^{d\times h}}\frac{1}{n}
   \sum_{i=1}^{n}\log\!
   \left(
   \frac{1+e^{\mathbf {x}_{t,i}^{\top}\mathbf{w}(\boldsymbol{\theta})}}{
    e^{y_{t,i}\mathbf {x}_{t,i}^{\top}\mathbf{w}(\boldsymbol{\theta})}
   }\right)
   \!+\!
   \frac{\lambda}{d}f(\boldsymbol{\theta},\boldsymbol{\theta}_{t-1})$
    }
   \vspace{0.1em}
   \STATE {
   \textbf{Output:} $\boldsymbol{\theta}_{T}$
   } 
\end{algorithmic}
\end{minipage}
\end{algorithm}
\vspace{-0.4em}
}
We consider regularizers that are convex in their first argument and decompose additively across coordinates. That is,
there exists $\tilde f:\mathbb R\times\mathbb R\to\mathbb R_{\geq 0}$ such that,
\[
f(\boldsymbol \theta,\mathbf v)=\sum_{j=1}^d \tilde f(\theta_j,v_j)
\quad
\text{for all }
\boldsymbol \theta,\mathbf v\in\mathbb R^d
\,.
\]

\paragraph{Notation.}
Throughout the paper, we write $\xrightarrow{d}$ for convergence in distribution, and define the softplus function by $\phi(z)=\log(1+e^{z})$. 
We write $a_n=\sameorder{x}{b_n}$ if there exist constants $0<c_x<C_x<\infty$, depending only on $x$, such that $c_x b_n\le a_n\le C_x b_n$ for all sufficiently large $n$. 
Similarly, $a_n=\lessorder{x}{b_n}$ and $a_n=\gtrorder{x}{b_n}$ denote the corresponding one-sided bounds.
A function $f:\R^d\to \R$ is called $\xi$ homogeneous if $f(\alpha\mathbf w)=\alpha^{\xi} f(\mathbf w)$ for every $\alpha\geq 0,$ and $\mathbf w\in \mathbb R^d$. Its proximal operator is defined by
\[
\prox_{\lambda f(\cdot)}(\mathbf y)=\argmin_{\mathbf x\in \mathbb R^d}\frac{1}{2\lambda}\Vert\mathbf y-\mathbf x \Vert_2^2+f(\mathbf x)
\,.
\]

\paragraph{Loss metric.} Following earlier theoretical work \citep{guo2017calibration,ovadia2019can}, we analyze the negative log-likelihood on an independent test sample, \ie
\begin{align*}
\mathcal{L}(\mathbf w)
\triangleq
\!\!\!\!
\mathop{\mathbb{E}}\limits_{\raisebox{-4pt}{$\scriptstyle \mathbf x \sim \mathcal{N}(\0,d^{-1}\mathbf I)$}}
\!\!\!\!
\big[
\log 
\bigprn{
    1+e^{\mathbf x^\top\mathbf w}
}
-
\pi_\rho(\mathbf x^\top\teacher)\mathbf x^\top\mathbf w
\big]
.
\end{align*}
where $\pi_\rho(u)=\rho+(1-2\rho)\frac{e^{u}}{1+e^{u}}$ is the link function (\ref{eq:logpol}).

\paragraph{The nature of our analysis.}
In this paper, we study regularized continual learning in the high-dimensional regime, where $n,d\to\infty$ with $\frac{n}{d}\to\delta>0$. 
Our main results characterize the performance of the predictors produced by \cref{schm:regularized} under different choices of regularization, as a function of the task horizon $T$, noise level $\rho$, aspect ratio $\delta$, and signal strength $\nu\triangleq\frac{\Vert\teacher\Vert}{\sqrt d}$.

For ease of presentation, throughout the paper we state simplified versions of our lemmas and theorems that capture the essence of the results, deferring the precise statements to the appendices.

\begin{figure*}[b]
    \centering
    \begin{subfigure}{.99\columnwidth}
        \centering
        \includegraphics[width=.88\linewidth]{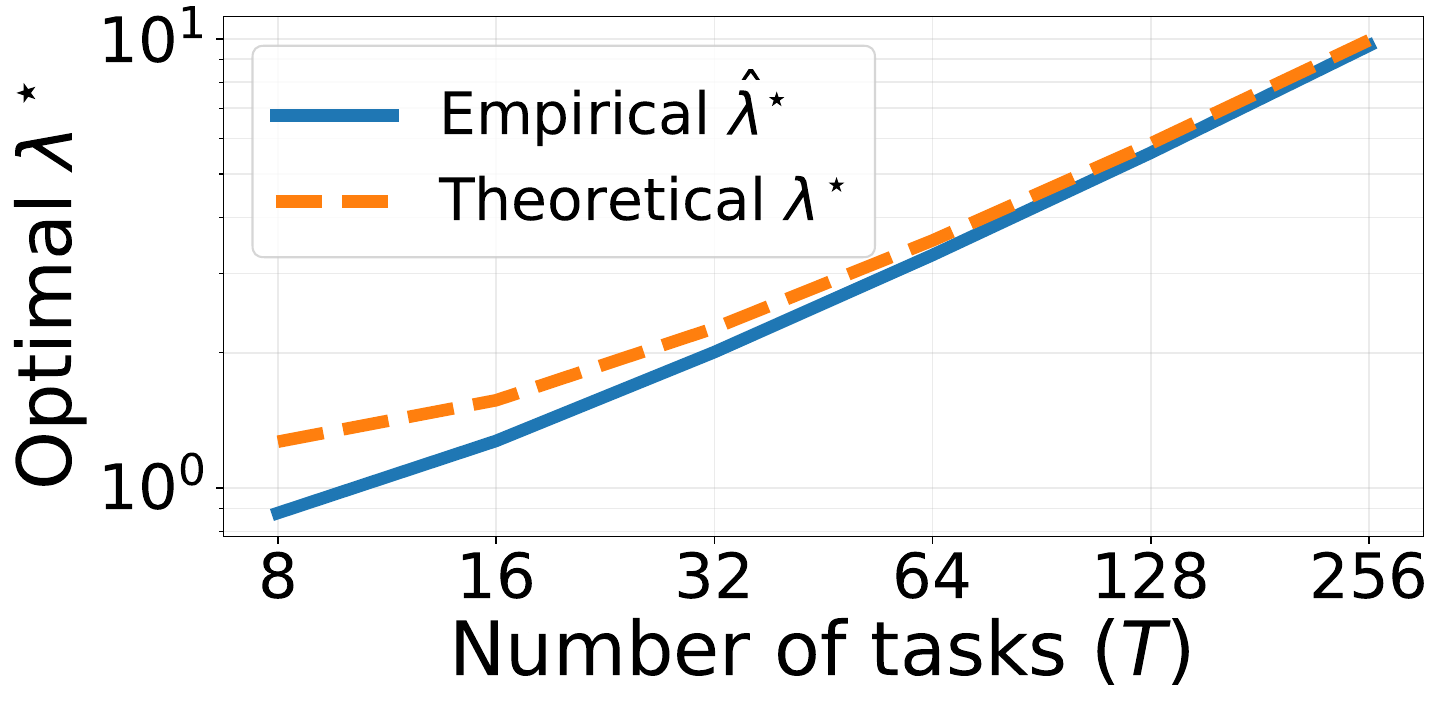}
        \caption{Optimal regularization strength grows as predicted by the full \cref{thm:2} (\cref{sec:optprf}).
        Theoretical $\lambda^\star$ is computed from \cref{thm:2}. 
        Empirical $\hat \lambda^\star$ is obtained by an exhaustive search (choosing $\delta\triangleq\frac{n}{d}=2$).
        }
    \end{subfigure}
    \hfill
    \begin{subfigure}{.99\columnwidth}
        \centering
        \includegraphics[width=.88\linewidth]{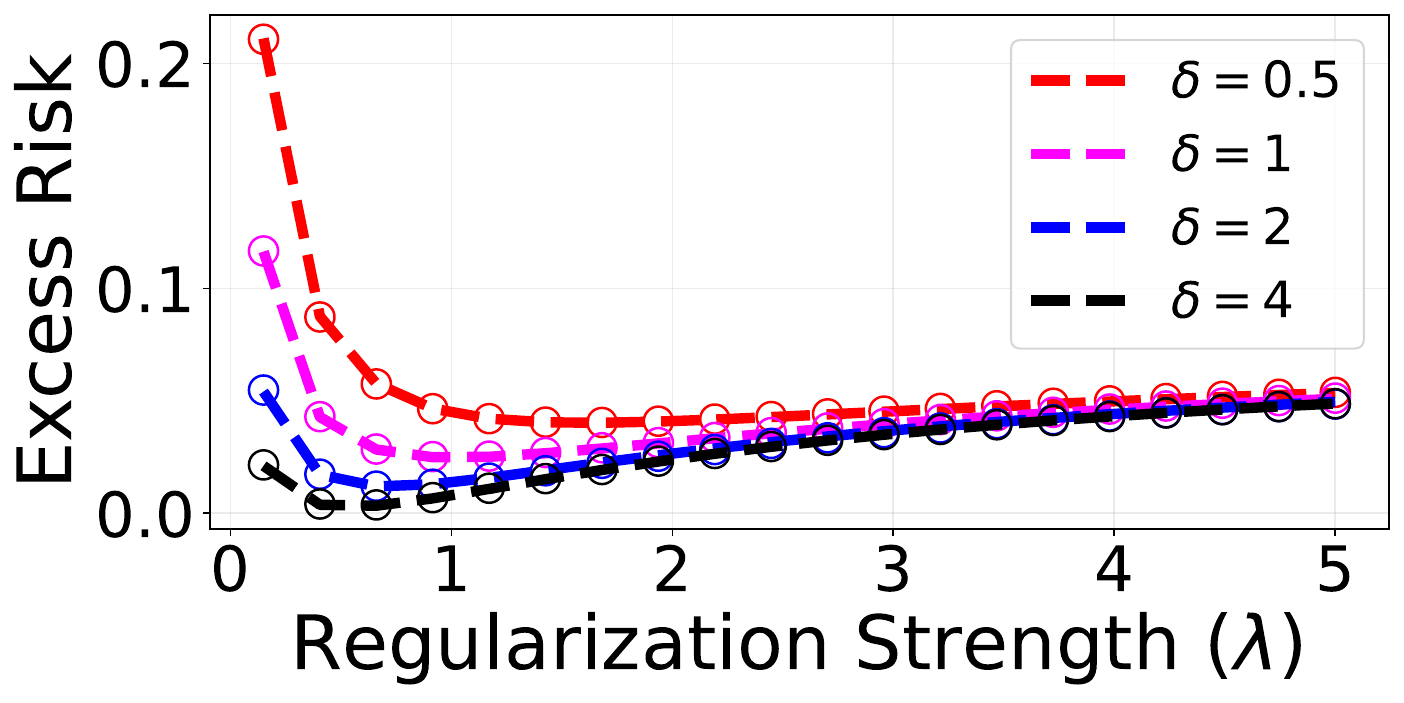}
        \caption{L2-regularized continual logistic regression over $10$ tasks.
        Dashed lines delineate the theoretical risk from the full \cref{thm:1} (\cref{sec:gensr}).
        Circles show empirical results under different aspect ratios $\delta\triangleq \frac{n}{d}$. 
        }
    \end{subfigure}
    \caption{\textbf{Qualitative validation of \cref{thm:1,thm:2}.}
    Empirical values are averaged over 100 runs.}
\label{fig:val}
\end{figure*}

\pagebreak

\section{Single Layer Analysis}

In this section we have a single layer ($h=1$), 
\ie a simple linear model where $\mathbf w(\boldsymbol{\theta})=\boldsymbol{\theta}$.
Thus, we simply denote $\mathbf w= \mathbf w(\boldsymbol{\theta})$ for brevity.

\subsection{High-Dimensional Characterization}
\label{sec:3}
In this section, we characterize the iterates of \cref{schm:regularized} asymptotically as $n,d\to \infty$ with $n/d\to\delta$. 
Our result shows that each coordinate of the learned predictor behaves as a scalar proximal estimator applied to an effective signal corrupted by Gaussian noise. This general expression provides a foundation for studying specific regularization functions.

For each task $t$, let $\eta_t$, $a_t$, and $b_t$ denote the effective regularization, signal, and noise parameters, respectively, given by the state evolution (\cref{eq:system}) in \cref{sec:gensr}. 
\begin{simplifiedtheorem}[Asymptotic Performance]
\label{thm:1}
Under the statistical model (\cref{sec:setting}), suppose that the state-evolution system (\ref{eq:system}) in \cref{sec:gensr} admits a unique solution ($\eta_t$, $a_t$, $b_t$) at each task $t\leq T$. Then, in the proportional high-dimensional limit $n,d\to\infty$, after sequentially learning $T$ tasks using \cref{schm:regularized}, the entries of $w_T$ converge as follows,
\[
w_{T,j}\xrightarrow{d}\prox_{\eta_T \tilde f(\cdot,F_{T-1})}\left(a_T w^{\star}+b_T Z\right),
\]
where $w^{\star}\sim\Pi$, $F_{T-1}$ are independent of $Z\sim\mathcal N(0,1)$, $\mathbf F_{T-1}$ is history statistic (that may depend on previous tasks), and $F_{T-1}$ is the limiting law of a uniformly sampled coordinate from $\mathbf F_{T-1}$.

\end{simplifiedtheorem}

The precise version of \cref{thm:1} in \cref{sec:gensr}. Throughout the remainder of the paper, unless stated otherwise, we assume that the system in \cref{eq:system} admits a unique saddle point solution for every $t\leq T$.

Next, we use our derived characterization to analyze common regularizers and their pitfalls, and to design a new regularizer that outperforms them under sparsity.

\subsection{Isotropic L2 Regularization:\linebreak Optimal Regularization Scaling Law}
\label{sec:3.1}

Following recent theoretical work
\citep[e.g.,][]{levinstein2026optimal,karpel2026optimal}, we consider the L2 (isotropic) regularization, where
\begin{align}
\label{eq:l2}
f(\mathbf{w},\mathbf{w}_{t-1})
=\norm{\mathbf{w}-\mathbf{w}_{t-1}}_2^2
=\sum_{j=1}^{d}(w_j-w_{t-1,j})^{2}\,.
\end{align}
In practice, \citet{Hsu2018Baselines,lubana2022quadratic} find that this simple regularizer often performs similarly to more sophisticated L2 regularizers that, \eg use Fisher information from previous tasks to weight deviations across coordinates \citep{kirkpatrick2017overcoming,zenke2017continual,aljundi2018memory,benzing2022unifying}.

First, we derive an explicit asymptotic characterization of the optimal regularization strength, clarifying how it should scale with the number of learned tasks.

\begin{simplifiedtheorem}[L2 regularization decays near-linearly in the task horizon $T$] 
\label{thm:2}
In the proportional high-dimensional limit $n,d\to\infty$, let $\lambda^\star$ denote the fixed regularization strength that minimizes the generalization loss after $T$ tasks. 
Then, there exist constants $C=C(\nu,\rho,\delta)>0$ and $\kappa=\kappa(\nu,\rho)>0$ such that, for all $\epsilon>0$ and $T>T_0=\bigO(1/\epsilon^{2})$, the optimal regularization strength is bounded as,
\begin{align*}   
(1-\epsilon)\frac{\kappa T}{\log\left(C\kappa T\right)}&<\lambda^{\star}<(1+\epsilon)\frac{\kappa T}{\log\left(C\kappa T\right)}
\,.
\end{align*}
Moreover, under this strength scaling the excess risk decreases in $T$ as, 
\[ 
\mathcal{L}(\mathbf w_T)-\inf_{\vw} \mathcal{L}(\vw)=\sameorder{\nu,\delta,\rho}{\frac{\log T}{T}}.
\]
\end{simplifiedtheorem}
In \cref{sec:optprf}, we prove \cref{thm:2} and give explicit expressions for $\kappa$, $C$, and the excess risk, making their dependence on $\nu,\delta,\rho$ explicit.

\paragraph{Empirical validation on synthetic data.}
\cref{fig:val} validates the predictions of \cref{thm:1,thm:2} using finite-dimensional data. 
We set dimension $d=50$ and noise level $\rho=0.1$, draw the teacher as $\mathbf{w}^\star\sim \mathcal N(\mathbf{1}_d,\I)$, and sample data points as
$\{\mathbf {x}_{t,i}\}_{i=1}^{n}\iidsim \mathcal N(\0,d^{-1}\mathbf I)$.

\paragraph{Near-optimal convergence under L2.}
Interestingly, L2 is already near-optimal: 
under the optimal scaling of $\lambda$ given in \cref{thm:2}, the generalization error decays as $\log(T)/T$.
This is within a logarithmic factor of the lower bound $\gtrorder{\nu,\delta,\rho}{1/T}$\footnote{
Under additional structural assumptions, such
as sparsity of the teacher, this lower bound can be strengthened.
} applicable to any learning algorithm as we establish in
\cref{sec:lowerproof}.

\paragraph{Is L2 always the optimal choice?}  
Across multiple tasks, L2 regularization does not take advantage of sparsity because it penalizes changes from the previous solution quadratically.
It therefore spreads each update across many coordinates rather than exploit a sparse underlying predictor.
In \cref{sec:l2_sparsity}, we empirically validate with synthetic data that across various sparsity levels the excess risk attained by regularizing with L2 remains approximately fixed (see \cref{fig:l2_sparse}).

\subsection{L1 Regularization Can Exploit Sparsity}
This lack of sparsity adaptation motivates us to study sparsity-inducing regularizers that can better exploit a sparse teacher and thereby improve generalization.

In single-task learning, L1 regularization is one of the most common approaches to recover sparse solutions \citep{vaddireddy2020feature,islam2024experiment}.
Such regularization has played a significant rule in compressed sensing \citep{yin2008bregman}, MRI \citep{lustig2007sparse}, medical image reconstruction \citep{zhu2013improved}, and biological image deconvolution \citep{ikoma2018convex}.

A natural, yet underexplored, continual-learning analogue is to set, at iteration $t$, 
\begin{align}
\label{eq:l1}
f(\mathbf{w},\mathbf{w}_{t-1})
=\norm{\mathbf{w}-\mathbf{w}_{t-1}}_1
=\sum_{j=1}^{d}|w_j-w_{t-1,j}|\,.
\end{align}
To examine the potential sparsity advantage of this regularizer, we specialize the teacher distribution to a standard teacher prior. This lets us to explicitly track how the excess risk depends on the sparsity level.

\paragraph{Sparsity modeling.}
Following earlier works \citep{SUN2022109246,jantre2024spike,kumar25a}, we consider a \emph{sparse} teacher prior. Specifically, we assume the entries of $\teacher$ are drawn \iid from a distribution,
\begin{align}
\label{eq:dis} 
\Pi\left(w\right)=(1-s)\delta_{0}(w)+s\mathcal{N}\left(w \Bigm|0,\frac{\nu^{2}}{s}\right), 
\end{align}
where $s\in (0,1)$ is the sparsity factor, $\mathcal{N}(w\mid 0,\sigma^2)$ is the density of the Gaussian distribution with zero mean and standard deviation $\sigma$, and $\delta_0(w)$ represents the Dirac delta at zero. Namely, each entry of $\teacher$ equals $0$ with probability $1-s$, while with probability $s$ it is nonzero and follows a Gaussian distribution.

\paragraph{Finite-horizon sparsity gains under L1.}
We first consider a fixed, finite task horizon and show that L1 regularization exploits the sparsity of the solution.
In this regime, the excess risk is governed by the number of active coordinates and vanishes at a rate of $s\log\left(\frac{1}{s}\right)$ as $s\to0$.
Thus, even after finitely many tasks, L1 regularization can achieve small error by exploiting the underlying sparse structure.
\begin{simplifiedlemma}[L1 regularization achieves near-linear decay
in the sparsity factor $s$]
\label{thm:finite-L1-bound}
For every fixed task horizon $T\geq1$ and every
$s\leq s_0=\sameorder{\nu,\delta}{1}$, the excess risk satisfies,
for all $\lambda>0$,
\[
\mathcal{L}(\mathbf w_T)-\inf_{\vw}\mathcal{L}(\mathbf w)
=
\gtrorder{\nu,\delta,\rho}
{s\log\left(\frac{1}{s}\right)}.
\]
Moreover, there exists $\lambda=\sameorder{\delta}{\sqrt{\log(\hfrac1s)+\log T}}$
such that
\[
\mathcal{L}(\mathbf w_T)-\inf_{\vw}\mathcal{L}(\mathbf w)
=
\lessorder{\nu,\delta,\rho}
{s\log\left(\frac{1}{s}\right)}.
\]
\end{simplifiedlemma}

\paragraph{Limited long-horizon gains under L1.}
\cref{thm:finite-L1-bound}  clarifies that the excess risk vanishes as the teacher becomes sparse.
However, this guarantee alone does not characterize the long-term behavior. 
Next, we show the (slow) convergence rate of the L1 regularizer, being almost logarithmic as the task horizon increases.

\begin{simplifiedlemma}[L1 regularization decays only poly-logarithmically in the task horizon $T$]
\label{thm:temporal_l1_large_T_upper}
For any fixed $s\leq s_0=\sameorder{\nu,\delta}{1}$ and every
$\lambda>0$,
\[
\mathcal{L}(\mathbf w_T)-\inf_{\vw}\mathcal{L}(\mathbf w)
=
\gtrorder{\nu,\delta,\rho}
{\frac{1}{\log^2 T}}.
\]
Moreover, there exists
$\lambda=\sameorder{\nu,\delta}{\sqrt{\log T}}$ such that
\[
\mathcal{L}(\mathbf w_T)-\inf_{\vw}\mathcal{L}(\mathbf w)
=
\lessorder{\nu,\delta}
{\frac{1}{\log T}}.
\]
\end{simplifiedlemma}

The upper bounds in \cref{thm:finite-L1-bound,thm:temporal_l1_large_T_upper} are proved in Appendices~\ref{prf:uniL1} and \ref{prf:temporal_l1_large_T_upper}, respectively. The corresponding lower bound is proved in Appendix~\ref{prf:finite-L1-bound}. 
These appendices provide explicit choices of $\lambda$ and retain the full dependence of the bounds on $\nu,\delta$.
In particular, we prove slightly stronger results, allowing the task horizon $T$ to grow while the sparsity factor $s$ decreases at the same time, unlike the results presented here, which assume that one of these parameters is fixed.

\pagebreak

\section{Two-Layer Analysis}

In this section we use two diagonal layers $h=2$, inducing a linear model parameterized as
\[
\hat{y}=\mathbf w^{\top}(\boldsymbol{\theta})\mathbf x
=
\prn{\boldsymbol{\theta}_{(1)}\odot\boldsymbol{\theta}_{(2)}}^{\top}
\mathbf x
\,.
\]

\subsection{From Parameter-Space L2 Regularization to Predictor-Space Sparsity}
We aim to analyze the behavior of this model with L2 regularization on the parameters $\boldsymbol{\theta}$. 
To do so, we will first show that the dynamics of \cref{schm:regularized} can be reduced to the dynamics of the same algorithm in a single layer, but with a different penalty $f$. 

Our main tool for this would be the following proposition (which is a slight generalization of the result in \citet[Theorem 2]{NEURIPS2021_e22cb9d6}, proved in \cref{prf:equ}):
\begin{proposition}[Weighted L2 equivalence for DLN]
\label{prop:deep-equivalence}
Denote by $\mathbf{w}(\boldsymbol{\theta})$ the linear predictor induced by a {depth-$h$} DLN parameterized by $\boldsymbol{\theta}$.
Then, for any predictor $\mathbf w\in\mathbb R^d$ and weighting function $c(\cdot,\cdot)>0$,
\begin{align*}
\min_{\substack{\boldsymbol{\theta}\in\mathbb{R}^{d\times h}:\\
\mathbf{w}(\boldsymbol{\theta})=\mathbf w}}
\frac{1}{h}
\sum_{j=1}^{d}c(w_j,\cdot) \!\sum_{\ell=1}^{h}|\theta_{(\ell),j}|^{2}
=
\sum_{j=1}^{d}c(w_j,\cdot) |w_{j}|^{2/h}
\,
.
\end{align*}
\end{proposition}

\begin{example}[Sparsity bias in L2-regularized DLNs]
In a single-task setup, 
learning a two-layer DLN under a standard L2 penalty on the weights is equivalent to an L1 penalty on the predictor.
That is, under common loss functions $\mathcal{L}(\vw)$ 
(\eg the logistic loss), 
the following equivalence hold:
\begin{align*}
\min_{\boldsymbol{\theta}\in\mathbb{R}^{d\times h}}
\mathcal{L}(\vw(\boldsymbol{\theta}))
+
\tfrac{\lambda}{2d}
\norm{\boldsymbol{\theta}}^2_{F}
~=~
\min_{\vw\in\mathbb{R}^{d}}
\mathcal{L}(\vw)
+
\tfrac{\lambda}{d}
\norm{\vw}_1
\,
.
\end{align*}
\end{example}
Such a sparsity bias can substantially improve generalization when the
ground-truth predictor is sparse. 
For example, L1-regularized learning requires sample complexity only logarithmic in the dimension $d$, whereas L2 regularization generally retains linear dependence on $d$ \citep{ng2004feature}.
Related sparsity-promoting biases have also been associated \citep{woodworth2020kernel} with the ``rich'' (a.k.a.\ active) feature-learning regime of NNs, which can generalize beyond standard kernel methods
\citep{NEURIPS2021_e22cb9d6,parhi2022kinds}. 
Sparsity bias may therefore be an important mechanism through which overparameterized models generalize well.

\paragraph{Can regularized CL also induce a sparsity bias?}
The analytical mechanism above does not extend to CL by simply centering the regularizer at previous-task parameters.
Specifically, the proof of \cref{prop:deep-equivalence} relies on the \emph{homogeneity} of the weighted L2 regularizer
while $\norm{\vw-\vw_{t-1}}_2^2$ (\cref{eq:l2}) is nonhomogeneous in $\vw$.
Next, we overcome this by shifting the preservation of past expertise to a weighting function $c(\cdot,\cdot)$.

\subsection{Sign Weighted L1 Regularization:
\linebreak
From Rich-Regime Sparsity to Stable CL}
\label{sec:4.2}

We aim to induce a sparsity bias that is both practical for continual learning and readily analyzable using linear dynamics (\eg with the high-dimensional characterization in \cref{sec:3}).
To this end, we employ the following weighted L2 regularizer:
\begin{align*}
f(\boldsymbol{\theta},\boldsymbol{\theta}_{t-1})
=
\frac{1}{2}
\sum_{j=1}^{d}c(w_j(\boldsymbol{\theta}),w_j(\boldsymbol{\theta}_{t-1}))\sum_{\ell=1}^{h}\theta_{(\ell),j}^{2}
\,,
\end{align*}
choosing  
$c(w_j,w_{t-1,j})
=
\frac{1}{[\mathrm{sign}(w_j)w_{t-1,j}]_++\epsilon}$
where $\epsilon>0$ is a fixed constant.

Importantly, by \cref{prop:deep-equivalence},
this regularizer, which acts in the DLN parameter space, induces a predictor-space \emph{sign weighted L1 regularization} (SW-L1):
\begin{align}
\label{eq:signed}
f(\mathbf{w},\mathbf{w}_{t-1})
&
=
\sum_{i=1}^{d}\frac{|w_{i}|}{[\mathrm{sign}(w_{i})w_{t-1,i}]_{+}+\epsilon}
\,.
\end{align}

Our sign weighted L1 regularizer $f$ is designed to satisfy two key properties.
\begin{enumerate}[leftmargin=1.7em,label=\Alph*)]
\item 
\textbf{Convexity and homogeneity in the first argument.}
We required convexity in \cref{schm:regularized} to facilitate the high-dimensional characterization of \cref{sec:3}.
Homogeneity implies a lower bound on the excess risk, as we will show in \cref{thm:one_homogeneous_error_floor}.

\item \textbf{Promotes sparse but stable continual learning.}
Common nonhomogeneous regularizers in CL
(\eg $\norm{\vw-\vw_{t-1}}_2^2$)
aim to preserve past expertise by centering norms at previous iterates.
In contrast, we encode this preservation through the weighting function $c(w_j,w_{t-1,j})$ while maintaining homogeneity.
As illustrated in \cref{fig:sw-l1-inactive}, this weighting promotes not only sparsity but also preservation of the coordinate-wise signs of the previous iterate.
\end{enumerate}

Our results in this section show that sign weighted L1 regularization achieves the optimal convergence rate in sparsity factor $s$ within the broader family of positively homogeneous regularizers.
Proofs are in \cref{prfs:4.2}.

\begin{theorem}[SW-L1 decays linearly in the sparsity factor $s$]
\label{thm:signed-weighted-l1-upper}
For any $s\le s_0=\sameorder{\nu,\delta}{1}$ and any $\lambda=\lambda_0>0$, set $\epsilon=\frac{\lambda_0}{\sqrt{\frac{2}{\delta}\log\left(\hfrac{64}{s}\right)}}.$
Then, for every task $T\geq 2$,
\[
\mathcal{L}(\mathbf w_T)-\inf_{\vw} \mathcal{L}(\vw)=\lessorder{\nu,\delta}{s}
\,.
\]
\end{theorem}

The rate above contrasts with L1 regularization which converges only as $s\log\left(\frac{1}{s}\right)$, as shown in \cref{thm:finite-L1-bound}. 
This also shows that the advantage of SW-L1 emerges only from the second task:
at the first task, where $\vw_0=\mathbf 0$, the sign-weighted penalty reduces to L1.

Under homogeneous weighting functions $c(\cdot,\cdot)$ (like ours), the penalties induced by \cref{prop:deep-equivalence} are overall positively homogeneous as well. 
This class has an intrinsic limitation: regardless of how the weights are chosen, its excess risk does not vanish with the task horizon for a \emph{fixed} sparsity level.
This shows that the upper bound of \cref{thm:signed-weighted-l1-upper} is \emph{tight}, and that sign weighted L1 is an optimal positively homogeneous regularizer.
\begin{lemma}[Non-vanishing error for positively homogeneous regularizers]
\label{thm:one_homogeneous_error_floor}
Under the statistical setting (\cref{sec:setting}).
Assume that the regularizer is order $\xi>0$ homogeneous.
Then, for every $T\geq 1$ and $s\in(0,1)$,
\[
\mathcal{L}(\mathbf{w}_{T})-\inf_{\vw} \mathcal{L}(\vw)=\gtrorder{\xi,\nu,\rho,\delta} {s}.
\]
\end{lemma}

\input{03_main_figure}

\section{Experiments: SW-L1 Best Exploits Sparsity}
\label{sec:exp}

\subsection{Setup}

In this section, we empirically evaluate the performance of the three regularizers considered in our paper:
L2, L1, and sign weighted L1 (\cref{eq:l2,eq:l1,eq:signed}, respectively).
We consider three increasingly complex settings: 
logistic regression on synthetic Gaussian data, logistic regression on MNIST, and a two-layer ReLU network on MNIST.
In each setting, a fixed sparse teacher generates binary labels with label-flip noise of $\rho=0.1$, and the student is updated sequentially. 
Performance is measured by the normalized loss
$(L(\mathbf w_t)-L_{\text{Bayes}})/L_{\text{Bayes}}$, where $L_{\text{Bayes}}$ denotes the entropy of the teacher-induced label distribution representing the irreducible minimum loss.

At each task, the student observes $n=1,000$ randomly selected samples and minimizes the regularized objective: using proximal FISTA for linear classification \citep{beck2009fast} and Adam for NNs \citep{kingma2015adam}. 
Full implementation details are in \cref{sec:experiments}.

\tparagraph{Linear classifier for Gaussian synthetic tasks.}
We consider data points sampled according to $\mathbf x_i\sim\mathcal{N}(\0,d^{-1}\mathbf I)$ with $d=500$.
Also, the teacher’s components are sampled from the distribution, as specified in \cref{eq:dis} with $\nu=5$. This experiment most closely matches our theoretical setting and provides direct empirical validation of our theoretical predictions.

\tparagraph{Linear classifier for MNIST.} 
We convert MNIST \citep{lecun1998gradient} into a binary classification problem by partitioning digits into 0--4 versus 5--9.
Each pixel is standardized using the training set statistics.
The teacher is a sparse logistic model whose support follows a Fisher-based pixel ranking \citep{Duda2000}.

\tparagraph{Two-layer ReLU for MNIST.}
We use the previously considered MNIST setting, with the student as a two-layer ReLU network with $m=2,000$ neurons trained with Adam. The teacher is a sparse subset of a pretrained dense network, where the active neurons are selected by neuron contribution score.
The sign weighted L1 regularizer is regularized by the equivalence established in \cref{prop:deep-equivalence}.

\subsection{Results}

\cref{fig:sparsity_all} reports the held-out test loss after learning a sequence of $T$ tasks across three sparsity levels.
Curves are averaged over 5 seeds, with shaded regions indicating one standard deviation.

Across all settings, regularizer performance depends on teacher sparsity.
With highly sparse teachers, L1 leverages sparsity and often beats L2, while SW-L1 provides even larger gain by reducing the loss rapidly and attaining the best performance.
As sparsity decreases, this advantage becomes less pronounced, the L1 plateaus early, whereas L2 keeps improving and eventually reaches lower loss.
Still, in the two linear model setups, SW-L1 converges quickly and remains better than L2 over the first tasks even at low sparsity.

\section{Related Work \& Discussion}
We propose a sign weighted L1 regularizer that exploits sparsity, remains stable over long task horizons, and preserves sparsity bias in continual learning with two-layer diagonal linear networks.
It converges substantially faster than existing regularizers, both theoretically and empirically.
Below, we relate our results to prior work and discuss future directions.

\tparagraph{CL theory beyond linear models.}
While a few theoretical works extend beyond linear models, their guarantees are limited to local convergence \citep{han2023convergence,schliserman26a} or neural networks in the lazy regime, where the model stays near its initialization \citep{bennani2020generalisationWithOGD,doan2021NTKoverlap,lee2021taskSimilarity,taheri2026on}.
We contribute a complementary direction by exploiting implicit sparsity biases of deeper models through their induced regularization in predictor space.
This allows us to derive \emph{global} convergence guarantees in a statistical setting while capturing rich-regime feature learning and sparsity bias.

\cite{graldi2025beinglazy} study the impact of model scale and feature learning (lazy and rich regimes), for a model trained sequentially via gradient descent. 
However, their analysis is specific to these dynamics, whereas ours applies to general decomposable convex regularizers used in practice \citep[e.g.,][]{kirkpatrick2017overcoming}.

\tparagraph{L1 variants.}
Improving the performance of L1 regularization has been the focus of several works in the \emph{single-task} setting.
Examples include reweighted L1 regularization \citep{candes2008enhancing,foucart2009sparsest,daubechies2010iteratively,zhao2012reweighted}, nonconcave penalties for variable selection \citep{fan2001variable}, the minimax concave penalty \citep{zhang2010nearly}, and, more recently, L$_{1-2}$ regularization \citep{hu2026ℓ}.

\tparagraph{Continual L2 regularization in linear settings.}
Several theoretical works have analyzed isotropic L2 regularization in continual learning \citep[e.g.,][]{li2023fixed,zhao2024statistical,levinstein2026optimal,karpel2026optimal}.
However, these works consider the more analytically tractable continual linear \emph{regression} setting, whereas ours is the first to rigorously characterize continual \emph{classification} under different regularization types and strengths.
Interestingly, our findings in \cref{sec:3.1} qualitatively mirror several results from the regression setting.
For example, \citet{karpel2026optimal}, whose statistical setup is closest to ours, also find that the optimal L2 regularization strength scales nearly linearly with the number of tasks, consistent with \cref{thm:2}.
In a related setting, \citet{karpel2026optimalE} find a similar inverse scaling law for the step size in a model trained continually using $N$ gradient steps and step size $\gamma.$
\citet{levinstein2026optimal} derive a \emph{near-optimal} regularization strength scaling as $\ln(T)$; however, other choices may perform better, since their result optimizes a worst-case \emph{upper bound}.
\citet{cai2025lastIterate} analyze a broader proximal framework, but require a cyclic ordering of tasks. We also note \citet{evron2023classification}, who analyze continual linear classification but require a regularization strength $\lambda\to 0$, primarily yielding a theoretical proxy to unregularized settings.

Finally, concurrent work \citep{guo2026score} studies sparse continual linear regression with an additional regularization matrix constructed from the empirical covariance matrix.
Their (global) iterates are essentially equivalent to
\[
\vw_{T}
=
\argmin_{\vw}
\prn{
\frac{1}{2n}
\sum_{m=1}^{T}
\norm{\X_{m}\vw - \y_{m}}^2_2
+
\lambda
\norm{\vw}_1
}
\,,
\]
\ie their method balances the cumulative loss over \emph{all} tasks seen so far against the L1 norm of the shared parameter vector.
However, even for their linear setup, this method requires the storage and continual updating of $d^2$ parameters, which can limit practical scalability \citep{kirkpatrick2017overcoming,schwarz2018progress,karpel2026optimalE}.
In contrast, we use a practical alternative that requires maintaining only two parameters and induces analyzable implicit biases beyond shallow linear models, specifically in diagonal linear networks.

\tparagraph{Open directions.}
Several assumptions could be relaxed, \eg by allowing multiple teachers, broader statistical feature models, or multiclass settings.
A particularly interesting direction is to identify other implicit biases, potentially in deeper or more practical models, and use them to design improved weighting schemes.
Such theory-guided designs could lead to better empirical continual regularization methods.

\bibliography{99_bibliography}
\bibliographystyle{ICLR/iclr2027_conference}

\clearpage
\appendix
\thispagestyle{empty}

\crefalias{section}{appendix}
\crefalias{subsection}{appendix}

\onecolumn
\aistatstitle{Supplementary Materials: Experiments and Proofs}

\section{Additional implementation details for the Experiments}
\label{sec:experiments}
 
\subsection{Setup}
 
All three experiments share the following design. A fixed sparse teacher generates binary labels at each task via a bit-flip model with noise parameter $\rho=0.1$. In the Gaussian and linear MNIST settings the teacher is a sparse linear model $\teacher$ and labels satisfy
\begin{equation}
  \PP(y=1\mid \mathbf x) = \rho + (1-2\rho)\sigma(\mathbf x^\top\teacher).
  \label{eq:app_labels_linear}
\end{equation}
In the ReLU MNIST setting the teacher is a sparse two-layer ReLU network
$f_{\theta^\star}$ with $k$ active neurons, and labels satisfy
\begin{equation}
  \PP(y=1\mid \mathbf x) = \rho + (1-2\rho)\sigma\left(\sum_{i=1}^{k}w_{2,i}^\star\mathrm{ReLU}({\mathbf w_{\star,1,i}}^\top\mathbf x)\right).
  \label{eq:app_labels_relu}
\end{equation}
The student is initialized at zero (or from a random Guassian for the ReLU model) and never revisits past data. Hyperparameters $(\lambda, \epsilon)$ are selected by pooled validation NLL: the configuration minimizing the mean validation loss across all seeds at the final task is applied to every seed. For all experiments we find the optimal hyperparameter $\lambda$ using an exhaustive search. The test set is withheld until final reporting. 
 
The three regularizers are:
\begin{align*}
  \text{L2:}&\quad
    \lambda\Vert \mathbf w  - \mathbf w_{t-1}\Vert_2^2, \quad
  \text{L1:}\quad 
    \lambda\Vert \mathbf w - \mathbf w_{t-1}\Vert_1,\quad
  \text{sign weighted L1:}\quad
    \lambda\sum_{i}\frac{|w_i|}
    {[\mathrm{sign}(w_i)\,w_{t-1,i}]_++\epsilon}.
\end{align*}

\subsection{Gaussian Synthetic Experiment}
 
\paragraph{Data.}
Date samples are drawn \iid as $\mathbf x\sim\mathcal{N}(\0,d^{-1}\mathbf I)$. A fixed training pool of $50,000$ points is drawn once; validation and test sets each contain $5,000$ independent points.
 
\paragraph{Teacher.}
$\teacher\in\mathbb{R}^d$ has exactly $k$ nonzero entries drawn using the spike-and-slab model as in \cref{eq:dis}
One teacher is used per sparsity level, shared across seeds.
 
\paragraph{Protocol.}
The dimension is $d=500$, $T=30$ tasks, $n=1,000$ samples per task, three seeds.
Sparsity levels: $s\in\{0.001, 0.01, 0.1\}$.
 
\paragraph{Student optimizer.}
Proximal FISTA~\citep{beck2009fast} with adaptive restart and backtracking
line search. Step size initialized from a power-iteration estimate of the Lipschitz constant (40 iterations, $1.05\times$ inflation).
Up to $1,000$ iterations; early stopping when the normalized
proximal-gradient residual falls below $10^{-6}$.
 
\subsection{Linear MNIST Experiment}
 
\paragraph{Data.}
MNIST is binarized as $\{0,\ldots,4\}$ vs.\ $\{5,\ldots,9\}$, $d=784$.
Pixels are standardized per-coordinate on the training split using the empirical mean and std (and divided by $\sqrt{d}$).
Training pool consisting of $50,000$ images.
Validation of $5,000$ images and Test with $10,000$ images.
 
\paragraph{Teacher.}
Pixels are ranked by a supervised Fisher score on all $60,000$ training images. For each sparsity level, the top-$k$ pixels are extracted and a logistic teacher is refitted on the full training set.

More specifically, for each pixel coordinate $j \in \{1, \dots, 784\}$, let
$\mu_{j,c} = \frac{1}{n_c}\sum_{i:\, y_i = c} x_{ij}$ and $\sigma_{j,c}^2 = \frac{1}{n_c}\sum_{i:\, y_i = c}(x_{ij} - \mu_{j,c})^2$
denote the class-conditional mean and variance of coordinate $j$ for class
$c \in \{0, 1\}$, computed once on the full training set.
The Fisher score of coordinate $j$ is then
\begin{equation}
    F_j = \frac{(\mu_{j,1} - \mu_{j,0})^2}{\sigma_{j,1}^2 + \sigma_{j,0}^2}.
\end{equation}
We rank all 784 coordinates in descending order of $F_j$ and retain the top $k = \lfloor s \cdot 784 \rceil$ coordinates from the teacher support.
\paragraph{Protocol.}
The dimension is $d=28\times 28=784$ with $T=30$ tasks, $n=1,000$ samples per task, and three seeds. Sparsity levels: $s\in\{0.001, 0.01, 0.1\}$

\paragraph{Student optimizer.}
Identical to the Gaussian experiment.

\subsection{Two-Layer ReLU MNIST Experiment}
 
\paragraph{Data.}
Same binarized MNIST as the linear experiment. Pixels are standardized per-coordinate using the empirical mean and std.

\paragraph{Protocol.}
The dimension is $d=28\times 28=784$ with $T=30$ tasks, $n=1,000$ samples per task, three seeds. Validation set: $2,000$ images (held out from the $10,000$ MNIST test images); test set: the remaining $8,000$.

\paragraph{Teacher.}
A two-layer ReLU network with $m=2,000$ neurons,
$f(x)=\sum_i w_{2,i}\mathrm{ReLU}(\mathbf w_{1,i}^\top x)$, is trained by Adam for 25 epochs (lr $10^{-3}$, weight decay $10^{-4}$, batch size 512) on the $50,000$ training images. Neurons are ranked by contribution score $|w_{2,i}|\cdot\mathrm{std}(\mathrm{ReLU}(\mathbf w_{1,i}^\top X))$. For each sparsity level, the top-$k$ neurons are retained and fine-tuned for 10 additional epochs; the remaining neurons are zeroed.
 
\paragraph{Sign weighted regularizer for ReLU.}
The sign weighted L1 penalty is applied in its weighted-L2 on the weights form. For neuron $i$ the coefficient $c_i = \frac{1}{[\mathrm{sign}(w^{(2)}_{i})\,w^{(2)}_{t-1,i}]_++ \epsilon}$ is computed with a stop-gradient on $\mathrm{sign}(w_{i}^{(2)})$, consistent with treating $c_i$ as fixed at the previous iterate, the penalty $\frac{1}{2}c_i(\Vert\mathbf w_{1,i}\Vert^2+w_{2,i}^2)$ is added to the loss.

\paragraph{Student optimizer.}
Both layers trained jointly by Adam (lr $3\times10^{-3}$, batch size 256, 300 epochs per task).
Student initialized with $w_{1,i}\sim\mathcal{N}(0,d^{-1})$ and
$w_{2,i}\sim\mathcal{N}(0,m^{-1})$.
Proximal step for L1/L2 applied once per epoch (after all mini-batches).

\subsection{Summary Table}
 
\begin{table}[h]
\centering
\caption{Key parameters across the three experimental settings.}
\label{tab:app_exp_summary}
\small
\begin{tabular}{lccc}
\toprule
& \textbf{Gaussian} & \textbf{Linear MNIST} & \textbf{ReLU MNIST} \\
\midrule
Dimension $d$       & 500      & 784      & 784   \\
Tasks $T$           & 30       & 30       & 30    \\
Samples/task $n$    & 1,000      & 1,000      & 1,000  \\
Training pool       & 50,000 & 50,000 & 50,000 \\
Validation          & 5,000  & 5,000  & 2,000  \\
Test                & 5,000  & 10,000 & 8,000  \\
Seeds               & 3        & 3        & 3     \\
Noise $\rho$        & 0.10     & 0.10     & 0.10  \\
Student optimizer   & FISTA    & FISTA    & Adam  \\
\bottomrule
\end{tabular}
\end{table}

\newpage
\section{L2 regularization does not exploit sparsity}
\label{sec:l2_sparsity}
\begin{figure}[h!]
    \centering
    \includegraphics[width=0.62\linewidth]{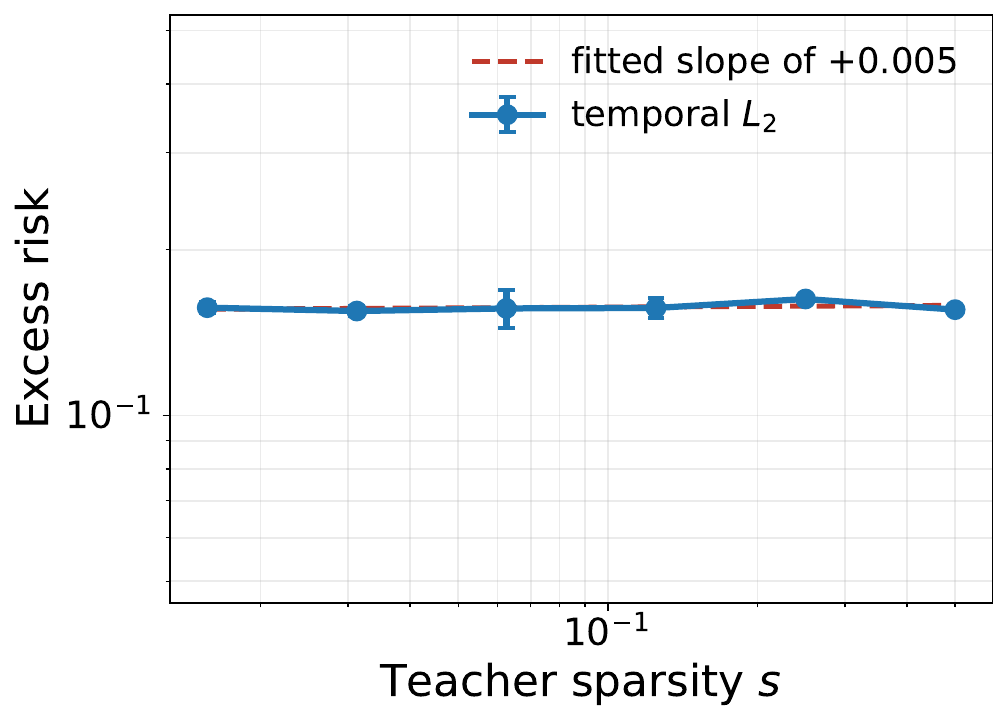}
    \caption{
    \textbf{Temporal L2 does not exploit sparsity of the teacher.}
    Excess risk $\mathcal{L}(\mathbf w_T)-\inf_{\mathbf w} \mathcal{L}(\mathbf w)$ after $T=8$ sequential tasks. The dimension $d=256$, the per-task sample size $n=128$ (hence $\delta=n/d=0.5$), the horizon $T$, the noise level $\rho=0$ and the
    signal strength $\nu=\lVert\teacher\rVert/\sqrt d=0.25$ are all held
    fixed. Error bars are one standard error over $3$ seeds. The dashed line is the least-squares fit on the log-log scale, with slope $+0.005$ ($R^2=0.11$).
    }
    \label{fig:l2_sparse}
\end{figure}
\paragraph{Experimental setup.}
We work in the teacher--student setting of \cref{schm:regularized}. Features are drawn i.i.d.\ as $\mathbf x\sim\mathcal N(\mathbf 0,\mathbf I)$ with $d=256$. The teacher $\teacher$ is $k$-sparse: a support of size $k=sd$ is drawn uniformly at random, the entries on it are drawn from $\Pi=\mathcal N(0,1)$, and $\teacher$ is then rescaled so that $\nu=\lVert\teacher\rVert/\sqrt d$ equals a constant. Labels are generated by the teacher through the noisy logistic model described in \cref{sec:setting} with label noise $\rho=0$.

The predictor is obtained from \cref{schm:regularized}. Each task is solved to convergence by full-batch Adam.
We report the excess risk of the \emph{last} predictor estimated on an independent test set of $8,000$ samples labeled by the same teacher, and $\inf_{\mathbf w} \mathcal{L}(\mathbf w)$ is the corresponding Bayes risk, computed in closed form from the teacher logits. The regularization strength $\lambda$ is selected for each method and each configuration separately, by minimizing the loss at the final task on a held-out validation set of $4,000$ samples labeled by the same teacher.

To isolate the effect of sparsity we sweep the teacher sparsity $s$ while holding every other quantity of the problem fixed, and fit $\log\big(\mathcal{L}(\mathbf{w}_{T})-\inf_{\mathbf{w}}\mathcal{L}(\mathbf{w})\big)\approx\beta\log s+\text{const}$. \cref{fig:l2_sparse} reports the outcome for temporal L2: the fitted slope is $\beta=+0.005$, and the excess risk stays within $0.155$-$0.163$ across the entire range. This indicates that the L2 penalty does not exploit sparsity of the teacher.

\newpage
\section{Proofs for Section \ref{sec:3}}
\subsection{CGMT proof for the bit-flip logistic teacher (general decomposable regularizer)}
\label{sec:prf3}
\begin{fuller}[\cref{thm:1}]
\label{sec:gensr}
Under the statistical model (\cref{sec:setting}), assume that the system (\ref{eq:system}) admits a unique solution $(\bar\alpha_t,\bar\sigma_t,\bar{r}_t,\bar\tau_t,\bar\gamma_t,\bar\theta_t)$ for every $t\leq T$. Then as $n,d\to\infty$, after learning $T$ tasks with \cref{schm:regularized}, for every locally Lipschitz function with at most quadratic growth $\Psi:\R^2\to\R$\footnote{$\Psi$ is called locally Lipschitz  if, for all $M>0$ there exists $L_M>0$, such that for all $\mathbf x,\mathbf y\in [-M,+M]^2$ we have $|\Psi(\mathbf x)-\Psi(\mathbf y)|\leq L_M\Vert \mathbf x-\mathbf y\Vert$. It has at most quadratic growth if there exists $C_\Psi>0$ such that $|\Psi(\mathbf x)|\leq C_\Psi(1+\Vert\mathbf x\Vert^2)$ for all $\mathbf x\in\R^2$.},
\begin{align*}
\frac{1}{d}\sum_{j=1}^{d}\Psi(w_{T,j},w_{\star,j})\xrightarrow{\PP}\E\left[\Psi\left(\prox_{\eta_T \tilde f(\cdot, F_{T-1}))}\left(a_T w^{\star}+b_T Z\right),w^{\star}\right)\right],
\end{align*}
where $F_{T-1}$ and $w^{\star}\sim\Pi$ are independent of $Z\sim\mathcal N(0,1)$, $\mathbf F_{t-1}$ is any coordinatewise state so that for $c>0$ bounded away from $0$, $M_{\lambda\tilde f(\cdot,F)}(x,c)$\footnote{The Moreau Envelope is defined as $M_{\lambda\widetilde{f}(\cdot,v)}(x,\tau)=\min_{y}\left\{ \lambda\widetilde{f}(y,v)+\frac{1}{2\tau}(y-x)^{2}\right\} .$} is continuous and satisfies $M_{\lambda\tilde f(\cdot,F)}(x,c)\leq C(1+x^2+F^2)$ for $C>0.$ Also, the past state has a deterministic empirical limit \ie $\frac{1}{d}\sum_{i=1}^d\delta_{(w_{\star,j}, F_{t-1,j})}\xrightarrow[]{\PP}\mu_{t-1}$ in the 2-Wasserstein distance. Moreover, the saddle-point parameters satisfy for all $t\leq T$ the nonlinear system
\begin{align}
\label{eq:system}
\left\{
\begin{aligned}
\nu^{2}\alpha_{t}&=\E\left[w^{\star}\prox_{\lambda\sigma_{t}\tau_{t}\widetilde{f}(\cdot, F_{t-1})}\left(\sigma_{t}\tau_{t}\left(\theta_{t}w^{\star}+\frac{r_{t}}{\sqrt{\delta}}Z\right)\right)\right],\\\gamma_{t}&=\frac{1}{r_{t}\sqrt{\delta}}\E\left[Z\,\prox_{\lambda\sigma_{t}\tau_{t}\widetilde{f}(\cdot, F_{t-1})}\left(\sigma_{t}\tau_{t}\left(\theta_{t}w^\star+\frac{r_{t}}{\sqrt{\delta}}Z\right)\right)\right],\\\nu^{2}\alpha_{t}^{2}+\sigma_{t}^{2}&=\E\left[\prox_{\lambda\sigma_{t}\tau_{t}\widetilde{f}(\cdot, F_{t-1})}\left(\sigma_{t}\tau_{t}\left(\theta_{t}w^{\star}+\frac{r_{t}}{\sqrt{\delta}}Z\right)\right)^{2}\right],\\\gamma_{t}^{2}&=\frac{2}{r_{t}^{2}}\E\left[\left(1-\pi_{\rho}(\nu Z_{1})\right)\left(\nu\alpha_{t}Z_{1}+\sigma_{t}Z_{2}-\prox_{\gamma_{t}\phi}\left(\nu\alpha_{t}Z_{1}+\sigma_{t}Z_{2}\right)\right)^{2}\right],\\\gamma_{t}\theta_{t}&=-2\E\left[\pi_{\rho}'(\nu Z_{1})\prox_{\gamma_{t}\phi}\left(\nu\alpha_{t}Z_{1}+\sigma_{t}Z_{2}\right)\right],\\1-\frac{\gamma_{t}}{\sigma_{t}\tau_{t}}&=\E\left[\frac{2\left(1-\pi_{\rho}(\nu Z_{1})\right)}{1+\gamma_{t}\phi''\left(\prox_{\gamma_{t}\phi}\left(\nu\alpha_{t}Z_{1}+\sigma_{t}Z_{2}\right)\right)}\right].
\end{aligned}
\right.
\end{align}
Here $Z,Z_1,Z_2\stackrel{\text{iid}}{\sim}\mathcal N(0,1)$, and $\eta_t=\lambda\bar\sigma_t\bar\tau_t$, $a_t=\bar\sigma_t\bar\tau_t\bar\theta_t$, $b_t=\frac{\bar\sigma_t\bar\tau_t\bar r_t}{\sqrt{\delta}}.$
\end{fuller}
\begin{remark}[About the uniqueness of the non-linear system (\ref{eq:system})]
Earlier CGMT \citep{thrampoulidis2015regularized} analyses of regularized logistic regression, such as \citet{salehi2019impact}, assume uniqueness of the associated nonlinear system rather than prove it. However, through numerical evaluations they observed that a fixed-point iterative method converges to a stable solution (see their Remark 2). More recent work has established existence and uniqueness of analogous state-equation systems for broad classes of convex regularized $M$-estimators in linear models \citep{bellec2023existence}, and for unregularized single-index models, including logistic regression, above some critical threshold $\delta_\infty$ \citep{koriyama2025phase}. Although their results do not directly cover our regularized single-index setting, they provide further support for the plausibility of the uniqueness assumption.
\end{remark}
\begin{proof}
The proof will consist of three main steps, as in \cite{salehi2019impact} (see Appendix C)
\begin{enumerate}
    \item We begin by expressing the optimization problem in bilinear form in order to apply CGMT and derive its corresponding auxiliary optimization problem.
    \item We then simplify the auxiliary optimization problem. Characterizing the problem via a scalar optimization.
    \item Finally, we solve the auxiliary optimization problem by deriving and solving the first-order optimality conditions.
\end{enumerate}

We begin with characterizing the limiting performance of a specific iteration $t$, for ease of presentation we will omit the dependence on $t$ and write $\mathbf X_t=\mathbf X$, $\mathbf y_t=\mathbf y$.
The optimization problem is
\begin{align}\label{eq:app_estimator_main_corr}
\widehat{\mathbf w}\in \argmin_{\mathbf w\in \mathbb R^d} \frac{1}{n}\left[\sum_{i=1}^n\log \left(1+e^{\mathbf x_i^\top\mathbf w}\right)-y_i\mathbf x_i^\top\mathbf w\right]+\frac{\lambda}{d}f(\mathbf w)
\end{align}
where $f$ is convex, and decomposable.

The optimization problem in \cref{eq:app_estimator_main_corr} can be re-written as
\begin{align}\label{eq:app_estimator_phi_corr}
\widehat{\mathbf{w}}\in\argmin_{\mathbf{w}\in\R^d}\left\{\frac{1}{n}\mathbf{1}^\top\phi(\mathbf X\mathbf{w})-\frac{1}{n}\mathbf y^\top\mathbf X\mathbf{w}+\frac{\lambda}{d}f(\mathbf{w})\right\}.
\end{align}
Under the statistical model, $\PP(y_i=1\mid x_i)=\rho+(1-2\rho)\sigma(x_i^\top\teacher),$ with $\sigma(x)=\frac{1}{1+e^{-x}}$
hence $\PP(y_i=1\mid x_i)=\pi_\rho(\mathbf x_i^\top\teacher),$ where $\pi_\rho(g)=\rho+(1-2\rho)\sigma(g).$
We therefore obtain the following identities,
\begin{align}\label{eq:app_pi_sym_corr}
\pi_\rho(-g)=1-\pi_\rho(g),\qquad\pi_\rho'(g)=(1-2\rho)\sigma(g)\sigma(-g)=(1-2\rho)\phi''(g).
\end{align}
Let
\[
\mathbf{H}=\sqrt{d}\begin{bmatrix}-\mathbf{x}_{1}^{\top}-\\
\vdots\\
-\mathbf{x}_{n}^{\top}-
\end{bmatrix}\in\R^{n\times d},\quad\delta=\frac{n}{d}.
\]
Then the entries of $\mathbf H$ are \iid standard Gaussian and
\cref{eq:app_estimator_phi_corr} is equivalent to
\begin{align}\label{eq:app_estimator_H_corr}
\widehat{\mathbf{w}}\in\argmin_{\mathbf{w}\in\R^{d}}\left\{ \frac{1}{n}\mathbf{1}^{\top}\phi\left(\frac{1}{\sqrt{d}}\mathbf{H}\mathbf{w}\right)-\frac{1}{n}\mathbf{y}^{\top}\frac{1}{\sqrt{d}}\mathbf{H}\mathbf{w}+\frac{\lambda}{d}f(\mathbf{w})\right\}.
\end{align}

In order to apply the CGMT, we need a min-max optimization, so we introduce $\mathbf u\in\R^n$ and rewrite \cref{eq:app_estimator_H_corr} as
\begin{align}
\min_{\mathbf{w}\in\R^{d},\mathbf{u}\in\R^{n}}\frac{1}{n}\mathbf{1}^{\top}\phi(\mathbf{u})-\frac{1}{n}\mathbf{y}^{\top}\mathbf{u}+\frac{\lambda}{d}f(\mathbf{w})\quad\text{s.t.}\quad\mathbf{u}=\frac{1}{\sqrt{d}}\mathbf{H}\mathbf{w}.
\end{align}
Introducing a Lagrange multiplier $\mathbf v\in\R^n$ gives us the desired min-max optimization,
\begin{align}
\min_{\mathbf{w},\mathbf{u}}\max_{\mathbf{v}}\frac{1}{n}\mathbf{1}^{\top}\phi(\mathbf{u})-\frac{1}{n}\mathbf{y}^{\top}\mathbf{u}+\frac{\lambda}{d}f(\mathbf{w})+\frac{1}{n}\mathbf{v}^{\top}\left(\mathbf{u}-\frac{1}{\sqrt{d}}\mathbf{H}\mathbf{w}\right).
\end{align}

\begin{remark}
To apply the CGMT, we would also need to restrict
the optimization variables $\mathbf{w}, \mathbf u, \mathbf v$ to some compact and convex sets $S_{\mathbf{w}}\subset\R^d,S_{\mathbf u}\subset\R^n,S_{\mathbf v}\subset\R^n,$ and perform the min--max optimization over these sets. These sets can be chosen to be arbitrary large and do not affect the optimization phase. For simplicity, we suppress the dependence on $(S_{\mathbf{w}},S_{\mathbf u},S_{\mathbf v})$
from the notation.
\end{remark}

Since $\tilde{\mathbf y}$ depends on $\mathbf H$ we can not directly apply CGMT. To deal with this problem we introduce $\mathbf P=\frac{\teacher \teacher^\top}{\Vert\teacher\Vert^2},$ and $\mathbf P^\perp=\mathbf I_d-\mathbf P .$
We can now decompose the matrix $\mathbf H$ to $\mathbf H=\mathbf H_1+\mathbf H_2$ where $\mathbf H_1=\mathbf H\mathbf P,$ and $\mathbf H_2=\mathbf H\mathbf P^\perp.$
Then the optimization becomes
\begin{align}
\min_{\mathbf{w},\mathbf{u}}\max_{\mathbf{v}}\frac{1}{n}\mathbf{1}^{\top}\phi(\mathbf{u})-\frac{1}{n}\mathbf{y}^{\top}\mathbf{u}+\frac{\lambda}{d}f(\mathbf{w})+\frac{1}{n}\mathbf{v}^{\top}\left(\mathbf{u}-\frac{1}{\sqrt{d}}\mathbf{H}_{1}\mathbf{w}\right)-\frac{1}{n\sqrt{d}}\mathbf{v}^{\top}\mathbf{H}_{2}\mathbf{w}.
\end{align}
Since $\mathbf P\teacher=\teacher,$ we have
\[
\frac{1}{\sqrt{d}}\mathbf{H}\teacher=\frac{1}{\sqrt{d}}\mathbf{H}_{1}\teacher=\frac{\Vert\teacher\Vert}{\sqrt{d}}\mathbf{q}=\nu\mathbf{q},\qquad\mathbf{q}\sim\mathcal{N}(\mathbf{0},\mathbf{I}_{n}).
\]
Under our statistical model \cref{sec:setting}, $y_i\mid q_i\sim\Ber(\pi_\rho(\nu q_i)),$ where $\pi_\rho(g)=\rho+(1-2\rho)\sigma(g).$
Hence $(\mathbf y,\mathbf H_1)$ is independent of $\mathbf H_2$,
so the CGMT applies. Since the optimization problem as well as the conditions of CGMT apply we write its corresponding auxiliary optimization problem
\begin{align*}
\min_{\mathbf{w},\mathbf{u}}\max_{\mathbf{v}}\frac{1}{n}\mathbf{1}^{\top}\phi(\mathbf{u})-\frac{1}{n}\mathbf{y}^{\top}\mathbf{u}+\frac{\lambda}{d}f(\mathbf{w})+\frac{1}{n}\mathbf{v}^{\top}\left(\mathbf{u}-\frac{1}{\sqrt{d}}\mathbf{H}_{1}\mathbf{w}\right)&\\-\frac{1}{n\sqrt{d}}\left(\mathbf{v}^{\top}\mathbf{h}\Vert\mathbf{P}^{\perp}\mathbf{w}\Vert+\Vert\mathbf{v}\Vert\mathbf{g}^{\top}\mathbf{P}^{\perp}\mathbf{w}\right),
\end{align*}
where $\mathbf g\sim\mathcal N(\mathbf 0,\mathbf I_d)$ and $\mathbf h\sim\mathcal N(\mathbf 0,\mathbf I_n)$.

We first maximize over $\mathbf v$. For fixed $(\mathbf{w},\mathbf u)$ we have,
\begin{align}
\label{eq:1}    
\max_{\mathbf{v}}\frac{1}{n\sqrt{d}}\Vert\mathbf{v}\Vert\mathbf{g}^{\top}\mathbf{P}^{\perp}\mathbf{w}+\frac{1}{n}\mathbf{v}^{\top}\left(\mathbf{u}-\frac{1}{\sqrt{d}}\mathbf{H}_{1}\mathbf{w}-\frac{\Vert\mathbf{P}^{\perp}\mathbf{w}\Vert}{\sqrt{d}}\mathbf{h}\right).
\end{align}

Let $r=\frac{\Vert\mathbf v\Vert}{\sqrt n},$ maximizing the objective would yield a vector $\mathbf v$ with the same direction of the vector it multiplies, therefore \cref{eq:1} is equivalent to 
\begin{align*}
\max_{r\ge0}\frac{1}{n}\mathbf{1}^{\top}\phi(\mathbf{u})-\frac{1}{n}\mathbf{y}^{\top}\mathbf{u}+\frac{\lambda}{d}f(\mathbf{w})+\frac{r}{\sqrt{nd}}\mathbf{g}^{\top}\mathbf{P}^{\perp}\mathbf{w}+r\left\Vert \frac{\mathbf{u}}{\sqrt{n}}-\frac{1}{\sqrt{nd}}\mathbf{H}_{1}\mathbf{w}-\frac{\Vert\mathbf{P}^{\perp}\mathbf{w}\Vert}{\sqrt{nd}}\mathbf{h}\right\Vert .
\end{align*}
Since $\mathbf H_1=\mathbf H\mathbf P$, and writing $\alpha=\frac{\mathbf{w}^\top\teacher}{\Vert\teacher\Vert^2}$, $\sigma=\frac{\Vert\mathbf P^\perp\mathbf{w}\Vert}{\sqrt d}$, $\mathbf q=\frac{1}{\Vert\teacher\Vert}\mathbf H\teacher,$
we have $\mathbf q\sim\mathcal N(\mathbf 0,\mathbf I_n)$ and
\[
\frac{1}{\sqrt{d}}\mathbf{H}(\mathbf{P}\mathbf{w})=\frac{1}{\sqrt{d}}\mathbf{H}(\alpha\teacher)=\alpha\frac{\Vert\teacher\Vert}{\sqrt{d}}\mathbf{q}=\nu\alpha\mathbf{q},
\]
with $\nu=\frac{\Vert\teacher\Vert}{\sqrt d}.$ Therefore the AO becomes
\begin{align}
\min_{\mathbf{w},\mathbf{u},\alpha,\sigma}\max_{r\ge0}\frac{1}{n}\mathbf{1}^{\top}\phi(\mathbf{u})-\frac{1}{n}\mathbf{y}^{\top}\mathbf{u}+\frac{\lambda}{d}f(\mathbf{w})+\frac{r}{\sqrt{nd}}\mathbf{g}^{\top}\mathbf{P}^{\perp}\mathbf{w}+r\left\Vert \frac{\mathbf{u}}{\sqrt{n}}-\frac{\nu\alpha}{\sqrt{n}}\mathbf{q}-\frac{\sigma}{\sqrt{n}}\mathbf{h}\right\Vert .
\label{eq:app_ao1_continue}
\end{align}
Finding the optimal direction of $\mathbf{w}$ is cumbersome due to the existence of $f(\mathbf{w})$ in the objective, therefore we now introduce
$\boldsymbol\mu,\mathbf s\in\mathbb R^d$ and rewrite the problem as
\begin{align} 
\begin{aligned}
\min_{\mathbf{w},\mathbf{u},\boldsymbol{\mu}}\max_{\mathbf{s},r\ge0}&\frac{1}{n}\mathbf{1}^{\top}\phi(\mathbf{u})-\frac{1}{n}\mathbf{y}^{\top}\mathbf{u}+\frac{\lambda}{d}f(\boldsymbol{\mu})+\frac{r}{\sqrt{nd}}\mathbf{g}^{\top}\mathbf{P}^{\perp}\mathbf{w}+r\left\Vert \frac{\mathbf{u}}{\sqrt{n}}-\frac{\nu\alpha}{\sqrt{n}}\mathbf{q}-\frac{\sigma}{\sqrt{n}}\mathbf{h}\right\Vert +\frac{1}{d}\mathbf{s}^{\top}(\boldsymbol{\mu}-\mathbf{w}).
\end{aligned}
\label{eq:app_mu_s_intro}
\end{align}

We then use the same asymptotically-valid min--max flip where flipping the order of optimizations in (AO) is justified whenever such a flipping is allowed in the (PO) (see Appendix C.2 of~\cite{salehi2019impact}),
and optimize first over the direction of $\mathbf P^\perp\mathbf{w}$. This yields
\begin{align} 
\begin{aligned}
\min_{\boldsymbol{\mu},\mathbf{u},\alpha,\sigma\ge0}\max_{\mathbf{s},r\ge0}&\frac{1}{n}\mathbf{1}^{\top}\phi(\mathbf{u})-\frac{1}{n}\mathbf{y}^{\top}\mathbf{u}+\frac{\lambda}{d}f(\boldsymbol{\mu})-\sigma\left\Vert \frac{1}{\sqrt{d}}\mathbf{P}^{\perp}\left(\frac{r}{\sqrt{\delta}}\mathbf{g}-\mathbf{s}\right)\right\Vert \\&+r\left\Vert \frac{\mathbf{u}}{\sqrt{n}}-\frac{\nu\alpha}{\sqrt{n}}\mathbf{q}-\frac{\sigma}{\sqrt{n}}\mathbf{h}\right\Vert \notag+\frac{1}{d}(\mathbf{P}\mathbf{s})^{\top}\boldsymbol{\mu}+\frac{1}{d}(\mathbf{P}^{\perp}\mathbf{s})^{\top}\boldsymbol{\mu}-\frac{1}{d}(\mathbf{P}\mathbf{s})^{\top}\mathbf{w}.
\end{aligned}
\label{eq:app_after_direction_w}
\end{align}

Next, we introduce $\tau,\upsilon\ge 0$ to replace the norms by squared norms which simplifies the preceding analysis
\begin{equation}   
\begin{aligned}
\min_{\boldsymbol{\mu},\mathbf{u},\alpha,\sigma,\upsilon\ge0}\max_{\mathbf{s},r,\tau\ge0}&\frac{1}{n}\mathbf{1}^{\top}\phi(\mathbf{u})-\frac{1}{n}\mathbf{y}^{\top}\mathbf{u}+\frac{\lambda}{d}f(\boldsymbol{\mu})-\frac{\sigma}{2\tau}-\frac{\sigma\tau}{2}\left\Vert \frac{1}{\sqrt{d}}\mathbf{P}^{\perp}\left(\frac{r}{\sqrt{\delta}}\mathbf{g}-\mathbf{s}\right)\right\Vert ^{2}+\frac{r}{2\upsilon}\\&\qquad+\frac{r\upsilon}{2}\left\Vert \frac{\mathbf{u}}{\sqrt{n}}-\frac{\nu\alpha}{\sqrt{n}}\mathbf{q}-\frac{\sigma}{\sqrt{n}}\mathbf{h}\right\Vert ^{2}+\frac{1}{d}(\mathbf{P}\mathbf{s})^{\top}\boldsymbol{\mu}+\frac{1}{d}(\mathbf{P}^{\perp}\mathbf{s})^{\top}\boldsymbol{\mu}-\frac{1}{d}(\mathbf{P}\mathbf{s})^{\top}\mathbf{w}.
\end{aligned}
\label{eq:app_witnuhau_upsilon}
\end{equation}

We now optimize over $\mathbf s$. Using a completion of squares,
\begin{align*}
&-\frac{\sigma\tau}{2}\left\Vert \frac{1}{\sqrt{d}}\mathbf{P}^{\perp}\left(\frac{r}{\sqrt{\delta}}\mathbf{g}-\mathbf{s}\right)\right\Vert ^{2}+\frac{1}{d}(\mathbf{P}^{\perp}\mathbf{s})^{\top}\boldsymbol{\mu}
\\&=-\frac{\sigma\tau}{2}\left\Vert \frac{1}{\sqrt{d}}\mathbf{P}^{\perp}\left(\frac{r}{\sqrt{\delta}}\mathbf{g}-\mathbf{s}+\frac{1}{\sigma\tau}\boldsymbol{\mu}\right)\right\Vert ^{2}
\quad+\frac{1}{2d\sigma\tau}\left\Vert \mathbf{P}^{\perp}\boldsymbol{\mu}+\sigma\tau\frac{r}{\sqrt{\delta}}\mathbf{P}^{\perp}\mathbf{g}\right\Vert ^{2}-\frac{\sigma\tau r^{2}}{2n}\Vert\mathbf{P}^{\perp}\mathbf{g}\Vert^{2}.
\end{align*}
Since $\mathbf g\sim\mathcal N(\mathbf 0,\mathbf I_d)$, we have
$\frac{\sigma\tau r^2}{2n}\Vert\mathbf P^\perp\mathbf g\Vert^2\asymp\frac{\sigma\tau r^2}{2\delta},$ and,
\begin{align*}
\frac{1}{d}\left\Vert \mathbf{P}^{\perp}\boldsymbol{\mu}+\sigma\tau\frac{r}{\sqrt{\delta}}\mathbf{P}^{\perp}\mathbf{g}\right\Vert ^{2}&=\frac{1}{d}\left\Vert \mathbf{P}^{\perp}\left(\boldsymbol{\mu}+\sigma\tau\frac{r}{\sqrt{\delta}}\mathbf{g}\right)\right\Vert ^{2}
=\frac{1}{d}\left\Vert \boldsymbol{\mu}+\sigma\tau\frac{r}{\sqrt{\delta}}\mathbf{g}\right\Vert ^{2}-\frac{1}{d}\left\Vert \mathbf{P}\left(\boldsymbol{\mu}+\sigma\tau\frac{r}{\sqrt{\delta}}\mathbf{g}\right)\right\Vert ^{2}\\&=\frac{1}{d}\left\Vert \boldsymbol{\mu}+\sigma\tau\frac{r}{\sqrt{\delta}}\mathbf{g}\right\Vert ^{2}-\frac{1}{d}\Vert\mathbf{P}\boldsymbol{\mu}\Vert^{2}-\frac{\sigma^{2}\tau^{2}r^{2}}{\delta d}\Vert\mathbf{P}\mathbf{g}\Vert^{2}-\frac{2\sigma\tau r}{\sqrt{\delta}d}(\mathbf{P}\mathbf{g})^{\top}\boldsymbol{\mu}.
\end{align*}   
And as $d\to\infty$, $\frac{1}{d}\Vert\mathbf P\mathbf g\Vert^2\to0$ and $\frac{1}{d}(\mathbf P\mathbf g)^\top\mu \to0$ almost surely. Since $\mathbf P\boldsymbol\mu$ is constrained by $\frac{1}{d}\teacher^\top\boldsymbol\mu=\alpha\nu^2,$ it follows that $\frac{1}{d}\Vert\mathbf P\boldsymbol\mu\Vert^2=\nu^2\alpha^2.$
Therefore \cref{eq:app_witnuhau_upsilon} reduces asymptotically to
\begin{align}
\min_{\boldsymbol{\mu},\mathbf{u},\alpha,\sigma,\upsilon\ge0}\max_{r,\tau\ge0}&\frac{1}{n}\mathbf{1}^{\top}\phi(\mathbf{u})-\frac{1}{n}\mathbf{y}^{\top}\mathbf{u}-\frac{\sigma}{2\tau}-\frac{\sigma\tau r^{2}}{2\delta}+\frac{r}{2\upsilon}-\frac{\nu^{2}\alpha^{2}}{2\sigma\tau}+\frac{r\upsilon}{2}\left\Vert \frac{\mathbf{u}}{\sqrt{n}}-\frac{\nu\alpha}{\sqrt{n}}\mathbf{q}-\frac{\sigma}{\sqrt{n}}\mathbf{h}\right\Vert ^{2}\notag\\&\qquad+\frac{1}{2d\sigma\tau}\left\Vert \boldsymbol{\mu}+\sigma\tau\frac{r}{\sqrt{\delta}}\mathbf{g}\right\Vert ^{2}+\frac{\lambda}{d}f(\boldsymbol{\mu})\quad\text{s.t. }\frac{1}{d}\mathbf{\mathbf{w}}^{\star\top}\boldsymbol{\mu}=\alpha\nu^{2}.\\&
\label{eq:app_before_theta}
\end{align}
We now optimize over $\boldsymbol\mu$ and attain the following optimization problem,
\begin{align*}
\min_{\boldsymbol\mu\in\mathbb R^d}\max_{\theta\in\mathbb R}\frac{1}{2d\sigma\tau}\left\Vert\boldsymbol\mu+\sigma\tau\frac{r}{\sqrt\delta}\mathbf g\right\Vert^2+\frac{\lambda}{d}f(\boldsymbol\mu) \quad
\text{s.t. }
\frac{1}{d}\mathbf\teacher^\top\boldsymbol\mu=\alpha\nu^2.
\end{align*}
Introducing a Lagrange multiplier $\theta\in\mathbb R$ for the linear constraint,
we obtain
\begin{align}
\min_{\boldsymbol\mu\in\mathbb R^d}\max_{\theta\in\mathbb R}\frac{1}{2d\sigma\tau}\left\Vert\boldsymbol\mu+\sigma\tau\frac{r}{\sqrt\delta}\mathbf g\right\Vert^2+\frac{\lambda}{d}f(\boldsymbol\mu)-\frac{\theta}{d}\teacher^\top\boldsymbol\mu+\alpha\theta\nu^2 .
\label{eq:app_mu_theta}
\end{align}
Completing the square,
\[
\frac{1}{2d\sigma\tau}\left\Vert \boldsymbol{\mu}+\sigma\tau\frac{r}{\sqrt{\delta}}\mathbf{g}\right\Vert ^{2}-\frac{\theta}{d}\teacher^\top\boldsymbol{\mu}=\frac{1}{2d\sigma\tau}\left\Vert \boldsymbol{\mu}+\sigma\tau\frac{r}{\sqrt{\delta}}\mathbf{g}-\sigma\tau\theta\teacher\right\Vert ^{2}-\frac{\sigma\tau\theta^{2}\nu_{d}^{2}}{2}.
\]
where we omitted the term $\frac{1}{d}\mathbf g^\top\teacher$ since $\frac{1}{d}\mathbf g^\top\teacher\to 0$ almost surely. Hence the $\boldsymbol\mu$-minimization is
\[
\max_{\theta\in\mathbb R}\left\{\frac{1}{d}M_{\lambda f(\cdot,\mathbf{F}_{t-1})}\left(\sigma\tau\left(\frac{r}{\sqrt\delta}\mathbf g+\theta\teacher\right),\sigma\tau\right)+\alpha\theta\nu^2-\frac{\sigma\tau\theta^2\nu^2}{2}\right\}.
\]
Substituting this into \cref{eq:app_before_theta}, we arrive at
\begin{align}
\min_{\mathbf u,\alpha,\sigma,\upsilon\ge 0}\max_{r,\tau\ge 0,\theta\in\mathbb R}&\frac{1}{n}\mathbf{1}^\top\phi(\mathbf u)-\frac{1}{n}\mathbf y^\top\mathbf u+\frac{r\upsilon}{2}\left\Vert\frac{\mathbf u}{\sqrt n}-\frac{\nu\alpha}{\sqrt n}\mathbf q-\frac{\sigma}{\sqrt n}\mathbf h\right\Vert^2-\frac{\sigma}{2\tau}-\frac{\sigma\tau r^2}{2\delta}+\frac{r}{2\upsilon}
\notag\\
&-\frac{\nu^2\alpha^2}{2\sigma\tau}+\nu^2\alpha\theta-\frac{\nu^2\sigma\tau\theta^2}{2}+\frac{1}{d}M_{\lambda f(\cdot,\mathbf{F}_{t-1})}\left(\sigma\tau\left(\frac{r}{\sqrt\delta}\mathbf g+\theta\teacher\right),\sigma\tau\right).
\label{eq:app_after_mu}
\end{align}
We next optimize over $\mathbf u$. Consider the $\mathbf u$-dependent term
\[
\frac{1}{n}\mathbf{1}^\top\phi(\mathbf u)-\frac{1}{n}\mathbf y^\top\mathbf u+\frac{r\upsilon}{2}\left\Vert\frac{\mathbf u}{\sqrt n}
-\frac{\nu\alpha}{\sqrt n}\mathbf q-\frac{\sigma}{\sqrt n}\mathbf h\right\Vert^2.
\]
Completing the square gives
\begin{align*}
-\frac{1}{n}\mathbf y^\top\mathbf u+\frac{r\upsilon}{2}\left\Vert\frac{\mathbf u}{\sqrt n}-\frac{\nu\alpha}{\sqrt n}\mathbf q-\frac{\sigma}{\sqrt n}\mathbf h\right\Vert^2&=\frac{r\upsilon}{2}\left\Vert\frac{\mathbf u}{\sqrt n}-\frac{\nu\alpha}{\sqrt n}\mathbf q-\frac{\sigma}{\sqrt n}\mathbf h-\frac{1}{r\upsilon\sqrt n}\mathbf y\right\Vert^2
\\&\quad
-\frac{1}{2r\upsilon}\frac{\Vert\mathbf y\Vert^2}{n}-\nu\alpha\frac{\mathbf y^\top\mathbf q}{n}-\sigma\frac{\mathbf y^\top\mathbf h}{n}.
\end{align*}
Using, $ y_i\mid q_i \sim \Ber(\pi_\rho(\nu q_i)).$ Therefore
\[
\frac{1}{n}\Vert\mathbf y\Vert^2=\frac{1}{n}\sum_{i=1}^n y_i\xrightarrow{\text{a.s.}}\E[\pi_\rho(\nu Z)]=\frac{1}{2},
\]
using $\pi_\rho(-t)=1-\pi_\rho(t)$, and
\[
\frac{1}{n}\mathbf y^\top\mathbf q=\frac{1}{n}\sum_{i=1}^n y_i q_i\xrightarrow{\text{a.s.}}\E[Z\pi_\rho(\nu Z)]=\nu\E[\pi_\rho'(\nu Z)],
\]
by Stein's lemma. Also, $\frac{1}{n}\mathbf y^\top\mathbf h\xrightarrow{\PP}0.$
Thus the $\mathbf u$-optimization is asymptotically equivalent to
\begin{align*}
\frac{1}{n}M_\phi\left(\nu\alpha\mathbf q+\sigma\mathbf h+\frac{1}{r\upsilon}\mathbf y,\frac{1}{r\upsilon}\right)-\frac{1}{4r\upsilon}-\nu^2\alpha\E[\pi_\rho'(\nu Z)].
\end{align*}

Hence \cref{eq:app_after_mu} becomes
\begin{align}
\min_{\alpha,\sigma,\upsilon\ge 0}\max_{r,\tau\ge 0,\theta\in\mathbb R}&-\frac{\sigma}{2\tau}-\frac{\sigma\tau r^2}{2\delta}+\frac{r}{2\upsilon}-\frac{\nu^2\alpha^2}{2\sigma\tau}+\nu^2\alpha\theta-\frac{\nu^2\sigma\tau\theta^2}{2}-\frac{1}{4r\upsilon}-\nu^2\alpha\E[\pi_\rho'(\nu Z)]
\notag\\
&\qquad
+\frac{1}{d} M_{\lambda f(\cdot,\mathbf{F}_{t-1})}\left(\sigma\tau\left(\frac{r}{\sqrt\delta}\mathbf g+\theta\teacher\right),\sigma\tau\right)+\frac{1}{n}M_\phi\left(\nu\alpha\mathbf q+\sigma\mathbf h+\frac{1}{r\upsilon}\mathbf y,\frac{1}{r\upsilon}\right).
\label{eq:app_scalar_finite}
\end{align}
Since the regularizer is coordinatewise decomposable,
\begin{align*}
\frac{1}{d}M_{\lambda f(\cdot,\boldsymbol{\mathbf{F}}_{t-1})}\left(\sigma\tau\left(\frac{r}{\sqrt{\delta}}\mathbf{g}+\theta\teacher\right),\sigma\tau\right)=\frac{1}{d}\sum_{j=1}^{d}M_{\lambda\widetilde{f}(\cdot,F_{t-1,j})}\left(\sigma\tau\left(\frac{r}{\sqrt{\delta}}g_{j}+\theta w_{\star,j}\right),\sigma\tau\right).
\end{align*}
Let $\Xi:\mathbb R^3\to\mathbb R$ be locally Lipschitz and have at most quadratic growth, so that
\[
|\Xi(u,w,F)|\leq C_\Xi(1+u^2+w^2+F^2).
\]
We first show the continuity and growth properties of the proximal operator that will be used later. Fix a compact parameter set $\zeta=(\sigma,\tau,\theta,r)\in K$ on which $\sigma\tau$ is bounded away from zero. By the Moreau-envelope identity,
\[
\nabla_{x}M_{\lambda\widetilde{f}(\cdot,F)}(x,c)=\frac{1}{c}\left(x-\prox_{\lambda c\widetilde{f}(\cdot,F)}(x)\right).
\]
For $c$ bounded away from zero, these gradients are uniformly Lipschitz in $x$. Because, 
\begin{align*}
&\left|\left(x-\prox_{\lambda c\widetilde{f}(\cdot,F)}(x)\right)-\left(y-\prox_{\lambda c\widetilde{f}(\cdot,F)}(y)\right)\right|^{2}\\&\leq\left|x-y\right|^{2}+\left|\prox_{\lambda c\widetilde{f}(\cdot,F)}(x)-\prox_{\lambda c\widetilde{f}(\cdot,F)}(y)\right|^{2}-2\left(\prox_{\lambda c\widetilde{f}(\cdot,F)}(x)-\prox_{\lambda c\widetilde{f}(\cdot,F)}(y)\right)(x-y)\leq|x-y|^{2}.
\end{align*}
from the firm non-expansiveness of proximal operator. Hence 
\begin{align*}&
\left|(x-\prox_{\lambda c\widetilde{f}(\cdot,F)}(x))-(y-\prox_{\lambda c\widetilde{f}(\cdot,F)}(y))\right|\leq |x-y|,\\&\left|\nabla_{x}M_{\lambda\widetilde{f}(\cdot,F)}(x,c)-\nabla_{x}M_{\lambda\widetilde{f}(\cdot,F)}(y,c)\right|\leq\frac{1}{c}|x-y|.\end{align*}
On compact sets, using $\left|\nabla_{x}M(x,z)-\frac{M(x+h,z)-M(x,z)}{h}\right|\leq\frac{1}{h}\int_{0}^{h}\left|\nabla_{x}M(x,z)-\nabla_{x}M(x+t,z)\right|\dd t\leq\frac{h}{2c_{\min}}.$
Let $(x_n,z_n)\to(x,z)$. For fixed $h$, joint continuity of $M$ gives $\frac{M(x+h,z_n)-M(x,z_n)}{h}\longrightarrow\frac{M(x+h,z)-M(x,z)}{h}$, together with the uniform Lipschitz of the Moreau gradient which handles $x_n\to x$ imply joint continuity of its gradient. Also the proximal map is jointly continuity when restricted to compact set $K$ by the Moreau envelope identity.

By the direct definition of the Moreau envelope $M_{\lambda\widetilde{f}(\cdot,v)}(x,\tau)=\min_{y}\left\{ \lambda\widetilde{f}(y,v)+\frac{1}{2\tau}(y-x)^{2}\right\} $, non negativity of $\widetilde f$ also gives,
\[
\frac{1}{2\sigma\tau}\left|\prox_{\lambda\sigma\tau\widetilde{f}(\cdot,F)}\left(\sigma\tau\left(\theta w+\frac{r}{\sqrt{\delta}}z\right)\right)-\sigma\tau\left(\theta w+\frac{r}{\sqrt{\delta}}z\right)\right|^{2}\leq M_{\lambda\widetilde{f}(\cdot,F)}\left(\sigma\tau\left(\theta w+\frac{r}{\sqrt{\delta}}z\right),\sigma\tau\right).
\]
The quadratic-growth bound on the Moreau envelope and compactness of $K$ therefore imply,
\[
\sup_{\zeta\in K}\left|\prox_{\lambda\sigma\tau\widetilde{f}(\cdot,F)}\left(\sigma\tau\left(\theta w+\frac{r}{\sqrt{\delta}}z\right)\right)\right|\leq C_{K}(1+|w|+|F|+|z|).
\]
Consequently,
\[
\begin{aligned}
\sup_{\zeta\in K}\Bigg|\Xi\Bigg(\prox_{\lambda\sigma\tau\widetilde{f}(\cdot,F)}\left(\sigma\tau\left(\theta w+\frac{r}{\sqrt{\delta}}z\right)\right),w,F\Bigg)\Bigg|\leq C_{K,\Xi}(1+w^{2}+F^{2}+z^{2}).
\end{aligned}
\]

We now freeze the scalar parameters at
$\bar\zeta_t=(\bar\sigma_t,\bar\tau_t,\bar r_t,\bar\theta_t)$.
Conditionally on the teacher and the first $t-1$ tasks, the pairs
$(w_{\star,j},F_{t-1,j})$ are fixed, while
$Z_1,\ldots,Z_d$ are independent standard Gaussians. Therefore,
the random variables $\Xi\left(y,w_{\star,j},F_{t-1,j}\right)-\mathbb{E}_{Z}\left[\Xi\left(y,w_{\star,j},F_{t-1,j}\right)\right]$
with $y=\prox_{\lambda\bar{\sigma}_{t}\bar{\tau}_{t}\widetilde{f}(\cdot,F_{t-1,j})}\left(\bar{\sigma}_{t}\bar{\tau}_{t}\left(\bar{\theta}_{t}w_{\star,j}+\frac{\bar{r}_{t}}{\sqrt{\delta}}Z_{j}\right)\right)$ are conditionally independent and centered.

We split the domain to two regimes, when $(w,F,Z)$ norm is bounded by some $R$ and when its norm is larger than R. Let $\chi_R:\mathbb R^3\to[0,1]$ be the indicator function that equals $1$ when $w^2+F^2+z^2\leq R^2$ and $0$ when
$w^2+F^2+z^2> R^2$. Conditionally on the teacher and the
first $t-1$ tasks, the centered summands
\[
\begin{aligned}
&\chi_{R}(w_{\star,j},F_{t-1,j},Z_{j})\Xi\left(y,w_{\star,j},F_{t-1,j}\right)-\mathbb{E}_{Z}\left[\chi_{R}(w_{\star,j},F_{t-1,j},Z)\Xi\left(y,w_{\star,j},F_{t-1,j}\right)\right]
\end{aligned}
\]
are independent and have conditional mean zero. The map
\[
(w,F,z)\longmapsto\Xi\left(\prox_{\lambda\bar{\sigma}_{t}\bar{\tau}_{t}\widetilde{f}(\cdot,F)}\left(\bar{\sigma}_{t}\bar{\tau}_{t}\left(\bar{\theta}_{t}w+\frac{\bar{r}_{t}}{\sqrt{\delta}}z\right)\right),w,F\right)
\]
is continuous, as it is implied by the local lipschitzness. It is therefore bounded on the compact set where
$\{(w,F,z):w^2+F^2+z^2\leq R^2\}$. In particular, there is a
constant $B_R<\infty$ such that
\[
\left|\chi_{R}(w,F,z)\Xi\left(\prox_{\lambda\bar{\sigma}_{t}\bar{\tau}_{t}\widetilde{f}(\cdot,F)}\left(\bar{\sigma}_{t}\bar{\tau}_{t}\left(\bar{\theta}_{t}w+\frac{\bar{r}_{t}}{\sqrt{\delta}}z\right)\right),w,F\right)\right|\leq B_{R}.
\]
Consequently, conditionally on $(\teacher,\mathbf F_{t-1})$, the variance of each term is at most $B_R^2$, since $\mathrm{Var}(X)\leq\mathbb E[X^2]$ and its absolute value is at most $B_R$. Subtracting its conditional expectation does not change its variance.

The centered terms are conditionally independent because each depends only on the corresponding independent Gaussian variable $Z_j$. Therefore, the variance of their average is the sum of their variances divided by $d^2$, and hence is bounded by $\frac{1}{d^{2}}\sum_{j=1}^{d}B_{R}^{2}=\frac{B_{R}^{2}}{d}\longrightarrow0.$ Conditional Chebyshev's inequality then implies that the empirical average of the centered terms converges to zero in probability.

Outside this region, when $w^2+F^2+z^2>R^2$ we use the bound
\[
\begin{aligned}
\Bigg|\Xi\Bigg(&\prox_{\lambda\bar{\sigma}_{t}\bar{\tau}_{t}\widetilde{f}(\cdot,F)}\left(\bar{\sigma}_{t}\bar{\tau}_{t}\left(\bar{\theta}_{t}w+\frac{\bar{r}_{t}}{\sqrt{\delta}}z\right)\right),w,F\Bigg)\Bigg|\leq C(1+w^{2}+F^{2}+z^{2}).
\end{aligned}
\]
Indeed, $\mathbb{E}\left[(1+Z^{2})\mathbf{1}_{\{|Z|>R\}}\right]=2R\varphi(R)+4\Phi(-R)\longrightarrow0,$ and the assumed 2-Wasserstein distance convergence of the empirical state distribution implies, for every $\epsilon>0$,
\begin{align}
\label{eq:20} 
\lim_{R\to\infty}\limsup_{d\to\infty}\mathbb{P}\left(\frac{1}{d}\sum_{j=1}^{d}(1+{w_{\star,j}}^{2}+F_{t-1,j}^{2})\mathbf{1}_{\{{w_{\star,j}}^{2}+F_{t-1,j}^{2}>R^{2}\}}>\epsilon\right)=0.
\end{align}
Since it follows from $w^2+F^2+Z^2>R^2$ that $w^2+F^2>\frac{R^2}{2}$ or $Z^2>\frac{R^2}{2},$
\[(1+w^2+F^2+Z^2)\mathbf 1_{\{w^2+F^2+Z^2>R^2\}}\leq(1+w^2+F^2+Z^2)\mathbf 1_{\{w^2+F^2>\frac{R^2}{2}\}}+(1+w^2+F^2+Z^2)\mathbf 1_{\{Z^2>\frac{R^2}{2}\}},
\]
The first term is controlled by the $2$-Wasserstein tail bound, and the second by the Gaussian tail together with uniform boundedness of the empirical second moments and $\PP(Z^2>\frac{R^2}{2})\to0$. It follows that,
\begin{equation}
\label{eq:37}
\begin{aligned}
&\frac{1}{d}\sum_{j=1}^{d}\Bigg\{\Xi\left(\prox_{\lambda\bar{\sigma}_{t}\bar{\tau}_{t}\widetilde{f}(\cdot,F_{t-1,j})}\left(\bar{\sigma}_{t}\bar{\tau}_{t}\left(\bar{\theta}_{t}w_{\star,j}+\frac{\bar{r}_{t}}{\sqrt{\delta}}Z_{j}\right)\right),w_{\star,j},F_{t-1,j}\right)-\\&\qquad\mathbb{E}_{Z}\left[\Xi\left(\prox_{\lambda\bar{\sigma}_{t}\bar{\tau}_{t}\widetilde{f}(\cdot,F_{t-1,j})}\left(\bar{\sigma}_{t}\bar{\tau}_{t}\left(\bar{\theta}_{t}w_{\star,j}+\frac{\bar{r}_{t}}{\sqrt{\delta}}Z\right)\right),w_{\star,j},F_{t-1,j}\right)\right]\Bigg\}\xrightarrow{\mathbb{P}}0.
\end{aligned}
\end{equation}

Next, for each fixed $z$ the map $(w,F)\longmapsto\Xi\left(\prox_{\lambda\bar{\sigma}_{t}\bar{\tau}_{t}\widetilde{f}(\cdot,F)}\left(\bar{\sigma}_{t}\bar{\tau}_{t}\left(\bar{\theta}_{t}w+\frac{\bar{r}_{t}}{\sqrt{\delta}}z\right)\right),w,F\right)$ is jointly continuous by the joint continuity of the proximal map and continuity of $\Xi$, this also shows that,
\[
(w,F)\longmapsto\mathbb{E}_{Z}\left[\Xi\left(\prox_{\lambda\bar{\sigma}_{t}\bar{\tau}_{t}\widetilde{f}(\cdot,F)}\left(\bar{\sigma}_{t}\bar{\tau}_{t}\left(\bar{\theta}_{t}w+\frac{\bar{r}_{t}}{\sqrt{\delta}}Z\right)\right),w,F\right)\right]
\]
is continuous. Indeed, for every series $(w_n,F_n)\to(w,F)$ we have (as $\forall n:\Vert(w_n,F_n)\Vert<R$ because it converges), \[\left|\Xi\left(\prox_{\lambda\bar{\sigma}_{t}\bar{\tau}_{t}\widetilde{f}(\cdot,F_{n})}\left(\bar{\sigma}_{t}\bar{\tau}_{t}\left(\bar{\theta}_{t}w_{n}+\frac{\bar{r}_{t}}{\sqrt{\delta}}z\right)\right),w_n,F_n\right)\right|\leq C_{R}\left(1+z^{2}\right).\] Thus $C_R(1+z^2)$ is integrable with $z\sim \mathcal N(0,1)$, and dominated convergence yields convergence of the expectations. The same bound gives
\begin{equation}
\label{eq:19}
\begin{aligned}
&\left|\mathbb{E}_{Z}\left[\Xi\left(\prox_{\lambda\bar{\sigma}_{t}\bar{\tau}_{t}\widetilde{f}(\cdot,F)}\left(\bar{\sigma}_{t}\bar{\tau}_{t}\left(\bar{\theta}_{t}w+\frac{\bar{r}_{t}}{\sqrt{\delta}}Z\right)\right),w,F\right)\right]\right|\\&\leq\mathbb{E}_{Z}\left[\left|\Xi\left(\prox_{\lambda\bar{\sigma}_{t}\bar{\tau}_{t}\widetilde{f}(\cdot,F)}\left(\bar{\sigma}_{t}\bar{\tau}_{t}\left(\bar{\theta}_{t}w+\frac{\bar{r}_{t}}{\sqrt{\delta}}Z\right)\right),w,F\right)\right|\right]\\&\leq C\mathbb{E}_{Z}\left[1+w^{2}+F^{2}+Z^{2}\right]=C\left(2+w^{2}+F^{2}\right)\leq C'\left(1+w^{2}+F^{2}\right).
\end{aligned}
\end{equation}
The weak convergence on a bounded set together with the second moment tail bound (by the assumed convergence in 2-Wasserstein distance) on the complement set, and \cref{eq:19,eq:20}, therefore give,
\begin{equation}
\label{eq:38}
\begin{aligned}
&\frac{1}{d}\sum_{j=1}^{d}\mathbb{E}_{Z}\left[\Xi\left(\prox_{\lambda\bar{\sigma}_{t}\bar{\tau}_{t}\widetilde{f}(\cdot,F_{t-1,j})}\left(\bar{\sigma}_{t}\bar{\tau}_{t}\left(\bar{\theta}_{t}w_{\star,j}+\frac{\bar{r}_{t}}{\sqrt{\delta}}Z\right)\right),w_{\star,j},F_{t- 1,j}\right)\right]\\&\qquad\xrightarrow{\mathbb{P}}\mathbb{E}\left[\Xi\left(\prox_{\lambda\bar{\sigma}_{t}\bar{\tau}_{t}\widetilde{f}(\cdot,F_{t-1})}\left(\bar{\sigma}_{t}\bar{\tau}_{t}\left(\bar{\theta}_{t}w^{\star}+\frac{\bar{r}_{t}}{\sqrt{\delta}}Z\right)\right),w^{\star},F_{t-1}\right)\right],
\end{aligned}
\end{equation}

where $Z\sim\mathcal N(0,1)$ is independent of $(w^\star,F_{t-1})$. 

It remains to replace the limiting scalar parameters by the random
AO scalar parameters. By the AO proximal representation,
\[
\widehat{w}_{j}^{\text{AO}}=\prox_{\lambda\sigma_{t,d}\tau_{t,d}\widetilde{f}(\cdot,F_{t-1,j})}\left(\sigma_{t,d}\tau_{t,d}\left(\theta_{t,d}w_{\star,j}+\frac{r_{t,d}}{\sqrt{\delta}}Z_{j}\right)\right),
\]
and, by scalar-optimizer convergence, $\zeta_{t,d}:=(\sigma_{t,d},\tau_{t,d},r_{t,d},\theta_{t,d})\xrightarrow{\mathbb P}
\bar\zeta_t.$
Since $\bar\sigma_t\bar\tau_t>0$, choose a compact neighborhood
$K$ of $\bar\zeta_t$ on which $\sigma\tau$ is bounded away
from zero. Then $\mathbb P(\zeta_{t,d}\notin K)\longrightarrow0,$ for all large $d.$ On the event $\{\zeta_{t,d}\in K\}$, truncate $(w_{\star,j},F_{t-1,j},Z_j)$ to a fixed compact set. Uniform continuity of the composed proximal map on this compact set and
$\zeta_{t,d}\xrightarrow{\mathbb P}\bar\zeta_t$ imply that the empirical average of the differences converges to zero. The contribution outside the compact set is uniformly controlled by $C_K(1+w^2+F^2+z^2)$ and is therefore negligible by the same 2-Wasserstein distance and Gaussian tail estimates. Hence,
\begin{align*}
\frac{1}{d}\sum_{j=1}^{d}\Bigg[\Xi\left(\widehat{w}_{j}^{\text{AO}},w_{\star,j},F_{t-1,j}\right)-\Xi\left(\prox_{\lambda\bar{\sigma}_{t}\bar{\tau}_{t}\widetilde{f}(\cdot,F_{t-1,j})}\left(\bar{\sigma}_{t}\bar{\tau}_{t}\left(\bar{\theta}_{t}w_{\star,j}+\frac{\bar{r}_{t}}{\sqrt{\delta}}Z_{j}\right)\right),w_{\star,j},F_{t-1,j}\right)\Bigg]\xrightarrow{\mathbb{P}}0.
\end{align*}

Combining the parameter-replacement step, the conditional law of
large numbers, and the empirical 2-Wasserstein distance convergence gives
\begin{align}
\label{eq:39}
\frac{1}{d}\sum_{j=1}^{d}\Xi\left(\widehat{w}_{j}^{\text{AO}},w_{\star,j},F_{t-1,j}\right)\xrightarrow{\mathbb{P}}\mathbb{E}\left[\Xi\left(\prox_{\lambda\bar{\sigma}_{t}\bar{\tau}_{t}\widetilde{f}(\cdot,F_{t-1})}\left(\bar{\sigma}_{t}\bar{\tau}_{t}\left(\bar{\theta}_{t}w^{\star}+\frac{\bar{r}_{t}}{\sqrt{\delta}}Z\right)\right),w^{\star},F_{t-1}\right)\right],
\end{align}
where $Z\sim\mathcal N(0,1)$ is independent of
$(w^\star,F_{t-1})$.
Likewise, setting
\[
\gamma=\frac{1}{r\upsilon},\qquad B\mid Z_1\sim\Ber(\pi_\rho(\nu Z_1)),
\]
we have
\[
\frac{1}{n}M_{\phi}\left(\nu\alpha\mathbf{q}+\sigma\mathbf{h}+\gamma\mathbf{y},\gamma\right)\xrightarrow{\text{a.s.}}\E\left[M_{\phi}\left(\nu\alpha Z_{1}+\sigma Z_{2}+\gamma B,\gamma\right)\right].
\]
Therefore, by the preceding convergence of the two empirical Moreau envelopes and~\cite[Appendix A, Lemma 9]{thrampoulidis2018precise}, the finite-dimensional scalar optimization converges to its deterministic limiting counterpart, which leaves us with, 
\begin{align}
\label{eq:app_scalar_limit_minmax}
\min_{\alpha\in\R,\sigma,\upsilon\ge 0}\max_{r,\tau\ge 0,\theta\in\R}\mathcal C(\alpha,\sigma,r,\tau,\upsilon,\theta)
\end{align}
where,
\begin{align}
\mathcal C(\alpha,\sigma,r,\tau,\upsilon,\theta)=&-\frac{\sigma}{2\tau}-\frac{\sigma\tau r^2}{2\delta}+\frac{r}{2\upsilon}-\frac{\nu^2\alpha^2}{2\sigma\tau}+\nu^2\alpha\theta-\frac{\nu^2\sigma\tau\theta^2}{2}-\frac{1}{4r\upsilon}-\nu^2\alpha\E[\pi_\rho'(\nu Z)]\notag\\&+\E\left[M_{\lambda \tilde f(\cdot,\mathbf{F}_{t-1})}\left(\sigma\tau\left(\frac{r}{\sqrt\delta}Z+\theta w^{\star}\right),\sigma\tau\right)\right]+\E\left[
M_\phi\left(\nu\alpha Z_1+\sigma Z_2+\frac{1}{r\upsilon}B,\frac{1}{r\upsilon}\right)\right],
\label{eq:app_scalar_limit}
\end{align}
and $Z,Z_1,Z_2\stackrel{\text{iid}}{\sim}\mathcal N(0,1)$, and $B\mid Z_1\sim\Ber(\pi_\rho(\nu Z_1))$.

We now turn to the first-order optimality conditions of \cref{eq:app_scalar_limit_minmax}. Let $\gamma=\frac{1}{r\upsilon},$
and write the scalar objective as
\begin{align*}
\mathcal{C}(\alpha,\sigma,r,\tau,\upsilon,\theta)=&-\frac{\sigma}{2\tau}-\frac{\sigma\tau r^{2}}{2\delta}+\frac{r}{2\upsilon}-\frac{\nu^{2}\alpha^{2}}{2\sigma\tau}+\nu^{2}\alpha\theta-\frac{\nu^{2}\sigma\tau\theta^{2}}{2}-\frac{1}{4r\upsilon}-\nu^{2}\alpha\E[\pi_{\rho}'(\nu Z)]\\&\qquad+\E\left[M_{\lambda\tilde{f}(\cdot,\mathbf{F}_{t-1})}\left(\sigma\tau\left(\frac{r}{\sqrt{\delta}}Z+\theta w^{\star}\right),\sigma\tau\right)\right]+\E\left[M_{\phi}\left(\nu\alpha Z_{1}+\sigma Z_{2}+\gamma B,\gamma\right)\right],
\end{align*}

Since the objective is smooth on $\{\sigma>0,r>0,\tau>0,\upsilon>0\}$, the unique saddle point
$(\bar\alpha,\bar\sigma,\bar{r},\bar\tau,\bar\upsilon,\bar\theta)$ satisfies the first-order optimality
conditions. We derive them one by one.
First we differentiate the objective with respect to $\theta$, only the terms
\[
\nu^2\alpha\theta-\frac{\nu^2\sigma\tau\theta^2}{2}+\E\left[M_{\lambda \tilde f(\cdot,\mathbf{F}_{t-1})}\left(
\sigma\tau\left(\frac{r}{\sqrt\delta}Z+\theta w^{\star}\right),\sigma\tau\right)\right]
\]
depend on $\theta$. Using the derivative formula for the Moreau envelope \citet[Lemma 2]{salehi2019impact},
\[
\frac{\partial}{\partial x}M_{\lambda f(\cdot,\mathbf{F}_{t-1})}(x,t)=\frac{1}{T}\left(x-\prox_{t\lambda f(\cdot,\mathbf{F}_{t-1})}(x)\right),
\]
we obtain
\begin{align*}
\frac{\partial\mathcal{C}}{\partial\theta}&=\nu^{2}\alpha-\nu^{2}\sigma\tau\theta+\E\left[\frac{1}{\sigma\tau}\left(\sigma\tau\left(\frac{r}{\sqrt{\delta}}Z+\theta w^{\star}\right)-\prox_{\lambda\sigma\tau\tilde{f}(\cdot,\mathbf{F}_{t-1})}\left(\sigma\tau\left(\frac{r}{\sqrt{\delta}}Z+\theta w^{\star}\right)\right)\right)\sigma\tau w^{\star}\right]\\&=\nu^{2}\alpha-\nu^{2}\sigma\tau\theta+\sigma\tau\theta\E[{w^{\star}}^{2}]+\frac{r\sigma\tau}{\sqrt{\delta}}\E[Zw^{\star}]-\E\left[w^{\star}\prox_{\lambda\sigma\tau\tilde{f}(\cdot,\mathbf{F}_{t-1})}\left(\sigma\tau\left(\theta w^{\star}+\frac{r}{\sqrt{\delta}}Z\right)\right)\right].
\end{align*}
Since $\E[{(w^{\star}})^2]=\nu^2$ and $\E[Z w^{\star}]=0$, the $\theta$-stationarity condition
$\partial_\theta\mathcal C=0$ becomes
\[
\nu^{2}\alpha=\E\left[w^{\star}\prox_{\lambda\sigma\tau\tilde  f(\cdot,\mathbf{F}_{t-1})}\left(\sigma\tau\left(\theta w^{\star}+\frac{r}{\sqrt{\delta}}Z\right)\right)\right].
\]
For fixed $(\alpha,\sigma)$ we define
$F_{\rho}(\gamma)=-\frac{\gamma}{4}+\E\left[M_{\phi}\left(\nu\alpha Z_{1}+\sigma Z_{2}+\gamma B,\gamma\right)\right]$ with $B\mid Z_1\sim\Ber(\pi_\rho(\nu Z_1))$.
Then the $(r,\upsilon)$-dependence of $\mathcal C$ is
\[
-\frac{\sigma\tau r^2}{2\delta}+\frac{r}{2\upsilon}+\E\left[M_\phi\left(\nu\alpha Z_1+\sigma Z_2+\frac{1}{r\upsilon}B,\frac{1}{r\upsilon}\right)\right]-\frac{1}{4r\upsilon}=-\frac{\sigma\tau r^2}{2\delta}+\frac{r}{2\upsilon}+F_\rho\left(\frac{1}{r\upsilon}\right).
\]
We now compute $F_\rho'(\gamma)$. By differentiation under the expectation and the Moreau-envelope
derivative formula,
\begin{align*}
F_\rho'(\gamma)&=-\frac{1}{4}+\frac{\dd}{\dd\gamma}\E\left[M_\phi\left(\nu\alpha Z_1+\sigma Z_2+\gamma B,\gamma\right)\right]\\&=-\frac{1}{4}-\E\left[\frac{1-B}{2\gamma^2}\left(
\nu\alpha Z_1+\sigma Z_2-\prox_{\gamma\phi}(\nu\alpha Z_1+\sigma Z_2)
\right)^2\right]\\&\qquad-\E\left[\frac{B}{2\gamma^2}\left(\nu\alpha Z_1+\sigma Z_2+\gamma-\prox_{\gamma\phi}(\nu\alpha Z_1+\sigma Z_2+\gamma)\right)^2\right]\\&\qquad+\E\left[\frac{B}{\gamma}\left(\nu\alpha Z_1+\sigma Z_2+\gamma-\prox_{\gamma\phi}(\nu\alpha Z_1+\sigma Z_2+\gamma)\right)\right].
\end{align*}
Using
\[
\PP(B=1\mid Z_1)=\pi_\rho(\nu Z_1),
\qquad
\PP(B=0\mid Z_1)=1-\pi_\rho(\nu Z_1),
\]
the symmetry identity $\prox_{\gamma\phi}(x+\gamma)=-\prox_{\gamma\phi}(-x),$
and $\pi_\rho(-t)=1-\pi_\rho(t)$, this simplifies to
\begin{align}\label{eq:app_Fprime_corr_full}
F_{\rho}'(\gamma)=-\frac{1}{\gamma^{2}}\E\left[\left(1-\pi_{\rho}(\nu Z_{1})\right)\left(\nu\alpha Z_{1}+\sigma Z_{2}-\prox_{\gamma\phi}(\nu\alpha Z_{1}+\sigma Z_{2})\right)^{2}\right].
\end{align}
Differentiating $\mathcal C$ with respect to $r$ and $\upsilon$, gives
\begin{align*}
\frac{\partial\mathcal{C}}{\partial r}&=-\frac{\sigma\tau r}{\delta}+\frac{1}{2\upsilon}-\frac{1}{\upsilon r^{2}}F_{\rho}'\left(\frac{1}{r\upsilon}\right)+\frac{\partial}{\partial r}\E\left[M_{\lambda \tilde f(\cdot,\mathbf{F}_{t-1})}\left(\sigma\tau\left(\frac{r}{\sqrt{\delta}}Z+\theta w^{\star}\right),\sigma\tau\right)\right],\\\frac{\partial\mathcal{C}}{\partial\upsilon}&=-\frac{r}{2\upsilon^{2}}-\frac{1}{r\upsilon^{2}}F_{\rho}'\left(\frac{1}{r\upsilon}\right).
\end{align*}
For the regularizer-envelope term,
\begin{align*}
&\frac{\partial}{\partial r}\E\left[M_{\lambda \tilde f(\cdot,\mathbf{F}_{t-1})}\left(\sigma\tau\left(\frac{r}{\sqrt{\delta}}Z+\theta w^{\star}\right),\sigma\tau\right)\right]
\\&=\E\left[\frac{1}{\sigma\tau}\left(\sigma\tau\left(\frac{r}{\sqrt{\delta}}Z+\theta w^{\star}\right)-\prox_{\lambda\sigma\tau \tilde f(\cdot,\mathbf{F}_{t-1})}\left(\sigma\tau\left(\frac{r}{\sqrt{\delta}}Z+\theta w^{\star}\right)\right)\right)\frac{\sigma\tau}{\sqrt{\delta}}Z\right]\\&=\frac{\sigma\tau r}{\delta}-\frac{1}{\sqrt{\delta}}\E\left[Z\prox_{\lambda\sigma\tau \tilde f(\cdot,\mathbf{F}_{t-1})}\left(\sigma\tau\left(\theta w^{\star}+\frac{r}{\sqrt{\delta}}Z\right)\right)\right].
\end{align*}
Substituting this into $\partial_r\mathcal C=0$, and combining with $\partial_\upsilon\mathcal C=0$,
we obtain (after setting $\gamma=1/(r\upsilon)$),
\[
\gamma=\frac{1}{r\sqrt{\delta}}\E\left[Z\prox_{\lambda\sigma\tau \tilde f(\cdot,\mathbf{F}_{t-1})}\left(\sigma\tau\left(\theta w^{\star}+\frac{r}{\sqrt{\delta}}Z\right)\right)\right]
\]
and
\[
\gamma^{2}=\frac{2}{r^{2}}\E\left[\left(1-\pi_{\rho}(\nu Z_{1})\right)\left(\nu\alpha Z_{1}+\sigma Z_{2}-\prox_{\gamma\phi}(\nu\alpha Z_{1}+\sigma Z_{2})\right)^{2}\right].
\]
Differentiating $\mathcal C$ with respect to $\tau$ yields
\begin{align*}
\frac{\partial\mathcal{C}}{\partial\tau}=\frac{\sigma}{2\tau^{2}}-\frac{\sigma r^{2}}{2\delta}+\frac{\nu^{2}\alpha^{2}}{2\sigma\tau^{2}}-\frac{\nu^{2}\sigma\theta^{2}}{2}+\frac{\partial}{\partial\tau}\E\left[M_{\lambda \tilde f(\cdot,\mathbf{F}_{t-1})}\left(\sigma\tau\left(\frac{r}{\sqrt{\delta}}Z+\theta w^{\star}\right),\sigma\tau\right)\right].
\end{align*}
Again by the Moreau-envelope derivative formula,
\begin{align*}
\frac{\partial}{\partial\tau}\E\left[M_{\lambda \tilde f(\cdot,\mathbf{F}_{t-1})}\left(\sigma\tau\left(\frac{r}{\sqrt{\delta}}Z+\theta w^{\star}\right),\sigma\tau\right)\right]
&=\frac{\sigma}{2}\left(\frac{r^{2}}{\delta}+\nu^{2}\theta^{2}\right)-\frac{1}{2\sigma\tau^{2}}\E\left[\prox_{\lambda\sigma\tau \tilde f(\cdot,\mathbf{F}_{t-1})}\left(\sigma\tau\left(\theta w^{\star}+\frac{r}{\sqrt{\delta}}Z\right)\right)^{2}\right].
\end{align*}
Substituting into $\partial_\tau\mathcal C=0$ gives
\[
\nu^{2}\alpha^{2}+\sigma^{2}=\E\left[\prox_{\lambda\sigma\tau \tilde f(\cdot,\mathbf{F}_{t-1})}\left(\sigma\tau\left(\theta w^{\star}+\frac{r}{\sqrt{\delta}}Z\right)\right)^{2}\right].
\]
Differentiating $\mathcal C$ with respect to $\sigma$, we obtain
\begin{align*}
&\frac{\partial\mathcal{C}}{\partial\sigma}=-\frac{1}{2\tau}-\frac{\tau r^{2}}{2\delta}+\frac{\nu^{2}\alpha^{2}}{2\sigma^{2}\tau}-\frac{\nu^{2}\tau\theta^{2}}{2}+\frac{\partial}{\partial\sigma}\E\left[M_{\lambda \tilde f(\cdot,\mathbf{F}_{t-1})}\left(\sigma\tau\left(\frac{r}{\sqrt{\delta}}Z+\theta w^{\star}\right),\sigma\tau\right)\right]\\&\hspace{14em}+\frac{\partial}{\partial\sigma}\E\left[M_{\phi}\left(\nu\alpha Z_{1}+\sigma Z_{2}+\gamma B,\gamma\right)\right].
\end{align*}
For the regularizer part,
\begin{align*}
\frac{\partial}{\partial\sigma}\E\left[M_{\lambda \tilde f(\cdot,\mathbf{F}_{t-1})}\left(\sigma\tau\left(\frac{r}{\sqrt{\delta}}Z+\theta w^{\star}\right),\sigma\tau\right)\right]
&=\frac{\tau}{2}\left(\frac{r^{2}}{\delta}+\nu^{2}\theta^{2}\right)-\frac{1}{2\sigma^{2}\tau}\E\left[\prox_{\lambda\sigma\tau \tilde f(\cdot,\mathbf{F}_{t-1})}\left(\sigma\tau\left(\theta w^{\star}+\frac{r}{\sqrt{\delta}}Z\right)\right)^{2}\right].
\end{align*}
For the data-fit part,
\begin{align*}
&\frac{\partial}{\partial\sigma}\E\left[M_{\phi}\left(\nu\alpha Z_{1}+\sigma Z_{2}+\gamma B,\gamma\right)\right]=\frac{\sigma}{\gamma}-\frac{2}{\gamma}\E\left[Z_{2}\left(1-\pi_{\rho}(\nu Z_{1})\right)\prox_{\gamma\phi}(\nu\alpha Z_{1}+\sigma Z_{2})\right]\\&\hspace{14em}=\frac{\sigma}{\gamma}\left(1-\E\left[\frac{2(1-\pi_{\rho}(\nu Z_{1}))}{1+\gamma\phi''(\prox_{\gamma\phi}(\nu\alpha Z_{1}+\sigma Z_{2}))}\right]\right),
\end{align*}
where in the second equality we used Stein's lemma together with
\[
\frac{\dd}{\dd x}\prox_{\gamma\phi}(x)=\frac{1}{1+\gamma\phi''(\prox_{\gamma\phi}(x))}.
\]
Substituting these formulas into $\partial_\sigma\mathcal C=0$, and using the $\tau$-equation already derived,
yields
\[
1-\frac{\gamma}{\sigma\tau}=\E\left[\frac{2(1-\pi_{\rho}(\nu Z_{1}))}{1+\gamma\phi''(\prox_{\gamma\phi}(\nu\alpha Z_{1}+\sigma Z_{2}))}\right].
\]
Differentiating $\mathcal C$ with respect to $\alpha$, we get
\begin{align*}
\frac{\partial\mathcal{C}}{\partial\alpha}=-\frac{\nu^{2}\alpha}{\sigma\tau}+\nu^{2}\theta-\nu^{2}\E[\pi_{\rho}'(\nu Z)]+\frac{\partial}{\partial\alpha}\E\left[M_{\phi}\left(\nu\alpha Z_{1}+\sigma Z_{2}+\gamma B,\gamma\right)\right].
\end{align*}
Now
\begin{align*}
\frac{\partial}{\partial\alpha}\E\left[M_{\phi}\left(\nu\alpha Z_{1}+\sigma Z_{2}+\gamma B,\gamma\right)\right]&=\frac{\nu}{\gamma}\E\left[Z_{1}\left(\nu\alpha Z_{1}+\sigma Z_{2}+\gamma B-\prox_{\gamma\phi}\left(\nu\alpha Z_{1}+\sigma Z_{2}+\gamma B\right)\right)\right]\\&=\frac{\nu^{2}\alpha}{\gamma}+\nu^{2}\E[\pi_{\rho}'(\nu Z_{1})]-\frac{\nu}{\gamma}\E\left[Z_{1}\prox_{\gamma\phi}\left(\nu\alpha Z_{1}+\sigma Z_{2}+\gamma B\right)\right].
\end{align*}
Hence $\partial_\alpha\mathcal C=0$ becomes
\begin{align}\label{eq:app_alpha_intermediate_full}
\gamma\nu\left(\theta-\frac{\alpha}{\sigma\tau}\right)+\nu\alpha=\E\left[Z_1\prox_{\gamma\phi}(\nu\alpha Z_1+\sigma Z_2+\gamma B)
\right].
\end{align}
Conditioning on $B$, using $\PP(B=1\mid Z_1)=\pi_\rho(\nu Z_1)$, the symmetry identity $\prox_{\gamma\phi}(x+\gamma)=-\prox_{\gamma\phi}(-x),$
and then Stein's lemma, the right-hand side of
\cref{eq:app_alpha_intermediate_full} simplifies to
\[
-2\nu\E\left[\pi_{\rho}'(\nu Z_{1})\prox_{\gamma\phi}(\nu\alpha Z_{1}+\sigma Z_{2})\right]+\nu\alpha\E\left[\frac{2(1-\pi_{\rho}(\nu Z_{1}))}{1+\gamma\phi''(\prox_{\gamma\phi}(\nu\alpha Z_{1}+\sigma Z_{2}))}\right].
\]
Using the $\sigma$-equation derived above, the last term equals $\nu\alpha\left(1-\frac{\gamma}{\sigma\tau}\right).$
Substituting this back into \cref{eq:app_alpha_intermediate_full} yields
\[
\gamma\theta=-2\E\left[\pi_\rho'(\nu Z_1)\prox_{\gamma\phi}(\nu\alpha Z_1+\sigma Z_2)\right].
\]
Collecting the six stationarity equations, we recover exactly the nonlinear system stated in the theorem.

Finally, by the Moreau-envelope representation used in the scalarization
step, the coordinatewise AO optimizer is
\[
\widehat{w}_{j}^{\text{AO}}=\prox_{\lambda\bar{\sigma}\bar{\tau}\widetilde{f}(\cdot,F_{t-1,j})}\left(\bar{\sigma}\bar{\tau}\left(\bar{\theta}w_{\star,j}+\frac{\bar{r}}{\sqrt{\delta}}Z_{j}\right)\right),
\]
with $Z_{j}\stackrel{\text{iid}}{\sim}\mathcal{N}(0,1).$ Let $\Xi:\mathbb R^3\to\mathbb R$ be locally Lipschitz with at most quadratic growth.
For $\zeta=(\sigma,\tau,r,\theta)$, define
\[
m_{\zeta}(w,F)=\mathbb{E}_{Z}\left[M_{\lambda\widetilde{f}(\cdot,F)}\left(\sigma\tau\left(\frac{r}{\sqrt{\delta}}Z+\theta w\right),\sigma\tau\right)\right].
\]
Let $K$ be a compact scalar-parameter set on which $\sigma\tau$ is bounded away from zero. The assumed continuity and uniform quadratic-growth bound of the Moreau envelope, together with dominated convergence in $Z$, imply that $(\zeta,w,F)\mapsto m_\zeta(w,F)$ is jointly continuous and that
\[
\sup_{\zeta\in K}|m_{\zeta}(w,F)|\le C_K(1+w^{2}+F^{2}).
\]
Conditionally on the teacher and the first $t-1$ tasks, the pairs $(w_{\star,j},F_{t-1,j})$ are fixed and the $g_j$'s are independent standard Gaussians. Fix $R>0$ and $\epsilon>0$. By continuity of the Moreau envelope and of $m_{\zeta}$, and compactness of $K$, there exists a finite net $\zeta_{1},\ldots,\zeta_{N}$ such that, for every $\zeta\in K$, some $\zeta_k$ approximates both $M_{\lambda\widetilde{f}(\cdot,F)}\left(\sigma\tau\left(\frac{r}{\sqrt{\delta}}z+\theta w\right),\sigma\tau\right)$ uniformly on $\{(w,F,g):w^{2}+F^{2}+g^{2}\leq R^{2}\}$ and $m_{\zeta}(w,F)$ uniformly on $\{(w,F):w^{2}+F^{2}\leq R^{2}\}$ to within $\epsilon$. Since the net is finite, the pointwise conditional law of large numbers gives convergence simultaneously at all its points. Hence, on the bounded region, the triangle inequality bounds the supremum over $\zeta\in K$ by the maximum over the finite net plus $2\epsilon$.
On the complementary region, the uniform quadratic-growth bound gives,
\begin{align*}
&\sup_{\zeta\in K}\frac{1}{d}\sum_{j=1}^{d}\left|M_{\lambda\widetilde{f}(\cdot,F_{t-1,j})}\left(\sigma\tau\left(\frac{r}{\sqrt{\delta}}g_{j}+\theta w_{\star,j}\right),\sigma\tau\right)\right|\mathbf{1}_{\{w_{\star,j}^{2}+F_{t-1,j}^{2}+g_{j}^{2}>R^{2}\}}\\&\qquad\leq\sup_{\zeta\in K}\frac{1}{d}\sum_{j=1}^{d}C\left(1+\sigma\tau\left(\frac{r}{\sqrt{\delta}}g_{j}+\theta w_{\star,j}\right)^{2}+F_{t-1,j}^{2}\right)\leq\frac{C_{K}}{d}\sum_{j=1}^{d}\left(1+w_{\star,j}^{2}+F_{t-1,j}^{2}+g_{j}^{2}\right)\mathbf{1}_{\{w_{\star,j}^{2}+F_{t-1,j}^{2}+g_{j}^{2}>R^{2}\}},
\end{align*}
and the analogous bound holds for the conditional expectations defining $m_{\zeta}$. The assumed convergence in $2$-Wasserstein distance and Gaussian tail integrability imply that these two bounds vanish in probability after first taking $d\to\infty$ and then $R\to\infty$. Therefore, first letting $d\to\infty$, then $\epsilon\to0$, and finally $R\to\infty$, we obtain,
\[
\sup_{\zeta\in K}\left|\frac{1}{d}\sum_{j=1}^{d}\left[M_{\lambda\widetilde{f}(\cdot,F_{t-1,j})}\left(\sigma\tau\left(\frac{r}{\sqrt{\delta}}g_{j}+\theta w_{\star,j}\right),\sigma\tau\right)-m_{\zeta}(w_{\star,j},F_{t-1,j})\right]\right|\xrightarrow{\mathbb{P}}0.
\]
We now use the same arguments to prove a similar result. Fix $R>0$ and $B_{R}=\{(w,F):w^2+F^2\leq R^2\}$, define $m_{\zeta,R}(w,F)=m_\zeta(w,F)\chi_{R}$ where $\chi_{R}=1$ on $B_{R},$ $\chi_{R}=0$ outside $B_{2R},$ $0\leq \chi_{R}\leq 1$ and is continuous. Due to joint continuity of $m_{\zeta}$ we may take (as before) finite net in $K$: $\zeta_1,\ldots\zeta_N$ such that for every $\zeta\in K,$ $\epsilon>0$ there exists some $\zeta_k$ satisfying $\sup_{(w,F)\in B_{2R}}|m_{\zeta,R}(w,F)-m_{\zeta_k,R}(w,F)|\leq \epsilon$. Thus,
\begin{align*}
\left|\int m_{\zeta,R}\dd\mu_{d}-\int m_{\zeta,R}\dd\mu_{t-1}\right|&\leq\left|\int m_{\zeta_{K},R}\dd\mu_{d}-\int m_{\zeta_{K},R}\dd\mu_{t-1}\right|+2\epsilon,
\end{align*}
where $\mu_d=\frac{1}{d}\sum_{i=1}^{d}\delta_{(w^\star_j, F_{t-1,j})}$ and $\mu_d\xrightarrow[]{\PP}\mu_{t-1}$ by the assumption of convergence in 2-Wasserstein distance.
Taking the supremum,
\begin{align}
\label{eq:18}
\sup_{\zeta\in K}\left|\int m_{\zeta,R}\dd\mu_{d}-\int m_{\zeta,R}\dd\mu_{t-1}\right|&\leq\max_{1\leq K\leq N}\left|\int m_{\zeta_{K},R}\dd\mu_{d}-\int m_{\zeta_{K},R}\dd\mu_{t-1}\right|+2\epsilon
\end{align}
There are now finitely many function $m_{\zeta_K, R}$ that are bounded and continuous (if $w^2+F^2\leq 4R^2$) since $\mu_d\xrightarrow[]{\PP}\mu_{t-1}$. For all fixed $k,$ 
\begin{align*}
\int m_{\zeta_{k},R}\dd\mu_{d}\xrightarrow[]{\PP}\int m_{\zeta_{k},R}\dd\mu_{t-1}\implies \max_{1\leq k\leq N}\left|\int m_{\zeta_{k},R}\dd\mu_{d}-\int m_{\zeta_{k},R}\dd\mu_{t-1}\right|\xrightarrow[\epsilon\to 0\text{ in \cref{eq:18}}]{\PP}0\\\implies \sup_{\zeta\in K}\left|\int m_{\zeta,R}\dd\mu_{d}-\int m_{\zeta,R}\dd\mu_{t-1}\right| \xrightarrow[]{\PP}0.
\end{align*}
Now we focus on the case where $w^2+F^2>R^2$, in this case the convergence is immediate as we show here. Due to the quadratic growth of the Moreau envelope,
\begin{align*}
\sup_{\zeta\in K}\left|\int\left(m_{\zeta}-m_{\zeta,R}\right)\dd\mu_{d}\right|&\underset{0\leq1-\chi_{R}\leq\mathbf{1}_{\left\{ \left\Vert (w,F)>R\right\Vert \right\} }}{\leq}\sup_{\zeta\in K}\int\left|m_{\zeta}\right|\mathbf{1}_{\left\{ \left\Vert (w,F)>R\right\Vert \right\} }\dd\mu_{d}\\&\leq C_{K}'\int\left(1+w^{2}+F^{2}\right)\mathbf{1}_{\{\left\Vert \left(w,F\right)\right\Vert >R\}}\dd\mu_{d},
\end{align*}

the same bound holds for $\mu_{t-1}$. The assumed convergence in 2-Wasserstein distance implies that for every $\epsilon>0$,
\[
\lim_{R\to\infty}\limsup_{d\to\infty}\mathbb{P}\left(\int\left(1+w^{2}+F^{2}\right)\mathbf{1}_{\{\|(w,F)\|>R\}}\,\dd\mu_{d}>\epsilon\right)=0.
\]
Consequently,
\[
\lim_{R\to\infty}\limsup_{d\to\infty}\mathbb{P}\left(\sup_{\zeta\in K}\left|\int\left(m_{\zeta}-m_{\zeta,R}\right)\,\dd\mu_{d}\right|>\epsilon\right)=0.
\]
Moreover, $\mu_{t-1}$ has finite second moment, so by dominated convergence,
\[
\int\left(1+w^{2}+F^{2}\right)\mathbf{1}_{\{\|(w,F)\|>R\}}\,\dd\mu_{t-1}\xrightarrow[R\to\infty]{\PP}0,
\]
The same quadratic-growth bound then gives
$\sup_{\zeta\in K}\left|\int\left(m_{\zeta}-m_{\zeta,R}\right)\,\dd\mu_{t-1}\right|\longrightarrow0.$
Combining these bounds with the convergence of the integrals, first letting $d\to\infty$ for fixed $R$, and then letting $R\to\infty$, yields $\sup_{\zeta\in K}\left|\int m_{\zeta}\,\dd\mu_{d}-\int m_{\zeta}\,\dd\mu_{t-1}\right|\xrightarrow{\mathbb{P}}0.$
By separability of $f$ and its Moreau envelope, we conclude that, uniformly in $\zeta$,
\[
\frac{1}{d}M_{\lambda f(\cdot;\mathbf F_{t-1})}\left(\sigma\tau\left(\frac{r}{\sqrt{\delta}}\mathbf{g}+\theta\teacher\right),\sigma\tau\right)\xrightarrow{\mathbb{P}}\mathbb{E}\left[M_{\lambda\widetilde{f}(\cdot,F_{t-1})}\left(\sigma\tau\left(\frac{r}{\sqrt{\delta}}Z+\theta w^{\star}\right),\sigma\tau\right)\right],
\]
where $Z\sim\mathcal N(0,1)$ is independent of $(w^\star,F_{t-1})$. For $\epsilon>0$, let,
\[
S_{\epsilon}:=\left\{ \mathbf{u}\in\mathbb{R}^{d}:\left|\frac{1}{d}\sum_{j=1}^{d}\Xi(u_{j},w_{\star,j},F_{t-1,j})-\mathbb{E}\left[\Xi\left(\prox_{\lambda\bar{\sigma}_{t}\bar{\tau}_{t}\widetilde{f}(\cdot,F_{t-1}))}\left(\bar{\sigma}_{t}\bar{\tau}_{t}\left(\bar{\theta}_{t}w^{\star}+\frac{\bar{r}_{t}}{\sqrt{\delta}}Z\right)\right),w^{\star},F_{t-1}\right)\right]\right|<\epsilon\right\} .
\]
By \cref{eq:37,eq:38}, $\mathbb{P}\left(\widehat{\mathbf{w}}^{\text{AO}}\in S_{\epsilon}\right)\longrightarrow1.$ Conditionally on the teacher and the previous tasks, $S_\epsilon$ is fixed. Recall the artificial bounded sets introduced at the proof and consider the variables belong to these bounded sets, the limiting AO objective is continuous and has a unique saddle point whose coordinatewise optimizer belongs to $S_\epsilon$ with probability tending to one. The CGMT theorem (see \citet[Corollary 3]{salehi2019impact}) therefore yields $\mathbb{P}\left(\widehat{\mathbf{w}}_{t}\in S_{\epsilon}\right)\longrightarrow1.$ It follows that,
\[
\frac{1}{d}\sum_{j=1}^{d}\Xi(\widehat{w}_{t,j},w_{\star,j},F_{t-1,j})\xrightarrow{\mathbb{P}}\mathbb{E}\left[\Xi\left(\prox_{\lambda\bar{\sigma}_{t}\bar{\tau}_{t}\widetilde{f}(\cdot,F_{t-1}))}\left(\bar{\sigma}_{t}\bar{\tau}_{t}\left(\bar{\theta}_{t}w^{\star}+\frac{\bar{r}_{t}}{\sqrt{\delta}}Z\right)\right),w^{\star},F_{t-1}\right)\right].
\]
In particular, taking  $\Xi(x,w,F)=\Psi(x,w)$, we obtain, for every locally Lipschitz $\Psi$ with at most quadratic growth,
\[
\frac{1}{d}\sum_{j=1}^{d}\Psi(\widehat{w}_{t,j},w_{\star,j})\xrightarrow{\mathbb{P}}\mathbb{E}\left[\Psi\left(\prox_{\lambda\bar{\sigma}_{t}\bar{\tau}_{t}\widetilde{f}(\cdot,F_{t-1})}\left(\bar{\sigma}_{t}\bar{\tau}_{t}\left(\bar{\theta}_{t}w^{\star}+\frac{\bar{r}_{t}}{\sqrt{\delta}}Z\right)\right),w^{\star}\right)\right].
\]
By induction and the uniqueness of the system solution, we know that $w_{t}$ is well-defined for all $t\leq T$, define for some iteration $t$ the following derived system of equations, 
\begin{align*}
\left\{ \begin{aligned}\nu^{2}\alpha_{t} & =\E\left[w^{\star}\prox_{\lambda\sigma_{t}\tau_{t}\tilde f(\cdot, F_{t-1})}\left(\sigma_{t}\tau_{t}\left(\theta_{t}w^{\star}+\frac{r_{t}}{\sqrt{\delta}}Z\right)\right)\right],\\
\gamma_{t} & =\frac{1}{r_{t}\sqrt{\delta}}\E\left[Z\prox_{\lambda\sigma_{t}\tau_{t}\tilde f(\cdot,F_{t-1})}\left(\sigma_{t}\tau_{t}\left(\theta_{t}w^{\star}+\frac{r_{t}}{\sqrt{\delta}}Z\right)\right)\right],\\
\nu^{2}\alpha_{t}^{2}+\sigma_{t}^{2} & =\E\left[\prox_{\lambda\sigma_{t}\tau_{t}\tilde f(\cdot,F_{t-1})}\left(\sigma_{t}\tau_{t}\left(\theta_{t}w^{\star}+\frac{r_{t}}{\sqrt{\delta}}Z\right)\right)^{2}\right],\\
\gamma_{t}^{2} & =\frac{2}{r_{t}^{2}}\E\left[\left(1-\pi_{\rho}(\nu Z_{1})\right)\left(\nu\alpha_{t}Z_{1}+\sigma_{t}Z_{2}-\prox_{\gamma_{t}\phi}\left(\nu\alpha_{t}Z_{1}+\sigma_{t}Z_{2}\right)\right)^{2}\right],\\
\gamma_{t}\theta_{t} & =-2\E\left[\pi_{\rho}'(\nu Z_{1})\prox_{\gamma_{t}\phi}\left(\nu\alpha_{t}Z_{1}+\sigma_{t}Z_{2}\right)\right],\\
1-\frac{\gamma_{t}}{\sigma_{t}\tau_{t}} & =\E\left[\frac{2\left(1-\pi_{\rho}(\nu Z_{1})\right)}{1+\gamma_{t}\phi''\left(\prox_{\gamma_{t}\phi}\left(\nu\alpha_{t}Z_{1}+\sigma_{t}Z_{2}\right)\right)}\right].
\end{aligned}
\right.
\end{align*}
where $Z,Z_1,Z_2\stackrel{\text{iid}}{\sim}\mathcal N(0,1).$

Therefore, by iterating the equations up to iteration  $T$, for every locally Lipschitz function $\Psi:\R^2\to\R$, we get that,
\begin{align*}
\frac{1}{d}\sum_{j=1}^{d}\Psi(w_{T,j},w_{T,j}^{\star})\xrightarrow{\PP}\mathbb{E}\left[\Psi\left(\prox_{\lambda\bar{\sigma}_{T}\bar{\tau}_{T}\widetilde{f}\left(\cdot,F_{T-1}\right)}\left(\bar{\sigma}_{T}\bar{\tau}_{T}\left(\bar{\theta}_{T}w^{\star}+\frac{\bar{r}_{T}}{\sqrt{\delta}}Z\right)\right),w^{\star}\right)\right],
\end{align*}
\end{proof}
\begin{remark}[Convergence of $\frac{1}{d}\sum_{j=1}^d \delta_{(w_{\star,j},F_{t-1,j}))}\to \mu_{t-1}$]
\label{rem:1}
Throughout the paper, we will mostly analyze the case in which $\mathbf F_{t-1}=\mathbf w_{t-1}$. We prove by induction that, for every fixed $t$, $\frac{1}{d}\sum_{j=1}^d\delta_{(w_j^\star,w_{t,j})}\xrightarrow[W_2]{\mathbb P}\mu_t.$\footnote{We will use $\xrightarrow[W_2]{\mathbb P}$ to denote convergence in probability in the 2-Wasserstein distance.}
At $t=0$, $\mathbf w_0=\mathbf0$, so
$\frac{1}{d}\sum_{j=1}^d\delta_{(w_j^\star,0)}
\xrightarrow[W_2]{\mathbb P}
\mu_0$
by the law of large numbers and the finite second moment of $w^\star$.
Suppose now that $\frac{1}{d}\sum_{j=1}^d\delta_{(w_j^\star,w_{t-1,j})}
\xrightarrow[W_2]{\mathbb P}\mu_{t-1}.$
Let $(w^\star,w_{t-1})\sim\mu_{t-1}$, and let $Z\sim N(0,1)$ be independent of this pair, and define $w_{t}=\prox_{\lambda\bar{\sigma}_{t}\bar{\tau}_{t}\widetilde{f}(\cdot,w_{t-1})}\left(\bar{\sigma}_{t}\bar{\tau}_{t}\left(\bar{\theta}_{t}w^{\star}+\frac{\bar{r}_{t}}{\sqrt{\delta}}Z\right)\right),$ 
where the scalar parameters are the unique limiting task-$t$ saddle-point parameters determined by $\mu_{t-1}$.

Conditionally on the teacher and the first $t-1$ tasks, the pairs $(w_j^\star,w_{t-1,j})$ are fixed, while the Gaussian variables in the task-$t$ AO representation are fresh and independent. Hence, for every bounded Lipschitz function $\psi$, the conditional law of large numbers, the convergence of the empirical distribution, and the convergence of the scalar AO parameters give $\frac{1}{d}\sum_{j=1}^{d}\psi(w_{\star,j},\widehat{w}_{t,j}^{\mathrm{AO}})\xrightarrow{\mathbb{P}}\E[\psi(w^{\star},w_{t})]$ (the same arguments that led to \cref{eq:39}).
Thus the AO empirical distribution converges weakly to $\mu_t$. Applying the same conditional truncation and tail argument as in the proof (see \cref{eq:20}) to $\psi(w,u)=w^2+u^2$ gives convergence of the second moment. Therefore, $\frac{1}{d}\sum_{j=1}^{d}\delta_{(w_{\star,j},\widehat{w}_{t,j}^{\mathrm{AO}})}\xrightarrow[W_{2}]{\mathbb{P}}\mu_{t}.$
The CGMT optimizer-transfer argument transfers both the bounded-Lipschitz convergence and the quadratic-moment convergence from the AO optimizer to the actual estimator, yielding $\frac{1}{d}\sum_{j=1}^{d}\delta_{(w_{\star,j},w_{t,j})}\xrightarrow[W_{2}]{\mathbb{P}}\mu_{t}.$
Since $\mathbf F_t=\mathbf w_t$, this is exactly the empirical-state condition required to analyze task $t+1$, completing the induction.
\end{remark}
\newpage

\section{Optimal L2 Regularization: Proofs for Section~\ref{sec:3.1}}
\label{sec:prf3.1}
\subsection{A closed form expression for the limiting performance of the final iterate under L2 regularization}
We are specifically interested in L2 regularization, where $\tilde{f}(w,w_{t-1})=(w-w_{t-1})^2.$ We verify the conditions of the full \cref{thm:1}. Its Moreau envelope is given by $M_{\lambda\tilde{f}(\cdot,v)}(x,c)=\frac{\lambda(x-v)^{2}}{1+2\lambda c}$ it is continuous and admits quadratic bound hold, furthermore, the joint convergence of the empirical distribution follows by \cref{rem:1}.

For this specific choice, we show that the the first three equations of the six-equation saddle-point system can be solved explicitly.
For every $\eta>0$,
\[
\prox_{\eta\tilde{f}}(x)=\argmin_{u\in\mathbb R}\left\{\frac{1}{2}(u-x)^2+\eta(u-w_{t-1})^2\right\}.
\]
Solving the optimization problem, stationary condition gives
\[
u-x+2\eta(u-w_{t-1})=0,
\]
hence $(1+2\eta)u=x+2\eta w_{t-1},$ and therefore
\[
\prox_{\eta \tilde{f}}(x)=\frac{x+2\eta w_{t-1}}{1+2\eta}.
\]
Let $(\alpha_t,\sigma_t,r_t,\tau_t,\gamma_t,\theta_t)$ be the unique solution for every $t\leq T$. In the scalar saddle-point system, 
\[
\prox_{\lambda\sigma_{t}\tau_{t}\tilde{f}}\left(\sigma_{t}\tau_{t}\left(\theta_{t}w^{\star}+\frac{r_{t}}{\sqrt{\delta}}Z\right)\right)=\frac{\sigma_{t}\tau_{t}\left(\theta_{t}w^{\star}+\frac{r_{t}}{\sqrt{\delta}}Z\right)+2\lambda\sigma_{t}\tau_{t}w_{t-1}}{1+2\lambda\sigma_{t}\tau_{t}}=\frac{\sigma_{t}\tau_{t}}{1+2\lambda\sigma_{t}\tau_{t}}\left(\theta_{t}w^{\star}+\frac{r_{t}}{\sqrt{\delta}}Z+2\lambda w_{t-1}\right).
\]
Then
\[
\left\{ \begin{aligned} & \nu^{2}\alpha_{t}=\E\left[w^{\star}\frac{\sigma_{t}\tau_{t}\left(\theta_{t}w^{\star}+\frac{r_{t}}{\sqrt{\delta}}Z\right)+2\lambda\sigma_{t}\tau_{t}w_{t-1}}{1+2\lambda\sigma_{t}\tau_{t}}\right],&\\&\gamma_{t}=\frac{1}{r_{t}\sqrt{\delta}}\E\left[Z\frac{\sigma_{t}\tau_{t}\left(\theta_{t}w^{\star}+\frac{r_{t}}{\sqrt{\delta}}Z\right)+2\lambda\sigma_{t}\tau_{t}w_{t-1}}{1+2\lambda\sigma_{t}\tau_{t}}\right],\\&\nu^{2}\alpha_{t}^{2}+\sigma_{t}^{2}=\E\left[\left(\frac{\sigma_{t}\tau_{t}\left(\theta_{t}w^{\star}+\frac{r_{t}}{\sqrt{\delta}}Z\right)+2\lambda\sigma_{t}\tau_{t}w_{t-1}}{1+2\lambda\sigma_{t}\tau_{t}}\right)^{2}\right].
\end{aligned}
\right.
\]
Using $\E[Z]=0$, $\E[Z^2]=1$, and the independence of $Z$ from $w^{\star},w_{t-1}$, this simplifies to
\[
\left\{
\begin{aligned}
&\nu^2\alpha_t=\frac{\sigma_t\tau_t}{1+2\lambda \sigma_t\tau_t}\left(\nu^2\theta_t+2\lambda \E[w^{\star} w_{t-1}]\right),\\&
\gamma_t=\frac{\sigma_t\tau_t}{\delta(1+2\lambda \sigma_t\tau_t)},
\\&\nu^2\alpha_t^2+\sigma_t^2=\left(\frac{\sigma_t\tau_t}{1+2\lambda \sigma_t\tau_t}\right)^2\left(\nu^2\theta_t^2+4\lambda\theta_t\E[w^{\star} w_{t-1}]+4\lambda^2\E[w_{t-1}^2]+\frac{r_t^2}{\delta}\right).
\end{aligned}
\right.
\]
Since $\frac{\sigma_t\tau_t}{1+2\lambda\sigma_t\tau_t}=\delta\gamma_t,$ the first identity becomes
\[
\alpha_t=\delta\gamma_t\theta_t+2\lambda\delta\gamma_t\frac{\E[w^{\star} w_{t-1}]}{\nu^2}.
\]
Using the proximal identity and the claim in \cref{thm:1},
\[
w_{t}\xrightarrow{d}\prox_{\lambda\sigma_{t}\tau_{t}\tilde{f}}\left(\sigma_{t}\tau_{t}\left(\theta_{t}w^{\star}+\frac{r_{t}}{\sqrt{\delta}}Z_{t}\right)\right)=\frac{\sigma_{t}\tau_{t}}{1+2\lambda\sigma_{t}\tau_{t}}\left(\theta_{t}w^{\star}+\frac{r_{t}}{\sqrt{\delta}}Z_{t}+2\lambda w_{t-1}\right)
\]
For notational convenience, we use $=$ to denote the deterministic values to which the random quantities converge as $n,d\to\infty$ and $n/d\to\delta$. Multiplying with $w^\star$ of previous equation and taking expectation,
\[
\mathbb{E}[w_{t}w^{\star}]=\delta\gamma_{t}\left(\theta_{t}\mathbb{E}[{w^{\star}}^{2}]+2\lambda\mathbb{E}\left[w_{t-1}w^{\star}\right]\right)=\delta\gamma_{t}\left(\theta_{t}\nu^{2}+2\lambda\mathbb{E}\left[w_{t-1}w^{\star}\right]\right)=\nu^{2}\alpha_{t},
\]
hence $\alpha_t=2\lambda\delta\gamma_t\alpha_{t-1}+\delta\gamma_t\theta_t.$
Likewise, subtracting $\nu^2\alpha_t^2=\nu^2\left(2\lambda\delta\gamma_t\alpha_{t-1}+\delta\gamma_t\theta_t\right)^2,$
the third equation yields
\[
\sigma_t^2=(2\lambda\delta\gamma_t)^2\sigma_{t-1}^2+\delta\gamma_t^2r_t^2.
\]
Then
\[
w_{t}\xrightarrow[]{d}2\lambda\delta\gamma_tw_{t-1}+\delta\gamma_t\theta_tw^{\star}+\sqrt{\delta}\gamma_t r_tZ_t,
\]
with $Z_t\stackrel{\text{iid}}{\sim}\mathcal N(0,1),$ and therefore
\[
\alpha_t=2\lambda\delta\gamma_t\alpha_{t-1}+\delta\gamma_t\theta_t,
\qquad\sigma_t^2=(2\lambda\delta\gamma_t)^2\sigma_{t-1}^2+{\delta}\gamma_t^2 r_t^2.
\]
Iterating back to $w_0=0$,
\[
w_{T}\xrightarrow{d}\sum_{s=1}^{T}\left(\prod_{j=s+1}^{T}2\lambda\delta\gamma_{j}\right)\left(\delta\gamma_{s}\theta_{s}w^{\star}+\sqrt{\delta}\gamma_{s}r_{s}Z_{s}\right).
\]
Hence
\begin{align}
\label{eq:21}   
\alpha_{T}=\sum_{s=1}^{T}\left(\prod_{j=s+1}^{T}2\lambda\delta\gamma_{j}\right)\delta\gamma_{s}\theta_{s},\qquad\sigma_{T}^{2}=\sum_{s=1}^{T}\left(\prod_{j=s+1}^{T}2\lambda\delta\gamma_{j}\right)^{2}\delta\gamma_{s}^{2}r_{s}^{2}.
\end{align}

From the sixth saddle equation,
\[
1-\frac{\gamma_t}{\sigma_t\tau_t}=\E\left[
\frac{2(1-\pi_\rho(\nu Z_1))}
{1+\gamma_t\phi''\left(\prox_{\gamma_t\phi}(\nu\alpha_tZ_1+\sigma_tZ_2)\right)}
\right].
\]
Since $\gamma_t=\frac{\sigma_t\tau_t}{\delta(1+2\lambda \sigma_t\tau_t)},$
we have $2\lambda\delta\gamma_t=\frac{2\lambda \sigma_t\tau_t}{1+2\lambda \sigma_t\tau_t},$
and hence
\[
\frac{\gamma_t}{\sigma_t\tau_t}=\frac{1}{\delta(1+2\lambda \sigma_t\tau_t)}=\frac{1-2\lambda\delta\gamma_t}{\delta}.
\]
Thus the sixth saddle equation gives the exact identity
\[
2\lambda\delta\gamma_t=1-\delta+\delta\E\left[
\frac{2(1-\pi_\rho(\nu Z_1))}
{1+\gamma_t\phi''\left(\prox_{\gamma_t\phi}(\nu\alpha_tZ_1+\sigma_tZ_2))\right)}
\right].
\]
For the bit-flip logistic model, $\pi_\rho(u)=\rho+(1-2\rho)\sigma(u).$ By Gaussian symmetry and $\sigma(-x)=1-\sigma(x)$, $\E[\sigma(\nu Z_1)]=\frac{1}{2},$ hence $\E[1-\pi_\rho(\nu Z_1)]=\frac{1}{2}.$ Since $\phi''\leq \frac{1}{4}$, $\gamma_t>0$ and $2\mathbb{E}\left[1-\pi_{\rho}\left(\nu Z_{1}\right)\right]=1$, \[\left|1-\mathbb{E}\left[\frac{2(1-\pi_{\rho}(\nu Z_{1}))}{1+\gamma_{t}\phi''(\prox_{\gamma_{t}\phi}(\nu\alpha_{t}Z_{1}+\sigma_{t}Z_{2}))}\right]\right|=\left|\gamma_{t}\mathbb{E}\left[\frac{2(1-\pi_{\rho}(\nu Z_{1}))\phi''(\prox_{\gamma_{t}\phi}(\nu\alpha_{t}Z_{1}+\sigma_{t}Z_{2}))}{1+\gamma_{t}\phi''(\prox_{\gamma_{t}\phi}(\nu\alpha_{t}Z_{1}+\sigma_{t}Z_{2}))}\right]\right|\leq\frac{\gamma_{t}}{4}.\] Therefore, $2\lambda\delta\gamma_t=1+O(\gamma_t).$ Using $\prox_{\gamma_t\phi}(x)=x-\gamma_t\phi'(x)+O(\gamma_t^2),$ and $\phi''(\prox_{\gamma_t\phi}(x))=\phi''(x)+O(\gamma_t),$ we get
\[
\frac{1}{1+\gamma_{t}\phi''(\prox_{\gamma_{t}\phi}(\nu\alpha_{t}Z_{1}+\sigma_{t}Z_{2}))}=1-\gamma_{t}\phi''(\nu\alpha_{t}Z_{1}+\sigma_{t}Z_{2})+O(\gamma_{t}^{2}).
\]
Hence
\[
\begin{aligned}
2\lambda\delta\gamma_t&=1-\delta+\delta\left[2\E[1-\pi_\rho(\nu Z_1)]-2\gamma_t\E\left[(1-\pi_\rho(\nu Z_1))\phi''(\nu\alpha_tZ_1+\sigma_t Z_2)\right]+O(\gamma_t^2)\right]\\&=1-2\delta\gamma_t\E\left[(1-\pi_\rho(\nu Z_1))\phi''(\nu\alpha_t Z_1+\sigma_tZ_2)\right]+O(\gamma_t^2).
\end{aligned}
\]
Since $\gamma_t=\frac{1}{2\delta\lambda}+O(\lambda^{-2}),$ then $2\lambda\delta\gamma_t=1-\frac{A_t}{\lambda}+O(\lambda^{-2}),$ where $A_t=\E\left[(1-\pi_\rho(\nu Z_1))\phi''(\nu\alpha_tZ_1+\sigma_tZ_2)\right].$ For the softplus function, $\phi''(x)=\frac{e^x}{(1+e^x)^2}.$ Writing $s(x)=\sigma(x)(1-\sigma(x)),$ we have $\phi''(x)=s(x)$, so $A_t=\E\left[(1-\pi_\rho(\nu Z_1))s(\nu\alpha_tZ_1+\sigma_tZ_2)\right].$ By \cref{lem:sym}, we obtain that $A_t=\frac{1}{2} h\left(\sqrt{\nu^2\alpha_t^2+\sigma_t^2}\right),$ with $h(q)=\E[s(qZ)].$ Next, using the fifth saddle equation, we have
\[
\delta\gamma_t\theta_t=-2\delta\E\left[\pi_\rho'(\nu Z_1)\prox_{\gamma_t\phi}(\nu\alpha_tZ_1+\sigma_tZ_2)
\right].
\]
Expanding the proximal map,
\[
\delta\gamma_t\theta_t=-2\delta\E[\pi_\rho'(\nu Z_1)(\nu\alpha_tZ_1+\sigma_tZ_2)]+2\delta\gamma_t
\E[\pi_\rho'(\nu Z_1)\phi'(\nu\alpha_tZ_1+\sigma_tZ_2)]+O(\gamma_t^2).
\]
Now $\pi_\rho'(\nu Z_1)$ is even in $Z_1$, while
$\nu\alpha_tZ_1+\sigma_tZ_2$ is odd in $Z_1$. Therefore $\E[\pi_\rho'(\nu Z_1)(\nu\alpha_tZ_1+\sigma_tZ_2)]=0.$
Consequently,
\[
\delta\gamma_t\theta_t=2\delta\gamma_t\E[\pi_\rho'(\nu Z_1)\phi'(\nu\alpha_tZ_1+\sigma_tZ_2)]+O(\gamma_t^2).
\]
Using again $\gamma_t=\frac{1}{2\delta\lambda}+O(\lambda^{-2}),$ we get $\delta\gamma_t\theta_t=\frac{B_t}{\lambda}+O(\lambda^{-2}),$ where
$B_t=\E\left[\pi_\rho'(\nu Z_1)\phi'(\nu\alpha_tZ_1+\sigma_tZ_2)\right].$
By \cref{lem:sym} we have that, $B_t\equiv B$ is constant, with
$B=\frac{1-2\rho}{2}h(\nu)>0.$
From the fourth saddle equation,
\[
\gamma_t^2=\frac{2}{r_t^2}\E\left[(1-\pi_\rho(\nu Z_1))\left(\nu\alpha_tZ_1+\sigma_tZ_2-\prox_{\gamma_t\phi}(\nu\alpha_tZ_1+\sigma_tZ_2)\right)^2\right].
\]
Using
\[
\nu\alpha_tZ_1+\sigma_tZ_2-\prox_{\gamma_t\phi}(\nu\alpha_tZ_1+\sigma_tZ_2)=\gamma_t\phi'(\nu\alpha_tZ_1+\sigma_tZ_2)+O(\gamma_t^2),
\]
we obtain
\[
\gamma_t^2=\frac{2\gamma_t^2}{r_t^2}\E\left[(1-\pi_\rho(\nu Z_1))\phi'(\nu\alpha_tZ_1+\sigma_tZ_2)^2\right]+O(\gamma_t^3).
\]
Thus
\[
r_t^2=2\E\left[(1-\pi_\rho(\nu Z_1))\phi'(\nu\alpha_tZ_1+\sigma_tZ_2)^2\right]+O(\lambda^{-1}).
\]
Therefore $\delta\gamma_t^2 r_t^2=\frac{V_t}{\lambda^2}+O(\lambda^{-3}),$ where
\[
V_t=\frac{1}{2\delta}\E\left[(1-\pi_\rho(\nu Z_1))\phi'(\nu\alpha_tZ_1+\sigma_tZ_2)^2\right].
\]
Overall, $2\lambda\delta\gamma_t=1-\frac{A_t}{\lambda}+O(\lambda^{-2}),$ $\delta\gamma_t\theta_t=\frac{B_t}{\lambda}+O(\lambda^{-2}),$ and $\delta\gamma_t^2r_t^2=\frac{V_t}{\lambda^2}+O(\lambda^{-3}),$ with $A_t=\frac{1}{2} h\left(\sqrt{\nu^2\alpha_t^2+\sigma_t^2}\right),$ $B_t=\E\left[\pi_\rho'(\nu Z_1)\phi'(\nu\alpha_tZ_1+\sigma_tZ_2)\right],$ and $V_t=\frac{1}{2\delta}\E\left[(1-\pi_\rho(\nu Z_1))\phi'(\nu\alpha_tZ_1+\sigma_tZ_2)^2\right].$
Hence,
\begin{equation}
\label{eq:17}
\begin{aligned}
&\alpha_{t}=\left(1-\frac{A_{t}}{\lambda}\right)\alpha_{t-1}+\frac{B_{t}}{\lambda}+O(\lambda^{-2}),\\&\sigma_{t}^{2}=\left(1-\frac{2A_{t}}{\lambda}\right)\sigma_{t-1}^{2}+\frac{V_{t}}{\lambda^{2}}+O(\lambda^{-3}),
\end{aligned}
\end{equation}
the last equation follows by \cref{eq:3} showing that $\sigma_t^2=O(\lambda^{-1})$. Since $\phi'(x)=\frac{e^x}{1+e^x},$ $\phi''(x)=\frac{e^x}{(1+e^x)^2}.$ Therefore $|\phi'(x)|<1,$ and $|\phi''(x)|\le \frac{1}{4}$ for all $x\in\mathbb R$, so $A_t, B_t, V_t$ are uniformly bounded by $\frac{1}{4}$, $\frac{1-2\rho}{4}$, and $\frac{1}{2\delta}$, respectively.
\subsection{Auxiliary Lemmas}
\label{sec:aux}
For this section we introduce the following notations,
\[
\sigma(x)=\frac{1}{1+e^{-x}},\hspace{2em}s(x)=\sigma(x)(1-\sigma(x))=\phi''(x),\hspace{2em}\pi_{\rho}(x)=\rho+(1-2\rho)\sigma(x),
\]
where $0\le \rho< \frac{1}{2}$, and $\sigma$ denotes the sigmoid function. Let the population logistic risk be
\[
\mathcal{L}_T(\lambda)=\mathbb E\left[\log(1+e^{\nu\alpha_tZ_1+\sigma_tZ_2})-\pi_\rho(\nu Z_1)(\nu\alpha_tZ_1+\sigma_tZ_2)\right].
\]
This definition follows because $\mathbf x^{\top}\teacher\xrightarrow[]{d}\nu Z_1,$ if we write $\mathbf w_t=\alpha_t\teacher+(\mathbf w_t-\alpha_t\teacher)$ with $\alpha_t=\frac{\mathbf w_t^\top\teacher}{\Vert\teacher\Vert^2}$ then $\mathbf x^{\top}\mathbf w_t\xrightarrow[]{d}\nu \alpha_tZ_1+\sigma_tZ_2.$  
Define the noiseless risk (where $\sigma_t=0$) as,
\[
\mathcal{L}_0(a)=\mathbb E_Z\left[\log(1+e^{\nu aZ})-\pi_\rho(\nu Z)\nu aZ\right],
\]
and let $\alpha^\star=\argmin_{a\ge0}\mathcal{L}_0(a)$. We introduce few other notations, gathered from the introduction,
\begin{align*}
&h(q)=\int_{-\infty}^{\infty}s(qz)\varphi(z)\dd z,\hspace{.5em}\varphi(z)=\frac{1}{\sqrt{2\pi}}e^{-\frac{z^2}{2}},\hspace{.5em} A_{t}=\frac{1}{2}h\left(\sqrt{\nu^{2}\alpha_{t}^{2}+\sigma_{t}^{2}}\right),\hspace{.5em} B_{t}=B=\frac{1-2\rho}{2}h(\nu),\hspace{.5em}r=\frac{B}{A},\\&\hspace{5em}A(a)=\frac{1}{2}h(\nu a),\hspace{.5em} A_{t}\xrightarrow{t\to\infty}A=\frac{1}{2}h(\nu r),\hspace{.5em} V_{t}=\frac{1}{2\delta}\mathbb{E}\left[(1-\pi_{\rho}(\nu Z_{1}))\phi'(\nu\alpha_{t}Z_{1}+\sigma_{t}Z_{2})^{2}\right].
\end{align*}
Few of the identities provided here are proved in the following Lemmas.
\begin{lemma}[Symmetry identities]
\label{lem:sym}
We have $A_t=\frac{1}{2}h\left(\sqrt{\nu^2\alpha_t^2+\sigma_t^2}\right),$ and $B_t=\frac{1-2\rho}{2}h(\nu).$ In particular $B_t\equiv B$ is constant.
\end{lemma}
\begin{proof}
By definition, $A_t=\mathbb E\left[(1-\pi_\rho(\nu Z_1))s(\nu\alpha_tZ_1+\sigma_t Z_2)\right].$
Using the symmetry
$(Z_1,Z_2)\stackrel{d}{=}(-Z_1,-Z_2),$
together with
$s(-u)=s(u),$ and $\pi_\rho(u)+\pi_\rho(-u)=1,$
we get
\[
\mathbb E[(1-\pi_\rho(\nu Z_1))s(\nu\alpha_tZ_1+\sigma_tZ_2)]=\mathbb E[\pi_\rho(\nu Z_1)s(\nu\alpha_tZ_1+\sigma_tZ_2)].
\]
Adding both identities gives
\[
2\mathbb E[(1-\pi_\rho(\nu Z_1))s(\nu\alpha_tZ_1+\sigma_tZ_2)]=\mathbb E[s(\nu\alpha_tZ_1+\sigma_tZ_2)].
\]
Therefore $A_t=\frac{1}{2}\mathbb E[s(\nu\alpha_tZ_1+\sigma_tZ_2)].$ Since $\nu\alpha_tZ_1+\sigma_tZ_2\sim N(0,\nu^2\alpha_t^2+\sigma_t^2),$ we get $A_t=\frac{1}{2}h\left(\sqrt{\nu^2\alpha_t^2+\sigma_t^2}\right).$ For $B_t$, using $\pi_\rho'(u)=(1-2\rho)s(u),$ we have $B_t=(1-2\rho)\mathbb E[s(\nu Z_1)\sigma(\nu\alpha_tZ_1+\sigma_tZ_2)]$ from symmetry, let $I_t=\mathbb E[s(\nu Z_1)\sigma(\nu\alpha_tZ_1+\sigma_tZ_2)]=\mathbb E[s(-\nu Z_1)\sigma(-(\nu\alpha_tZ_1+\sigma_tZ_2))].$ Since $\sigma(-(\nu\alpha_tZ_1+\sigma_tZ_2))=1-\sigma(\nu\alpha_tZ_1+\sigma_tZ_2)$, $s(\nu Z_1)=s(-\nu Z_1)$ we have $I_t=\mathbb E[s(\nu Z_1)(1-\sigma(\nu\alpha_tZ_1+\sigma_tZ_2))].$ Adding this to the definition of $I_t$, we have $2I_t=\mathbb E[s(\nu Z_1)]=h(\nu).$ Therefore $B_t=\frac{1-2\rho}{2}h(\nu).$
\end{proof}
\begin{lemma}[Variance bounds]
\label{lem:varbound}
For $1\lesssim \frac{AT}{\lambda}=o(T)$, we have $\sigma_t^2\asymp\frac{1}{\lambda}.$
\end{lemma}
\begin{proof}
Since $A_t\to A>0$ and $V_t\to V>0$ by \cref{lem:posinf}, for all sufficiently large $t$\footnote{Note that \cref{lem:posinf} relies on the upper bound on $\sigma_t$ established in \cref{lem:varbound}, which makes the argument circular. However, the upper bound on $\sigma_t$ itself follows from the conditions $V_t\leq v_+$ and $A_t \leq a_{+}$ for some $a_{+}$ and $v_{+}$ (see \cref{eq:3}). We have already proven these bounds independently, with $a_{+}=\frac{1}{4\delta}$ and $v_{+}=\frac{1}{2\delta}.$},
\[
0<a_{-}\le A_{t}\le a_{+},\qquad0<v_{-}\le V_{t}\le v_{+}.
\]
Using the developed recursion $\sigma_T$ in \cref{eq:21},
\begin{align}
\label{eq:3}
\sigma_T^2=\sum_{s=1}^T\left(\frac{V_s}{\lambda^2}+O\left(\lambda^{-3}\right)\right)\prod_{j=s+1}^T \left(1-\frac{2A_j}{\lambda}+O(\lambda^{-2})\right)\underset{v_{-}\leq V_t\leq v_{+},a_{-}\leq A_t\leq a_{+}}{\leq}C\sum_{s=1}^T\frac{1}{\lambda^2}\exp\left(-c\frac{T-s}{\lambda}\right)\le\frac{C}{\lambda},
\end{align}
for some $C,c>0$. To establish the lower bound, take only the last $\lfloor\eta\lambda\rfloor$ terms. For these terms we have $\prod_{j=s+1}^T \left(1-\frac{2A_j}{\lambda}\right)\geq \left(1-\frac{2a_{+}}{\lambda}\right)^{\eta\lambda}\geq c>0$. Hence
\[\sigma_T^2\underset{V_t\ge v_{-}}{\geq}c\sum_{s=T-\lfloor\eta\lambda\rfloor}^T\frac{1}{\lambda^2}\ge\frac{c}{\lambda},
\]
for some $c>0$, the existence of uniform lower and upper bounds follows from compactness of the trajectory $(\alpha_t,\sigma_t)$, derived in \cref{eq:3} and \cref{lem:posinf}.
\end{proof}
\begin{lemma}[Positivity of the limiting coefficients]
\label{lem:posinf}
Assume $\lambda\to\infty,$ $\frac{T}{\lambda}\to\tau\in[0,\infty].$ Let $\alpha(\tau)$ be the solution of $\dot\alpha=B-\alpha A(\alpha),$ with $\alpha(0)=0,$ where $A(a)=\frac{1}{2}h(\nu a),$ $B=\frac{1-2\rho}{2}h(\nu).$ If $\tau=\infty$, define $\alpha(\infty)=r,$ where $r$ is the unique fixed point $B=rA(r).$ Then $A_T\to A_\tau=\frac{1}{2}h(\nu\alpha(\tau))>0,$ and $V_T\to V_\tau=\frac{1}{2\delta}\E\left[(1-\pi_\rho(\nu Z))\phi'(\nu\alpha(\tau)Z)^2\right]>0.$ 
\end{lemma}

\begin{proof}
Recall that $A_t=\frac{1}{2}h\left(\sqrt{\nu^2\alpha_t^2+\sigma_t^2}\right),$ and the signal recursion (\cref{eq:17}),
\[
\alpha_t=\left(1-\frac{A_t}{\lambda}\right)\alpha_{t-1}+\frac{B}{\lambda}+O(\lambda^{-2}),
\]
with $A_t=\frac{1}{2} h\left(\sqrt{\nu^2\alpha_t^2+\sigma_t^2}\right),$ $B=\frac{1-2\rho}{2}h(\nu).$ Hence
\[
\alpha_t-\alpha_{t-1}=\frac{1}{\lambda}\left(B-A_t\alpha_{t-1}\right)+O(\lambda^{-2}).
\]
Since $\sigma_t^2=O(\lambda^{-1})$, we have $A_t=A(\alpha_{t-1})+o(1),$ with $A(a)=\frac{1}{2} h(\nu a)$, and therefore
\[
\alpha_t-\alpha_{t-1}=\frac{1}{\lambda}\left(B-\alpha_{t-1}A(\alpha_{t-1})\right)+o(\lambda^{-1}).
\]
Thus $\alpha_t$ is the explicit Euler discretization, with step size $h=\frac{1}{\lambda}$ of the ODE $\dot\alpha=B-\alpha A(\alpha),$ with $\alpha(0)=0.$ The corresponding ODE time is $Th=\frac{T}{\lambda}.$ Therefore, if $\frac{T}{\lambda}\to\tau\in[0,\infty]$. Using \cref{lem:varbound} we have $\sigma_t^2\to0.$ Therefore $\nu^2\alpha_t^2+\sigma_t^2\to\nu^2\alpha(\tau)^2.$ By the continuity of $h$,
\[
A_t=\frac{1}{2}h\left(\sqrt{\nu^2\alpha_t^2+\sigma_t^2}\right)\to\frac{1}{2}h(\nu\alpha(\tau))=A_\tau.
\]
Since $h(q)=\E[\sigma(qZ)(1-\sigma(qZ))]$ and $0<\sigma(qZ)(1-\sigma(qZ))$ almost surely for every $q$, we have $A_\tau>0.$
Next,
\[
V_t=\frac{1}{2\delta}\E\left[(1-\pi_\rho(\nu Z_1))\phi'(\nu\alpha_tZ_1+\sigma_tZ_2)^2
\right].
\]
Since $\alpha_t\to\alpha(\tau),$ with $\sigma_t\to0,$ and $\phi'$ is bounded and continuous, dominated convergence gives
\[
V_t\to\frac{1}{2\delta}\E\left[(1-\pi_\rho(\nu Z))\phi'(\nu\alpha(\tau)Z)^2\right]= V_\tau.
\]
Finally, $1-\pi_\rho(\nu Z)>0$ almost surely, and $\phi'(\nu\alpha(\tau)Z)^2>0$ almost surely. Hence $V_\tau>0.$
\end{proof}

\begin{lemma}[Monotonicity]
\label{lem:mon}
The map $q\mapsto qh(q)$ is strictly increasing on $(0,\infty)$. Also, $h'(q)<0.$
\end{lemma}

\begin{proof}
Changing variables $u=qz$,
\[
qh(q)=\int_{-\infty}^{\infty}s\left(u\right)\varphi\left(\frac{u}{q}\right)\dd u.
\] For fixed $u\neq0$, the map $q\mapsto \varphi\left(\frac{u}{q}\right)$ is strictly increasing. Since $s(u)>0$, $qh(q)$ is strictly increasing. For the derivative, $h'(q)=\mathbb E[Zs'(qZ)],$ because $s'(x)=-\tanh\left(\frac{x}{2}\right)s(x),$ we get
\[
Zs'(qZ)=-Z\tanh\left(\frac{qZ}{2}\right)s(qZ)\le0,
\]
strictly on a set of positive probability. Hence $h'(q)<0$.
\end{proof}

\begin{lemma}[Relation between $r$ and $\alpha^\star$]
\label{lem:rel}
The limiting fixed point $r=\frac{B}{A(r)}$ satisfies $r h(\nu r)=(1-2\rho)h(\nu).$ The noiseless logistic optimum satisfies $\alpha^\star h(\nu\alpha^\star)=(1-2\rho)h(\nu).$ Consequently, $r=\alpha^\star$ in this regime.
\end{lemma}

\begin{proof}
By \cref{lem:posinf} we let $r$ be the the solution of $r=\frac{B}{A(r)}$, or equivalently, $r$ is the solution of $rh(\nu r)=(1-2\rho)h(\nu),$ recall that
\[
\mathcal{L}_0'(a)=\nu\mathbb E\left[Z(\sigma(\nu aZ)-\pi_\rho(\nu Z))\right].
\]
Using Stein's identity,
\[
\mathbb E[Z\sigma(qZ)]=q\mathbb E[\sigma(qZ)(1-\sigma(qZ))]=qh(q).
\]
Thus
\[
\mathcal{L}_0'(a)=\nu^2\left[a h(\nu a)-(1-2\rho)h(\nu)\right].
\]
Therefore $\mathcal{L}_0'(\alpha^\star)=0$ gives
\begin{align*}
\alpha^\star h(\nu\alpha^\star)=(1-2\rho)h(\nu).
\end{align*}
Since $a\mapsto ah(\nu a)$ is strictly increasing (see \cref{lem:mon}), comparing the equations for $r$ and $\alpha^\star$ proves the claim.
\end{proof}

\begin{lemma}[Small-noise expansion of the logistic risk]
\label{lem:exp}
If $\alpha_t$ remains bounded and $\sigma_t\to0$, then
\[
\mathcal{L}_t(\lambda)=\mathcal{L}_0(\alpha_t)+\frac{\sigma_t^2}{2}h(\nu \alpha_t)+O(\sigma_t^4).
\]
\end{lemma}

\begin{proof}
For fixed $Z_1$, define
\[
f_{Z_1}(m)=\log(1+e^m)-\pi_\rho(\nu Z_1)m.
\]
Then
\[
f_{Z_1}''(m)=\sigma(m)(1-\sigma(m))=s(m).
\]
Expanding in the Gaussian perturbation $\sigma_tZ_2$,
\[
\mathbb E_{Z_2}
f_{Z_1}(\nu\alpha_tZ_1+\sigma_tZ_2)=f_{Z_1}(\nu\alpha_tZ_1)+\frac{\sigma_t^2}{2}f_{Z_1}''(\nu\alpha_tZ_1)+O(\sigma_t^4).
\]
The odd powers vanish from symmetry of $Z_2$, and the higher derivatives are bounded. Taking expectation over $Z_1$ yields the claim. 
\end{proof}

\begin{lemma}[Bootstrap lower bound]
\label{lem:1}
There exists $c_0>0$ such that, if $\frac{AT}{\lambda}\le c_0\log\left(T\right),$ then $r-\alpha_t\ge c e^{-C\frac{AT}{\lambda}}.$ 
\end{lemma}

\begin{proof}
Let $e_t=r-\alpha_t.$ Using $B=rA(r)$ in \cref{lem:posinf},
\[
e_t=\left(1-\frac{A_t}{\lambda}\right)e_{t-1}+\frac{r(A_t-A)}{\lambda}+O(\lambda^{-2}).
\]
Since $A_t=\frac{1}{2}h\left(\sqrt{\nu^2\alpha_t^2+\sigma_t^2}\right),$ $A=A(r)=\frac{1}{2}h(\nu r),$ and $h$ is smooth on bounded sets. Hence, on the bounded trajectory,
$\sqrt{\nu^2\alpha_t^2+\sigma_t^2}-\nu r\le-\nu e_t+C\sigma_t^2.$
Since $h$ is decreasing and Lipschitz on bounded sets, $A_t-A=\frac{1}{2}\{h\left(\sqrt{\nu^2\alpha_t^2+\sigma_t^2}\right)-h(\nu r)\}\ge-C\sigma_t^2.$
Therefore $A_t\ge A-C\sigma_t^2.$ Hence $\frac{r(A_t-A)}{\lambda}\gtrsim-\lambda^{-2},$
so
\[
e_T\ge\left(1-\frac{A_T}{\lambda}\right)e_{T-1}-C\lambda^{-2}.
\]
for some $C>0$. Iterating,
\[
e_T\ge e_0\prod_{j=1}^T\left(1-\frac{A_j}{\lambda}\right)-C\sum_{s=1}^T\lambda^{-2}\prod_{j=s+1}^T\left(1-\frac{A_j}{\lambda}\right).
\]
Since $A_j\le A_+$, we have $\prod_{j=1}^T\left(1-\frac{A_j}{\lambda}\right)\ge\exp\left(-C\frac{T}{\lambda}\right),$ and since $A_j\geq A_->0$ the accumulated error is
\[
\sum_{s=1}^T\lambda^{-2}\prod_{j=s+1}^T\left(1-\frac{A_j}{\lambda}\right)\le\frac{1}{\lambda^2}\sum_{k=0}^{T}e^{-\frac{ck}{\lambda}}\lesssim\frac{1}{\lambda}.
\]
for some $c>0$. Thus
\[
e_T\ge e_0 e^{-C\frac{AT}{\lambda}}-C\frac{1}{\lambda}.
\]
If $\frac{AT}{\lambda}\le c_0\log\left(T\right)$ and $c_0>0$ is sufficiently small, then $e^{-C\frac{AT}{\lambda}}>2C\frac{1}{\lambda}.$
Hence
\[
r-\alpha_T=e_T\ge c e^{-C\frac{AT}{\lambda}}.
\]
Moreover, the variance bound in \cref{eq:3} gives
\[
\sigma_T^2\le \frac{C}{\lambda}\lesssim e^{-C\frac{AT}{\lambda}}\lesssim e_T.
\]
Local strong convexity of $\mathcal{L}_0$ gives
\[
\mathcal{L}_0(\alpha_T)-\mathcal{L}_0(\alpha^\star)\ge c(r-\alpha_T)^2\ge c e^{-2C\frac{AT}{\lambda}}.
\]
For $c_0$ sufficiently small, this is much larger than $\frac{\log\left(T\right)}{T}$.
\end{proof}

\begin{lemma}[Optimal Scaling]
\label{lem:opts}
There exist constants $T_0,C,c>0$ such that for all $T\geq T_0$ the optimal regularizer satisfies $c\frac{T}{\log\left(T\right)}\leq \lambda^\star\leq C\frac{T}{\log\left(T\right)}.$
\end{lemma}

\begin{proof}
We prove the equivalent statement $x^\star=\Theta(\log\left(T\right)),$ with $x=\frac{AT}{\lambda}$. First choose a comparison point $x_0=C_0\log\left(T\right)$ with $C_0>0$ sufficiently large. By linearizing the ode around $r$ and the recursion in \cref{eq:17} gives $\frac{\dd}{\dd x}(\alpha-r)=-\beta(\alpha-r)+O((\alpha-r)^2)+O(\frac{1}{T})$, with $\beta=1+\frac{\nu r h'(\nu r)}{h(\nu r)}.$ Solving this yields $\alpha(x)-r=C_\alpha e^{-\beta x}+O(e^{-2\beta x})+O(\frac{x}{T}),$ where $C_\alpha$ is the solution to $C_\alpha=\lim_{\tau\to\infty}e^{\kappa\tau}\left(r-\alpha(\tau)\right)$ and $\kappa=\frac{1}{2}\left(h(\nu r)+\nu r h'(\nu r)\right),$ therefore for a sufficiently large $C$ and $c$,
\[
|\alpha_t(x_0)-r|\le C e^{-cx_0}+C\frac{x_0}{T}.
\]
Choosing $C_0$ large enough,
\[
|\alpha_t(x_0)-r|\le C\frac{\log\left(T\right)}{T}.
\]
Also, $\sigma_t^2(x_0)=O\left(\frac{\log\left(T\right)}{T}\right).$
Therefore in \cref{lem:exp},
\[
\mathcal{L}_T(x_0)-\mathcal{L}_0(r)=O\left(\frac{\log\left(T\right)}{T}\right),
\]
Now we rule out $x\gg\log\left(T\right)$, since $\mathcal{L}_0(\alpha_t)\ge \mathcal{L}_0(\alpha^\star),$
so the variance lower bound in \cref{lem:varbound} gives
\[
\mathcal{L}_T(x)-\mathcal{L}_0(\alpha^\star)\ge c\frac{x}{T}.
\]
Thus $x\gg\log\left(T\right)$ is worse than the comparison point,
for $x\gg\log\left(T\right)$, $x=o(T)$. If $x$ is not $o(T)$, then
$\lambda$ does not diverge, so the variance does not vanish and the risk stays a fixed positive distance above the noiseless floor $\mathcal{L}_0(r)$.
This is again worse than the comparison point. Hence $x^\star\lesssim \log\left(T\right).$
Finally, \cref{lem:1} rules out $x\le c_0\log\left(T\right)$ for a sufficiently small $c_0>0$. Hence $x^\star\gtrsim\log\left(T\right).$ Therefore $x^\star=\Theta(\log\left(T\right)).$ Since $\lambda^\star=\frac{AT}{x^\star},$ we get $\lambda^\star=\Theta\left(\frac{T}{\log\left(T\right)}\right).$
\end{proof}

\newpage

\subsection{Proof of \cref{thm:2}}
\label{sec:optprf}
After the simplifications and auxiliary Lemmas established above we are ready to prove \cref{thm:2}.
\begin{fuller}[\cref{thm:2}]
As $n\to\infty$ the optimal fixed step size $\lambda^\star$ that minimizes the resulting generalization loss of the regularized logistic regression after $T$ tasks with non-zero teacher\footnote{Since $\mathbf w_0=\mathbf 0_d$ then additionally having $\teacher=\mathbf 0_d$, directly implies $\lambda^\star\to\infty$ (independently of $T$).} satisfies $\lambda^\star\asymp \frac{T}{\log T}$. Specifically, for all sufficiently large $T>T_0=O\left(\frac{1}{\epsilon^2}\right)$,
the optimal parameters satisfy
\begin{align*}   
(1-\epsilon)\frac{\kappa T}{\log\Big(\frac{\kappa\nu^{2}\E[Z^{2}\phi''(\nu rZ)]C_{\alpha}^{2}T}{\frac{1}{2\delta}\E\left[(1-\pi_{\rho}(\nu Z))\frac{1}{(1+e^{-\nu rZ})^{2}}\right]}\Big)}&<\lambda^{\star}<(1+\epsilon)\frac{\kappa T}{\log\Big(\frac{\kappa\nu^{2}\E[Z^{2}\phi''(\nu rZ)]C_{\alpha}^{2}T}{\frac{1}{2\delta}\E\left[(1-\pi_{\rho}(\nu Z))\frac{1}{(1+e^{-\nu rZ})^{2}}\right]}\Big)},
\end{align*}
where $\pi_\rho(u)=\rho+(1-2\rho)\frac{e^{u}}{1+e^{u}}$, $\kappa=\E[\phi''(\nu rZ)]+\nu r\E[Z\phi'''(\nu rZ)],$ $r$ solves the equation $r=\frac{(1-2\rho)\E[\phi''(\nu Z)]}{\E[\phi''(\nu r Z)]}$, and $C_\alpha$ is determined by the limiting ODE
\[
2\dot{\alpha}=(1-2\rho)\E\left[\phi''(\nu Z)\right]-\alpha\E\left[\phi''(\nu\alpha Z)\right],\qquad\alpha(0)=0,
\]
through
\[
C_\alpha=\lim_{\tau\to\infty}e^{\frac{1}{2}\kappa\tau}(r-\alpha(\tau))\in(0,\infty),
\]
and $Z\sim \mathcal N(0,1).$ Moreover, under this scaling the excess risk decreases in $T$ as, 
\[ 
\mathcal{L}(\mathbf w_T)-\inf_{\vw} \mathcal{L}(\vw)=\sameorder{\nu,\delta,\rho}{\frac{\log(T)}{T}}.
\]
\end{fuller}

\medskip

Before presenting the proof, we direct the reader to the notations introduced in \cref{sec:aux}.

The positivity of $\kappa=h(\nu r)+\nu rh'(\nu r)>0$ follows from the strict monotonicity of $q\mapsto qh(q)$ by \cref{lem:mon}.
Throughout the proof we set $\beta=\frac{\kappa}{A}=1+\frac{\nu r h'(\nu r)}{h(\nu r)}$ and define $x=\frac{AT}{\lambda}$.

Note that, by linearizing the ode around $r$ gives $\frac{\dd}{\dd x}(\alpha-r)=-\beta(\alpha-r)+O((\alpha-r)^2)+O(\frac{1}{T})$ solving this yields $\alpha(x)-r=C_\alpha e^{-\beta x}+O(e^{-2\beta x})+O(\frac{x}{T}).$
Its stable fixed point $r$ satisfies $rh(\nu r)=(1-2\rho)h(\nu)$ by \cref{lem:rel}, and taking Taylor expansion in \cref{eq:17} leads,
\[
\sigma_T^2=\frac{V}{2A\lambda}+O\left(\frac{x}{T}e^{-\beta x}+\frac{x^2}{T^2}\right)=\frac{V}{2A^2}\frac{x}{T}+O\left(\frac{x}{T}e^{-\beta x}+\frac{x^2}{T^2}\right),
\]
where $V=\frac{1}{2\delta}\E\left[(1-\pi_\rho(\nu Z))\phi'(\nu rZ)^2\right].$
Since $r=\alpha^\star$ by \cref{lem:rel}, which implies $\mathcal{L}_0'(r)=0$, by performing a small-noise expansions  about $r=\alpha^\star$, since $\mathcal{L}_0'(r)=0$ we obtain
\[
\mathcal{L}_T(\alpha_T)-\mathcal{L}_0(r)=\frac{1}{2}\mathcal{L}_0''(r)(\alpha_T-r)^2+O(|\alpha_T-r|^3)
\]
then using the ode linearization,
\[
\mathcal{L}_T(\alpha_T)-\mathcal{L}_0(r)=\frac{1}{2}\mathcal{L}_0''(r)C_\alpha^2e^{-2\beta x}+O\left(e^{-3\beta x}+\frac{x}{T}e^{-\beta x}+\frac{x^2}{T^2}\right),
\]
and because by \cref{lem:exp},
\[
\mathcal{L}_T(\lambda)-\mathcal{L}_0(\alpha_T)=\frac{\sigma_T^2}{2}h(\nu \alpha_T)+O(\sigma_T^4),
\]
we finally get, since $r=\alpha^\star$ and $h(\nu\alpha_{T})=\underbrace{h(\nu r)}_{=2A}+\nu h'(\nu r)(\alpha_{T}-r)+O((\alpha_{T}-r)^{2})=2A+O(e^{-\beta x}+\frac{x}{T})$,
\[
\mathcal{L}_T(\lambda)-\mathcal{L}_0(\alpha^\star)=\frac{1}{2}\mathcal{L}_0''(r)C_\alpha^2e^{-2\beta x}+\frac{V}{2A}\frac{x}{T}+O\left(e^{-3\beta x}+\frac{x}{T}e^{-\beta x}+\frac{x^2}{T^2}\right),
\]
The leading two-term approximation is therefore
\[
R_{\rm lead}(x)=\frac{1}{2}\mathcal{L}_0''(r)C_\alpha^2e^{-2\beta x}+\frac{V}{2A}\frac{x}{T}.
\]
Its minimizer $x_0$ satisfies $-2\beta \frac{1}{2}\mathcal{L}_0''(r)C_\alpha^2e^{-2\beta x_0}+\frac{V}{2AT}=0,$ hence $e^{-2\beta x_0}=\frac{V}{2A\beta\mathcal{L}_0''(r)C_\alpha^2}\frac{1}{T}.$
Then
\[
x_0=\frac{1}{2\beta}\log\left(\frac{2TA\beta \mathcal{L}_0''(r)C_\alpha^2}{V}\right),
\]
and the centered approximation for the optimal regularization is
\[
\lambda_0=\frac{AT}{x_0}=\frac{2\beta AT}
{\log\left(\frac{2TA\beta\mathcal{L}_0''(r)C_\alpha^2}{V}\right)}=\frac{\left(h(\nu r)+\nu r h'(\nu r)\right)T}{\log\left(\frac{T\left(h(\nu r)+\nu r h'(\nu r)\right)\mathcal{L}_0''(r)C_\alpha^2}{V}\right)}.
\]
This is indeed the optimal regularization strength even in a regime where the expansion does not hold (i.e. remainder term explodes), this is established in \cref{lem:opts}.
We now verify the accuracy threshold of the centered approximation. Let
$x_T^\star$ be the minimizer. At $x=x_0$, $e^{-\beta x_0}=O\left(\sqrt{\frac{1}{T}}\right).$ The second derivative of the leading risk satisfies $R_{\rm lead}''(x_0)=2\beta^2\mathcal{L}_0''(r)C_\alpha^2e^{-2\beta x_0}.$ The derivative of the remainder at $x_0$ is bounded by
$O\left(e^{-3\beta x_0}+\frac{x_0}{T}e^{-\beta x_0}+\frac{x_0}{T^2}\right).$ Since $R_{\rm lead}'(x_0)=0.$ Taylor expansion around $x_0$ gives
\[
0=R_{\rm lead}''(x_0)(x_t^\star-x_0)+R_{\rm mem}'(x_0)+o(x_t^\star-x_0)+o(R_{\rm mem}'(x_0))
\]
Dividing by $R_{\rm lead}''(x_0)$, we obtain
$x_T^\star-x_0=O\left(x_0\sqrt{\frac{1}{T}}\right).$
Consequently, $\frac{x_T^\star}{x_0}=1+O\left(\sqrt{\frac{1}{T}}\right),$ and since $\lambda=\frac{AT}{x}$, $\frac{\lambda^\star}{\lambda_0}=1+O\left(\sqrt{\frac{1}{T}}\right).$
Therefore, for any fixed $0<\epsilon<1$,
\[
(1-\epsilon)\lambda_0\le\lambda^\star\le(1+\epsilon)\lambda_0
\]
whenever $\sqrt{\frac{1}{T}}\lesssim\epsilon.$ Equivalently, $T\gtrsim\frac{1}{\epsilon^2},$ the optimal regularizer as appears in the Theorem follow by noting that by definition $V=\frac{1}{2\delta}\E\left[(1-\pi_\rho(\nu Z))\phi'(\nu rZ)^2\right]$, $\kappa=\frac{1}{2}\left(h(\nu r)+\nu r h'(\nu r)\right)$, and $\mathcal{L}_0''(r)=\nu^2 \E[Z^{2}\phi''(\nu rZ)].$ Now, we develop the optimal rate achieved, recall that the leading excess risk is
\begin{align*}
R_{\rm lead}(x)=\frac{1}{2}\mathcal{L}_0''(r)C_\alpha^2e^{-2\beta x}+\frac{V}{2A}\frac{x}{T}.
\end{align*}
From this expression it follows immediately that for all $\lambda=\sameorder{\nu,\delta,\rho}{\frac{T}{\log(T)}}$ the optimal excess risk satisfies $\mathcal{L}(\mathbf w_T)-\inf_{\vw}\mathcal{L}(\vw)=\sameorder{\nu,\delta,\rho}{\frac{\log(T)}{T}}.$ 
\newpage
\section{Continual sparse classification}
\label{sec:prf4}
We verify the condition in full \cref{thm:1} for the regularizers considered in this section: L1 and sign weighted L1. For L1 the Moreau envelope is given by (see \cref{lem:2}) \[M_{\lambda\tilde{f}(\cdot,v)}(x,c)=\begin{cases}
\frac{(x-v)^{2}}{2c} & |x-v|\leq\lambda c\\
\lambda|x-v|-\frac{\lambda^{2}c}{2} & |x-v|>\lambda c
\end{cases}\] and for sign weighted L1 (see \cref{lem:3}),
\[M_{\lambda\widetilde{f}(\cdot,v)}(x,c)=\begin{cases}
-\frac{\lambda}{\epsilon+(-v)_{+}}x-\dfrac{c}{2}\left(\frac{\lambda}{\epsilon+(-v)_{+}}\right)^{2}, & x<-c\frac{\lambda}{\epsilon+(-v)_{+}},\\[5pt]
\dfrac{x^{2}}{2c}, & -c\frac{\lambda}{\epsilon+(-v)_{+}}\le x\le c\frac{\lambda}{\epsilon+v_{+}},\\
\frac{\lambda}{\epsilon+v_{+}}x-\dfrac{c}{2}\left(\frac{\lambda}{\epsilon+v_{+}}\right)^{2}, & x>c\frac{\lambda}{\epsilon+v_{+}}.
\end{cases}\] these are both continuous and admit quadratic growth, furthermore, the joint convergence of the empirical distribution follows by \cref{rem:1}.
\subsection{Auxiliary Lemmas}
\begin{lemma}[Proximal map of the scalar L1 penalty]
\label{lem:2}
Let $\lambda>0$ and define $f(w,z)=|w-z|$. Then
\[\prox_{\lambda f(\cdot,z)}(v)=\argmin_{w\in\mathbb R}\left\{|w-z|+\frac{1}{2\lambda}(w-v)^2\right\}=z+\mathcal S_\lambda(v-z),\] where \[\mathcal{S}_{\lambda}(u)=\sign(u)(|u|-\lambda)_{+}=\begin{cases}
u-\lambda, & u>\lambda,\\
0, & |u|\leq\lambda,\\
u+\lambda, & u<-\lambda.
\end{cases}\]
\end{lemma}

\begin{proof}
Consider $g(w)=|w-z|+\frac{1}{2\lambda}(w-v)^2.$ If $w-z>0$, then $|w-z|=w-z$, and $g'(w)=1+\frac{w-v}{\lambda}.$ Setting $g'(w)=0$ gives $w=v-\lambda,$ which is valid only if $v>z+\lambda$. If $w-z<0$, then $|w-z|=z-w$, and $g'(w)=-1+\frac{w-v}{\lambda}.$ Setting $g'(w)=0$ gives $w=v+\lambda,$ which is valid only if $v<z-\lambda$.

Finally, at $w=z$ we use the subgradient condition $0\in\partial |w-z|+\frac{1}{\lambda}(w-v).$ Since $\partial |w-z|\big|_{w=z}=[-1,1]$, this becomes $0\in[-1,1]+\frac{z-v}{\lambda},$ which is equivalent to $|v-z|\leq\lambda.$ Therefore $w=z$ whenever $|v-z|\leq\lambda$. Combining the three cases gives $\prox_{\lambda f(\cdot,z)}(v)=z+\mathcal S_\lambda(v-z).$
\end{proof}
\begin{lemma}[Proximal map of the sign weighted L1 regularizer]
\label{lem:3}
Fix $u\in\mathbb R$ and $\epsilon>0$, and define
\[
f(w;u)=\begin{cases}
\dfrac{|w|^{2}}{[wu]_{+}+\epsilon|w|}, & w\neq0,\\
0, & w=0.
\end{cases}
\]
Then $f(\cdot;u)$ is convex and positively one-homogeneous. Moreover, for
every $q>0$, $\lambda_0>0$, and $x\in\mathbb R$,
\[
\prox_{\lambda_{0}qf(\cdot,u)}(qx)=q\mathcal{T}_{\frac{\lambda_{0}}{[u]_{+}+\epsilon},\frac{\lambda_{0}}{[-u]_{+}+\epsilon}}(x),
\]
where
\[
\mathcal{T}_{a,b}(x)=\begin{cases}
x-a, & x>a,\\
0, & -b\leq x\leq a,\\
x+b, & x<-b.
\end{cases}
\]
\end{lemma}

\begin{proof}
For $w>0$, one has $[wu]_+=w[u]_+$, while for $w<0$,
$[wu]_+=(-w)[-u]_+$. Hence
\[
f(w,u)=\begin{cases}
\dfrac{w}{[u]_{+}+\epsilon}, & w>0,\\
\dfrac{-w}{[-u]_{+}+\epsilon}, & w<0,\\
0, & w=0.
\end{cases}
\]
Thus $f(\cdot,u)$ is piecewise linear, with left derivative $-\frac{1}{[-u]_++\epsilon}$ and right derivative $\frac{1}{[u]_++\epsilon}.$
Since the left derivative is no larger than the right derivative, $f(\cdot,u)$ is convex. Also, $f(cw,u)=c f(w,u)$ for every $c\geq0$, so $f$ is positively
one-homogeneous.
Let $y=\prox_{\lambda_0 q f(\cdot,u)}(qx).$ Writing $y=qv$ and using $f(qv,u)=qf(v,u)$, the proximal objective becomes
\[
\frac{1}{2}(qv-qx)^2+\lambda_0q f(qv,u)=q^2\left(\frac{1}{2}(v-x)^2+\lambda_0 f(v,u)\right).
\]
Therefore $y=q\prox_{\lambda_0 f(\cdot;u)}(x).$
On $v>0$, the minimization problem is
\[
\min_{v>0}\left\{\frac{1}{2}(v-x)^2+\frac{\lambda_0}{[u]_++\epsilon}v\right\},
\]
whose minimizer is $v=x-\frac{\lambda_0}{[u]_++\epsilon}$
when $x>\frac{\lambda_0}{[u]_++\epsilon}$. Similarly, on $v<0$, the minimizer is $v=x+\frac{\lambda_0}{[-u]_++\epsilon}$ when $x<-\frac{\lambda_0}{[-u]_++\epsilon}$. In the remaining interval the minimizer is $v=0$. 
\end{proof}
\begin{proposition}[Connection between coefficient error and excess risk]
\label{prop:coefficient_error_to_excess_risk}
Suppose that for some universal constant $c>0$, $\E\left[(w_{t}-rw^{\star})^2\right]\geq c$
then there is a universal constant $c'>0$, such that, $\mathcal{L}(\mathbf w_t)-\inf_{\vw} \mathcal{L}(\vw)\geq c'\frac{1}{(1+\nu)^6}$
for every $t$, 
\[
\mathcal{L}(\mathbf w_t)-\inf_{\vw} \mathcal{L}(\vw)\leq\frac{1}{8}\E\left[(w_{t}-rw^{\star})^{2}\right].\]
If we further assume that $E=\E[(w_{t}-rw^{\star})^2]<\infty$. Then,
\begin{equation}
\begin{aligned}
\frac{1}{8}\min\left\{ \E\left[e^{-(r\nu+\sqrt E)|Z|}\right],\E\left[Z^{2}e^{-(r\nu+\sqrt E)|Z|}\right]\right\} \E\left[(w_{t}-rw^{\star})^{2}\right]
\leq\mathcal{L}(\mathbf w_t)-\inf_{\vw} \mathcal{L}(\vw).
\end{aligned}
\label{eq:local_risk_equivalence}
\end{equation}
\end{proposition}
\begin{proof}
For the scalar parameters $(\alpha_t,\sigma_t)$, define the population
logistic risk
\[
\mathcal{L}(\mathbf w_t)=\E\left[\phi\left(\nu\alpha_tZ_1+\sigma_tZ_2\right)-\pi_\rho(\nu Z_1)\left(\nu\alpha_tZ_1+\sigma_tZ_2\right)\right],
\]
and define the optimal population risk
\[
\mathcal{L}_\star=\E\left[\phi(r\nu Z_1)-\pi_\rho(\nu Z_1)r\nu Z_1\right],
\]
where $Z_1,Z_2\iidsim\mathcal N(0,1)$.  
Indeed, $\mathcal{L}_\star$ is the optimal population risk. By conditional Jensen's inequality (on $Z_2$) and independence, $\mathcal{L}(\alpha,\sigma)\geq \mathcal{L}(\alpha,0),$ so the optimum satisfies $\sigma_\star=0$. Moreover,
\[
\frac{\partial}{\partial\alpha}\mathcal{L}(\alpha,0)
=\nu^2\left[\alpha h(\nu\alpha)-(1-2\rho)h(\nu)\right],
\]
and hence its unique zero is $\alpha=r$, since
$rh(\nu r)=(1-2\rho)h(\nu)$ by \cref{lem:rel}. Finally, strict convexity of the logistic loss implies that this stationary point is the unique global minimizer, and therefore $\inf_{\vw} \mathcal{L}(\vw)=\mathcal{L}(r,0).$

By the definitions of $\alpha_t$ and $\sigma_t$, $\E[w^{\star} w_{t}]=\nu^2\alpha_t,$  $\E[w_{t}^2]=\nu^2\alpha_t^2+\sigma_t^2,$ $\E[({w^{\star}})^2]=\nu^2.$
Therefore,
\begin{equation}
\begin{aligned}
\E\left[(w_{t}-rw^{\star})^{2}\right]=\E[w_{t}^{2}]-2r\E[w^{\star}w_{t}]+r^{2}\E[{w^{\star}}^{2}]&=\nu^{2}\alpha_{t}^{2}+\sigma_{t}^{2}-2r\nu^{2}\alpha_{t}+r^{2}\nu^{2}\\&=\nu^{2}\left(\alpha_{t}-r\right)^{2}+\sigma_{t}^{2}.
\end{aligned}
\label{eq:7}
\end{equation}
The defining equation for $r$ (\cref{lem:rel}) implies $\E\left[Z_{1}\left(\phi'(r\nu Z_{1})-\pi_{\rho}(\nu Z_{1})\right)\right]=0.$ Since $Z_2$ is independent of $Z_1$ and has zero mean,
\begin{align*}
&\E\left[\left(\phi'(r\nu Z_{1})-\pi_{\rho}(\nu Z_{1})\right)\left(\nu(\alpha_{t}-r)Z_{1}+\sigma_{t}Z_{2}\right)\right]\\&=\nu(\alpha_{t}-r)\E\left[Z_{1}\left(\phi'(r\nu Z_{1})-\pi_{\rho}(\nu Z_{1})\right)\right]+\sigma_{t}\E\left[\phi'(r\nu Z_{1})-\pi_{\rho}(\nu Z_{1})\right]\E[Z_{2}]=0.
\end{align*}
Consequently,
\begin{align}
\mathcal{L}(\mathbf w_t)-\inf_{\vw} \mathcal{L}(\vw)&=\E\Big[\phi\left(\nu\alpha_{t}Z_{1}+\sigma_{t}Z_{2}\right)-\phi(r\nu Z_{1})-\phi'(r\nu Z_{1})\left(\nu(\alpha_{t}-r)Z_{1}+\sigma_{t}Z_{2}\right)\Big].
\label{eq:risk_bregman_identity}
\end{align}
Taylor's theorem gives
\begin{align}
&\phi\left(\nu\alpha_{t}Z_{1}+\sigma_{t}Z_{2}\right)-\phi(r\nu Z_{1})-\phi'(r\nu Z_{1})\left(\nu(\alpha_{t}-r)Z_{1}+\sigma_{t}Z_{2}\right)\notag\\&=\left(\nu(\alpha_{t}-r)Z_{1}+\sigma_{t}Z_{2}\right)^{2}\int_{0}^{1}(1-u)\phi''\left(r\nu Z_{1}+u\left(\nu(\alpha_{t}-r)Z_{1}+\sigma_{t}Z_{2}\right)\right)\dd u.
\label{eq:taylor_bregman}
\end{align}
Since $0<\phi''(x)\leq\frac{1}{4},$
equations \cref{eq:risk_bregman_identity,eq:taylor_bregman} imply
\begin{align*}
\mathcal{L}(\mathbf w_T)-\inf_{\vw} \mathcal{L}(\vw)&\leq\frac{1}{4}\int_{0}^{1}(1-u)\dd u\E\left[\left(\nu(\alpha_{t}-r)Z_{1}+\sigma_{t}Z_{2}\right)^{2}\right]
=\frac{1}{8}\left(\nu^{2}\left(\alpha_{t}-r\right)^{2}+\sigma_{t}^{2}\right).
\end{align*}
We now prove the lower bound on bounded-error regions. For every $x$,
\[
\phi''(x)=\frac{e^x}{(1+e^x)^2}\geq\frac{1}{4}e^{-|x|}.
\]
Suppose that $\nu^2(\alpha_t-r)^2+\sigma_t^2\leq E.$ Then
\begin{align*}
\sqrt{\nu^2\alpha_t^2+\sigma_t^2}\leq r\nu+\sqrt{\nu^2(\alpha_t-r)^2+\sigma_t^2}\leq r\nu+\sqrt E.
\end{align*}
Therefore, for every $u\in[0,1]$, $r\nu Z_1+u\left(\nu(\alpha_t-r)Z_1+\sigma_tZ_2\right)$ is a centered Gaussian random variable whose standard deviation is at most $r\nu+\sqrt E$. By rotational invariance of $(Z_1,Z_2)$, for every centered Gaussian linear
form with standard deviation at most $r\nu+\sqrt E$, and every other Gaussian linear
form $aZ_1+bZ_2$,
\begin{align}
&\E\left[\phi''\left(cZ_1+d Z_2\right)(aZ_1+bZ_2)^2\right]\notag\geq\frac{1}{4}\min\left\{ \E\left[e^{-(r\nu+\sqrt E)|Z|}\right],\E\left[Z^{2}e^{-(r\nu+\sqrt E)|Z|}\right]\right\} 
(a^2+b^2).
\label{eq:gaussian_curvature_lower_bound}
\end{align}
To see this, rotate the Gaussian coordinates so that $cZ_1+d Z_2=q\widetilde Z_1,$ with $q^2=c^2+d^2.$ Writing $aZ_1+bZ_2=a_\parallel\widetilde Z_1+a_\perp\widetilde Z_2,$ with some $a_\parallel, a_\perp$ (that make an orthogonal rotation) the cross term vanishes, and
\begin{align*}
&\E\left[\phi''(q\widetilde Z_{1})(a_{\parallel}\widetilde Z_{1}+a_{\perp}\widetilde Z_{2})^{2}\right]=a_{\parallel}^{2}\E[\widetilde Z_{1}^{2}\phi''(q\widetilde Z_{1})]+a_{\perp}^{2}\E[\phi''(q\widetilde Z_{1})]\\&\qquad\geq\frac{1}{4}a_{\parallel}^{2}\E[Z^{2}e^{-(r\nu+\sqrt E)|Z|}]+\frac{1}{4}a_{\perp}^{2}\E[e^{-(r\nu+\sqrt E)|Z|}]\\&\qquad\geq\frac{1}{4}\min\left\{ \E\left[e^{-(r\nu+\sqrt E)|Z|}\right],\E\left[Z^{2}e^{-(r\nu+\sqrt E)|Z|}\right]\right\} (a_{\parallel}^{2}+a_{\perp}^{2}).
\end{align*}
Since,
\[
\E\left[\left(\nu(\alpha_{t}-r)Z_{1}+\sigma_{t}Z_{2}\right)^{2}\right]=\nu^{2}\left(\alpha_{t}-r\right)^{2}+\sigma_{t}^{2},
\]
we obtain
\begin{align*}
\mathcal{L}(\mathbf w_T)-\inf_{\vw} \mathcal{L}(\vw)&\geq\frac{1}{4}\min\left\{ \E\left[e^{-(r\nu+\sqrt E)|Z|}\right],\E\left[Z^{2}e^{-(r\nu+\sqrt E)|Z|}\right]\right\} \int_{0}^{1}(1-u)\dd u\left(\nu^{2}\left(\alpha_{t}-r\right)^{2}+\sigma_{t}^{2}\right)\\&=\frac{1}{8}\min\left\{\E\left[e^{-(r\nu+\sqrt E)|Z|}\right],\E\left[Z^{2}e^{-(r\nu+\sqrt E)|Z|}\right]\right\} \left(\nu^{2}\left(\alpha_{t}-r\right)^{2}+\sigma_{t}^{2}\right).
\end{align*}
This proves \cref{eq:local_risk_equivalence}.
It remains to show that a uniform positive coefficient-error floor implies a positive risk floor even when the iterates are not known to remain bounded,
\[
\min\{\pi_\rho(u),1-\pi_\rho(u)\}=\rho+\frac{1-2\rho}{1+e^{|u|}}
\]
For fixed $u$ the right-hand side is increasing in $\rho$, therefore its minimum over $0\leq\rho<\frac{1}{2}$ is attained at $\rho=0$, hence
\[
\min\{\pi_\rho(u),1-\pi_\rho(u)\}\geq \frac{1}{1+e^{|u|}}\geq \frac{1}{2}e^{-|u|}
\]
Thus for $a^2+b^2=1$,
\[
\E\left[\min\left\{\pi_\rho(\nu Z_1),1-\pi_\rho(\nu Z_1)\right\}|aZ_1+bZ_2|\right]\geq\frac{1}{2}\E\left[e^{-\nu|Z_1|}|aZ_1+bZ_2|\right]
\]
First, if $a\geq \frac{1}{\sqrt 2}$ then,
\begin{align*} 
\E\left[e^{-\nu|Z_{1}|}\left|aZ_{1}+bZ_{2}\right|\right]&=\E_{Z_{1}}\left[e^{-\nu|Z_{1}|}\E\left[\left|aZ_{1}+bZ_{2}\right|\mid Z_{1}\right]\right]\\&\ge\E_{Z_{1}}\left[e^{-\nu|Z_{1}|}\left|\E\left[aZ_{1}+bZ_{2}\mid Z_{1}\right]\right|\right]\geq\frac{1}{\sqrt{2}}\E\left[e^{-\nu|Z_{1}|}|Z_{1}|\right].
\end{align*}
which evaluates to,
\[
\E\left[a\left|Z\right|e^{-\nu|Z|}\right]\geq 2a\varphi\left(1\right)e^{-1}\int_{0}^{\frac{1}{1+\nu}}\left|Z\right|\dd z=\frac{\varphi\left(1\right)e^{-1}}{\sqrt{2}(1+\nu)^{2}}
\]
Otherwise $b\geq \frac{1}{\sqrt 2}$ and then since $\E|x+bZ_2|$ is minimized at $x=0.$
\[
\E\left[e^{-\nu|Z_{1}|}|aZ_{1}+bZ_{2}|\right]=\E_{Z_{1}}\E[e^{-\nu|Z_{1}|}|aZ_{1}+bZ_{2}|\mid Z_{1}]\geq|b|\E|e^{-\nu|Z_{1}|}\left|Z_{2}\right||\geq\frac{2\varphi\left(1\right)e^{-1}}{\sqrt{\pi}(1+\nu)}.
\]
Then,
\begin{align*}
\mathcal{L}_{t}
&=\E\left[\phi\left(\nu\alpha_{t}Z_{1}+\sigma_{t}Z_{2}\right)-\pi_{\rho}(\nu Z_{1})\left(\nu\alpha_{t}Z_{1}+\sigma_{t}Z_{2}\right)\right]\\&\geq\E\left[\min\{\pi_{\rho}(\nu Z_{1}),1-\pi_{\rho}(\nu Z_{1})\}\left|\nu\alpha_{t}Z_{1}+\sigma_{t}Z_{2}\right|\right]
\geq\frac{1}{2}\E\left[e^{-\nu|Z_{1}|}|aZ_{1}+bZ_{2}|\right]\\&\geq\frac{\varphi(1)e^{-1}}{2(1+\nu)^{2}}\sqrt{\nu^{2}\alpha_{t}^{2}+\sigma_{t}^{2}}.
\end{align*}
Moreover,
\begin{align*}
\sqrt{\nu^2\alpha_t^2+\sigma_t^2}&\geq\sqrt{\nu^2(\alpha_t-r)^2+\sigma_t^2}-r\nu.
\end{align*}
Since the zero predictor has population risk $\log 2$, $\mathcal{L}_\star\leq\log 2.$ Set $E_0=\left(r\nu+\frac{2\left(1+\nu\right)^{2}(\log(2)+1)}{\varphi\left(1\right)e^{-1}}\right)^{2}.$
If $\nu^2(\alpha_t-r)^2+\sigma_t^2\geq E_0,$
then
\begin{align*}
\mathcal{L}(\mathbf{w}_{t})\geq\frac{\varphi(1)e^{-1}}{2\left(1+\nu\right)^{2}}\left(r\nu+\frac{2\left(1+\nu\right)^{2}(\log(2)+1)}{\varphi\left(1\right)e^{-1}}-r\nu\right)=\log2+1\geq\mathcal{L}_{\star}+1.
\end{align*}
Hence
$\mathcal{L}(\mathbf w_T)-\inf_{\vw} \mathcal{L}(\vw)\geq1.$
Combining both cases,
\begin{align}
\mathcal{L}(\mathbf w_T)-\inf_{\vw} \mathcal{L}(\vw)\geq\min\left\{1,\frac{c}{8}\min\left\{ \E\left[e^{-(r\nu+\sqrt E_0)|Z|}\right],\E\left[Z^{2}e^{-(r\nu+\sqrt E_0)|Z|}\right]\right\}\right\}.
\label{eq:coefficient_floor_to_risk_floor}
\end{align}
Restricting $0\leq Z\leq \left(1+r\nu+\sqrt{E_{0}}\right)^{-1}$, we have $\varphi(Z)\geq\varphi(1)$ and $e^{-\left(r\nu+\sqrt{E_{0}}\right)Z}\geq e^{-1}$, so,
\begin{align*}
\E\left[e^{-\left(r\nu+\sqrt{E_{0}}\right)|Z|}\right]&=\E\left[e^{-\left(2r\nu+\frac{2\left(1+\nu\right)^{2}(\log(2)+1)}{\varphi\left(1\right)e^{-1}}\right)|Z|}\right]\\&\geq2\varphi\left(1\right)e^{-1}\int_{0}^{\frac{1}{1+2r\nu+\frac{2\left(1+\nu\right)^{2}(\log(2)+1)}{\varphi\left(1\right)e^{-1}}}}\dd z=\frac{2\varphi\left(1\right)e^{-1}}{1+2r\nu+\frac{2\left(1+\nu\right)^{2}(\log(2)+1)}{\varphi\left(1\right)e^{-1}}},\\\E\left[Z^{2}e^{-\left(r\nu+\sqrt{E_{0}}\right)|Z|}\right]&=\E\left[Z^{2}e^{-\left(2r\nu+\frac{2(1+\nu)^2(\log(2)+1)}{4\varphi\left(1\right)e^{-1}}\right)|Z|}\right]\\&\geq2\varphi\left(1\right)e^{-1}\int_{0}^{\frac{1}{1+2r\nu+\frac{2\left(1+\nu\right)^{2}(\log(2)+1)}{\varphi\left(1\right)e^{-1}}}}z^{2}\dd z=\frac{2\varphi\left(1\right)e^{-1}}{3\left(1+2r\nu+\frac{2\left(1+\nu\right)^{2}(\log(2)+1)}{\varphi\left(1\right)e^{-1}}\right)^{3}}.
\end{align*}
Since $r\leq 1$, we have $\frac{2\varphi\left(1\right)e^{-1}}{3\left(1+2r\nu+\frac{2\left(1+\nu\right)^{2}(\log(2)+1)}{\varphi\left(1\right)e^{-1}}\right)^{3}}\leq C_1\frac{1}{(1+\nu)^6}$ for some universal constant $C_1>0$.
\end{proof}

\newpage
\section{Proofs for Section \ref{sec:4.2}}
\label{prfs:4.2}
\subsection{Proof of \cref{thm:one_homogeneous_error_floor}}
\label{prf:one_homogeneous_error_floor}
\begin{recall}[\cref{thm:one_homogeneous_error_floor}]
Assume that for every $t$, the regularizer is an order $p>0$ homogeneous, \ie $f(c\mathbf w,\mathbf v)=c^{p} f(\mathbf w,\mathbf v)$ for every $c\geq 0,$ $\mathbf w\in \mathbb R^n$ and a fixed $\mathbf v\in \mathbb R^n$.
Then there exists a universal constant $C>0$ such that
for every $T\geq 1$ and $s\in(0,1)$,
\[
\mathcal{L}(\mathbf w_T)-\inf_{\vw} \mathcal{L}(\vw)\geq C\min\left\{1,\frac{\left(1-2\rho\right)^{2}}{\left(1+\nu\right)^{2}\max\left\{ 1,(p-1)\max\left\{ 2^{2-p},\left(\frac{3}{2}\right)^{p-2}\right\} \right\} ^{2}}\min\left\{ \nu^{2},\frac{s}{\delta}\right\} \right\}.
\]
\end{recall}

\begin{proof}
The first scalar equation gives
\[
w_{t}\xrightarrow[]{d}\prox_{\lambda\sigma_{t}\tau_{t}f}\left(\sigma_{t}\tau_{t}\left(\theta_{t}w^{\star}+\frac{r_{t}}{\sqrt{\delta}}Z\right)\right).
\]
We first show directly from the fourth and sixth scalar equations that the fresh Gaussian noise inside this proximal map has non-vanishing variance. The optimality condition for the proximal map appearing in the fourth,
fifth, and sixth equations is
\begin{align*}
&\nu\alpha_{t}Z_{1}+\sigma_{t}Z_{2}-\prox_{\gamma_{t}\phi}\left(\nu\alpha_{t}Z_{1}+\sigma_{t}Z_{2}\right)=\gamma_{t}\phi'\left(\prox_{\gamma_{t}\phi}\left(\nu\alpha_{t}Z_{1}+\sigma_{t}Z_{2}\right)\right).
\end{align*}
where the last transition comes from,
\[
\prox_{\gamma_t\phi}\left(\nu\alpha_tZ_1+\sigma_tZ_2\right)=\argmin_{u\in \mathbb R}\left\{\gamma_t \phi(u)+\frac{1}{2}\left(u-\nu\alpha_tZ_1-\sigma_tZ_2\right)^2\right\}
\]
whose first-order stationary implies 
\[
0=\gamma_t\phi'(\prox_{\gamma_t\phi}\left(\nu\alpha_tZ_1+\sigma_tZ_2\right))+\prox_{\gamma_t\phi}(\nu\alpha_tZ_1+\sigma_tZ_2)-(\nu\alpha_tZ_1+\sigma_tZ_2)
\]
Substituting this identity into the fourth scalar equation yields
\begin{align}
r_{t}^{2}=2\E\left[\left(1-\pi_{\rho}(\nu Z_{1})\right)\phi'\left(\prox_{\gamma_{t}\phi}\left(\nu\alpha_{t}Z_{1}+\sigma_{t}Z_{2}\right)\right)^{2}\right].
\label{eq:rt_identity}
\end{align}
The sixth scalar equation is
\begin{align*}
1-\frac{\gamma_{t}}{\sigma_{t}\tau_{t}}=\E\left[\frac{2\left(1-\pi_{\rho}(\nu Z_{1})\right)}{1+\gamma_{t}\phi''\left(\prox_{\gamma_{t}\phi}\left(\nu\alpha_{t}Z_{1}+\sigma_{t}Z_{2}\right)\right)}\right].
\end{align*}
Since $\mathbb E\left[2\left(1-\pi_\rho(\nu Z_1)\right)\right]=1,$ subtract it from both sides, we obtain
\begin{align}
\frac{1}{\sigma_{t}\tau_{t}}&=2\E\left[\left(1-\pi_{\rho}(\nu Z_{1})\right)\frac{\phi''\left(\prox_{\gamma_{t}\phi}\left(\nu\alpha_{t}Z_{1}+\sigma_{t}Z_{2}\right)\right)}{1+\gamma_{t}\phi''\left(\prox_{\gamma_{t}\phi}\left(\nu\alpha_{t}Z_{1}+\sigma_{t}Z_{2}\right)\right)}\right]\label{eq:25}\\&\leq2\E\left[\left(1-\pi_{\rho}(\nu Z_{1})\right)\phi''\left(\prox_{\gamma_{t}\phi}\left(\nu\alpha_{t}Z_{1}+\sigma_{t}Z_{2}\right)\right)\right].\notag
\end{align}
For $\phi(u)=\log(1+e^u),$ we have $0\leq\phi''(u)=\phi'(u)\left(1-\phi'(u)\right)\leq\phi'(u).$
Therefore,
\begin{equation}
\label{eq:10}
\begin{aligned}
\frac{1}{\sigma_{t}\tau_{t}}&\leq2\E\left[\left(1-\pi_{\rho}(\nu Z_{1})\right)\phi'\left(\prox_{\gamma_{t}\phi}\left(\nu\alpha_{t}Z_{1}+\sigma_{t}Z_{2}\right)\right)\right]\\&\leq2\sqrt{\E\left[1-\pi_{\rho}(\nu Z_{1})\right]\E\left[\left(1-\pi_{\rho}(\nu Z_{1})\right)\phi'\left(\prox_{\gamma_{t}\phi}\left(\nu\alpha_{t}Z_{1}+\sigma_{t}Z_{2}\right)\right)^{2}\right]}\\&=2\sqrt{\frac{1}{2}\cdot\frac{r_{t}^{2}}{2}}=r_{t},
\end{aligned}
\end{equation}

where the second inequality is Cauchy--Schwarz and the penultimate equality uses \cref{eq:rt_identity}. Consequently, $\sigma_t\tau_t r_t\geq 1.$ It follows that $\frac{\sigma_t\tau_t r_t}{\sqrt{\delta}}\geq\frac{1}{\sqrt{\delta}}.$ We next use the assumption on $f$. By $p$-homogeneity,
\[
f(w,w_{t-1})=\begin{cases}
w^{p}f(1,w_{t-1}), & w\geq0,\\
(-w)^{p}f(-1,w_{t-1}), & w<0.
\end{cases}
\]
Condition on $w^{\star}$ and on all previous tasks. 
First, we assume that $w^{\star}>0$. Since $r>0$ (see \cref{lem:rel}), the target $rw^{\star}$ is positive. By the optimality condition for the proximal map, whenever $w_t>0$,
\[
\sigma_{t}\tau_{t}\left(\theta_{t}w^{\star}+\frac{r_{t}}{\sqrt{\delta}}Z\right)=w_{t}+\lambda\sigma_{t}\tau_{t}pf(1,w_{t-1})w_{t}^{p-1}.
\]
In particular, the input corresponding to the output $rw^{\star}$ is $rw^{\star}+\lambda\sigma_{t}\tau_{t}pf(1,w_{t-1})(rw^{\star})^{p-1}.$ Then, for every $u\in\left[\frac{rw^{\star}}{2},\frac{3rw^{\star}}{2}\right],$ we have
\begin{align*}
1+\lambda\sigma_{t}\tau_{t}p(p-1)f(1,w_{t-1})u^{p-2}&\leq1+\lambda\sigma_{t}\tau_{t}p(p-1)f(1,w_{t-1})\max\left\{ 2^{2-p},\left(\frac{3}{2}\right)^{p-2}\right\} (rw^{\star})^{p-2}\\&\leq\max\left\{ 1,\,(p-1)\max\left\{ 2^{2-p},\left(\frac{3}{2}\right)^{p-2}\right\} \right\} \left(1+\lambda\sigma_{t}\tau_{t}pf(1,w_{t-1})(rw^{\star})^{p-2}\right)\\&=\max\left\{ 1,\,(p-1)\max\left\{ 2^{2-p},\left(\frac{3}{2}\right)^{p-2}\right\} \right\} \frac{rw^{\star}+\lambda\sigma_{t}\tau_{t}pf(1,w_{t-1})(rw^{\star})^{p-1}}{rw^{\star}}.
\end{align*}
Hence, if $|w_t-rw^{\star}|<\frac{rw^{\star}}{2},$ then $w_t>0$, and the mean-value theorem gives
\begin{align*}
|w_{t}-rw^{\star}|&\geq\frac{rw^{\star}}{\max\left\{ 1,\,(p-1)\max\left\{ 2^{2-p},\left(\frac{3}{2}\right)^{p-2}\right\} \right\} }\frac{\left|\sigma_{t}\tau_{t}\left(\theta_{t}w^{\star}+\frac{r_{t}}{\sqrt{\delta}}Z\right)-rw^{\star}-\lambda\sigma_{t}\tau_{t}pf(1,w_{t-1})(rw^{\star})^{p-1}\right|}{rw^{\star}+\lambda\sigma_{t}\tau_{t}pf(1,w_{t-1})(rw^{\star})^{p-1}}.
\end{align*}
If instead $|w_{t}-rw^{\star}|\geq\frac{rw^{\star}}{2},$ then, since the minimum below is at least $\frac{rw^{\star}}{2}$, combining the two cases yields
\begin{align}
|w_{t}-rw^{\star}|&\geq\frac{rw^{\star}}{\max\left\{ 2,\,(p-1)\max\left\{ 2^{2-p},\left(\frac{3}{2}\right)^{p-2}\right\} \right\} }\notag\\&\quad\times\min\Bigg\{1,\,\frac{\left|\sigma_{t}\tau_{t}\left(\theta_{t}w^{\star}+\frac{r_{t}}{\sqrt{\delta}}Z\right)-rw^{\star}-\lambda\sigma_{t}\tau_{t}pf(1,w_{t-1})(rw^{\star})^{p-1}\right|}{rw^{\star}+\lambda\sigma_{t}\tau_{t}pf(1,w_{t-1})(rw^{\star})^{p-1}}\Bigg\}.
\label{eq:prox_distortion}
\end{align}
The case $w^{\star}<0$ is identical, using $f(-1,w_{t-1})$.

We next use the following bound. For every
$a\in\mathbb R$, $b>0$, and $q>0$,
\[
\E\left[\min\left\{ 1,\frac{(a+bZ-q)^{2}}{q^{2}}\right\} \right]\geq\frac{1}{24}\frac{b^{2}}{a^{2}+b^{2}}.
\]
It follows by writing $Y=a+bZ-q$, so we have $\E[Y^2]=b^2+(a-q)^2$ and
\[
\E[Y^4]=3b^4+6b^2(a-q)^2+(a-q)^4\leq 3\left(b^2+(a-q)^2\right)^2.
\]
Therefore, by the Paley--Zygmund inequality, $\PP\left(Y^{2}\geq\frac{1}{2}\E[Y^{2}]\right)\geq\frac{1}{12},$ and hence
\begin{align*}
\E\left[\min\left\{ 1,\frac{Y^{2}}{q^{2}}\right\} \right]\geq\frac{1}{12}\min\left\{ 1,\frac{b^{2}+(a-q)^{2}}{2q^{2}}\right\} \geq\frac{1}{24}\min\left\{ 1,\frac{b^{2}+(a-q)^{2}}{q^{2}}\right\} .
\end{align*}
Moreover,
\[
\frac{b^{2}+(a-q)^{2}}{q^{2}}\geq\inf_{x>0}\frac{b^{2}+(a-x)^{2}}{x^{2}}=\frac{b^{2}}{a^{2}+b^{2}},
\]
and the latter quantity is at most one, proving the claim.
Squaring \cref{eq:prox_distortion} and applying this bound conditionally on
$w^{\star}$ and all previous tasks gives
\begin{align*}
\E\left[(w_{t}-rw^{\star})^{2}\mid w^{\star},\text{ previous tasks}\right]
\geq\frac{r^{2}{w^{\star}}^{2}}{24\max\left\{ 2,\,(p-1)\max\left\{2^{2-p},\left(\frac{3}{2}\right)^{p-2}\right\} \right\} ^{2}}\frac{\left(\frac{\sigma_{t}\tau_{t}r_{t}}{\sqrt{\delta}}\right)^{2}}{(\sigma_{t}\tau_{t}\theta_{t}w^{\star})^{2}+\left(\frac{\sigma_{t}\tau_{t}r_{t}}{\sqrt{\delta}}\right)^{2}}.
\end{align*}
Furthermore,
\begin{align*}
\theta_{t}&=2\E\left[\pi_{\rho}'(\nu Z_{1})\phi'\left(\prox_{\gamma_{t}\phi}\left(\nu\alpha_{t}Z_{1}+\sigma_{t}Z_{2}\right)\right)\right]\\&\leq2\E\left[\left(1-\pi_{\rho}(\nu Z_{1})\right)\phi'\left(\prox_{\gamma_{t}\phi}\left(\nu\alpha_{t}Z_{1}+\sigma_{t}Z_{2}\right)\right)\right]\\&\leq2\sqrt{\E\left[1-\pi_{\rho}(\nu Z_{1})\right]\E\left[\left(1-\pi_{\rho}(\nu Z_{1})\right)\phi'\left(\prox_{\gamma_{t}\phi}\left(\nu\alpha_{t}Z_{1}+\sigma_{t}Z_{2}\right)\right)^{2}\right]}=r_{t},
\end{align*}
where we used $\pi_{\rho}'(u)=(1-2\rho)\sigma(u)(1-\sigma(u))\leq(1-2\rho)\sigma(u)=1-\pi_{\rho}(u),$ $\E\left[1-\pi_{\rho}(\nu Z_{1})\right]=\frac{1}{2},$ and \cref{eq:rt_identity}. Consequently, because $\theta_t\leq r_t,$
\begin{align*}
&\frac{\left(\frac{\sigma_{t}\tau_{t}r_{t}}{\sqrt{\delta}}\right)^{2}}{(\sigma_{t}\tau_{t}\theta_{t}w^{\star})^{2}+\left(\frac{\sigma_{t}\tau_{t}r_{t}}{\sqrt{\delta}}\right)^{2}}=\frac{r_{t}^{2}}{\delta\theta_{t}^{2}{w^{\star}}^{2}+r_{t}^{2}}\geq\frac{1}{1+\delta{w^{\star}}^{2}}.
\end{align*}
Thus
\begin{align*}
\E\left[(w_{t}-rw^{\star})^{2}\mid w^{\star},\text{ previous tasks}\right]
&\geq\frac{r^{2}{w^{\star}}^{2}}{24\max\left\{ 2,\,(p-1)\max\left\{ 2^{2-p},\left(\frac{3}{2}\right)^{p-2}\right\} \right\} ^{2}\left(1+\delta{w^{\star}}^{2}\right)}\\&\geq\frac{r^{2}}{48\max\left\{ 2,\,(p-1)\max\left\{ 2^{2-p},\left(\frac{3}{2}\right)^{p-2}\right\} \right\} ^{2}}\min\left\{ {w^{\star}}^{2},\frac{1}{\delta}\right\} .
\end{align*}
Averaging over the previous tasks and over $w^{\star}$ yields
\[
\E\left[(w_{t}-rw^{\star})^{2}\right]\geq\frac{r^{2}}{48\max\left\{ 2,\,(p-1)\max\left\{ 2^{2-p},\left(\frac{3}{2}\right)^{p-2}\right\} \right\} ^{2}}\E\left[\min\left\{ {w^{\star}}^{2},\frac{1}{\delta}\right\} \right].
\]
Finally, since $w^{\star}=S\frac{\nu}{\sqrt{s}}G,$ where $S\sim \mathrm{Ber}(s),$ $G\sim \mathcal N(0,1)$ we have $\E\left[\min\left\{ {w^{\star}}^{2},\frac{1}{\delta}\right\} \right]=s\E_{G}\left[\min\left\{ \frac{\nu^{2}}{s}G^{2},\frac{1}{\delta}\right\} \right].$ On the event $\{|G|\geq1\}$, we have $\min\left\{ \frac{\nu^{2}}{s}G^{2},\frac{1}{\delta}\right\} \geq\min\left\{ \frac{\nu^{2}}{s},\frac{1}{\delta}\right\} ,$ and therefore
\begin{align*}
s\E_{G}\left[\min\left\{ \frac{\nu^{2}}{s}G^{2},\frac{1}{\delta}\right\} \right]&\geq s\PP(|G|\geq1)\min\left\{ \frac{\nu^{2}}{s},\frac{1}{\delta}\right\}
\geq\frac{3}{10}\min\left\{ \nu^{2},\frac{s}{\delta}\right\}.
\end{align*}
Hence,
\[
\E\left[(w_{t}-rw^{\star})^{2}\right]\geq\frac{r^{2}}{160\max\left\{ 2,\,(p-1)\max\left\{ 2^{2-p},\left(\frac{3}{2}\right)^{p-2}\right\} \right\} ^{2}}\min\left\{ \nu^{2},\frac{s}{\delta}\right\} .
\]
It remains to consider $0<p<1$. Condition on $w^{\star}$ and on all previous tasks. Suppose first that $w_t>0$. Since $w_t$ is a global minimizer of the proximal objective,
\begin{align*}
\frac{1}{2}\left(w_{t}-\sigma_{t}\tau_{t}\left(\theta_{t}w^{\star}+\frac{r_{t}}{\sqrt{\delta}}Z\right)\right)^{2}+\lambda\sigma_{t}\tau_{t}f(1,w_{t-1})w_{t}^{p}\leq\frac{1}{2}\left(\sigma_{t}\tau_{t}\left(\theta_{t}w^{\star}+\frac{r_{t}}{\sqrt{\delta}}Z\right)\right)^{2}.
\end{align*}
Moreover, since $w_t>0$, the first-order condition gives
\[
\sigma_{t}\tau_{t}\left(\theta_{t}w^{\star}+\frac{r_{t}}{\sqrt{\delta}}Z\right)=w_{t}+\lambda\sigma_{t}\tau_{t}pf(1,w_{t-1})w_{t}^{p-1}.
\]
Substituting this identity into the preceding inequality yields $w_{t}^{2-p}\geq2\lambda\sigma_{t}\tau_{t}(1-p)f(1,w_{t-1}).$
Consequently, 
\[
\frac{\dd}{\dd w_{t}}\left[w_{t}+\lambda\sigma_{t}\tau_{t}pf(1,w_{t-1})w_{t}^{p-1}\right]=1-\lambda\sigma_{t}\tau_{t}p(1-p)f(1,w_{t-1})w_{t}^{p-2}
\geq1-\frac{p}{2}>0.
\]
Hence the proximal map is increasing,
\[
\frac{\dd}{\dd x}\prox_{\lambda\sigma_{t}\tau_{t}f}(x)=\frac{1}{1-\lambda\sigma_{t}\tau_{t}p(1-p)f(1,w_{t-1})w_{t}^{p-2}}\geq1.
\]
The same conclusion holds on the negative nonzero branch, with
$f(-1,w_{t-1})$ in place of $f(1,w_{t-1})$. On the other hand if $\PP(w_t=0\mid w^{\star},\text{ previous tasks})\geq\frac{1}{2},$
then
\[
\E\left[(w_{t}-rw^{\star})^{2}\mid w^{\star},\text{ previous tasks}\right]\geq\frac{1}{2}r^{2}{w^{\star}}^{2}.
\]
Otherwise, $\PP(w_t\neq0\mid w^{\star},\text{ previous tasks})>\frac{1}{2},$
so at least one of the events $\{w_t>0\}$ or $\{w_t<0\}$ has conditional
probability at least $\frac{1}{4}$. Let $H$ denote such an event. Since on each
nonzero branch the proximal map is increasing with derivative at least
one,
\[
\Var(w_{t}\mid H)\geq\Var\left(\sigma_{t}\tau_{t}\left(\theta_{t}w^{\star}+\frac{r_{t}}{\sqrt{\delta}}Z\right)\middle|H\right).
\]
Since $H$ corresponds to a one-sided Gaussian event of probability at
least $\frac{1}{4}$,
\[
\PP(H\mid w^{\star},\text{ previous tasks})\Var(Z\mid H)\geq\frac{1}{20}.
\]
Therefore,
\begin{align*}
\E\left[(w_{t}-rw^{\star})^{2}\mid w^{\star},\text{ previous tasks}\right]&\geq\PP(H\mid w^{\star},\text{ previous tasks})\Var(w_{t}\mid H)
\geq\frac{1}{20}\left(\frac{\sigma_{t}\tau_{t}r_{t}}{\sqrt{\delta}}\right)^{2}\geq\frac{1}{20\delta}.
\end{align*}
Thus, for every $0<p\leq1$,
\[
\E\left[(w_{t}-rw^{\star})^{2}\mid w^{\star},\text{ previous tasks}\right]\geq\frac{1}{20}\min\left\{ r^{2}{w^{\star}}^{2},\frac{1}{\delta}\right\} .
\]
Since $h(\nu r)\leq h(0)=\frac{1}{4},$ and $r=\frac{(1-2\rho)h(\nu)}{h(\nu r)}$ implying $r\geq 4h(\nu)(1-2\rho).$ Also from \cref{eq:28} $h(\nu)\gtrsim\frac{1}{1+\nu},$ and therefore $r\gtrsim\frac{1-2\rho}{1+\nu}.$
Then \cref{prop:coefficient_error_to_excess_risk} proves the claim.
\end{proof}
\subsection{Proof of \cref{thm:signed-weighted-l1-upper}}
\label{prf:signed-weighted-l1-upper}
\begin{recall}[\cref{thm:signed-weighted-l1-upper}]
Let $\lambda>0,$ $\epsilon=\frac{\lambda}{\sqrt{\frac{2}{\delta}\log\left(\frac{64}{s}\right)}}.$
Then there exist universal constants $C,c>0$ such that, if $s\leq c\min\big\{\frac{\delta}{(1+\nu)^{4}\log_+\left(\frac{1+\nu}{\delta}\right)},\frac{\delta(1-2\rho)^{2}\nu^{2}}{(1+\nu)^{2}\log_+^{3}\left(\frac{(1+\nu)^{2}}{\delta(1-2\rho)^{2}\nu^{2}}\right)},\frac{\delta}{(1+\nu)^{24}}\big\},$\footnote{Here, and in all following results $\log_+(x)=\max\{1,\log(x)\}.$}
then for every task $t\geq 2$ the following holds.
\[
\mathcal{L}(\mathbf w_T)-\inf_{\vw} \mathcal{L}(\vw)\leq C\frac{s(1+\nu)^{8}}{\delta}
\]
\end{recall}
\begin{proof}

Recall that we aim to bound $\mathcal{L}(\mathbf w_t)-\inf_{\vw} \mathcal{L}(\vw)$. 
By \cref{prop:coefficient_error_to_excess_risk} we know that \[\mathcal{L}(\mathbf w_t)-\inf_{\vw} \mathcal{L}(\vw)\leq\frac{1}{8}\E[(w_{t}-rw^{\star})^2],\] 
therefore it is sufficient to minimize $\E[(w_{t}-rw^{\star})^2]$.
Since by \cref{eq:7},
\[
\E[(w_{t}-rw^{\star})^2]=\nu^2(\alpha_t-r)^2+\sigma_t^2
\]
It is sufficient to bound each one of the terms $\alpha_t-r$ and $\sigma_t^2$.
The regularizer is
\[
f(w)=\begin{cases}
\dfrac{w}{[w_{t-1}]_{+}+\epsilon_{s}}, & w>0,\\
\dfrac{-w}{[-w_{t-1}]_{+}+\epsilon_{s}}, & w<0.
\end{cases}
\]
By \cref{lem:3} it is convex and positively one-homogeneous. In particular, the first scalar equation can be written as $w_{t}=\sigma_t\tau_tV_t,$ where $V_t=\mathcal T_{\frac{\lambda_0}{[w_{t-1}]_++\epsilon_s},\frac{\lambda_0}{[-w_{t-1}]_++\epsilon_s}}\left(\theta_tw^{\star}+\frac{r_t}{\sqrt{\delta}}Z_t\right)$
and
\[
\mathcal{T}_{a,b}(x)=\begin{cases}
x-a, & x>a,\\
0, & -b\leq x\leq a,\\
x+b, & x<-b.
\end{cases}
\]
Both thresholds are at most $\frac{\lambda_0}{\epsilon_s}=\sqrt{\frac{2}{\delta}\log\left(\frac{64}{s}\right)}.$
Define $m_t=\frac{\mathbb E[w^{\star} V_t]}{\nu^2}.$ The first and third equations give $\alpha_t=\sigma_t\tau_tm_t,$ and $\sigma_t^2=(\sigma_t\tau_t)^2\left(\mathbb E[V_t^2]-\nu^2m_t^2\right).$
Moreover, conditionally on $(w^{\star},w_{t-1})$,
\[
\frac{\partial V_t}{\partial Z_t}
=\frac{r_t}{\sqrt{\delta}}\mathbf{1}_{\{V_t\neq0\}}
\]
almost surely. Stein's Lemma therefore gives $\gamma_t=\frac{\sigma_t\tau_t}{\delta}\mathbb P(V_t\neq0).$ The equation for $r_t$ is (see \cref{eq:rt_identity} and derivations above it),
\begin{align}
\label{eq:34}  
r_t^2=2\mathbb E\left[(1-\pi_\rho(\nu Z_1))\phi'(\prox_{\gamma_t\phi}\left(\nu\alpha_tZ_1+\sigma_tZ_2\right))^2\right].
\end{align}

Since $0<\phi'(u)<1$ and $2\mathbb E[1-\pi_\rho(\nu Z_1)]=1$, we have $r_t\leq1.$ Using
\begin{align}
\label{eq:23}
\prox_{\gamma_t\phi}\left(\nu\alpha_tZ_1+\sigma_tZ_2\right)+\gamma_t\phi'(\prox_{\gamma_t\phi}\left(\nu\alpha_tZ_1+\sigma_tZ_2\right))=\nu\alpha_tZ_1+\sigma_tZ_2,
\end{align}
in the equation for $\theta_t$, and using the fact that $\pi_\rho'(\nu Z_1)$ is even, gives
\begin{align}
\label{eq:22}
\theta_{t}=2\mathbb{E}\left[\pi_{\rho}'(\nu Z_{1})\left(\phi'(\prox_{\gamma_{t}\phi}\left(\nu\alpha_{t}Z_{1}+\sigma_{t}Z_{2}\right))-\frac{\nu\alpha_{t}Z_{1}+\sigma_{t}Z_{2}}{\gamma_{t}}\right)\right]=2\E\left[\pi_{\rho}'(\nu Z_{1})\phi'\left(\prox_{\gamma_{t}\phi}\left(\nu\alpha_{t}Z_{1}+\sigma_{t}Z_{2}\right)\right)\right]
\end{align}
Consequently, $0<\theta_t\leq2(1-2\rho)h(\nu).$
The sixth scalar equation can be rewritten as
\[
\frac{1}{\sigma_t\tau_t}=\mathbb E\left[\frac{2(1-\pi_\rho(\nu Z_1))\phi''(\prox_{\gamma_t\phi}\left(\nu\alpha_tZ_1+\sigma_tZ_2\right))}{1+\gamma_t\phi''(\prox_{\gamma_t\phi}\left(\nu\alpha_tZ_1+\sigma_tZ_2\right))}\right].
\]
Since $0\leq\phi''(u)=\phi'(u)\left(1-\phi'(u)\right)\leq\phi'(u)$ and 
\[
\gamma_t=\frac{\sigma_t\tau_t}{\delta}\PP\left(V_t\neq 0\right)\geq\frac{\sigma_t\tau_t}{\delta}\PP\left(V_t\neq 0,w^{\star}=0\right)\geq (1-s)\frac{\sigma_t\tau_t}{\delta}\PP\left(\left|\frac{r_t}{\sqrt{\delta}}Z_t\right|>\frac{\lambda_0}{\epsilon_s}\mid w^{\star}=0\right)>0.
\] 
Cauchy--Schwarz yields
\[
\frac{1}{\sigma_t\tau_t}\leq2\mathbb E[(1-\pi_\rho(\nu Z_1))\phi'(\prox_{\gamma_t\phi}\left(\nu\alpha_tZ_1+\sigma_tZ_2\right))]\leq r_t.
\]
Thus $\sigma_t\tau_tr_t\geq1$ and $\sigma_t\tau_t\geq1.$
We next control false coordinates. Put $\zeta_t=\mathbb P(w_{t}\neq0\mid w^{\star}=0).$
Suppose $w_{t-1}>0$. The probability that $w_{t}$ has the same sign is $\Phi\left(-\frac{\frac{\lambda_0}{w_{t-1}+\epsilon_s}\sqrt{\delta}}{r_t}\right)$ since the expression inside $\Phi(\cdot)$ is negative the probability is at most $\frac{1}{2}$. An output having the opposite sign requires
\[
\frac{r_t}{\sqrt{\delta}}Z_t<-\sqrt{\frac{2}{\delta}\log\left(\frac{64}{s}\right)},
\]
whose probability is at most $\Phi\left(-\sqrt{2\log\left(\frac{64}{s}\right)}\right)$. The case $w_{t-1}<0$ is identical. If $w_{t-1}=0$, both directions have threshold $\sqrt{\frac{2}{\delta}\log\left(\frac{64}{s}\right)}$, and therefore
\[
\mathbb P(w_{t}\neq0\mid w^{\star}=0,w_{t-1}=0)\leq2\Phi\left(-\sqrt{2\log\left(\frac{64}{s}\right)}\right).
\]
It follows that
\[
\zeta_t\leq\frac{1}{2}\zeta_{t-1}+2\Phi\left(-\sqrt{2\log\left(\frac{64}{s}\right)}\right).
\]
Since $\zeta_0=0$, and $2\Phi(-u)\leq e^{-\frac{u^{2}}{2}}$,
we obtain, uniformly in $t$, $\zeta_t\leq\frac{s}{32}.$
Therefore $\mathbb P(V_t\neq0)\leq s+(1-s)\zeta_t\leq\frac{33s}{32},$
and hence $\frac{\gamma_t}{\sigma_t\tau_t}=\frac{\mathbb P(V_t\neq0)}{\delta}\leq\frac{33s}{32\delta}.$ 
Next, we prove that $\sigma_t\tau_t$ is uniformly bounded. For $q>0$ and $\eta\geq0$, define $g(x)=\phi'\left(\prox_{q\eta\phi}(qx)\right).$
Then $g$ increases from zero to one and
\[
g'(x)=\frac{q\phi''(\prox_{q\eta\phi}(qx))}{1+q\eta\phi''(\prox_{q\eta\phi}(qx))}.
\]
To prove that $\sigma_t\tau_t$ is bounded we first prove a lower estimate on $\mathbb E[2(1-\pi_\rho(\nu Z_1))g'(\sqrt{\E[V_{t}^{2}]}Z)]$ which appears in equation six this allows us to reach a contradiction. This form follows by orthogonal transformation, we introduce an independent $G\sim \mathcal N(0,1)$, and \[
\left(\begin{array}{c}
Z\\
G
\end{array}\right)=\left(\begin{array}{cc}
\frac{\nu m_{t}}{\sqrt{\E[V_{t}^{2}]}} & \frac{\sqrt{\E[V_{t}^{2}]-\nu^{2}m_{t}^{2}}}{\sqrt{\E[V_{t}^{2}]}}\\
\frac{\sqrt{\E[V_{t}^{2}]-\nu^{2}m_{t}^{2}}}{\sqrt{\E[V_{t}^{2}]}} & -\frac{\nu m_{t}}{\sqrt{\E[V_{t}^{2}]}}
\end{array}\right)\left(\begin{array}{c}
Z_{1}\\
Z_{2}
\end{array}\right)
\] so $Z_1=\frac{\nu m_t}{\E[V_{t}^{2}]}h+\sqrt{1-\frac{\nu^2m_t^2}{\E[V_{t}^{2}]}}G,$ with $G\sim \mathcal N(0,1),$ conditionally on $\sqrt{\E[V_{t}^{2}]}Z=h$, and from $\alpha_t=\sigma_t\tau_tm_t$ and $\sigma_t^2=(\sigma_t\tau_t)^2(\E[V_t^2]-\nu^2m_t^2)$ which follows by the first and third equations, respectively, so 
\[
\nu\alpha_{t}Z_{1}+\sigma_{t}Z_{2}=\nu\alpha_{t}\left(\frac{\nu m_{t}}{\sqrt{\E[V_{t}^{2}]}}Z+\frac{\sqrt{\E[V_{t}^{2}]-\nu^{2}m_{t}^{2}}}{\sqrt{\E[V_{t}^{2}]}}G\right)+\sigma_{t}\left(\frac{\sqrt{\E[V_{t}^{2}]-\nu^{2}m_{t}^{2}}}{\sqrt{\E[V_{t}^{2}]}}Z-\frac{\nu m_{t}}{\sqrt{\E[V_{t}^{2}]}}G\right)=\sqrt{\E[V_{t}^{2}]}Z.
\]
Define
\[
\omega_{t}(x)=\E_{G}\left[2\left(1-\pi_{\rho}\left(\nu\left(\frac{\nu m_{t}}{\E[V_{t}^{2}]}x+\sqrt{1-\frac{\nu^{2}m_{t}^{2}}{\E[V_{t}^{2}]}}G\right)\right)\right)\right]
\]
from symmetry, $\omega_t(0)=1,$ and, uniformly in $t$, $|\omega_t(x)-1|\le C|x|,$ because \[\omega_{t}'(x)=-\frac{2\nu(\nu m_{t})}{\E[V_{t}^{2}]}\E_{G}\left[\pi_{\rho}'\left(\nu\left(\frac{\nu m_{t}}{\E[V_{t}^{2}]}x+\sqrt{1-\frac{\nu^{2}m_{t}^{2}}{\E[V_{t}^{2}]}}G\right)\right)\right]\] and because $\sup_u\pi_\rho'(u)=(1-2\rho)\sup_u\phi''(u)\leq \frac{1-2\rho}{4}\leq\frac{1}{4}$ and $\omega_t(0)=1$ with $\nu m_t<\sqrt{\E[V_{t}^{2}]}$ the statement holds with $C=\frac{\nu}{2\sqrt{\E[V_{t}^{2}]}}.$ It will be convenient to use the inverse relation $a=g(x)=\phi'(u)=\frac{1}{1+e^{-u}}$ as the variable. The first order optimality enforces $u+q\eta\phi'(u)=qx$ (where $q=\sigma_t\tau_t$).
The inverse relation implies $u=\log\left(\frac{a}{1-a}\right)$, and
\[
x(a)=\frac{1}{q}\log\left(\frac{a}{1-a}\right)+\eta a.
\]
Let $a_0=g(0)$. Then
\[
\log\left(\frac{a_0}{1-a_0}\right)+q\eta a_0=0.
\]
At $a=\frac{1}{2}$ we have $\log\left(1\right)+\frac{1}{2}q\eta\geq 0$ and at $a=\frac{1}{q\eta+2}$ we have $\log\left(\frac{\frac{1}{q\eta+2}}{1-\frac{1}{q\eta+2}}\right)+\frac{q\eta}{q\eta+2}\leq-\frac{q\eta}{q\eta+1}+\frac{q\eta}{q\eta+2}<0$ the monotonicity of the function therefore gives $\frac{1}{q\eta+2}\leq a_0\leq\frac{1}{2}.$
For $a\in[a_0,a_0+\frac{1}{4}]$,
\[
\frac{\dd x}{\dd a}=\frac{1}{qa(1-a)}+\eta\underset{\frac{1}{q\eta+2}\leq a\leq \frac{3}{4}}{\leq} \frac{4}{q}(q\eta+2)+\eta\leq 8(\eta+q^{-1}).
\]
If $R<2(1+\nu)(\eta+q^{-1})$, the interval
$a_0\leq a\leq a_0+\frac{R}{8(1+\nu)(\eta+q^{-1})}$
is mapped into $0\leq x\leq \frac{R}{1+\nu}$ as
$x(a)-x(a_0)=\int_{a_0}^a\frac{\dd x}{\dd b}\dd b$, so
\[
x(a)\leq8(\eta+q^{-1})(a-a_{0})\leq8(\eta+q^{-1})\frac{R}{8(1+\nu)(\eta+q^{-1})}=\frac{R}{1+\nu}.
\]
Then, if $R<2(1+\nu)(\eta+q^{-1})$, since $\dd a=g'(x)\dd x$ we have
\begin{align*}
\E\left[2(1-\pi_{\rho}(\nu Z_{1}))g'(H_{t})\right]
&\geq\int_{a_{0}}^{a_{0}+\frac{R}{8(1+\nu)(\eta+q^{-1})}}\frac{1}{2\sqrt{2\pi}R}e^{-\frac{x(a)^{2}}{2R^{2}}}\dd a\\&\geq\frac{e^{-\frac{1}{2}}}{\sqrt{2\pi}R}\frac{R}{16(1+\nu)(\eta+q^{-1})}=\frac{e^{-\frac{1}{2}}}{16\sqrt{2\pi}(1+\nu)(\eta+q^{-1})}.\end{align*}
Similarly, if $R\geq2(1+\nu)(\eta+q^{-1})$, the values
$a\in[a_0,a_0+\frac{1}{4}]$ are mapped into an interval $0\leq x(a)\leq8(\eta+q^{-1})\frac{1}{4}=2(\eta+q^{-1})\leq\frac{R}{1+\nu}.$
Hence $\E\left[2(1-\pi_{\rho}(\nu Z_{1}))g'(H_t)\right]\geq\int_{a_0}^{a_0+\frac{1}{4}}\frac{1}{2\sqrt{2\pi}R}e^{-\frac{x(a)^2}{2R^2}}\dd a\geq\frac{e^{-\frac{1}{2}}}{8\sqrt{2\pi}R}.$
The two cases prove $
\E\left[2(1-\pi_{\rho}(\nu Z_{1}))g'(H_t)\right]\geq\frac{e^{-\frac{1}{2}}}{16\sqrt{2\pi}(1+\nu)\left(R+\eta+q^{-1}\right)}.$

Suppose now, for contradiction, that no uniform upper bound on $\sigma_t\tau_t$ exists for a sufficiently small $s$. Then there are $s_n\downarrow0$, tasks $t_n$, and scalar solutions such that $\sigma_{t_n}\tau_{t_n}\to\infty$. Let
$H_n=\nu m_nZ_1+\sqrt{\E[V_{n}^2]-\nu^2m_{n}^2}Z_2,$ this definition follows from $\alpha_t=\sigma_t\tau_tm_t$ and $\sigma_t^2=(\sigma_t\tau_t)^2(\E[V_t^2]-\nu^2m_t^2)$ by the first and third equations, respectively. Therefore,
\[
\nu\alpha_{n}Z_{1}+\sigma_{n}Z_{2}=\sigma_{n}\tau_{n}\left(\nu m_{n}Z_{1}+\sqrt{\E[V_{n}^{2}]-\nu^{2}m_{n}^{2}}Z_{2}\right)=\sigma_{n}\tau_{n}H_{n}
\]
The previous bound give $\PP(V_n\neq 0)\leq \frac{33s_n}{32}\xrightarrow[n\to\infty]{}0$. Moreover, $\E[H_n^2]=\nu^2m_n^2+\E[V_{n}^2]-\nu^2m_{n}^2=\mathbb E[V_n^2],$
which is uniformly bounded because
\[
\E[V_n^2]\leq\E\left[\left(\theta_nw^{\star}+\frac{r_n}{\sqrt{\delta}}Z\right)^2\right]\underset{r_n\leq 1, \theta_n\leq 2(1-2\rho)h(\nu),w^{\star}\perp Z}{\leq} 2\nu^2(1-2\rho)h(\nu)+\frac{1}{\delta}.
\]
Define $g_n(h)=\phi'\left(\prox_{\sigma_n\tau_n\frac{\PP(V_n\neq 0)}{\delta}\phi}(\sigma_n\tau_nh)\right).$ Multiplying the sixth scalar equation by $\sigma_n\tau_n$ gives \[1=\mathbb E\left[2(1-\pi_\rho(\nu Z_1))g_n'(H_n)\right].\] From the lower bound proved for $\mathbb E[2(1-\pi_\rho(\nu Z_1))g'(\sqrt{\E[V_n^2]}Z)]$ we get,
\begin{align}
&1\geq\frac{\frac{e^{-\frac{1}{2}}}{16\sqrt{2\pi}(1+\nu)}}{\sqrt{\E[H^2_n]}+\frac{\PP(V_n\neq 0)}{\delta}+(\sigma_n\tau_n)^{-1}}\label{eq:26}
\\&\implies
\sqrt{\mathbb E[H^2_n]}+\frac{\PP(V_n\neq 0)}{\delta}+(\sigma_n\tau_n)^{-1}\geq \frac{e^{-\frac{1}{2}}}{16\sqrt{2\pi}(1+\nu)}>0.\notag
\end{align}

Since $\E[H^2_n]$ is uniformly bounded (from above and below) and $|m_n|\leq \frac{\sqrt{\E[H^2_n]}}{\nu}$ by Cauchy-Schwartz then by Bolzano–Weierstrass, there exists a subsequence, still denoted by $n$,
\[
m_n\to m_\infty,\qquad\sqrt{\E[V_{n}^2]-\nu^2m_{n}^2}\to\xi_\infty,\qquad\nu^2m_\infty^2+\xi_\infty^2>0.
\]
For every bounded continuous $\psi$,
\[
\int_{\mathbb R}\psi(h)g_n'(h)\dd h=\int_0^1\psi\left(\frac{1}{(\sigma_n\tau_n)^{-1}}\log\left(\frac{a}{1-a}\right)+\frac{\PP(V_n\neq 0)}{\delta}a\right)\dd a\longrightarrow\psi(0).
\]
Conditionally on $H_n=h$, $Z_{1}\overset{d}{=}\frac{\nu m_{n}}{\E[V_{n}^{2}]}h+\sqrt{1-\frac{\nu^{2}m_{n}^{2}}{\E[V_{n}^{2}]}}G,$ where $G\sim\mathcal N(0,1)$. In particular, conditional on
$H_n=0$, $Z_1$ is symmetric, and therefore
\[
\mathbb E[2(1-\pi_\rho(\nu Z_1))\mid H_n=0]=1.
\]
The conditional Gaussian formula and the lower bound on $\sqrt{\E[V_n^2]}$ imply local uniform convergence of the conditional expectations. Hence the weak convergence of $g_n'(h)\dd h$ gives
\begin{align*}
1&=\lim_{n\to\infty}\mathbb{E}\left[2(1-\pi_{\rho}(\nu Z_{1}))g_{n}'(H_{n})\right]\\&=\lim_{n\to\infty}\int\E[2(1-\pi_{\rho}\left(\nu Z_{1}\right)\mid H_{n}=h]\frac{1}{\sqrt{2\pi\E[V_{n}^{2}]}}e^{-\frac{h^{2}}{2\sqrt{\E[V_{n}^{2}]}}}g_{n}'(h)\dd h=\frac{1}{\sqrt{2\pi\E[V_{\infty}^{2}]}},
\end{align*}
Thus $\E[V_\infty^2]=\frac{1}{\sqrt{2\pi}}.$ For every $h\neq0$, $g_n(h)\longrightarrow\mathbf{1}_{\{h>0\}}.$ Dominated convergence in the equation for $\theta_n$ gives
\[
\theta_n\longrightarrow2\mathbb E\left[\pi_\rho'(\nu Z_1)\mathbf{1}_{\{\nu m_\infty Z_1+\xi_\infty Z_2>0\}}\right].
\]
The joint law is invariant under $(Z_1,Z_2)\mapsto(-Z_1,-Z_2)$, while $\pi_\rho'(\nu Z_1)$ is even. Therefore $\theta_n\longrightarrow(1-2\rho)h(\nu).$ Since $s_n\downarrow0$, the bounds give as $\nu|m_n-\theta_n|\leq \sqrt{\E[(V_n-\theta_nw^\star)^2]}\longrightarrow0$ the inequality follows because $m_nw^{\star}$ is the $L^2$-projection of $V_t$ onto $w^{\star}$\footnote{It follows directly by noting that $\E[(V_n-a w^{\star})^2]=\E[(V_n)^2]-2a\E[V_n w^{\star}]+a^2\nu^2$, first order stationary condition would enforce $a=\frac{\E[V_t w^{\star}]}{\nu^2}=m_t$} and the convergence is proven later independently, see conclusion in \cref{eq:refined_normalized_channel}, similarly $\xi_n\xrightarrow[]{n\to\infty}0$. It follows that $\sqrt{\E[V_\infty^2]}=\nu(1-2\rho)h(\nu)$. This contradicts $\nu(1-2\rho)h(\nu)<\frac{1}{\sqrt{2\pi}}.$
Indeed, $\frac{\dd}{\dd a}\left(ah(a)\right)=\mathbb E[Z^2\phi'(aZ)]>0$ and $\lim_{a\to\infty}ah(a)=\frac{1}{\sqrt{2\pi}}.$
Therefore there exist $s_0>0$ and $Q_{\nu,\rho,\delta}<\infty$ such
that $\sigma_t\tau_t\leq Q_{\nu,\rho,\delta}$ for every $s\leq s_0$ and every task $t$. It is enough to consider tasks for which, $\sigma_t\tau_t>64\sqrt{2\pi}e^{\frac{1}{2}}(1+\nu)$, otherwise it immediately follows that $Q_{\nu,\rho,\delta}=64\sqrt{2\pi}e^{\frac{1}{2}}(1+\nu)$.
The main contradiction of the previous argument lies in the fact that $\nu(1-2\rho)h(\nu)<\frac{1}{\sqrt{2\pi}}$, we investigate when it holds
\begin{equation}
\begin{aligned}
&\frac{1}{\sqrt{2\pi}}-\nu(1-2\rho)h(\nu)\\&\leq\left|\frac{1}{\sqrt{2\pi}}-\sqrt{\E[V_{t}^{2}]}\right|+\underbrace{\left|\sqrt{\E[V_{t}^{2}]}-\nu\theta_{t}\right|}_{=\left|\left\Vert \left(\nu m_{t},\sqrt{\E[V_{t}^{2}]-\nu^{2}m_{t}^{2}}\right)\right\Vert _{2}-\left\Vert (\nu\theta_{t},0)\right\Vert _{2}\right|}+\nu\left|\theta_{t}-(1-2\rho)h(\nu)\right|\\&\stackrel{[1]}{\leq}\frac{2}{\sqrt{2\pi}}\left|\int_{\mathbb{R}}\left(e^{-\frac{h^{2}}{2\E[V_{t}^{2}]}}\E\left[1-\pi_{\rho}(\nu Z_{1})\mid H_{t}=h\right]-\E\left[1-\pi_{\rho}(\nu Z_{1})\mid H_{t}=0\right]\right)g_{t}'(h)\dd h\right|\\&\hspace{4em}+\sqrt{\E\left[(V_{t}-m_{t}w^{\star})^{2}\right]}+\nu|m_{t}-\theta_{t}|+2\nu\sup_{u}\pi_{\rho}'(u)\E\left[\left|g_{t}(H_{t})-\mathbf{1}_{\{H_{t}>0\}}\right|\right]\\&\stackrel{[2]+[3]}{\leq}\left[\frac{2}{\sqrt{2\pi}\frac{e^{-\frac{1}{2}}}{16\sqrt{2\pi}(1+\nu)}}\left(e^{-\frac{1}{2}}+\frac{\nu(1-2\rho)}{4}\right)+\frac{\nu(1-2\rho)}{2\sqrt{2\pi}\frac{e^{-\frac{1}{2}}}{16\sqrt{2\pi}(1+\nu)}}\right]\left(\frac{2\log2}{\sigma_{t}\tau_{t}}+\frac{\underbrace{\PP(V_{t}\neq0)}_{\leq\frac{33}{32}s}}{2\delta}\right)\\&\hspace{4em}+2\sqrt{\frac{2s}{\delta}\left(1+2\log\frac{64}{s}\right)}.
\end{aligned}
\label{eq:6}
\end{equation}
Here [1] follows from the sixth equation. Indeed,
\[
1=\E\left[2(1-\pi_{\rho}(\nu Z_{1}))g_{t}'(H_{t})\right]=\frac{2}{\sqrt{2\pi\E[V_{t}^{2}]}}\int_{\mathbb{R}}e^{-\frac{h^{2}}{2\E[V_{t}^{2}]}}\E\left[1-\pi_{\rho}(\nu Z_{1})\mid H_{t}=h\right]g_{t}'(h)\dd h.
\]
Since $\E\left[1-\pi_\rho(\nu Z_1)\mid H_t=0\right]=\frac{1}{2}$ from symmetry, and $\int_{\mathbb R}g_t'(h)\dd h=1,$ multiplying the sixth equation by $\sqrt{\E[V_{t}^{2}]}$ gives
\[
\sqrt{\E[V_{t}^{2}]}-\frac{1}{\sqrt{2\pi}}=\frac{2}{\sqrt{2\pi}}\int_{\mathbb{R}}\left(e^{-\frac{h^{2}}{2\E[V_{t}^{2}]}}\E\left[1-\pi_{\rho}(\nu Z_{1})\mid H_{t}=h\right]-\frac{1}{2}\right)g_{t}'(h)\dd h.
\]
Moreover, by the opposite triangle inequality,
\[
\left|\sqrt{\E[V_{t}^{2}]}-\nu\theta_{t}\right|\leq\sqrt{\E[(V_{t}-m_{t}w^{\star})^{2}]}+\nu|m_{t}-\theta_{t}|.
\]
Also, $\E[\pi_{\rho}'(\nu Z_{1})\mathbf{1}_{\{H_{t}>0\}}]=\E[\pi_{\rho}'(\nu Z_{1})\mathbf{1}_{\{H_{t}<0\}}],$
because $\pi_\rho'(\nu Z_1)$ is even and
$(Z_1,Z_2)\stackrel{d}{=}(-Z_1,-Z_2)$.
Since $\sqrt{\E[V_{t}^{2}]}>0$, $\PP(H_t=0)=0$, and therefore
\[
2\E[\pi_{\rho}'(\nu Z_{1})\mathbf{1}_{\{H_{t}>0\}}]=\E[\pi_{\rho}'(\nu Z_{1})]=(1-2\rho)h(\nu).
\]
For [2], define $F_{t}(h)=e^{-\frac{h^2}{2\E[V_{t}^{2}]}}\E\left[1-\pi_\rho(\nu Z_1)\mid H_t=h\right].$ Therefore for a sufficiently small $s\leq\frac{\delta e^{-\frac{1}{2}}}
{66\sqrt{2\pi}(1+\nu)}$,
\begin{align*}
|F_{t}'(h)|\leq\frac{|h|}{\E[V_{t}^{2}]}e^{-\frac{h^{2}}{2\E[V_{t}^{2}]}}+\nu\frac{|\nu m_{t}|}{\E[V_{t}^{2}]}\sup_{u}\pi_{\rho}'(u)\leq\frac{e^{-\frac{1}{2}}}{\sqrt{\E[V_{t}^{2}]}}+\frac{\nu(1-2\rho)}{4\sqrt{\E[V_{t}^{2}]}}\leq\frac{1}{\frac{e^{-\frac{1}{2}}}{32\sqrt{2\pi}(1+\nu)}}\left(e^{-\frac{1}{2}}+\frac{\nu(1-2\rho)}{4}\right),
\end{align*}
where we used $|\nu m_t|\leq \sqrt{\E[V_{t}^{2}]}$ and
$\sup_u\pi_\rho'(u)=\frac{1-2\rho}{4}$.
Consequently,
\[
\left|\sqrt{\E[V_{t}^{2}]}-\frac{1}{\sqrt{2\pi}}\right|\leq\frac{64(1+\nu)}{e^{-\frac{1}{2}}}\left(e^{-\frac{1}{2}}+\frac{\nu(1-2\rho)}{4}\right)\int_{\mathbb{R}}|h|g_{t}'(h)\dd h.
\]

For [3], using the inverse relation
\[
h(a)=\frac{1}{\sigma_{t}\tau_{t}}\log\left(\frac{a}{1-a}\right)+\frac{\PP(V_{t}\neq0)}{\delta}a,
\]
we obtain
\begin{align*}
\int_{\mathbb{R}}|h|g_{t}'(h)\dd h
&
=\int_{0}^{1}\left|\frac{1}{\sigma_{t}\tau_{t}}\log\left(\frac{a}{1-a}\right)+\frac{\PP(V_{t}\neq0)}{\delta}a\right|\dd a\\&\leq\frac{1}{\sigma_{t}\tau_{t}}\int_{0}^{1}\left|\log\left(\frac{a}{1-a}\right)\right|\dd a+\frac{\PP(V_{t}\neq0)}{\delta}\int_{0}^{1}a\dd a=\frac{2\log2}{\sigma_{t}\tau_{t}}+\frac{\PP(V_{t}\neq0)}{2\delta}.
\end{align*}
Furthermore,
\begin{align*}
\E\left[\left|g_{t}(H_{t})-\mathbf{1}_{\{H_{t}>0\}}\right|\right]\leq\frac{1}{\sqrt{2\pi\E[V_{t}^{2}]}}\underbrace{\int_{\mathbb{R}}\left|g_{t}(h)-\mathbf{1}_{\{h>0\}}\right|\dd h}_{=\int_{\mathbb{R}}|h|g_{t}'(h)\dd h}\leq\frac{64(1+\nu)}{e^{-\frac{1}{2}}}\left(\frac{2\log2}{\sigma_{t}\tau_{t}}+\frac{\PP(V_{t}\neq0)}{2\delta}\right).
\end{align*}

Note that,
\[
\frac{2}{\sqrt{2\pi}\frac{e^{-\frac{1}{2}}}{64\sqrt{2\pi}(1+\nu)}}\left(e^{-\frac{1}{2}}+\frac{\nu(1-2\rho)}{4}\right)+\frac{\nu(1-2\rho)}{2\sqrt{2\pi}\frac{e^{-\frac{1}{2}}}{64\sqrt{2\pi}(1+\nu)}}=64e^{\frac{1}{2}}(1+\nu)\left(2e^{-\frac{1}{2}}+\nu(1-2\rho)\right).
\]
Thus \cref{eq:6} becomes
\[
\frac{1}{\sqrt{2\pi}}-\nu(1-2\rho)h(\nu)\leq 64e^{\frac{1}{2}}(1+\nu)\left(2e^{-\frac{1}{2}}+\nu(1-2\rho)\right)\left(\frac{2\log2}{\sigma_{t}\tau_{t}}+\frac{33s}{64\delta}\right)+2\sqrt{\frac{2s}{\delta}\left(1+2\log\frac{64}{s}\right)}.
\]
It is sufficient to decrease $s_0$ so that, for every $s\leq s_0$,
\[
64e^{\frac{1}{2}}(1+\nu)\left(2e^{-\frac{1}{2}}+\nu(1-2\rho)\right)\frac{33s}{64\delta}+2\sqrt{\frac{2s}{\delta}\left(1+2\log\frac{64}{s}\right)}\leq\frac{1}{2}\left(\frac{1}{\sqrt{2\pi}}-\nu(1-2\rho)h(\nu)\right)
\]
and $\frac{33s}{32\delta}\leq\frac{e^{-\frac{1}{2}}}{64\sqrt{2\pi}(1+\nu)}.$ Then,
\[
\frac{1}{2}\left(\frac{1}{\sqrt{2\pi}}-\nu(1-2\rho)h(\nu)\right)\leq64e^{\frac{1}{2}}(1+\nu)\left(2e^{-\frac{1}{2}}+\nu(1-2\rho)\right)\frac{2\log2}{\sigma_{t}\tau_{t}},
\]
which implies $\sigma_{t}\tau_{t}\leq\frac{128e^{\frac{1}{2}}(1+\nu)\left(2e^{-\frac{1}{2}}+\nu(1-2\rho)\right)\log2}{\frac{1}{\sqrt{2\pi}}-\nu(1-2\rho)h(\nu)}.$  Recall $h(\nu)=\frac{1}{\sqrt{2\pi}}\int_{\mathbb{R}}\phi''\left(\nu z\right)e^{-\frac{z^{2}}{2}}\dd z.$ We aim to upper bound $\frac{1}{\frac{1}{\sqrt{2\pi}}-\nu(1-2\rho)h(\nu)}$, note that \[\sqrt{2\pi}\left(\frac{1}{\sqrt{2\pi}}-\nu(1-2\rho)h(\nu)\right)=2\rho+\nu(1-2\rho)\int_{\mathbb{R}}\phi''\left(\nu z\right)\left(1-e^{-\frac{z^{2}}{2}}\right)\dd z\] because $\int_\R \phi''(u)\dd u =1$, and
\begin{align*}
\int_{\mathbb{R}}\phi''\left(u\right)\left(1-e^{-\frac{u^{2}}{2\nu^{2}}}\right)\dd u\geq\left(1-e^{-\frac{1}{2\nu^{2}}}\right)\int_{\left|u\right|\geq1}\phi''\left(u\right)\dd u\geq2\underbrace{\left(1-e^{-\frac{1}{2\nu^{2}}}\right)}_{\geq\frac{1}{1+2\nu^{2}}}\underbrace{\left(1-\phi'\left(1\right)\right)}_{=\frac{1}{1+e}},
\end{align*}
we can therefore take $Q=\frac{128\sqrt{2\pi}e^{\frac{1}{2}}(1+\nu)\left(2e^{-\frac{1}{2}}+\nu(1-2\rho)\right)\log2}{2\rho+\frac{2\left(1-2\rho\right)}{\left(1+e\right)\left(1+2\nu^{2}\right)}}.$

We now bound the term $\E[(V_t-\theta_tw^{\star})^2]$ for every $t\geq 2$. Since $\gamma_t=\frac{\sigma_t\tau_t\mathbb P(V_t\neq0)}{\delta}\leq Q\frac{33s}{32\delta},$ we get,
\begin{equation}
\label{eq:16}
\begin{aligned}
0<|\theta_{t}-(1-2\rho)h(\nu)|&=2\left|\E\left[\pi_{\rho}'(\nu Z_{1})\left(\phi'(\nu\alpha_{t}Z_{1}+\sigma_{t}Z_{2})-\phi'\left(\prox_{\gamma_{t}\phi}(\nu\alpha_{t}Z_{1}+\sigma_{t}Z_{2})\right)\right)\right]\right|\\&\underset{\phi''(u)\leq\frac{1}{4}}{\leq}\frac{1}{2}\E\left[\left|\nu\alpha_{t}Z_{1}+\sigma_{t}Z_{2}-\prox_{\gamma_{t}\phi}(\nu\alpha_{t}Z_{1}+\sigma_{t}Z_{2})\right|\right]|\\&\underset{\mathrm{\cref{eq:23}}}{=}\frac{\gamma_{t}}{2}\E\left[\phi'(\prox_{\gamma_{t}\phi}(\nu\alpha_{t}Z_{1}+\sigma_{t}Z_{2}))\right]\leq\frac{\gamma_{t}}{2}(1-2\rho)h(\nu)\end{aligned}\end{equation} where $2\E[\pi_\rho'(\nu Z_1)\phi'(\nu\alpha_tZ_1+\sigma_tZ_2))]=(1-2\rho)h(\nu)$ follows because $(Z_1,Z_2)\stackrel{d}{=}(-Z_1,-Z_2)$ and $\pi_\rho'(\nu Z_1)$ is even.
Explicitly,
\begin{equation}
\label{eq:24}
\begin{aligned}
2\E\left[\pi_{\rho}'\left(\nu Z_{1}\right)\phi'\left(\nu\alpha_{t}Z_{1}+\sigma_{t}Z_{2}\right)\right]&=\E\left[\pi_{\rho}'\left(\nu Z_{1}\right)\left(\phi'\left(\nu\alpha_{t}Z_{1}+\sigma_{t}Z_{2}\right)+\phi'\left(-\nu\alpha_{t}Z_{1}-\sigma_{t}Z_{2}\right)\right)\right]\\&=\E\left[\pi_{\rho}'\left(\nu Z_{1}\right)\right]=\left(1-2\rho\right)\E\left[\phi''\left(\nu Z_{1}\right)\right]=\left(1-2\rho\right)h\left(\nu\right),
\end{aligned}
\end{equation}
and by \cref{eq:22} $\theta_t=2\mathbb E[\pi_\rho'(\nu Z_1)\phi'(\prox_{\gamma_t\phi}\left(\nu\alpha_tZ_1+\sigma_tZ_2\right))].$
Therefore, we have $\theta_t=(1-2\rho)h(\nu)+O(s).$ So, there exists $s_0$, such that for every  $s\leq s_0$ and every $t$ we have $\theta_t\geq \frac{1}{2}(1-2\rho)h(\nu)=\underline{\theta}>0.$
We first refine the contribution of the active coordinates. From symmetry it is enough to consider $w^{\star}>0$.
Recall that, on $w^{\star}\neq 0$, $w^{\star}=\frac{\nu}{\sqrt{s}}G,$ with $G\sim N(0,1)$. Put $a=\frac{8\sqrt{\frac{2}{\delta}\log\left(\frac{64}{s}\right)}\sqrt{s}}{(1-2\rho)h(\nu)\nu}.$
The coordinates with $|G|<a$ have probability $\PP(|G|<a)=2\int_{0}^{a}\sqrt\frac{1}{2\pi}e^{-\frac{t^2}{2}}\dd t\leq\sqrt{\frac{2}{\pi}}a.$
Using the crude bound as both thresholds are at most $\frac{\lambda_0}{\epsilon_s}=\sqrt{\frac{2}{\delta}\log\left(\frac{64}{s}\right)},$
\[
\left|V_t-\theta_tw^{\star}\right|^2\leq2\frac{r_t^2}{\delta}Z_t^2+\left(2\sqrt{\frac{2}{\delta}\log\left(\frac{64}{s}\right)}\right)^2\leq\frac{2}{\delta}Z_t^2+\frac{8}{\delta}\log\left(\frac{64}{s}\right),
\]
therefore
\begin{align*}
s\E[(V_{t}-\theta_{t}w^{\star})^{2}\mathbf{1}_{\{w^{\star}\neq0,|G|<a\}}]\leq2s\sqrt{\frac{2}{\pi}}a\left(\frac{1}{\delta}+\frac{2}{\delta}\log\left(\frac{64}{s}\right)\right)=\frac{16s\sqrt{\frac{4s}{\delta\pi}\log\left(\frac{64}{s}\right)}}{(1-2\rho)h(\nu)\nu}\left(\frac{1}{\delta}+\frac{2}{\delta}\log\left(\frac{64}{s}\right)\right)=o(s)
\,.
\end{align*}
For $|G|\geq a$, in a 'good' event  (large enough to not destroy the positive teacher signal) $Z_{t-1}\geq -\frac{\underline{\theta} w^{\star}\sqrt{\delta}}{4},$ we want to find the threshold in the correct direction, in that case we analyze the case $w^{\star}=\frac{\nu}{\sqrt s}G\geq\frac{8\sqrt{\frac{2}{\delta}\log\left(\frac{64}{s}\right)}}{(1-2\rho)h(\nu)}$ and $Z_{t-1}\geq-\frac{\frac{1}{2}(1-2\rho)h(\nu)w^{\star}\sqrt{\delta}}{4}.$ Since $r_{t-1}\leq1$ and $\theta_{t-1}\geq\underline{\theta}$,
\[
\theta_{t-1}w^{\star}+\frac{r_{t-1}}{\sqrt{\delta}}Z_{t-1}\geq\underline{\theta}w^{\star}-\frac{\underline{\theta}}{4}w^{\star}=\frac38(1-2\rho)h(\nu)w^{\star}.
\]
The positive threshold at task $t-1$ is at most $\sqrt{\frac{2}{\delta}\log\left(\frac{64}{s}\right)}$, and hence
\[
V_{t-1}\geq\frac{3}{8}(1-2\rho)h(\nu)w^{\star}-\sqrt{\frac{2}{\delta}\log\left(\frac{64}{s}\right)}\geq\frac{3}{8}(1-2\rho)h(\nu)w^{\star}-\frac{1}{8}(1-2\rho)h(\nu)w^{\star}=\frac{1}{4}(1-2\rho)h(\nu)w^{\star}.
\]
Since $\sigma_{t-1}\tau_{t-1}\geq1$,
\[
w_{t-1}=\sigma_{t-1}\tau_{t-1}V_{t-1}\geq\frac{1}{4}(1-2\rho)h(\nu)w^{\star}.
\]
Consequently, the positive threshold at task $t$ satisfies
\[
\frac{\lambda_0}{[w_{t-1}]_++\epsilon_s}\leq\frac{2\lambda_0}{\frac{1}{2}(1-2\rho)h(\nu)w^{\star}}.
\]
Since  $\E[(V_{t}-\theta_{t}w^{\star})^{2}]\leq2\frac{r_{t}^{2}Z_{t}^{2}}{\delta}+2\left(\frac{\lambda_{0}}{\left[w_{t-1}\right]_{+}+\epsilon_{s}}\right)^{2}$. The second term is bounded by $\frac{32\lambda_0^2s}{(1-2\rho)^2h(\nu)^2\nu^2G^2}$ because an active coordinate occurs with probability $s$, its contribution is $\frac{32\lambda_0^2s^2}{(1-2\rho)^2h(\nu)^2\nu^2}\E\left[\frac{\mathbf{1}_{\{G\geq a\}}}{G^2}\right]$ at most, and 
\[
\E\left[\frac{\mathbf{1}_{\{|G|\geq a\}}}{G^2}\right]=2\int_a^\infty\frac{\varphi(g)}{g^2}\dd g\leq 2\frac{\varphi(0)}{a}\leq\sqrt{\frac{2}{\pi}}\frac{1}{a},
\]
this is $O\left(\frac{s^{\frac{3}{2}}}{\sqrt{\frac{2}{\delta}\log\left(\frac{64}{s}\right)}}\right)=o(s).$
It remains to control the other events. We are left with the event in which $|G|\geq a$ and $Z_{t-1}<-\frac{\underline{\theta}w^{\star}\sqrt{\delta}}{4}$, because $w^{\star}>0$, and $w^{\star}=\frac{\nu}{\sqrt{s}}G$ this event simply reduces to $G\geq a$ and $Z_{t-1}<-\frac{(1-2\rho)h(\nu)\nu\sqrt{\delta}G}{8\sqrt{s}}$, which we denote by $B_{t-1}$. For every fixed $b>0$, 
\[
\E\left[\Phi\left(-\frac{b|G|}{\sqrt{s}}\right)\right]=\frac{1}{\pi}\arctan\left(\frac{\sqrt{s}}{b}\right)\leq\frac{\sqrt{s}}{\pi b},
\]
So,
\begin{align*}
\mathbb{E}\left[(V_{t}-\theta_{t}w^{\star})^{2}\mathbf{1}_{\{w^{\star}\neq0\}}\mathbf{1}_{\{B_{t-1}\}}\right]&\le s\left(\frac{2}{\delta}+\frac{4}{\delta}\log\frac{64}{s}\right)\PP\left(G\ge a,Z_{t-1}<-\frac{bG}{\sqrt{s}}\middle|w^{\star}\neq0\right)\\&\le s\left(\frac{2}{\delta}+\frac{4}{\delta}\log\frac{64}{s}\right)\mathbb{E}\left[\Phi\left(-\frac{b|G|}{\sqrt{s}}\right)\right].
\end{align*}
We can drop $G \ge a$ because doing so only increases the probability.
Thus both the event contributes at most $O\left(s^{\frac{3}{2}}\log\left(\frac{64}{s}\right)\right)=o(s).$
The first term of the Gaussian noise on active coordinates contributes at most $\E\left[\frac{2r_t^2}{\delta}Z_t^2\mathbf{1}_{\{w^{\star}\neq 0\}}\right]\leq\frac{2s}{\delta}$. Therefore, after decreasing $s_0>0$ we can require that for all $s\leq s_0$, $o(s)\leq\frac{s}{2\delta}$, then,
\begin{align}
\E\left[(V_t-\theta_tw^{\star})^2\mathbf{1}_{\{w^{\star}\neq0\}}\right]\leq \frac{5s}{2\delta},
\label{eq:active_refined_Os}
\end{align}
for all $t\geq 2$. We now look at the null-coordinate contribution.
On $\{w^{\star}=0,w_{t-1}\neq0\}$, the previous argument already gives
\[
\E\left[(V_{t}-\theta_{t}w^{\star})^{2}\mathbf{1}_{\{w^{\star}=0,w_{t-1}\neq0\}}\right]=\E\left[V_{t}^{2}\mathbf{1}_{\{w^{\star}=0,w_{t-1}\neq0\}}\right]\underset{r_{t}\leq1}{\leq}\frac{1}{\delta}\E[Z_{t}^{2}]\underbrace{\PP(w^{\star}=0,w_{t-1}\neq0)}_{\leq\zeta_{t-1}\leq\frac{s}{32}}\leq\frac{s}{32\delta}.
\]
On $\{w^{\star}=0,w_{t-1}=0\}$, $V_t=\frac{r_t}{\sqrt{\delta}}\mathcal S_{\frac{\sqrt{2\log(\frac{64}{s})}}{r_t}}(Z_t).$ For $u>0$,
\begin{align*}
\E[\mathcal{S}_u(Z)^2]=\E[(|Z|-u)_+^2]&=2\int_u^\infty(z-u)^2\varphi(z)\dd z=2\varphi(u)\int_0^\infty y^2e^{-uy-\frac{y^2}{2}}\dd y\\&\leq2\varphi(u)\int_0^\infty y^2e^{-uy}\dd y=\frac{4\varphi(u)}{u^3}.
\end{align*}
Since $r_t\leq1$ (see conclusion beneath \cref{eq:34}), then $\frac{\sqrt{2\log(\frac{64}{s})}}{r_t}\geq\sqrt{2\log(\frac{64}{s})}.$ Consequently,
\begin{align*}
&\E\left[V_t^2\mathbf{1}_{\{w^{\star}=0,w_{t-1}=0\}}\right]\leq
\frac{4}{\delta}\frac{\varphi\left(\sqrt{2\log(\frac{64}{s})}\right)}{\left(2\log(\frac{64}{s})\right)^{\frac{3}{2}}}=\frac{4}{\delta}\frac{\frac{1}{\sqrt{2\pi}}e^{-\log(\frac{64}{s})}}{\left(2\log(\frac{64}{s})\right)^{\frac{3}{2}}}=\frac{s}{16\sqrt{2\pi}\delta\left(2\log(\frac{64}{s})\right)^{\frac{3}{2}}}=o(s).
\end{align*}
Combining the active and null contributions gives, after decreasing $s_0>0$ we can require that for all $s\leq s_0$, $o(s)\leq\frac{s}{32\delta}$, and then, uniformly in $t\geq2$,
\begin{align}
\E[(V_t-\theta_tw^{\star})^2]
\leq\left(\frac{5}{2}+\frac{1}{32}\right)\frac{s}{\delta}=\frac{81}{32}\frac{s}{\delta}<\frac{3s}{\delta}.
\label{eq:refined_normalized_channel}
\end{align}
Consequently, since $m_t-\theta_t=\frac{\E[w^{\star} V_t]}{\nu^2}-\theta_t=\frac{1}{\nu^2}\E[w^{\star}(V_t-\theta_tw^{\star})]$, we have by Cauchy-Schwartz,
\[
|m_t-\theta_t|\leq \frac{1}{\nu^2}\sqrt{\E[{w^{\star}}^2]}\sqrt{\E[(V_t-\theta_t w^{\star})^2]}=\frac{1}{\nu}\sqrt{\E[(V_t-\theta_tw^{\star})^2]}\leq\sqrt{\frac{3s}{\nu^2\delta}},
\]
and, combining both cases and because $m_tw^{\star}$ is the $L^2$-projection of $V_t$ onto
$w^{\star}$\footnote{It follows directly by noting that $\E[(V_t-a w^{\star})^2]=\E[(V_t)^2]-2a\E[V_t w^{\star}]+a^2\nu^2$, first order stationary condition would enforce $a=\frac{\E[V_t w^{\star}]}{\nu^2}=m_t$.},
\[
\mathbb E[V_t^2]-\nu^2m_t^2\leq \frac{3s}{\delta}.
\]
The estimates now imply from the first equation,
\[
|\alpha_t-\sigma_t\tau_t\theta_t|=\sigma_t\tau_t|m_t-\theta_t|\leq Q\sqrt{\frac{3s}{\nu^2\delta}}
\]
and from the third equation,
\[
\sigma_t^2=(\sigma_t\tau_t)^2\left(\mathbb E[V_t^2]-\nu^2m_t^2\right)\leq \frac{3s}{\delta}Q^2.
\]
Also, $\gamma_t=\frac{\sigma_t\tau_t\mathbb P(V_t\neq0)}{\delta}\leq Q\frac{33s}{32\delta}.$
Put $r_t=\sqrt{\nu^2\alpha_t^2+\sigma_t^2}.$ The proximal identity and the bounded derivatives of $\phi'$ and $\phi''$ with \cref{eq:24} give $\theta_t=(1-2\rho)h(\nu)+O(\gamma_t)$
and from symmetry \[2\E[(1-\pi_\rho(\nu Z_1))\phi''(\nu\alpha_tZ_1+\sigma_tZ_2)]=\E[\phi''(\nu\alpha_tZ_1+\sigma_tZ_2)]],\] and $\frac{1}{\sigma_t\tau_t}=h\left(\sqrt{\E[V_{t}^2}]\right)+O(\gamma_t).$
Since $\frac{\alpha_t}{\sigma_t\tau_t}=m_t$, it follows that
\begin{align*}
&\left|\alpha_{t}h\left(\sqrt{\nu^2\alpha_t^2+\sigma_t^2}\right)-(1-2\rho)h(\nu)\right|\\&\leq\alpha_{t}\left|h\left(\sqrt{\nu^2\alpha_t^2+\sigma_t^2}\right)-\frac{1}{\sigma_t\tau_t}\right|+\left|\frac{\alpha_{t}}{\sigma_t\tau_t}-\theta_{t}\right|+\left|\theta_{t}-(1-2\rho)h(\nu)\right|\\&=\alpha_{t}\left|h\left(\sqrt{\nu^2\alpha_t^2+\sigma_t^2}\right)-\frac{1}{\sigma_t\tau_t}\right|+|m_{t}-\theta_{t}|+\left|\theta_{t}-(1-2\rho)h(\nu)\right|\\&\stackrel{[1]}{=}\alpha_{t}\left|\E\left[2(1-\pi_{\rho}(\nu Z_{1}))\phi''\left(\nu\alpha_{t}Z_{1}+\sigma_{t}Z_{2}\right)\right]
-\E\left[\frac{2(1-\pi_{\rho}(\nu Z_{1}))\phi''\left(\prox_{\gamma_{t}\phi}(\nu\alpha_{t}Z_{1}+\sigma_{t}Z_{2})\right)}{1+\gamma_{t}\phi''\left(\prox_{\gamma_{t}\phi}(\nu\alpha_{t}Z_{1}+\sigma_{t}Z_{2})\right)}\right]\right|\\&\quad+|m_{t}-\theta_{t}|+\left|2\E\left[\pi_{\rho}'(\nu Z_{1})\phi'\left(\prox_{\gamma_{t}\phi}(\nu\alpha_{t}Z_{1}+\sigma_{t}Z_{2})\right)\right]-(1-2\rho)h(\nu)\right|\\&\stackrel{[2]}{\leq}\alpha_{t}\E\left[2(1-\pi_{\rho}(\nu Z_{1}))\left|\phi''(\nu\alpha_{t}Z_{1}+\sigma_{t}Z_{2})-\phi''\left(\prox_{\gamma_{t}\phi}(\nu\alpha_{t}Z_{1}+\sigma_{t}Z_{2})\right)\right|\right]
\\&\quad+\alpha_{t}\E\left[2(1-\pi_{\rho}(\nu Z_{1}))\phi''\left(\prox_{\gamma_{t}\phi}(\nu\alpha_{t}Z_{1}+\sigma_{t}Z_{2})\right)\left|1-\frac{1}{1+\gamma_{t}\phi''\left(\prox_{\gamma_{t}\phi}(\nu\alpha_{t}Z_{1}+\sigma_{t}Z_{2})\right)}\right|\right]\\&\quad+|m_{t}-\theta_{t}|+{2\E\left[\pi_{\rho}'(\nu Z_{1})\left|\phi'\left(\prox_{\gamma_{t}\phi}(\nu\alpha_{t}Z_{1}+\sigma_{t}Z_{2})\right)-\phi'(\nu\alpha_{t}Z_{1}+\sigma_{t}Z_{2})\right|\right]}\\&\stackrel{[3]}{\leq}\frac{\alpha_{t}\gamma_{t}}{6\sqrt{3}}+\frac{\alpha_{t}\gamma_{t}}{16}+|m_{t}-\theta_{t}|+\frac{\gamma_{t}}{2}\E[\pi_{\rho}'(\nu Z_{1})]\\&=\alpha_{t}\gamma_{t}\left(\frac{1}{6\sqrt{3}}+\frac{1}{16}\right)+|m_{t}-\theta_{t}|+\frac{\gamma_{t}}{2}(1-2\rho)h(\nu).
\end{align*}
Here, [1] follows from the sixth scalar equation
\[
\frac{1}{\sigma_t\tau_t}=\E\left[
\frac{2(1-\pi_\rho(\nu Z_1))\phi''\left(\prox_{\gamma_t\phi}(\nu\alpha_tZ_1+\sigma_tZ_2)\right)}{1+\gamma_t\phi''\left(\prox_{\gamma_t\phi}(\nu\alpha_tZ_1+\sigma_tZ_2)\right)}\right],
\]
together with
\begin{align*}
2\E\left[(1-\pi_\rho(\nu Z_1))\phi''(\nu\alpha_tZ_1+\sigma_tZ_2)\right]=\E\left[\phi''(\nu\alpha_tZ_1+\sigma_tZ_2)\right]=h\left(\sqrt{\nu^2\alpha_t^2+\sigma_t^2}\right).
\end{align*}
and by \cref{eq:22} $\theta_t=2\mathbb E[\pi_\rho'(\nu Z_1)\phi'(\prox_{\gamma_t\phi}\left(\nu\alpha_tZ_1+\sigma_tZ_2\right))].$ Furthermore, since $\pi_\rho'(u)$ is even, $\phi'(u)+\phi'(-u)=1,$
\begin{align*}
2\E\left[\pi_{\rho}'(\nu Z_{1})\phi'(\nu\alpha_{t}Z_{1}+\sigma_{t}Z_{2})\right]=\E[\pi_{\rho}'(\nu Z_{1})]=(1-2\rho)h(\nu).
\end{align*}
This gives [2]. For [3], since $0<\phi'(u)<1,$ and the proximal optimality give,
\[
\left|\nu\alpha_tZ_1+\sigma_tZ_2-\prox_{\gamma_t\phi}(\nu\alpha_tZ_1+\sigma_tZ_2)\right|=\left|\gamma_t\phi'\left(\prox_{\gamma_t\phi}(\nu\alpha_tZ_1+\sigma_tZ_2)\right)\right|\leq\gamma_t.
\]
Since $\sup_{u\in\mathbb R}|\phi'''(u)|=\frac{1}{6\sqrt{3}},$ so the function $\phi''(u)$ is $\frac{1}{6\sqrt{3}}$-Lipschitz and,
\begin{align*}
\left|\phi''(\nu\alpha_tZ_1+\sigma_tZ_2)-\phi''\left(\prox_{\gamma_t\phi}(\nu\alpha_tZ_1+\sigma_tZ_2)\right)\right|\leq\frac{\gamma_t}{6\sqrt{3}}.
\end{align*}
Moreover, $2\E[1-\pi_\rho(\nu Z_1)]=1,$
and hence
\[
\alpha_t\E\left[2(1-\pi_\rho(\nu Z_1))\left|\phi''(\nu\alpha_tZ_1+\sigma_tZ_2)-
\phi''\left(\prox_{\gamma_t\phi}(\nu\alpha_tZ_1+\sigma_tZ_2)\right)\right|\right]\leq\frac{\alpha_t\gamma_t}{6\sqrt{3}}.
\]
For the second term, since $0<\phi''(u)\leq\frac{1}{4},$
we have
\[
\phi''\left(\prox_{\gamma_{t}\phi}(\nu\alpha_{t}Z_{1}+\sigma_{t}Z_{2})\right)\left|1-\frac{1}{1+\gamma_{t}\phi''\left(\prox_{\gamma_{t}\phi}(\nu\alpha_{t}Z_{1}+\sigma_{t}Z_{2})\right)}\right|=\frac{\gamma_{t}\phi''\left(\prox_{\gamma_{t}\phi}(\nu\alpha_{t}Z_{1}+\sigma_{t}Z_{2})\right)^{2}}{1+\gamma_{t}\phi''\left(\prox_{\gamma_{t}\phi}(\nu\alpha_{t}Z_{1}+\sigma_{t}Z_{2})\right)}\leq\frac{\gamma_{t}}{16}.
\]
Therefore
\begin{align*}
&\alpha_t
\E\left[2(1-\pi_\rho(\nu Z_1))\phi''\left(\prox_{\gamma_t\phi}(\nu\alpha_tZ_1+\sigma_tZ_2)\right)\left|1-\frac{1}{1+\gamma_t\phi''\left(\prox_{\gamma_t\phi}(\nu\alpha_tZ_1+\sigma_tZ_2)\right)}\right|\right]\leq\frac{\alpha_t\gamma_t}{16}.
\end{align*}
Finally, since $\sup_{u\in\mathbb R}\phi''(u)=\frac{1}{4},$ which implies that $\phi'(u)$ is $\frac{1}{4}$-Lipschitz, we have,
\begin{align*}
&2\E\left[\pi_\rho'(\nu Z_1)\left|\phi'\left(\prox_{\gamma_t\phi}(\nu\alpha_tZ_1+\sigma_tZ_2)\right)-\phi'(\nu\alpha_tZ_1+\sigma_tZ_2)\right|\right]\\&
\leq \frac{1}{2}\E\left[\pi_\rho'(\nu Z_1)\left|\prox_{\gamma_t\phi}(\nu\alpha_tZ_1+\sigma_tZ_2)-(\nu\alpha_tZ_1+\sigma_tZ_2)\right|\right]\leq \frac{\gamma_t}{2}\E[\pi_\rho'(\nu Z_1)]=\frac{\gamma_t}{2}(1-2\rho)h(\nu).
\end{align*}
Consequently,
\[
\left|\alpha_{t}h\left(\sqrt{\nu^2\alpha_t^2+\sigma_t^2}\right)-(1-2\rho)h(\nu)\right|\leq\alpha_{t}\gamma_{t}\left(\frac{1}{6\sqrt{3}}+\frac{1}{16}\right)+|m_{t}-\theta_{t}|+\frac{\gamma_{t}}{2}(1-2\rho)h(\nu).
\]
Assume $s$ is sufficiently small so that
\[
\frac{3s}{\delta}\leq\min\left\{1,
\frac{16}{33Q},\frac{(1-2\rho)^2\nu^2h(\nu)^2}{16Q^2}
\right\}.
\]
Since $\gamma_t\leq\frac{33Q}{32}\frac{s}{\delta},$
we have $\gamma_t\leq\frac{1}{2}.$ Therefore
\begin{equation}
\label{eq:36}
\begin{aligned}
\left|\theta_{t}-(1-2\rho)h(\nu)\right|&\leq2\E\left[\pi_{\rho}'(\nu Z_{1})\left|\phi'(\prox_{\gamma_{t}\phi}(\nu\alpha_{t}Z_{1}+\sigma_{t}Z_{2}))-\phi'(\nu\alpha_{t}Z_{1}+\sigma_{t}Z_{2})\right|\right]\\&\leq\frac{\gamma_{t}}{2}(1-2\rho)h(\nu)\leq\frac{1}{4}(1-2\rho)h(\nu).
\end{aligned}
\end{equation}
Moreover, since $\alpha_t-\sigma_t\tau_t\theta_t=\sigma_t\tau_t(m_t-\theta_t),$ and $\sigma_t\tau_t\geq1$, the previously established bound $|\alpha_t-\sigma_t\tau_t\theta_t|\leq Q\sqrt{\frac{3s}{\nu^2\delta}}$
implies $|m_t-\theta_t|\leq Q\sqrt{\frac{3s}{\nu^2\delta}}\leq\frac{1}{4}(1-2\rho)h(\nu).$ Consequently, $\frac{1}{2}(1-2\rho)h(\nu)\leq m_t\leq\frac{3}{2}(1-2\rho)h(\nu).$ Since $1\leq\sigma_t\tau_t\leq Q$ and $\alpha_t=\sigma_t\tau_t m_t,$
we obtain $\frac{1}{2}(1-2\rho)h(\nu)\leq\alpha_t\leq\frac{3}{2}Q(1-2\rho)h(\nu).$
Using the bound obtained above,
\begin{align*}
&\left|
\alpha_th\left(
\sqrt{\nu^2\alpha_t^2+\sigma_t^2}\right)-(1-2\rho)h(\nu)\right|\\&\leq\alpha_t\gamma_t\left(\frac{1}{6\sqrt3}+\frac{1}{16}\right)+|m_t-\theta_t|+\frac{\gamma_t}{2}(1-2\rho)h(\nu)\\&\leq \sqrt{
\frac{3Q^2s}{\nu^2\delta}
}+\left[\frac{99Q^2(1-2\rho)h(\nu)}{64}\left(\frac{1}{6\sqrt3}+\frac{1}{16}\right)+\frac{33Q(1-2\rho)h(\nu)}{64}\right]\frac{3s}{\delta}.
\end{align*}

Furthermore,
\begin{align*}
&\sqrt{\nu^2\alpha_t^2+\sigma_t^2}-\nu\alpha_t=\frac{\sigma_t^2}{
\sqrt{\nu^2\alpha_t^2+\sigma_t^2}+\nu\alpha_t}\stackrel{[4]}{\leq}\frac{\sigma_t^2}{\nu(1-2\rho)h(\nu)}\stackrel{[5]}{\leq}
\frac{Q^2}{\nu(1-2\rho)h(\nu)}\frac{3s}{\delta}.
\end{align*}
Here, [4] follows from
\[
\sqrt{\nu^2\alpha_t^2+\sigma_t^2}+\nu\alpha_t\geq2\nu\alpha_t\geq\nu(1-2\rho)h(\nu),
\]
and [5] follows from $\sigma_t^2\leq Q^2\frac{3s}{\delta}.$
Since $h'(a)=\E[Z\phi'''(aZ)]$ and $\sup_{u\in\mathbb R}|\phi'''(u)|=\frac{1}{6\sqrt3},$ we have $|h'(a)|\leq\sqrt{\frac{2}{\pi}}\frac{1}{6\sqrt3}.$ Hence,
\begin{align*}
&\alpha_{t}\left|h(\nu\alpha_{t})-h\left(\sqrt{\nu^{2}\alpha_{t}^{2}+\sigma_{t}^{2}}\right)\right|\leq\alpha_{t}\sqrt{\frac{2}{\pi}}\frac{1}{6\sqrt{3}}\left(\sqrt{\nu^{2}\alpha_{t}^{2}+\sigma_{t}^{2}}-\nu\alpha_{t}\right)\\&\leq\frac{3}{2}Q(1-2\rho)h(\nu)\sqrt{\frac{2}{\pi}}\frac{1}{6\sqrt{3}}\frac{Q^{2}}{\nu(1-2\rho)h(\nu)}\frac{3s}{\delta}=\frac{Q^{3}}{4\nu}\sqrt{\frac{2}{3\pi}}\frac{3s}{\delta}.
\end{align*}
Combining the last two estimates,
\begin{align*}
&\left|\alpha_{t}h(\nu\alpha_{t})-(1-2\rho)h(\nu)\right|
\leq \sqrt{\frac{3sQ^2}{\delta\nu^2}}+\left[\frac{99Q^{2}(1-2\rho)h(\nu)}{64}\left(\frac{1}{6\sqrt{3}}+\frac{1}{16}\right)+\frac{33Q(1-2\rho)h(\nu)}{64}+\frac{Q^{3}}{4\nu}\sqrt{\frac{2}{3\pi}}\right]\frac{3s}{\delta},
\end{align*}
We aim to use the mean-value theorem to bound the quantity $|\alpha_t-r|$, we need to find in which region $\alpha_t$ is located. If we decrease $s_0$ so that $Q^3\frac{s}{\delta}\leq C_1\sqrt{\frac{s}{\delta}}$ for all $s\leq s_0$, then, for some $C_2>0$, $\alpha_{t}h(\nu\alpha_{t})\leq(1-2\rho)h(\nu)+\frac{C_2}{\nu}\sqrt{\frac{s}{\delta}}$, if also $s_0$ is small so that $\frac{C_2}{\nu}\sqrt{\frac{s}{\delta}}\leq \frac{\frac{1}{\sqrt{2\pi}}-\nu(1-2\rho)h(\nu)}{2}$ then $\left|\alpha_{t}h(\nu\alpha_{t})-(1-2\rho)h(\nu)\right|\leq\frac{\frac{1}{\sqrt{2\pi}}-\nu(1-2\rho)h(\nu)}{2\nu}$, also,
\begin{align*}
\frac{1}{\nu\sqrt{2\pi}}-\alpha_{t}h(\nu\alpha_{t})&=\frac{1}{\nu\sqrt{2\pi}}\int_{\R}\phi''(u)\left(1-e^{-\frac{u^{2}}{2\nu^{2}\alpha_t^{2}}}\right)\dd u\\&\leq\frac{1}{\nu\sqrt{2\pi}}\frac{1}{2\nu^{2}\alpha_t^{2}}\int_{\R}u^{2}\phi''(u)\dd u
=\frac{\pi^{2}}{6\sqrt{2\pi}\nu^{3}\alpha_t^{2}},
\end{align*}
$\frac{\frac{1}{\sqrt{2\pi}}-\nu(1-2\rho)h(\nu)}{2\nu}\leq\frac{\pi^{2}}{6\sqrt{2\pi}\nu^{3}\alpha_{t}^{2}}$ implies $\alpha_{t}\leq\sqrt{\frac{2\nu\pi^{2}}{6\sqrt{2\pi}\nu^{3}\left(\frac{1}{\sqrt{2\pi}}-\nu(1-2\rho)h(\nu)\right)}}$ as $\frac{1}{\sqrt{2\pi}\left(\frac{1}{\sqrt{2\pi}}-\nu(1-2\rho)h(\nu)\right)}\leq \frac{1}{2\rho+\frac{2\left(1-2\rho\right)}{\left(1+e\right)\left(1+2\nu^{2}\right)}}$. 
As function of $\rho$, $2\rho+\frac{2\left(1-2\rho\right)}{\left(1+e\right)\left(1+2\nu^{2}\right)}$ is increasing indeed taking the derivative with respect to $\rho$ yields that, $2-\frac{4}{(1+e)(1+2\nu^2)}>0$, then $\frac{1}{\sqrt{2\pi}\left(\frac{1}{\sqrt{2\pi}}-\nu(1-2\rho)h(\nu)\right)}\leq \frac{\left(1+e\right)\left(1+2\nu^{2}\right)}{2}$ so $\alpha_{t}\leq\sqrt{\frac{\nu\pi^{2}}{6\nu^{3}}\left(1+e\right)\left(1+2\nu^{2}\right)}$, and since $\nu r\leq \nu$, thus every $a$ between $\alpha_t$ and $r$ satisfies $\nu a<\sqrt{\frac{\pi^{2}}{6}\left(1+e\right)\left(1+2\nu^{2}\right)}$.
Now $\frac{\dd}{\dd a}\left(ah(\nu a)\right)=\E[Z^2\phi''(\nu aZ)],$ therefore we have for all $0<b\leq \min\left\{1,\frac{1}{\nu a}\right\}$, 
\begin{align}
\label{eq:14}
\E[Z^{2}\phi''(\nu aZ)]\stackrel{[6]}{\geq}\E\left[Z^{2}\mathbf{1}_{\left\{ |Z|\leq b\right\} }\right]\phi''\left(1\right)\stackrel{[7]}{\geq}\frac{2\varphi\left(1\right)}{3}b^{3}\phi''\left(1\right)
\end{align}
Here, [6] follows from $\phi''(u)=\frac{1}{4\cosh^2\left(\frac{u}{2}\right)}$ (which is ascending in $\R^{-}$ and descending in $\R^{+}$) therefore on $\left\{ |Z|\leq b\right\} $ we have $\phi''(\nu a Z)>\phi''(1)$,
since $|\nu aZ|\leq1.$
Also [7] follows since for all $a\leq 1$,
\[\E\left[Z^{2}\mathbf{1}_{\{|Z|\leq a\}}\right]=2\int_{0}^{a}z^{2}\varphi\left(z\right)\geq\frac{2\varphi\left(1\right)}{3}a^{3}.
\]
Since $rh(\nu r)=(1-2\rho)h(\nu),$ with
$b=\min\left\{ 1,\frac{1}{\max\left\{ \sqrt{\frac{\pi^{2}}{6}\left(1+e\right)\left(1+2\nu^{2}\right)},\nu r\right\} }\right\},$
the mean-value theorem gives
\begin{align*}
|\alpha_{t}-r|&\leq\frac{3}{2\varphi(1)\phi''(1)}\max\left\{ 1,\left(\sqrt{\frac{\pi^{2}}{6}\left(1+e\right)\left(1+2\nu^{2}\right)}\right)^{3}\right\} \\&\quad\left[\sqrt{\frac{3sQ^{2}}{\delta\nu}}+\left[\frac{99Q^{2}(1-2\rho)h(\nu)}{64}\left(\frac{1}{6\sqrt{3}}+\frac{1}{16}\right)+\frac{33Q(1-2\rho)h(\nu)}{64}+\frac{Q^{3}}{4\nu}\sqrt{\frac{2}{3\pi}}\right]\frac{3s}{\delta}\right]
\end{align*}
Finally since by \cref{eq:7},
\[
\E[(w_{t}-rw^{\star})^2]=\nu^2(\alpha_t-r)^2+\sigma_t^2,
\]
and therefore using \cref{prop:coefficient_error_to_excess_risk},
\begin{align*}
\mathcal{L}(\mathbf{w}_{T})-\inf_{\vw}\mathcal{L}(\vw)&\leq\frac{3Q^{2}}{8}\frac{s}{\delta}+\frac{\nu^{2}}{8}\left[\frac{3}{2\varphi(1)\phi''(1)}\right]^{2}\max\left\{ 1,\left(\sqrt{\frac{\pi^{2}}{6}\left(1+e\right)\left(1+2\nu^{2}\right)}\right)^{6}\right\} \\&\left[\sqrt{\frac{3sQ^{2}}{\delta\nu^{2}}}+\left[\frac{99Q^{2}(1-2\rho)h(\nu)}{64}\left(\frac{1}{6\sqrt{3}}+\frac{1}{16}\right)+\frac{33Q(1-2\rho)h(\nu)}{64}+\frac{Q^{3}}{4\nu}\sqrt{\frac{2}{3\pi}}\right]\frac{3s}{\delta}\right]^{2}.
\end{align*}
with $Q=\frac{128\sqrt{2\pi}e^{\frac{1}{2}}(1+\nu)\left(2e^{-\frac{1}{2}}+\nu(1-2\rho)\right)\log2}{2\rho+\frac{2\left(1-2\rho\right)}{\left(1+e\right)\left(1+2\nu^{2}\right)}}.$
Note that, $(1-2\rho)h(\nu)\leq\frac{1}{4}$ and $\nu(1-2\rho)h(\nu)\leq \frac{1}{\sqrt{2\pi}}$ since $r\leq 1$, overall there exists a universal constant $C>0$ such that
\[
\mathcal{L}(\mathbf w_T)-\inf_{\vw} \mathcal{L}(\vw)\leq C(1+\nu)^{8}\frac{s}{\delta}
\]
Below, we summarize all conditions used to make $s$ sufficiently small. 

For some $C_1,C_2,C_3,C_4,C_5,C_6>0$,
\begin{equation}
\begin{aligned}
&\frac{33s}{32\delta}\leq\frac{C_{1}}{1+\nu},\\&\frac{s}{\delta}\leq C_{2}\frac{\frac{1}{\sqrt{2\pi}}-\nu(1-2\rho)h(\nu)}{(1+\nu)^{2}},\\&\frac{s}{\delta}\left(1+2\log\frac{64}{s}\right)\leq C_{3}\left(\frac{1}{\sqrt{2\pi}}-\nu(1-2\rho)h(\nu)\right)^{2},\\&\frac{s}{\delta}\log\frac{64}{s}\left(1+2\log\frac{64}{s}\right)^{2}\leq C_{4}(1-2\rho)^{2}h(\nu)^{2}\nu^{2},\\&\frac{s}{\delta}\leq\frac{C_{5}}{Q^{6}}.
\end{aligned}
\label{eq:s0_conditions}
\end{equation}
Using 
\begin{align}
&\frac{1}{\sqrt{2\pi}}-\nu(1-2\rho)h(\nu)\gtrsim\frac{1}{(1+\nu)^{2}},\notag&\\&h(\nu)=\E[\phi''(\nu Z)]\geq\phi''(1)\PP\left(|Z|\leq\min\left\{1,\frac{1}{\nu}\right\}\right)\geq2\phi''(1)\varphi(1)\min\left\{1,\frac{1}{\nu}\right\} \geq\frac{2\phi''(1)\varphi(1)}{1+\nu},\label{eq:28}
\end{align} and, since $\rho\leq \frac{1}{2}$, $Q\lesssim (1+\nu)^4,$ the conditions in \cref{eq:s0_conditions} are satisfied provided \[s_0\leq c \min\left\{\frac{\delta} {(1+\nu)^4\log_+\left(\frac{1+\nu}{\delta}\right)}, \frac{\delta(1-2\rho)^2\nu^2} {(1+\nu)^2 \log_+^3\left( \frac{(1+\nu)^2} {\delta(1-2\rho)^2\nu^2}\right)}, \frac{\delta}{(1+\nu)^{24}} \right\}, \]
for some universal constant $c>0$.

\end{proof}

\newpage
\begin{lemma}
\label{lem:4}
For all $x\ge 0$, we have 
\[
\frac{\Phi(-x)}{\varphi(x)}\geq\frac{1}{x+1}
\]
\end{lemma}

\begin{proof}
Define $F(x)=\Phi(-x)-\frac{\varphi(x)}{x+1}.$
First, we evaluate the function at the boundary $x=0$:
\[
F(0)=\frac{1}{2}-\frac{1}{\sqrt{2\pi}}>0
\]
Next, we compute the derivative $F'(x)$ using the quotient rule. Recall that $\Phi'(-x)=-\varphi(x)$ and $\varphi'(x)=-x\varphi(x)$:
\[
F'(x)=-\varphi(x)-\frac{\varphi'(x)(x+1)-\varphi(x)}{(x+1)^2}
\]
Substituting $\varphi'(x)=-x\varphi(x)$ into the numerator:
\begin{align*}
F'(x)&=-\varphi(x)-\frac{-x\varphi(x)(x+1)-\varphi(x)}{(x+1)^{2}}
=-\varphi(x)+\frac{\varphi(x)[x(x+1)+1]}{(x+1)^{2}}\\&=\varphi(x)\left[-1+\frac{x^{2}+x+1}{(x+1)^{2}}\right]
=\varphi(x)\left[\frac{-(x^{2}+2x+1)+(x^{2}+x+1)}{(x+1)^{2}}\right]\\&=-\frac{x}{(x+1)^{2}}\varphi(x)
\,.
\end{align*}
Therefore, $F'(x) \leq 0$ for all $x \geq 0$, meaning $F(x)$ is monotonically decreasing on this interval.
Finally, we examine the asymptotic behavior as $x \to \infty$:
\[
\lim_{x \to \infty} F(x) = \lim_{x \to \infty} \Phi(-x) - \lim_{x \to \infty} \frac{\varphi(x)}{x+1} = 0 - 0 = 0
\]
Since $F(0) > 0$, and $F(x)$ decreases monotonically toward $0$ as $x \to \infty$, it must remain non-negative over the entire positive domain.
\end{proof}
\begin{lemma}[Shifted Gaussian tail bound]
\label{lem:5}
For every $x,q\in\mathbb R$ and every $a\in(0,1)$,
\[
\Phi(-(x-q))\leq\exp\left(\frac{aq^2}{2(1-a)}\right)\Phi(-x)^a.
\]
\end{lemma}

\begin{proof}
Let $Z\sim\mathcal N(0,1)$. By a change of variables,
\begin{align*}
\Phi(-(x-q))&=\int_{x-q}^{\infty}\varphi(z)\dd z=\int_x^\infty\varphi(u-q)\dd u
=\int_x^\infty \exp\left(qu-\frac{q^2}{2}\right)\varphi(u)\dd u=\E\left[\exp\left(qZ-\frac{q^2}{2}\right)\mathbf{1}_{\{Z\geq x\}}\right].
\end{align*}
Applying H\"older's inequality with exponents
$\frac{1}{1-a}$ and $\frac{1}{a}$ gives
\begin{align*}
\Phi(-(x-q))&\leq\left(\E\exp\left(\frac{qZ-\frac{q^2}{2}}{1-a}\right)\right)^{1-a}\PP(Z\geq x)^a
=\exp\left(\frac{aq^2}{2(1-a)}\right)\Phi(-x)^a,
\end{align*}
where we used $\E[\exp\left(tZ\right)]=\exp\left(\frac{t^2}{2}\right)$.
\end{proof}
\begin{lemma}[Null-coordinate excursion bound]
\label{lem:6}
Let $T\geq2.$
Suppose that
\[
\prod_{j=1}^T\left[1-2\Phi\left(-\left(1-\frac{1}{8\log\left(2T\right)}\right)\frac{\lambda\sqrt\delta}{r_j}\right)\right]\leq\frac{1}{2}.
\]
Then
\[
\nu^2(\alpha_T-r)^2+\sigma_T^2\geq\frac{(1-s)\lambda^2}{4096\log^2(2T)}.
\]
\end{lemma}

\begin{proof}
Condition on $w^{\star}=0$.  We write for convinience, 
\begin{align*}
a_j=\Phi\left(-\left(1-\frac{1}{8\log\left(2T\right)}\right)\frac{\lambda\sqrt\delta}{r_j}\right),
\\
b_j=\Phi\left(-\left(1+\frac{1}{8\log\left(2T\right)}\right)\frac{\lambda\sqrt\delta}{r_j}\right).
\end{align*}
For $j=1,\ldots,T$, define
\[
A_j^+=\left\{Z_j\geq\left(1+\frac{1}{8\log\left(2T\right)}\right)\frac{\lambda\sqrt\delta}{r_j},\quad|Z_k|\leq\left(1-\frac{1}{8\log\left(2T\right)}\right)\lambda\frac{\sqrt{\delta}}{r_{k}}\text{ for every }k>j
\right\},
\]
and similarly
\[
A_j^-=\left\{Z_j\leq-\left(1+\frac{1}{8\log\left(2T\right)}\right)\frac{\lambda\sqrt\delta}{r_j},\quad|Z_k|\leq\left(1-\frac{1}{8\log\left(2T\right)}\right)\lambda\frac{\sqrt{\delta}}{r_{k}}\text{ for every }k>j
\right\}.
\]
The events $\{A_j^+,A_j^-:1\leq j\leq T\}$ are pairwise disjoint. We first show $A_j^+\subseteq\{w_{t}\geq\frac{\lambda}{8\log\left(2T\right)}\}.$ Indeed, at task $j$ the temporal L1 projection interval (see \cref{lem:2}) is $\left[\sigma_j\tau_j\left(\frac{r_j}{\sqrt\delta}Z_j-\lambda\right),\sigma_j\tau_j\left(\frac{r_j}{\sqrt\delta}Z_j+\lambda\right)\right].$
On $A_j^+$, $\frac{r_j}{\sqrt\delta}Z_j\geq\left(1+\frac{1}{8\log\left(2T\right)}\right)\lambda,$
and hence the lower endpoint is at least $(\sigma_j\tau_j)\frac{\lambda}{8\log\left(2T\right)}\geq\frac{\lambda}{8\log\left(2T\right)},$ where we used $\sigma_j\tau_j\geq1$. Therefore $w_{j}\geq\frac{\lambda}{8\log\left(2T\right)}.$ Now let $k>j$. On $A_j^+$, $Z_k\geq-\left(1-\frac{1}{8\log\left(2T\right)}\right)\frac{\lambda\sqrt{\delta}}{r_k},$
so the upper endpoint at task $k$ satisfies
\[
\sigma_k\tau_k\left(\frac{r_k}{\sqrt\delta}Z_k+\lambda\right)\geq(\sigma_k\tau_k)\frac{\lambda}{8\log\left(2T\right)}\geq\frac{\lambda}{8\log\left(2T\right)}.
\]
Thus, if $w_{k-1}\geq\frac{\lambda}{8\log\left(2T\right)}$, projecting $w_{k-1}$ onto the task-$k$ interval cannot produce a value below $\frac{\lambda}{8\log\left(2T\right)}$. By induction, $w_{t}\geq\frac{\lambda}{8\log\left(2T\right)}.$ Likewise, $A_j^-\subseteq\{w_{t}\leq-\frac{\lambda}{8\log\left(2T\right)}\}.$ Consequently,
\begin{align}
\PP\left(|w_{t}|\geq\frac{\lambda}{8\log\left(2T\right)}\mid w^{\star}=0\right)\geq2\sum_{j=1}^Tb_j\prod_{k=j+1}^T(1-2a_k).
\label{eq:null_excursion_events}
\end{align}
We next lower-bound the right-hand side.
Set $P_j=\prod_{k=j}^T(1-2a_k),$ and $P_{T+1}=1.$ Then $P_{j+1}-P_j=2a_jP_{j+1}.$
Hence, by telescoping and the assumption of the lemma,
\begin{align}
2\sum_{j=1}^Ta_jP_{j+1}=1-P_1\geq\frac{1}{2}.
\label{eq:null_telescoping}
\end{align}
Call an index $j$ good if $a_j\geq\frac{1}{16T}.$
The contribution of the remaining indices satisfies
\[
2\sum_{\{j:a_j<\frac{1}{16T}\}}a_jP_{j+1}\leq2\sum_{\{j:a_j<\frac{1}{16T}\}}a_j\leq\frac{1}{8}.
\]
Therefore \cref{eq:null_telescoping} gives $2\sum_{\{j:a_j\geq\frac{1}{16T}\}}a_jP_{j+1}\geq\frac38.$
We claim that on every good index, $b_j\geq\frac{1}{16}a_j.$
To see this, put $y=\left(1-\frac{1}{8\log\left(2T\right)}\right)\frac{\lambda\sqrt\delta}{r_j},$ and $z=\left(1+\frac{1}{8\log\left(2T\right)}\right)\frac{\lambda\sqrt\delta}{r_j}.$
Since $a_j=\Phi(-y)\geq\frac{1}{16T}$ and $\Phi(-y)\leq\frac{1}{2}e^{-\frac{y^2}{2}},$ we have $y^2\leq2\log(8T).$
If $y\leq1$, then, since for $T\geq 2$, $\frac{1+\frac{1}{8\log\left(2T\right)}}{1-\frac{1}{8\log\left(2T\right)}}\leq\frac65,$ we have $\frac{\Phi(-z)}{\Phi(-y)}\geq\frac{\Phi(-\frac{6}{5})}{\frac{1}{2}}\geq2\Phi\left(-\frac65\right)>\frac{1}{16}.$ 
If $y>1$, the Gaussian Mills bounds $\Phi(-z)\geq\frac{z\varphi(z)}{1+z^2},$ and $\Phi(-y)\leq\frac{\varphi(y)}{y}$
give
\[
\frac{\Phi(-z)}{\Phi(-y)}\geq\frac{zy}{1+z^2}\exp\left(-\frac{z^2-y^2}{2}\right).
\]
Here $\frac{zy}{1+z^2}\geq\frac{5z^2}{6(1+z^2)}\geq\frac{5}{12}\geq\frac{1}{3},$
while
\[
\frac{z^2-y^2}{2}=\frac{2}{8\log\left(2T\right)}\left(\frac{\lambda\sqrt\delta}{r_j}\right)^2=\frac{\frac{2}{8\log\left(2T\right)}}{\left(1-\frac{1}{8\log\left(2T\right)}\right)^2}y^2\leq\frac{3}{2}.
\]
Thus
\[
\frac{\Phi(-z)}{\Phi(-y)}\geq\frac{1}{3}e^{-\frac{3}{2}}>\frac{1}{16}.
\]
This proves $b_j\geq\frac{1}{16}a_j.$
Combining
\cref{eq:null_excursion_events}, with $b_j\geq\frac{1}{16}a_j$
and $2\sum_{\{j:a_j\geq\frac{1}{16T}\}}a_jP_{j+1}\geq\frac38.$
\begin{align*}
\PP\left(|w_{t}|\geq\frac{\lambda}{8\log\left(2T\right)}\mid w^{\star}=0\right)
&
\geq2\sum_{\{j:a_j\geq\frac{1}{16T}\}}b_jP_{j+1}
\geq\frac{2}{16}\sum_{\{j:a_j\geq\frac{1}{16T}\}}a_jP_{j+1}\geq\frac{3}{128}\geq\frac{1}{64}.
\end{align*}
Finally, on $\{w^{\star}=0\}$ the target $rw^{\star}$ is zero, so
\begin{align*}
\nu^{2}\left(\alpha_{t}-r\right)^{2}+\sigma_{t}^{2}&=\E\left[(w_{t}-rw^{\star})^{2}\right]
\geq(1-s)\E\left[w_{t}^{2}\mid w^{\star}=0\right]\\&\geq(1-s)\left(\frac{\lambda}{8\log\left(2T\right)}\right)^{2}\PP\left(|w_{t}|\geq\frac{\lambda}{8\log\left(2T\right)}\mid w^{\star}=0\right)\geq(1-s)\frac{\left(\frac{1}{8\log\left(2T\right)}\right)^{2}\lambda^{2}}{64}.
\end{align*}
We conclude that
\[
\nu^2(\alpha_T-r)^2+\sigma_T^2\geq\frac{(1-s)\lambda^2}
{4096\log^2(2T)}.
\]
\end{proof}

\newpage

\subsection{A proof of a uniform lower bound for the L1 regularizer}
\label{prf:uniL1}
\begin{lemma}[Uniform lower bound for L1 regularization]
There exist universal constants $C,c>0$, such that for every $s\leq s_0=c\min\left\{\frac{\frac{\delta}{1+\nu}}{\log_+^2\left(\frac{e(1+\nu)}{\delta}\right)},\frac{\frac{1}{\delta(1+\nu)}}{\log_+^4\left(e\delta(1+\nu)\right)},\frac{\delta\nu^2(1-2\rho)^2}{(1+\nu)^4}\right\},$ every $T\geq2$, and every $\lambda>0$,
\[
\mathcal{L}(\mathbf w_T)-\inf_{\vw} \mathcal{L}(\vw)\geq C\frac{\nu^7(1-2\rho)^7}{(1+\nu)^{32}(1+\delta)^{\frac{3}{2}}}\frac{s\log^{3}\left(\frac{1}{s}\right)}{\left(\log\left(\frac{1}{s}\right)+\log\left(2T\right)\right)^{2}}.
\]
Moreover, if $T=1$ then there exist universal constants $c,C>0$ such that, for all $s\leq c\frac{\delta\nu^2(1-2\rho)^2}{(1+\nu)^2\log\left(\frac{1+\nu}{\delta\nu^2(1-2\rho)^2}\right)}$ and for every $\lambda>0$,
\[
\mathcal{L}(\mathbf w_1)-\inf_{\vw} \mathcal{L}(\vw)\geq \frac{C}{(1+\nu)^6}\frac{s}{\delta}\log\left(\frac{1}{s}\right).
\]
\end{lemma}
\begin{proof}
Note that for the special case of $T=1$ uses the lower bound developed in \cref{thm:single_L1}.

For temporal L1 by \cref{lem:2},
\[
w_{t}=w_{t-1}+\mathcal S_{\lambda\sigma_t\tau_t}\left(\sigma_t\tau_t\left(\theta_tw^{\star}+\frac{r_t}{\sqrt\delta}Z_t\right)-w_{t-1}\right).
\]
Equivalently we may write $w_{t}$ as the projection of $w_{t-1}$ onto the interval $\sigma_t\tau_t\left(\theta_tw^{\star}+\frac{r_t}{\sqrt\delta}Z_t\pm\lambda\right)$,
\[
w_{t}=\Pi_{\left[\sigma_t\tau_t\left(\theta_tw^{\star}+\frac{r_t}{\sqrt\delta}Z_t-\lambda\right),\sigma_t\tau_t\left(\theta_tw^{\star}+\frac{r_t}{\sqrt\delta}Z_t+\lambda\right)\right]}(w_{t-1}).
\]
Hence $\left|\frac{w_{t}}{\sigma_t\tau_t}-\frac{r_t}{\sqrt\delta}Z_t-\theta_tw^{\star}\right|\leq\lambda.$
We assume by contradiction that for arbitrarily small $s$ there exist $T\geq2$ and $\lambda>0$ such that,
\begin{align}
\nu^2(\alpha_t-r)^2+\sigma_t^2<c_\star s\frac{\log^3\left(\frac{1}{s}\right)}{\left(\log\left(\frac{1}{s}\right)+\log\left(2T\right)\right)^2},
\label{eq:contradiction_temporal}
\end{align}
here $c_\star>0$ is a sufficiently small constant whose value be established at the end of the proof.
Since
\[
\frac{s\log^3\left(\frac{1}{s}\right)}{\left(\log\left(\frac{1}{s}\right)+\log\left(2T\right)\right)^2}\leq s\log\left(\frac{1}{s}\right)\longrightarrow0,
\]
we must have $\alpha_t\longrightarrow r$, and $\sigma_t\longrightarrow0.$
Then $\nu\alpha_{t}Z_1+\sigma_{t}Z_2\longrightarrow\nu rZ_1$. Furthermore, the second equation gives $\gamma_t=\frac{1}{r_t\sqrt\delta}\E[Z_tw_{t}].$ Since $Z_t$ is independent of $w^{\star}$, by Cauchy-Schwartz,
\[
|\E[Z_tw_{t}]|=|\E[Z_t(w_{t}-rw^{\star})]|\leq\sqrt{\nu^2(\alpha_t-r)^2+\sigma_t^2}.
\]
Using $\sigma_t\tau_tr_t\geq1$ (see below \cref{eq:22}), we obtain
\[
0\leq\frac{\gamma_t}{\sigma_t\tau_t}\leq\sqrt{\frac{\nu^2(\alpha_t-r)^2+\sigma_t^2}{\delta}}\longrightarrow0.
\]
The sixth scalar equation is
\[
1-\frac{\gamma_t}{\sigma_t\tau_t}=\E\left[\frac{2(1-\pi_\rho(\nu Z_1))}{1+\gamma_t\phi''(\prox_{\gamma_t\phi}(\nu\alpha_{t}Z_{1}+\sigma_{t}Z_{2}))}\right].
\]
The left-hand side tends to $1$. This forces $\gamma_t\longrightarrow0.$ Indeed, if $\gamma_t$ converged to a positive finite number, then the denominator on the right would be strictly larger than one with positive probability, since in that case by the proximal relation (\cref{eq:prox_relation_large_T}) $\nu r Z_{1}-\gamma_\infty\leq \prox_{\gamma_t\phi}(\nu r Z_{1}))\leq \nu r Z_{1}$, then since $Z_1$ is finite almost surely and $\phi''(\prox_{\gamma_t\phi}(\nu\alpha_{t}Z_{1}+\sigma_{t}Z_{2})))=\frac{e^{\prox_{\gamma_t\phi}(\nu\alpha_{t}Z_{1}+\sigma_{t}Z_{2}))}}{(1+e^{\prox_{\gamma_t\phi}(\nu\alpha_{t}Z_{1}+\sigma_{t}Z_{2}))})^2}>0$ so $1+\gamma_\infty\phi''(\prox_{\gamma_t\phi}(\nu\alpha_{t}Z_{1}+\sigma_{t}Z_{2})))>1$ almost surely, and from symmetry $\E[\pi_\rho(\nu Z_1)]=\frac{1}{2}$ implying $\E[2(1-\pi_\rho(-\nu Z_1))]=1$ so its expectation would have limit strictly smaller than one.  If $\gamma_t\to\infty$, then from the proximal relation \begin{align}
\label{eq:prox_relation_large_T}
\nu\alpha_{t}Z_1+\sigma_{t}Z_2=\prox_{\gamma_t\phi}(\nu\alpha_{t}Z_{1}+\sigma_{t}Z_{2})+\gamma_t\phi'(\prox_{\gamma_t\phi}(\nu\alpha_{t}Z_{1}+\sigma_{t}Z_{2})),
\end{align}
and $\alpha_t\to r$, $\sigma_t\to 0$, one gets $\prox_{\gamma_t\phi}(\nu\alpha_{t}Z_{1}+\sigma_{t}Z_{2})\to-\infty$ on every event $\{|\nu\alpha_{t}Z_1+\sigma_{t}Z_2|\leq M\}$. Indeed, if you suppose there exists $B>0$ such that $\prox_{\gamma_t\phi}(\nu\alpha_{t}Z_{1}+\sigma_{t}Z_{2})>-B$ then by 
\begin{align*}
\nu\alpha_{t}Z_1+\sigma_{t}Z_2=\prox_{\gamma_t\phi}(\nu\alpha_{t}Z_{1}+\sigma_{t}Z_{2}))&+\gamma_t\phi'(\prox_{\gamma_t\phi}(\nu\alpha_{t}Z_{1}+\sigma_{t}Z_{2})))
>-B+\gamma_t\phi'(-B)\xrightarrow[]{\gamma_t\to \infty}\infty
\end{align*}
which contradicts $|\nu\alpha_{t}Z_1+\sigma_{t}Z_2|\leq M$, also we have
\begin{align*} 
\gamma_{t}\phi''(\prox_{\gamma_{t}\phi}(\nu\alpha_{t}Z_{1}+\sigma_{t}Z_{2})))=\gamma_{t}\phi'(\prox_{\gamma_{t}\phi}(\nu\alpha_{t}Z_{1}+\sigma_{t}Z_{2}))(1-\phi'(\prox_{\gamma_{t}\phi}(\nu\alpha_{t}Z_{1}+\sigma_{t}Z_{2})))\\\stackrel{\mathrm{\ref{eq:prox_relation_large_T}}}{=}\left(\nu\alpha_{t}Z_{1}+\sigma_{t}Z_{2}-\prox_{\gamma_{t}\phi}(\nu\alpha_{t}Z_{1}+\sigma_{t}Z_{2})\right)(1-\phi'(\prox_{\gamma_{t}\phi}(\nu\alpha_{t}Z_{1}+\sigma_{t}Z_{2})))\xrightarrow[]{\gamma_t\to\infty}\infty,
\end{align*}
the expectation in the sixth equation would then tend to zero on events whose probability tends to one, again a contradiction.
Thus
\[
|\prox_{\gamma_t\phi}(\nu\alpha_{t}Z_{1}+\sigma_{t}Z_{2})-\nu\alpha_{t}Z_1+\sigma_{t}Z_2|=\gamma_t\phi'(\prox_{\gamma_{t}\phi}(\nu\alpha_{t}Z_{1}+\sigma_{t}Z_{2}))\leq\gamma_t\longrightarrow0,
\]
and consequently $\prox_{\gamma_t\phi}(\nu\alpha_{t}Z_{1}+\sigma_{t}Z_{2})\longrightarrow\nu rZ_1$ in probability. From
\[
r_t^2=2\E\left[(1-\pi_\rho(\nu Z_1))\phi'(\prox_{\gamma_t\phi}(\nu\alpha_{t}Z_{1}+\sigma_{t}Z_{2}))^2\right],
\]
dominated convergence gives
\[
r_t^2\longrightarrow2\E\left[(1-\pi_\rho(\nu Z_1))\phi'(\nu rZ_1)^2\right]>0.
\]
Hence $r_t\geq c>0$ for a sufficiently small $s$. Now we aim to lower bound $\E[(w_{t}-rw^{\star})^2]=\nu^2(\alpha_t-r)^2+\sigma_t^2$, note that 
\begin{align*}
\E[(w_{t}-rw^{\star})^2]]&\geq\E[\E[(w_{t}-rw^{\star})^2]\mid w^{\star},w_{t-1}]]
\\&=\mathrm{Var}(w_{t}\mid w^{\star},w_{t-1})+(\E[w_{t}\mid w^{\star}, w_{t-1}]-rw^{\star})^2\geq\mathrm{Var}(w_{t}\mid w^{\star},w_{t-1}) 
\end{align*}
So it is sufficient to lower bound $\mathrm{Var}(w_{t}\mid w^{\star},w_{t-1}).$ By Cauchy-Schwartz, \[\mathrm{Cov}(w_{t},Z_t\mid w^\star,w_{t-1})^2\leq \mathrm{Var}(w_{t}\mid w^\star,w_{t-1})\mathrm{Var}(Z_t\mid w^\star,w_{t-1}).\] Since $\mathrm{Var}(Z_t\mid w^\star,w_{t-1})=1$ then $\mathrm{Var}(w_{t}\mid w^{\star},w_{t-1})\geq \mathrm{Cov}(w_{t},Z_t\mid w^{\star},w_{t-1})^2$ (due to the independence of $Z_t$ in $w^\star$ and $w_{t-1}$). If $w_{t}\neq w_{t-1}$ that means that soft threshold has magnitude larger than the threshold i.e. $\left|Z_{t}+\underbrace{\frac{\sqrt{\delta}}{r_{t}}\left(\theta_{t}w^{\star}-\frac{w_{t-1}}{\sigma_{t}\tau_{t}}\right)}_{=c}\right|>\frac{\sqrt{\delta}}{r_{t}}\lambda$ the probability of this event is at least as large when $c=0$. Explicitly, \[\PP\left(|Z_t+c|>\frac{\sqrt \delta}{r_t}\lambda\right)\geq \PP\left(|Z_t|>\frac{\sqrt \delta}{r_t}\lambda\right)=2\Phi\left(-\frac{\lambda\sqrt \delta}{r_t}\right)\]
By Gaussian integration by parts and since $\frac{\partial w_{t}}{\partial Z_t}=\frac{\sigma_t\tau_tr_t}{\sqrt \delta}\mathbf{1}_{\{|Z_t+c|>\frac{\sqrt \delta}{r_t}\lambda\}}$, this gives
\begin{align*}
&\nu^2(\alpha_t-r)^2+\sigma_t^2\geq\mathrm{Cov}(w_{t},Z_t\mid w^{\star},w_{t-1})^2=\E\left[w_tZ_t\mid w^{\star},w_{t-1}\right]^2\\&=\E\left[\frac{\sigma_t\tau_tr_t}{\sqrt \delta}\mathbf{1}_{\{|Z_t+c|>\frac{\sqrt \delta}{r_t}\lambda\}}\mid w^{\star},w_{t-1}\right]^2\geq\frac{4(\sigma_t\tau_t)^2r_t^2}{\delta}\Phi\left(-\frac{\lambda\sqrt\delta}{r_t}\right)^2\geq\frac4\delta\Phi\left(-\frac{\lambda\sqrt\delta}{r_t}\right)^2.
\end{align*}
Together with the fact that $r_t>0$ and by the assumption, this implies $\lambda\to\infty.$
Condition on $w^{\star}=0$. If $w_{t-1}=0$, then $w_{t}=\sigma_t\tau_t\mathcal S_\lambda\left(\frac{r_t}{\sqrt\delta}Z_t\right),$ and therefore $w_{t}\neq0$ if and only if $|Z_t|>\frac{\lambda\sqrt\delta}{r_t}.$
Once a null coordinate becomes nonzero it remains nonzero almost surely: if $w_{t-1}\neq0$, a later projection can equal exactly zero only if one of the continuously distributed interval endpoints is exactly zero. Hence
\begin{align}
\PP(w_{t}\neq0\mid w^{\star}=0)=1-\prod_{j=1}^t\left[1-2\Phi\left(-\frac{\lambda\sqrt\delta}{r_j}\right)\right].
\label{eq:null_support_exact}
\end{align}
We first claim that $\PP(w_{t}\neq0\mid w^{\star}=0)<\frac{1}{2}.$ Otherwise, $\prod_{j=1}^t\left[1-2\Phi\left(-\frac{\lambda\sqrt\delta}{r_j}\right)\right]\leq\frac{1}{2}.$
Since $\frac{1}{8\log\left(2T\right)}>0,$ this implies
\[
\prod_{j=1}^{t}\left[1-2\Phi\left(-\left(1-\frac{1}{8\log\left(2t\right)}\right)\frac{\lambda\sqrt{\delta}}{r_{j}}\right)\right]\leq\prod_{j=1}^{t}\left[1-2\Phi\left(-\frac{\lambda\sqrt{\delta}}{r_{j}}\right)\right]\leq\frac{1}{2}
\]
Since it must hold $\lambda\to\infty$ it means also that $\lambda\geq 1$ then by \cref{lem:6}, 
\[
\nu^2(\alpha_t-r)^2+\sigma_t^2\geq\frac{1-s}{4096\log^2(2T)}.
\]
But $\frac{s\log^3\left(\frac{1}{s}\right)}{\left(\log\left(\frac{1}{s}\right)+\log(2T)\right)^2}\leq\frac{s\log^3\left(\frac{1}{s}\right)}{\log^2(2T)},$ so it is enough to take $s$ sufficiently small that $c_\star s\log^3\left(\frac{1}{s}\right)<\frac{1}{8192}$. This contradicts \cref{eq:contradiction_temporal}. Thus $\PP(w_t\neq0\mid w^{\star}=0)<\frac{1}{2}$.
Moreover, \cref{eq:null_support_exact} implies
$2\Phi\left(-\frac{\lambda\sqrt{\delta}}{r_j}\right)<\frac{1}{2}$ for every $j\leq T$. Hence, we have $\frac{\lambda\sqrt{\delta}}{r_j}\geq \Phi^{-1}\left(\frac{3}{4}\right)>0$, and therefore, using $r_j\geq\sqrt{\frac{h(\nu)}{8}}$ (see \cref{eq:35}), then $\lambda\geq\Phi^{-1}\left(\frac{3}{4}\right)\sqrt{\frac{h(\nu)}{8\delta}}.$
From \cref{eq:null_support_exact}, and since $1-\prod_{j=1}^t z_j \leq \sum_{j=1}^t (1-z_j)$ for all $z_j \in [0,1]$ where $j\in [1,t]$, therefore, $\PP(w_{t}\neq0\mid w^{\star}=0)\leq\sum_{j=1}^t2\Phi\left(-\frac{\lambda\sqrt\delta}{r_j}\right).$
Also by \cref{eq:null_support_exact},
\begin{align*} 
-\log\left(1-\PP(w_{t}\neq0\mid w^{\star}=0)\right)
= -\sum_{j=1}^t\log\left(1-2\Phi\left(-\frac{\lambda\sqrt\delta}{r_j}\right)\right)\underset{-\log(1-x)\geq x \mathrm{ }\forall x\in [0,1]}{\geq}\sum_{j=1}^T2\Phi\left(-\frac{\lambda\sqrt\delta}{r_j}\right).
\end{align*}

Since the probability on the left is smaller than $\frac{1}{2}$, and $-\log(1-x)\leq2x$ for all $0\leq x\leq\frac{1}{2}$ so
\begin{align}
\PP(w_{t}\neq0\mid w^{\star}=0)\leq\sum_{j=1}^T2\Phi\left(-\frac{\lambda\sqrt\delta}{r_j}\right)\leq2\PP(w_{t}\neq0\mid w^{\star}=0).
\label{eq:null_tail_equivalence}
\end{align}
We also need a quantitative relation between this probability and the null-coordinate squared error. On $\{w^{\star}=0,w_{t}\neq0\}$ let $J$ be the last update time. Then after projecting the update to one of the endpoints,
\[
w_{t}=\sigma_J\tau_J\left(\frac{r_J}{\sqrt\delta}Z_J-\epsilon_J\lambda\right).
\]
with $\epsilon_J\in\{-1,+1\}.$ Thus, if $0<|w_{t}|\leq a$, then,
$\left|\sigma_J\tau_J\left(\frac{r_J}{\sqrt\delta}Z_J-\epsilon_J\lambda\right)\right|\leq a$ implies $\left|Z_J-\epsilon_J\frac{\lambda\sqrt\delta}{r_J}\right|\leq a\sqrt\delta,$ because $\sigma_J\tau_Jr_J\geq1$ (see \cref{eq:10}).
Denote for simplicity $a=\frac{1}{64\sqrt{\delta}\left(1+L\right)}$ where $L=\sqrt{2\log\left(\frac{4T}{\sum_{j=1}^T2\Phi\left(-\frac{\lambda\sqrt\delta}{r_j}\right)}\right)}.$ Since $a\sqrt{\delta}=\frac{1}{64(1+L)}\leq\frac{1}{64}<\Phi^{-1}\left(\frac{3}{4}\right)$, we have $a\leq\frac{\lambda}{r_j}.$ Thus, using the Gaussian density bound $\PP(|Z-x|\leq q)\leq2q\varphi(x-q),$ whenever $x\geq q>0$ together with the identity $\varphi(x)\leq(x+1)\Phi(-x),$ for all $x\geq 0$ which follows from \cref{lem:4}, we have, if $\frac{\lambda\sqrt\delta}{r_j}\leq L$, 
\begin{align*}
\PP\left(\left|Z-\frac{\lambda\sqrt{\delta}}{r_{j}}\right|\leq a\sqrt{\delta}\right)&\leq2a\sqrt{\delta}\varphi\left(\frac{\lambda\sqrt{\delta}}{r_{j}}-a\sqrt{\delta}\right)=2a\sqrt{\delta}e^{\frac{a\delta\lambda}{r_{j}}-\frac{(a\sqrt{\delta})^{2}}{2}}\varphi\left(\frac{\lambda\sqrt{\delta}}{r_{j}}\right)\\&\leq2a\sqrt{\delta}e^{La\sqrt{\delta}}(L+1)\Phi\left(-\frac{\lambda\sqrt{\delta}}{r_{j}}\right)\\&\underset{a\sqrt{\delta}=\frac{1}{64(1+L)}}{\leq}\frac{1}{16}\Phi\left(-\frac{\lambda\sqrt{\delta}}{r_{j}}\right),\end{align*}
On the other hand, if $\frac{\lambda\sqrt\delta}{r_j}>L$ since $a\sqrt\delta\leq\frac{\lambda\sqrt{\delta}}{r_{j}}$, then
\[
\PP\left(\left|Z-\frac{\lambda\sqrt\delta}{r_j}\right|\leq a\sqrt\delta\right)\leq2a\sqrt\delta\varphi(L-a\sqrt\delta).
\]
Consequently, using the union bound over the last update time and the two possible signs,
\begin{align*}
\PP\left(0<|w_{t}|\leq a\mid w^{\star}=0\right)&\leq2\sum_{j=1}^{t}\PP\left(\left|Z-\frac{\lambda\sqrt{\delta}}{r_{j}}\right|\leq a\sqrt{\delta}\right)\\&\leq\frac{1}{16}\sum_{\{\frac{\lambda\sqrt{\delta}}{r_{j}}\leq L\}}2\Phi\left(-\frac{\lambda\sqrt{\delta}}{r_{j}}\right)+4ta\sqrt{\delta}\varphi(L-a\sqrt{\delta})\\&\leq\frac{1}{16}\sum_{j=1}^{t}2\Phi\left(-\frac{\lambda\sqrt{\delta}}{r_{j}}\right)+\frac{4ta\sqrt{\delta}}{\sqrt{2\pi}}e^{La\sqrt{\delta}-\frac{(a\sqrt{\delta})^{2}}{2}}e^{-\frac{L^{2}}{2}}\\&\stackrel{[1]}{=}\frac{1}{16}\sum_{j=1}^{t}2\Phi\left(-\frac{\lambda\sqrt{\delta}}{r_{j}}\right)+\frac{a\sqrt{\delta}}{\sqrt{2\pi}}e^{La\sqrt{\delta}-\frac{(a\sqrt{\delta})^{2}}{2}}\sum_{j=1}^{t}2\Phi\left(-\frac{\lambda\sqrt{\delta}}{r_{j}}\right)\\&\leq\frac{1}{16}\sum_{j=1}^{t}2\Phi\left(-\frac{\lambda\sqrt{\delta}}{r_{j}}\right)+\frac{1}{16}\sum_{j=1}^{t}2\Phi\left(-\frac{\lambda\sqrt{\delta}}{r_{j}}\right)=\frac{1}{8}\sum_{j=1}^{t}2\Phi\left(-\frac{\lambda\sqrt{\delta}}{r_{j}}\right),
\end{align*}
where [1] follows from $e^{-\frac{L^2}{2}}=\frac{\sum_{j=1}^t2\Phi\left(-\frac{\lambda\sqrt\delta}{r_j}\right)}{4t}.$ Therefore $\PP(0<|w_{t}|\leq a\mid w^{\star}=0)\leq\frac{1}{8}\sum_{j=1}^t2\Phi\left(-\frac{\lambda\sqrt\delta}{r_j}\right).$
By \cref{eq:null_tail_equivalence} it follows that $\PP(0<|w_{t}|\leq a\mid w^{\star}=0)\leq\frac{1}{4}\PP(w_{t}\neq0\mid w^{\star}=0)$, so
\begin{align*}
\PP(|w_{t}|>a\mid w^{\star}=0)&=\PP(w_{t}\neq0\mid w^{\star}=0)-\PP(0<|w_{t}|\leq a\mid w^{\star}=0)\\&\geq\PP(w_{t}\neq0\mid w^{\star}=0)-\frac{1}{4}\PP(w_{t}\neq0\mid w^{\star}=0)\geq\frac{1}{2}\PP(w_{t}\neq0\mid w^{\star}=0)   
\end{align*}
Therefore by Markov's inequality,
\begin{align*}
\nu^{2}(\alpha_{t}-r)^{2}+\sigma_{t}^{2}&=\E\left[\left(w_{t}-rw^{\star}\right)^{2}\right]
\geq\left(1-s\right)\E\left[w_{t}^{2}\mid w^{\star}=0\right]\geq\left(1-s\right)a^{2}\PP\left(\left|w_{t}\right|>a\mid w^{\star}=0\right).
\end{align*}
Using $\left(1+\sqrt{2x}\right)^{2}\leq10x$ for all $x\geq1$, then since $\log\left(\frac{4t}{\PP(w_{t}\neq0\mid w^{\star}=0)}\right)\geq 1$,
\begin{align} 
\nu^{2}(\alpha_{t}-r)^{2}+\sigma_{t}^{2}\geq\frac{3\left(1-s\right)}{163840\delta}\frac{\PP(w_{t}\neq0\mid w^{\star}=0)}{\log\left(\frac{4T}{\PP(w_{t}\neq0\mid w^{\star}=0)}\right)}.
\label{eq:null_quantitative_short}
\end{align}
If $\PP(w_{t}\neq0\mid w^{\star}=0)\leq s,$ there is nothing to prove. Otherwise,
\[
\log\left(\frac{4T}{\PP(w_{t}\neq0\mid w^{\star}=0)}\right)\leq C\left(\log\left(\frac{1}{s}\right)+\log\left(2T\right)\right).
\]
Thus \cref{eq:null_quantitative_short,eq:contradiction_temporal} imply
\begin{align}
\PP(w_{t}\neq0\mid w^{\star}=0)\leq s+Cc_\star s\frac{\log^3\left(\frac{1}{s}\right)}{\log\left(\frac{1}{s}\right)+\log\left(2T\right)}\leq C s\log^2\left(\frac{1}{s}\right).
\label{eq:null_support_small}
\end{align}
Combining this with \cref{eq:null_tail_equivalence}, we get $\log\left[\left(\sum_{j=1}^t2\Phi\left(-\frac{\lambda\sqrt\delta}{r_j}\right)\right)^{-1}\right]\geq\log\left(\frac{1}{s}\right)-2\log\log\left(\frac{1}{s}\right)-\log(C)$, therefore,
\begin{align}
\log\left[\left(\sum_{j=1}^t2\Phi\left(-\frac{\lambda\sqrt\delta}{r_j}\right)\right)^{-1}\right]\geq\frac{1}{2}\log\left(\frac{1}{s}\right)
\label{eq:tail_log_lower}
\end{align}
for a sufficiently small $s$. For a null coordinate, any update implies $w_{t}\neq0$ almost surely.
Hence
\[
\PP(w_{t}\neq w_{t-1})\leq s+\PP(w_{t}\neq0\mid w^{\star}=0)\leq Cs\log^2\left(\frac{1}{s}\right).
\]
Stein's Lemma gives $\frac{\gamma_t}{\sigma_t\tau_t}=\frac{1}{\delta}\PP(w_{t}\neq w_{t-1}).$ By the proximal relation (\cref{eq:prox_relation_large_T})
we have $\nu\alpha_{t}Z_1+\sigma_{t}Z_2\geq\prox_{\gamma_t\phi}(\nu\alpha_{t}Z_{1}+\sigma_{t}Z_{2})$. For the logistic sigmoid,
\begin{align}
\label{eq:15}
0\leq\phi'(\nu\alpha_{t}Z_{1}+\sigma_{t}Z_{2})-\phi'\left(\prox_{\gamma_{t}\phi}(\nu\alpha_{t}Z_{1}+\sigma_{t}Z_{2})\right)\leq\frac{2\gamma_{t}\phi''\left(\prox_{\gamma_{t}\phi}(\nu\alpha_{t}Z_{1}+\sigma_{t}Z_{2})\right)}{1+\gamma_{t}\phi''\left(\prox_{\gamma_{t}\phi}(\nu\alpha_{t}Z_{1}+\sigma_{t}Z_{2})\right)}.
\end{align}
Indeed, if $\gamma_t\phi''(\prox_{\gamma_t\phi}(\nu\alpha_{t}Z_{1}+\sigma_{t}Z_{2}))\geq1$ this is trivial, because $0<\phi'<1$. If $\gamma_t\phi''(\prox_{\gamma_t\phi}(\nu\alpha_{t}Z_{1}+\sigma_{t}Z_{2}))<1$ and $\phi'(\prox_{\gamma_t\phi}(\nu\alpha_{t}Z_{1}+\sigma_{t}Z_{2})))\geq\frac{1}{2}$, so $\nu\alpha_{t}Z_{1}+\sigma_{t}Z_{2}\geq\prox_{\gamma_{t}\phi}(\nu\alpha_{t}Z_{1}+\sigma_{t}Z_{2})\geq0$ concavity of the sigmoid ($\phi'''(v)=\phi''(v)(1-2\phi'(v))\leq 0$ for $v\geq 0$), with the proximal equation, gives
\begin{align*}
&\phi'(\nu\alpha_{t}Z_{1}+\sigma_{t}Z_{2})-\phi'(\prox_{\gamma_{t}\phi}(\nu\alpha_{t}Z_{1}+\sigma_{t}Z_{2})))\\&\leq\gamma_{t}\phi''(\prox_{\gamma_{t}\phi}(\nu\alpha_{t}Z_{1}+\sigma_{t}Z_{2}))\underbrace{\phi'(\prox_{\gamma_{t}\phi}(\nu\alpha_{t}Z_{1}+\sigma_{t}Z_{2}))}_{=\nu\alpha_{t}Z_{1}+\sigma_{t}Z_{2}-\prox_{\gamma_{t}\phi}(\nu\alpha_{t}Z_{1}+\sigma_{t}Z_{2})}\leq\gamma_{t}\phi''(\prox_{\gamma_{t}\phi}(\nu\alpha_{t}Z_{1}+\sigma_{t}Z_{2})).
\end{align*}
If $\phi'(\prox_{\gamma_t\phi}(\nu\alpha_{t}Z_{1}+\sigma_{t}Z_{2})))<\frac{1}{2}$, then
\begin{align*}
&\nu\alpha_{t}Z_1+\sigma_{t}Z_2-\prox_{\gamma_t\phi}(\nu\alpha_{t}Z_{1}+\sigma_{t}Z_{2})=\gamma_t\phi'(\prox_{\gamma_t\phi}(\nu\alpha_{t}Z_{1}+\sigma_{t}Z_{2}))
\\&\hspace{8em}=\frac{\gamma_t\phi''(\prox_{\gamma_t\phi}(\nu\alpha_{t}Z_{1}+\sigma_{t}Z_{2}))}{1-\phi'(\prox_{\gamma_t\phi}(\nu\alpha_{t}Z_{1}+\sigma_{t}Z_{2}))}\leq2\gamma_t\phi''(\prox_{\gamma_t\phi}(\nu\alpha_{t}Z_{1}+\sigma_{t}Z_{2})),
\end{align*}
and the $\frac{1}{4}$-Lipschitz property of $\phi'$ gives the same conclusion. From symmetry,
\begin{align*} 
&2\E[\pi_\rho'(\nu Z_1)\phi'(\nu\alpha_tZ_1+\sigma_tZ_2)]\\&=\E[\pi_\rho'(\nu Z_1)\underbrace{(\phi'(\nu\alpha_tZ_1+\sigma_tZ_2)+\phi'(-\nu\alpha_tZ_1-\sigma_tZ_2)}_{=1})]=\E[\pi_\rho'(\nu Z_1)]=(1-2\rho)h(\nu).
\end{align*}
Therefore, since $\pi_\rho'(u)\leq\frac{1-2\rho}{4},$ and by \cref{eq:22} $\theta_t=2\E[\pi_\rho'(\nu Z_1)\phi'(\prox_{\gamma_t\phi}(\nu\alpha_tZ_1+\sigma_tZ_2)]$ with the previous inequality,
\begin{equation}
\label{eq:33}    
\begin{aligned} 
0\leq(1-2\rho)h(\nu)-\theta_{t}&=2\E\left[\pi_{\rho}'(\nu Z_{1})\left(\phi'(\nu\alpha_{t}Z_{1}+\sigma_{t}Z_{2})-\phi'(\prox_{\gamma_{t}\phi}(\nu\alpha_{t}Z_{1}+\sigma_{t}Z_{2}))\right)\right]\\&\leq(1-2\rho)\E\left[\frac{\gamma_t\phi''(\prox_{\gamma_t\phi}(\nu\alpha_{t}Z_{1}+\sigma_{t}Z_{2}))}{1+\gamma_t\phi''(\prox_{\gamma_t\phi}(\nu\alpha_{t}Z_{1}+\sigma_{t}Z_{2}))}\right].
\end{aligned}
\end{equation}

The sixth scalar equation, along with the equation for $\gamma_t$ and Stein's Lemma (note that, $\frac{\partial w_{t}}{\partial Z_t}=\frac{\sigma_t\tau_tr_t}{\sqrt\delta}\mathbf{1}_{\{w_{t}\neq w_{t-1}\}}$) then $\gamma_t=\frac{1}{r_t\sqrt \delta}\E[Z_tw_{t}]=\frac{1}{r_t\sqrt\delta}\frac{\sigma_t\tau_tr_t}{\sqrt\delta}\PP(w_{t}\neq w_{t-1})=\frac{\sigma_t\tau_t}{\delta}\PP(w_{t}\neq w_{t-1})$, this gives,
\[
\frac{1}{\delta}\PP(w_{t}\neq w_{t-1})=\frac{\gamma_t}{\sigma_t\tau_t}=\E\left[2(1-\pi_\rho(\nu Z_1))\frac{\gamma_t\phi''(\prox_{\gamma_t\phi}(\nu\alpha_{t}Z_{1}+\sigma_{t}Z_{2}))}{1+\gamma_t\phi''(\prox_{\gamma_t\phi}(\nu\alpha_{t}Z_{1}+\sigma_{t}Z_{2}))}\right].
\]
From \cref{eq:15},
\[
0\leq\phi'(\nu\alpha_{t}Z_{1}+\sigma_{t}Z_{2})-\phi'\left(\prox_{\gamma_{t}\phi}(\nu\alpha_{t}Z_{1}+\sigma_{t}Z_{2})\right)\leq\frac{2\gamma_{t}\phi''\left(\prox_{\gamma_{t}\phi}(\nu\alpha_{t}Z_{1}+\sigma_{t}Z_{2})\right)}{1+\gamma_{t}\phi''\left(\prox_{\gamma_{t}\phi}(\nu\alpha_{t}Z_{1}+\sigma_{t}Z_{2})\right)}.
\]
By plugging this upper bound we get $(1-2\rho)h(\nu)-\theta_t\leq \E\left[\frac{4\pi_\rho'(\nu Z_1)\gamma_t\phi''(\prox_{\gamma_t\phi}(\nu\alpha_{t}Z_{1}+\sigma_{t}Z_{2}))}{1+\gamma_t\phi''(\prox_{\gamma_t\phi}(\nu\alpha_{t}Z_{1}+\sigma_{t}Z_{2}))}\right]$. Now we bound $\pi_\rho'(\nu Z_1)$ from above, by definition $\pi_\rho'(\nu Z_1)=(1-2\rho)\phi'(\nu Z_1)(1-\phi'(\nu Z_1)),$ $1-\pi_\rho(\nu Z_1)=\rho+(1-2\rho)(1-\phi'(\nu Z_1))$. By these, it follows that for every $0\leq\rho<\frac{1}{2}$, 
\begin{align*}   
\underbrace{\rho+(1-2\rho)(1-\phi'(\nu Z_{1}))}_{=1-\pi_{\rho}(\nu Z_{1})}-\underbrace{\phi'(\nu Z_{1})(1-\phi'(\nu Z_{1}))}_{=\frac{\pi_{\rho}'(\nu Z_{1})}{1-2\rho}}&=(1-\phi'(\nu Z_{1}))(1-\phi'(\nu Z_{1})-2\rho)+\rho\\&=(1-\phi'(\nu Z_{1})-\rho)^{2}+\rho(1-\rho)\geq0
\end{align*}
and therefore,
\[
\underbrace{(1-2\rho)\phi'(\nu Z_{1})(1-\phi'(\nu Z_{1}))}_{\pi_{\rho}'(\nu Z_{1})}\leq (1-2\rho)\underbrace{\left(\rho+(1-2\rho)(1-\phi'(\nu Z_{1}))\right)}_{1-\pi_{\rho}(\nu Z_{1})}.
\]
So,
\begin{align} 
\label{eq:32}  
(1-2\rho)h(\nu)-\theta_t
\leq 4(1-2\rho)\E\left[(1-\pi_\rho(\nu Z_1))\frac{\gamma_t\phi''(\prox_{\gamma_t\phi}(\nu\alpha_{t}Z_{1}+\sigma_{t}Z_{2}))}{1+\gamma_t\phi''(\prox_{\gamma_t\phi}(\nu\alpha_{t}Z_{1}+\sigma_{t}Z_{2}))}\right]=2(1-2\rho)\frac{\gamma_t}{\sigma_t\tau_t}.
\end{align}
Combining the null-coordinate support bound with $\{w_t\neq w_{t-1},w^{\star}=0\}\subseteq \{w_t\neq0,w^{\star}=0\},$ almost surely, we obtain
\[
\PP(w_t\neq w_{t-1})\le s+\PP(w_t\neq0\mid w^{\star}=0)\le Cs\log^2\left(\frac{1}{s}\right).
\]
Since the contradiction argument above implies $\lambda\to\infty$, for all sufficiently small $s$ we have $Cs\log^2\left(\frac{1}{s}\right)\le \lambda\sqrt{s}.$
Therefore $\PP(w_t\neq w_{t-1})\le \lambda\sqrt{s}.$
By \cref{eq:32} and $\frac{1}{\delta}\PP(w_{t}\neq w_{t-1})=\frac{\gamma_t}{\sigma_t\tau_t}$, we have $\frac{1}{2}(1-2\rho)h(\nu)\leq\theta_t$ uniformly in $t$ for a sufficiently small $s$. Also using \cref{eq:22},
\[
\theta_t=2\E[\pi_\rho'(\nu Z_1)\phi'(\prox_{\gamma_t\phi}(\nu\alpha_{t}Z_{1}+\sigma_{t}Z_{2})]\leq\frac{1-2\rho}{2}\E[\phi'(\prox_{\gamma_t\phi}(\nu\alpha_{t}Z_{1}+\sigma_{t}Z_{2})].
\]
Thus, since $\frac{1}{2}(1-2\rho)h(\nu)\leq \theta_t$, and $\theta_t=2(1-2\rho)\E[\phi''(\nu Z_1)\phi'(\prox_{\gamma_t\phi}(\nu\alpha_tZ_1+\sigma_tZ_2))]$, hence for $0\leq\rho\leq\frac{1}{2}$, $\E[\phi''(\nu Z_1)\phi'(\prox_{\gamma_t\phi}(\nu\alpha_tZ_1+\sigma_t Z_2))]\geq\frac{1}{4}h(\nu)$. Now, we have uniformly for all $0\leq\rho\leq\frac{1}{2}$, $1-\pi_\rho(u)\geq \phi''(u)$ since $1-\pi_\rho(u)-\phi''(u)=\rho+(1-2\rho)(1-\phi'(u))-\phi'(u)(1-\phi'(u))=(1-\phi'(u)-\rho)^2+\rho(1-2\rho)\geq 0$ as we have already established.
Hence
\begin{equation}
\label{eq:35}
\begin{aligned}
r_{t}^{2}&=2\E\left[(1-\pi_{\rho}(\nu Z_{1}))\phi'\left(\prox_{\gamma_{t}\phi}(\nu\alpha_{t}Z_{1}+\sigma_{t}Z_{2})\right)^{2}\right]\geq2\E\left[\phi''(\nu Z_{1})\phi'\left(\prox_{\gamma_{t}\phi}(\nu\alpha_{t}Z_{1}+\sigma_{t}Z_{2})\right)^{2}\right]\\&\geq\frac{2\left(\E\left[\phi''(\nu Z_{1})\phi'\left(\prox_{\gamma_{t}\phi}(\nu\alpha_{t}Z_{1}+\sigma_{t}Z_{2})\right)^{2}\right]\right)^{2}}{\E\left[\phi''(\nu Z_{1})\right]}=\frac{2(\frac{1}{4}h(\nu))^{2}}{h(\nu)}=\frac{h(\nu)}{8}\end{aligned}
\end{equation}
for every $t\leq T$. We also have the universal upper bound $0\leq\theta_t\leq(1-2\rho)h(\nu).$ Indeed $\prox_{\gamma_t\phi}(\nu\alpha_{t}Z_{1}+\sigma_{t}Z_{2})\leq \nu\alpha_tZ_1+\sigma_tZ_2$ by the prox relation \cref{eq:prox_relation_large_T}, and from symmetry
\[
2\E[\pi_\rho'(\nu Z_1)\phi'(\nu\alpha_tZ_1+\sigma_tZ_2)]=(1-2\rho)h(\nu).
\]
The bound in \cref{eq:tail_log_lower} implies two useful events.
First,
\begin{align}
\PP\left(\forall j\leq T:|Z_{j}|\leq\frac{\lambda\sqrt{\delta}}{r_{j}}-\frac{\log\left[\left(\sum_{k=1}^{T}2\Phi\left(-\frac{\lambda\sqrt{\delta}}{r_{k}}\right)\right)^{-1}\right]}{4\sqrt{\log\left(2T\right)+\log\left[\left(\sum_{k=1}^{T}2\Phi\left(-\frac{\lambda\sqrt{\delta}}{r_{k}}\right)\right)^{-1}\right]}}\right)\geq\frac{1}{2}.
\label{eq:additive_gaussian_gap}
\end{align}
To verify this, for $q>0$ and $a\in(0,1)$, \cref{lem:5} gives $\Phi(-(x-q))\leq\exp\left(\frac{aq^2}{2(1-a)}\right)\Phi(-x)^a.$
Note that,
\begin{align}
\label{eq:8}  
\PP\left(\exists j\leq T:\left|Z_{j}\right|>\frac{\lambda\sqrt{\delta}}{r_{j}}-q\right)\leq\sum_{j=1}^{T}2\Phi\left(-\left(\frac{\lambda\sqrt{\delta}}{r_{j}}-q\right)\right)
\end{align}
Summing over $j$ and using Jensen's inequality,
\[
T\frac{1}{T}\sum_{j=1}^{T}2\Phi\left(-\frac{\lambda\sqrt{\delta}}{r_{j}}\right)^{a}\leq T\left(\frac{1}{T}\sum_{j=1}^{T}2\Phi\left(-\frac{\lambda\sqrt{\delta}}{r_{j}}\right)\right)^{a}=T^{1-a}\left(\sum_{j=1}^{T}2\Phi\left(-\frac{\lambda\sqrt{\delta}}{r_{j}}\right)\right)^{a},
\]
From that inequality, starting from the quantity in \cref{eq:8},
\begin{align*}
&\sum_{j=1}^{T}2\Phi\left(-\left(\frac{\lambda\sqrt{\delta}}{r_{j}}-q\right)\right)\leq2\exp\left(\frac{aq^{2}}{2\left(1-a\right)}\right)\sum_{j=1}^{T}\Phi\left(-\frac{\lambda\sqrt{\delta}}{r_{j}}\right)^{a}\\&=2^{1-a}\exp\left(\frac{aq^{2}}{2\left(1-a\right)}\right)\sum_{j=1}^{T}2\Phi\left(-\frac{\lambda\sqrt{\delta}}{r_{j}}\right)^{a}\leq2^{1-a}\exp\left(\frac{aq^{2}}{2\left(1-a\right)}\right)T^{1-a}\left(\sum_{j=1}^{T}2\Phi\left(-\frac{\lambda\sqrt{\delta}}{r_{j}}\right)\right)^{a}
\end{align*}
Taking logarithm,
\begin{align*}
&\log\left(2^{1-a}\exp\left(\frac{aq^{2}}{2\left(1-a\right)}\right)T^{1-a}\left(\sum_{j=1}^{T}2\Phi\left(-\frac{\lambda\sqrt{\delta}}{r_{j}}\right)\right)^{a}\right)\\&=\left(1-a\right)\log\left(2T\right)+\frac{aq^{2}}{2\left(1-a\right)}+a\log\left(\sum_{j=1}^{T}2\Phi\left(-\frac{\lambda\sqrt{\delta}}{r_{j}}\right)\right)\\&=\log\left(\sum_{j=1}^{T}2\Phi\left(-\frac{\lambda\sqrt{\delta}}{r_{j}}\right)\right)+\left(1-a\right)\left(\log\left(2T\right)-\log\left(\sum_{j=1}^{T}2\Phi\left(-\frac{\lambda\sqrt{\delta}}{r_{j}}\right)\right)\right)+\frac{aq^{2}}{2\left(1-a\right)}
\end{align*}
this is minimized with respect to $a$, if $1-a=\frac{q}{\sqrt{2\left(\log\left(2T\right)+\log\left(\left(\sum_{j=1}^{T}2\Phi\left(-\frac{\lambda\sqrt{\delta}}{r_{j}}\right)\right)^{-1}\right)\right)}}$
therefore the logarithm of the resulting upper bound is at most
\begin{align*} 
&\log\left(\PP\left(\exists j\leq T:\left|Z_{j}\right|>\frac{\lambda\sqrt{\delta}}{r_{j}}-q\right)\right)
\\& \leq -\log\left[\left(\sum_{j=1}^T2\Phi\left(-\frac{\lambda\sqrt\delta}{r_j}\right)\right)^{-1}\right]-\frac{q^2}{2}+
q\sqrt{2\left(\log\left(2T\right)+\log\left[\left(\sum_{j=1}^T2\Phi\left(-\frac{\lambda\sqrt\delta}{r_j}\right)\right)^{-1}\right]\right)}.
\end{align*}
Choosing $q$ equal to the quantity in \cref{eq:additive_gaussian_gap} makes this at most $-\left(1-\frac{\sqrt2}{4}\right)\log\left[\left(\sum_{j=1}^T2\Phi\left(-\frac{\lambda\sqrt\delta}{r_j}\right)\right)^{-1}\right],$
which tends to $-\infty$ by \cref{eq:tail_log_lower} as $s\to0.$ Therefore, for a sufficiently small $s$, $\PP\left(\exists j\leq T:\left|Z_{j}\right|>\frac{\lambda\sqrt{\delta}}{r_{j}}-q\right)\leq \frac{1}{2}$, taking the complement of the event proves the claim.
Second,
\begin{align}
\PP\left(\forall j\leq T:|Z_{j}|\leq\left[1-\frac{\log\left[\left(\sum_{k=1}^{T}2\Phi\left(-\frac{\lambda\sqrt{\delta}}{r_{k}}\right)\right)^{-1}\right]}{8\left(\log\left(2T\right)+\log\left[\left(\sum_{k=1}^{T}2\Phi\left(-\frac{\lambda\sqrt{\delta}}{r_{k}}\right)\right)^{-1}\right]\right)}\right]\frac{\lambda\sqrt{\delta}}{r_{j}}\right)\geq\frac{1}{2}.
\label{eq:multiplicative_gaussian_gap}
\end{align}
To show this we prove the following identity, for every $x\geq0$ and $\frac{7}{8}\leq a<1$, $2\Phi(-ax)\leq C\left(2\Phi(-x)\right)^{a^2}$ for a universal constant $C>0$. Indeed, by Mill's inequality $\frac{\varphi(y)}{1+y}\underset{y\geq 1}{\leq}\frac{y\varphi(y)}{1+y^2}\leq\Phi(-y)\leq\frac{\varphi(y)}{y}\underset{y\geq 1}{\leq}\frac{2\varphi(y)}{1+y}.$ For $0\leq y\leq1$, $2\Phi(-y)\geq2\Phi(-1)>0$, so the same inequality follows after increasing the universal constant $C$. Apply the upper bound with $y=ax$, $2\Phi(-ax)\leq C\frac{e^{-\frac{a^2x^2}{2}}}{1+ax},$ and apply the lower bound at $y=x$ and raise it to the power of $a^2$, $\left(2\Phi(-x)\right)^{a^2}\geq c^{a^2}\frac{e^{-\frac{a^2x^2}{2}}}{(1+x)^{a^2}}.$ Consequently, $\frac{2\Phi(-ax)}{\left(2\Phi(-x)\right)^{a^2}}\leq Cc^{-a^2}\frac{(1+x)^{a^2}}{1+ax}.$
Since $a\leq1$, $(1+x)^{a^2}\leq1+x,$ and since $a\geq\frac{7}{8}$, $1+ax\geq a(1+x)\geq\frac78(1+x).$ Therefore, $\frac{(1+x)^{a^2}}{1+ax}\leq\frac{1}{a}\leq\frac87.$ Also $c^{-a^2}$ is uniformly bounded for $a\in[\frac{7}{8},1]$. Hence $2\Phi(-ax)\leq C\left(2\Phi(-x)\right)^{a^2}$. Summing and using Jensen's inequality,
\[
\sum_{j=1}^{T}2\Phi\left(-a\frac{\lambda\sqrt{\delta}}{r_{j}}\right)\leq CT\left(\frac{1}{T}\sum_{j=1}^{T}2\Phi\left(-\frac{\lambda\sqrt{\delta}}{r_{j}}\right)^{a^{2}}\right)\leq CT^{1-a^{2}}\left(\sum_{j=1}^{T}2\Phi\left(-\frac{\lambda\sqrt{\delta}}{r_{j}}\right)\right)^{a^{2}}
\]
With $a$ equal to the multiplicative factor in
\cref{eq:multiplicative_gaussian_gap}, the logarithm is at most $-\frac34\log\left[\left(\sum_{j=1}^T2\Phi\left(-\frac{\lambda\sqrt\delta}{r_j}\right)\right)^{-1}\right],$  which tends to $-\infty$ by \cref{eq:tail_log_lower}. Once again as before, a union bound proves \cref{eq:multiplicative_gaussian_gap} (note that $\frac{7}{8}\leq a\leq1$).

We now split into two cases.

\textbf{Case 1: $s\left(\log\left(\frac{1}{s}\right)+\log\left(2T\right)\right)\geq1$}.
On the event (\ref{eq:additive_gaussian_gap}), using
$r_t\geq\sqrt{\frac{h(\nu)}{8}}$ (see \cref{eq:35}),
\[
\frac{r_{j}}{\sqrt{\delta}}|Z_{j}|\leq\lambda-\frac{\sqrt{h(\nu)}\log\left[\left(\sum_{k=1}^{T}2\Phi\left(-\frac{\lambda\sqrt{\delta}}{r_{k}}\right)\right)^{-1}\right]}{4\sqrt{8\delta}\sqrt{\log\left(2T\right)+\log\left[\left(\sum_{k=1}^{T}2\Phi\left(-\frac{\lambda\sqrt{\delta}}{r_{k}}\right)\right)^{-1}\right]}}.
\]
Using $0\leq\theta_t\leq(1-2\rho)h(\nu)$, every positive active coordinate satisfying
\[
0<w^{\star}\leq \frac{1}{4\sqrt{8\delta h(\nu)}(1-2\rho)}\frac{\log\left[\left(\sum_{k=1}^{T}2\Phi\left(-\frac{\lambda\sqrt{\delta}}{r_{k}}\right)\right)^{-1}\right]}{\sqrt{\log\left(2T\right)+\log\left[\left(\sum_{k=1}^{T}2\Phi\left(-\frac{\lambda\sqrt{\delta}}{r_{k}}\right)\right)^{-1}\right]}}
\]
satisfies $\left|\theta_{j}w^{\star}+\frac{r_{j}}{\sqrt{\delta}}Z_{j}\right|<\lambda$ for every $j$.
Thus every projection interval contains zero. Since $w_0=0$, then $w_{j}=0$ for all $j\leq T$. Conditionally on coordinate being active, $w^{\star}=\frac{\nu}{\sqrt s}G.$
Therefore, since we know this event happens with probability of at least $\frac{1}{2}$ by \cref{eq:additive_gaussian_gap},
\begin{align*}
&\nu^{2}\left(\alpha_{t}-r\right)^{2}+\sigma_{t}^{2}\stackrel{\mathrm{\ref{eq:7}}}{=}\E\left[\left(w_{t}-rw^{\star}\right)^{2}\right]\\&\geq\frac{s}{2}\E\left[r^{2}\left(\frac{\nu}{\sqrt{s}}G\right)^{2}\mathbf{1}_{\left\{ 0<G<\frac{\sqrt{s}}{4\nu\sqrt{8\delta h(\nu)}(1-2\rho)}\frac{\log\left[\left(\sum_{k=1}^{T}2\Phi\left(-\frac{\lambda\sqrt{\delta}}{r_{k}}\right)\right)^{-1}\right]}{\sqrt{\log\left(2T\right)+\log\left[\left(\sum_{k=1}^{T}2\Phi\left(-\frac{\lambda\sqrt{\delta}}{r_{k}}\right)\right)^{-1}\right]}}\right\} }\right]\\&\stackrel{[1]}{\geq}\frac{r^{2}\nu^{2}}{2}\min\Bigg\{-\varphi\left(1\right)+\Phi\left(1\right)-\Phi\left(0\right),\\&\hspace{4em}\frac{\varphi\left(1\right)}{3}\biggprn{\frac{\sqrt{s}}{4\nu\sqrt{8\delta h(\nu)}(1-2\rho)}\frac{\log\left[\left(\sum_{k=1}^{T}2\Phi\left(-\frac{\lambda\sqrt{\delta}}{r_{k}}\right)\right)^{-1}\right]}{\sqrt{\log\left(2T\right)+\log\left[\left(\sum_{k=1}^{T}2\Phi\left(-\frac{\lambda\sqrt{\delta}}{r_{k}}\right)\right)^{-1}\right]}}}^{3}\Bigg\}
\end{align*}
where [1] follows by,
\begin{align*}
&\E\left[G^{2}\mathbf{1}_{\{0\leq G\leq u\}}\right]=\int_{0}^{u}z^{2}\varphi\left(z\right)\dd z\\&\geq\begin{cases}
\varphi\left(1\right)\int_{0}^{u}z^{2}\dd z=\frac{\varphi\left(1\right)}{3}u^{3} & 0<u\leq1\\
\int_{0}^{1}z^{2}\varphi\left(z\right)\dd z=\left[-z\varphi\left(z\right)\right]_{0}^{1}+\int_{0}^{1}\varphi\left(z\right)\dd z=-\varphi\left(1\right)+\Phi\left(1\right)-\Phi\left(0\right) & u\geq1
\end{cases}
\end{align*}
Recall that we attained a lower bound on $\log\left[\left(\sum_{k=1}^{T}2\Phi\left(-\frac{\lambda\sqrt{\delta}}{r_{k}}\right)\right)^{-1}\right]$ in \cref{eq:tail_log_lower}.
To replace this quantity with the lower bound, we define $f(x)=\frac{x^3}{\left(\log\left(2T\right)+x\right)^{\frac{3}{2}}}$, then 
\begin{align*}
f'(x)=\frac{3x^{2}\left(\log\left(2T\right)+x\right)^{\frac{3}{2}}-\frac{3}{2}x^{3}\left(\log\left(2T\right)+x\right)^{\frac{1}{2}}}{\left(\log\left(2T\right)+x\right)^{3}}=\frac{3x^{2}\log\left(2T\right)+\frac{3}{2}x^{3}}{\left(\log\left(2T\right)+x\right)^{\frac{5}{2}}}\underset{x>0}{>}0
\end{align*}
so the function is increasing for all $x>0$, this justifies the replacement with the lower bound. For a sufficiently small $s$, the constant in the lower bound is harmless. Hence 
\[
\nu^{2}\left(\alpha_{t}-r\right)^{2}+\sigma_{t}^{2}\geq\frac{r^{2}\nu^{2}\varphi\left(1\right)}{6}\left(\frac{\sqrt{s}}{\nu}\frac{1}{4\sqrt{8\delta h(\nu)}(1-2\rho)}\frac{\frac{1}{2}\log\left(\frac{1}{s}\right)}{\sqrt{\log\left(2T\right)+\frac{1}{2}\log\left(\frac{1}{s}\right)}}\right)^{3}
\]
But in the present case, $s\left(\log\left(\frac{1}{s}\right)+\log\left(2T\right)\right)\geq1,$ so
\[
s^{\frac{3}{2}}\frac{\log^{3}\left(\frac{1}{s}\right)}{\left(\log\left(\frac{1}{s}\right)+\log\left(2T\right)\right)^{\frac{3}{2}}}=s\frac{\log^{3}\left(\frac{1}{s}\right)}{\left(\log\left(\frac{1}{s}\right)+\log\left(2T\right)\right)^{2}}\sqrt{s\left(\log\left(\frac{1}{s}\right)+\log\left(2T\right)\right)}\geq s\frac{\log^3\left(\frac{1}{s}\right)}{\left(\log\left(\frac{1}{s}\right)+\log\left(2T\right)\right)^2}.
\]
Therefore,
\[
\nu^{2}\left(\alpha_{t}-r\right)^{2}+\sigma_{t}^{2}\geq\frac{r^{2}\varphi\left(1\right)}{384\sqrt{8}\nu}\frac{1}{\delta^{\frac{3}{2}}(1-2\rho)^{3}}\frac{s\log^{3}\left(\frac{1}{s}\right)}{\left(\log\left(\frac{1}{s}\right)+\log\left(2T\right)\right)^{2}},
\]
contradicting \cref{eq:contradiction_temporal} if $c_\star$ is small
enough.

\textbf{Case 2 $s\left(\log\left(\frac{1}{s}\right)+\log\left(2T\right)\right)<1$.}

We first show that $\lambda\sqrt s$ must be small.
On the event (\ref{eq:multiplicative_gaussian_gap}), every positive active
coordinate satisfying
\[
0<w^{\star}\leq\frac{\lambda}{(1-2\rho)h(\nu)}\frac{\log\left[\left(\sum_{k=1}^{T}2\Phi\left(-\frac{\lambda\sqrt{\delta}}{r_{k}}\right)\right)^{-1}\right]}{8\left(\log\left(2T\right)+\log\left[\left(\sum_{k=1}^{T}2\Phi\left(-\frac{\lambda\sqrt{\delta}}{r_{k}}\right)\right)^{-1}\right]\right)}
\]
remains identically zero, because $0\leq\theta_t\leq(1-2\rho)h(\nu)$ and
\cref{eq:multiplicative_gaussian_gap} imply
$\left|\theta_{j}w^{\star}+\frac{r_{j}}{\sqrt{\delta}}Z_{j}\right|\leq\lambda$ for every $j$.
Integrating $G^2$ and repeating the same calculations from above
\begin{equation} 
\begin{aligned}
&\nu^{2}\left(\alpha_{t}-r\right)^{2}+\sigma_{t}^{2}=\E\left[\left(w_{t}-rw^{\star}\right)^{2}\right]\\&\geq
\frac{s}{2}\E\left[r^{2}\left(\frac{\nu}{\sqrt{s}}G\right)^{2}\mathbf{1}_{\left\{0<G<\frac{\sqrt s}{\nu}\frac{\lambda}{(1-2\rho)h(\nu)}\frac{\log\left[\left(\sum_{k=1}^{T}2\Phi\left(-\frac{\lambda\sqrt{\delta}}{r_{k}}\right)\right)^{-1}\right]}{8\left(\log\left(2T\right)+\log\left[\left(\sum_{k=1}^{T}2\Phi\left(-\frac{\lambda\sqrt{\delta}}{r_{k}}\right)\right)^{-1}\right]\right)}\right\}}\right]\\&\geq \frac{r^{2}\nu^{2}}{2}\min\Bigg\{-\varphi\left(1\right)+\Phi\left(1\right)-\Phi\left(0\right),\\&\hspace{4em}\frac{\varphi(1)}{3}\left(\frac{\lambda\sqrt{s}}{\nu(1-2\rho)h(\nu)}\frac{\log\left[\left(\sum_{k=1}^{T}2\Phi\left(-\frac{\lambda\sqrt{\delta}}{r_{k}}\right)\right)^{-1}\right]}{8\left(\log(2T)+\log\left[\left(\sum_{k=1}^{T}2\Phi\left(-\frac{\lambda\sqrt{\delta}}{r_{k}}\right)\right)^{-1}\right]\right)}\right)^{3}.\Bigg\}.
\end{aligned}
\label{eq:weak_active_lambda}
\end{equation}
Since the left side tends to zero, the constant term is harmless. Using \cref{eq:tail_log_lower,eq:contradiction_temporal} and the fact that $x\mapsto \frac{x}{\log(2T)+x}$ is increasing, we obtain
\[
(\lambda\sqrt{s})^{3}\leq\frac{49152\nu(1-2\rho)^{3}h(\nu)^{3}}{r^{2}\varphi(1)}c_{\star}s\left(\log\left(\frac{1}{s}\right)+\log(2T)\right).
\]
Therefore,
\begin{align}
\lambda\sqrt{s}\leq(1-2\rho)h(\nu)\left(\frac{49152\nu}{r^{2}\varphi(1)}\right)^{\frac{1}{3}}c_{\star}^{\frac{1}{3}}\left[s\left(\log\left(\frac{1}{s}\right)+\log(2T)\right)\right]^{\frac{1}{3}}.
\label{eq:lambdamall}
\end{align}
We next control the normalized feature channel directly, that is, $\left|\frac{w_{t}}{\sigma_t\tau_t}-\theta_tw^{\star}-\frac{r_t}{\sqrt\delta}Z_t\right|\leq\lambda.$
Since $w^{\star}$ and $Z_t$ are independent and $Z_t$ has mean 0, multiplying by $|w^{\star}|$ and taking expected value with $|\E[X]|\leq\E[|X|]$ we get,
\begin{align}
\left|\frac{\E[w^{\star}w_{t}]}{\nu^{2}\sigma_{t}\tau_{t}}-\theta_{t}\right|\leq\frac{\lambda}{\nu^{2}}\E|w^{\star}|=\frac{\sqrt{\frac{2}{\pi}}}{\nu}\lambda\sqrt{s}.
\label{eq:normalized_mean_temporal}
\end{align}

Moreover,
\[
\frac{\E[w_{t}^{2}]}{(\sigma_{t}\tau_{t})^{2}}-\nu^{2}\left(\frac{\E[w^{\star}w_{t}]}{\nu^{2}\sigma_{t}\tau_{t}}\right)^{2}=\E\left[\left(\frac{w_{t}}{\sigma_{t}\tau_{t}}-\frac{\E[w^{\star}w_{t}]}{\nu^{2}\sigma_{t}\tau_{t}}w^{\star}\right)^{2}\right].
\]
The coefficient $\frac{\E[w^{\star} w_{t}]}{\nu^2\sigma_t\tau_t}$ is the $L^2$-projection coefficient of $\frac{w_{t}}{\sigma_t\tau_t}$ onto the span of $w^{\star}$.\footnote{Indeed, for $f(a)=\E\left[\left(\frac{w_{t}}{\sigma_t\tau_t}-aw^{\star}\right)^2\right],$
we have $f'(a)=-2\E\left[w^{\star}\left(\frac{w_{t}}{\sigma_t\tau_t}-aw^{\star}\right)\right].$
Thus the unique stationary point satisfies $a=\frac{\E[w^{\star} w_{t}]}{(\sigma_t\tau_t)\E[(w^{\star})^2]}=\frac{\E[w^{\star} w_{t}]}{\nu^2\sigma_t\tau_t}.$
Since $f''(a)=2\E[(w^{\star})^2]=2\nu^2>0$, this stationary
point is the unique minimizer.}
Consequently,
\begin{align}
\frac{\E[w_{t}^2]}{(\sigma_t\tau_t)^2}-\nu^2\left(\frac{\E[w^{\star} w_{t}]}{\nu^2\sigma_t\tau_t}\right)^2\leq\E\left[\left(\frac{w_{t}}{\sigma_t\tau_t}-\theta_t w^{\star}\right)^2\right].
\label{eq:projection_variance_temporal}
\end{align}
We now bound the right-hand side. On active coordinates,
\[
\left|\frac{w_{t}}{\sigma_t\tau_t}-\theta_t w^{\star}\right|\leq\lambda+\frac{r_t}{\sqrt\delta}|Z_t|\leq\lambda+\frac{|Z_t|}{\sqrt\delta},
\]
where we used $r_t\leq1$ (see conclusion beneath \cref{eq:34}). Since $\PP(w^{\star}\neq0)=s$ and the activity indicator is independent of $Z_t$,
\begin{align}
\E\left[\left(\frac{w_{t}}{\sigma_{t}\tau_{t}}-\theta_{t}w^{\star}\right)^{2}\mathbf{1}_{\{w^{\star}\neq0\}}\right]\leq s\E\left[\left(\lambda+\frac{|Z_{t}|}{\sqrt{\delta}}\right)^{2}\right]=s\left(\lambda^{2}+2\lambda\sqrt{\frac{2}{\pi\delta}}+\frac{1}{\delta}\right)\leq2s\left(\lambda^{2}+\frac{1}{\delta}\right).
\label{eq:active_normalized_variance}
\end{align}
On null coordinates that were already nonzero at time $t-1$, the event $\{w^{\star}=0,w_{t-1}\neq0\}$ depends only on the previous Gaussian variables and is therefore independent of the fresh $Z_t$. Hence
\begin{align}
&\E\left[\left(\frac{w_{t}}{\sigma_t\tau_t}\right)^2\mathbf{1}_{\{w^{\star}=0,w_{t-1}\neq0\}}\right]\leq2(1-s)\PP(w_{t-1}\neq0\mid w^{\star}=0)\left(\lambda^2+\frac{1}{\delta}\right).
\label{eq:old_null_variance}
\end{align}
For a newly activated null coordinate, $w^{\star}=0$ and $w_{t-1}=0$, we get $\frac{w_{t}}{\sigma_t\tau_t}=\mathrm{sign}(Z_t)\left(\frac{r_t}{\sqrt\delta}|Z_t|-\lambda\right)_+$  by the soft-thresholding formula. Therefore,
\begin{align*}
\E\left[\left(\frac{w_{t}}{\sigma_t\tau_t}\right)^2\mathbf{1}_{\{w^{\star}=0,w_{t-1}=0,w_{t}\neq0\}}\right]\leq2(1-s)\frac{r_t^2}{\delta}\PP(w_{t-1}=0\mid w^{\star}=0)\int_{\frac{\lambda\sqrt\delta}{r_t}}^{\infty}\left(z-\frac{\lambda\sqrt\delta}{r_t}\right)^2\varphi(z)\dd z.
\end{align*}
For every $x\geq0$ using integration by parts,
\begin{align*}
\int_x^\infty(z-x)^2\varphi(z)\dd z=(1+x^2)\Phi(-x)-x\varphi(x)
=\Phi(-x)+x\left(x\Phi(-x)-\varphi(x)\right)\leq\Phi(-x),
\end{align*}
where we used $x\Phi(-x)\leq\varphi(x)$ because $f'(x)=-\Phi(-x)<0$ where $f(x)=\varphi(x)-x\Phi(-x)$ with $\lim_{x\to\infty}f(x)=0$. Since $r_t\leq1$ (see conclusion beneath \cref{eq:34}),
\begin{align}
&\E\left[\left(\frac{w_{t}}{\sigma_t\tau_t}\right)^2\mathbf{1}_{\{w^{\star}=0,w_{t-1}=0,w_{t}\neq0\}}\right]\leq\frac{2(1-s)}{\delta}\PP(w_{t-1}=0,w_{t}\neq0\mid w^{\star}=0).
\label{eq:new_null_variance}
\end{align}

Since null support is persistent, the events $\{w_{t-1}\neq0\}$ and $\{w_{t-1}=0,w_{t}\neq0\}$ partition $\{w_{t}\neq0\}$ almost surely. Combining
\cref{eq:projection_variance_temporal,eq:active_normalized_variance,eq:old_null_variance,eq:new_null_variance}, we obtain
\begin{align}
\frac{\E[w_{t}^2]}{(\sigma_t\tau_t)^2}-\nu^2\left(\frac{\E[w^{\star}w_{t}]}{\nu^2\sigma_t\tau_t}\right)^2\leq2\left[s+\PP(w_{t}\neq0\mid w^{\star}=0)\right]\left(\lambda^2+\frac{1}{\delta}\right).
\label{eq:normalized_variance_temporal}
\end{align}
By the scalar identities $\nu^2\alpha_t=\E[w^{\star} w_{t}]$ and $\nu^2\alpha_t^2+\sigma_t^2=\E[w_{t}^2]$, the left-hand side is exactly $\frac{1}{\tau_t^2}.$
From \cref{eq:null_quantitative_short}, for $s\leq\frac{1}{2}$,
\[
\nu^2(\alpha_t-r)^2+\sigma_t^2\geq\frac{3}{327680\delta}\frac{\PP(w_{t}\neq0\mid w^{\star}=0)}{\log\left(\frac{4t}{\PP(w_{t}\neq0\mid w^{\star}=0)}\right)}.
\]
Under the contradiction assumption,
\[
\nu^2(\alpha_t-r)^2+\sigma_t^2<c_\star s\frac{\log^3\left(\frac{1}{s}\right)}{\left(\log\left(\frac{1}{s}\right)+\log\left(2t\right)\right)^2}.
\]
If $\PP(w_{t}\neq0\mid w^{\star}=0)\geq s$, then
\[
\log\left(\frac{4t}{\PP(w_{t}\neq0\mid w^{\star}=0)}\right)\leq\log\left(\frac{4t}{s}\right)\leq2\left(\log\left(\frac{1}{s}\right)+\log\left(2t\right)\right),
\]
and hence
\[
\PP(w_{t}\neq0\mid w^{\star}=0)\leq c_{\star}s\frac{655360\delta\log^{3}\left(\frac{1}{s}\right)}{3\left(\log\left(\frac{1}{s}\right)+\log\left(2t\right)\right)}
\]
If instead $\PP(w_{t}\neq0\mid w^{\star}=0)<s$, this probability is already bounded by $s$. Since null support is persistent (almost surely, unless the endpoints are exactly 0), for every
$t\leq T$,
\begin{align}
\PP(w_{t}\neq0\mid w^{\star}=0)\leq s+\frac{655360}{3}\delta c_\star s\frac{\log^3\left(\frac{1}{s}\right)}{\log\left(\frac{1}{s}\right)+\log\left(2T\right)}\underset{\log\left(\frac{1}{s}\right)+\log\left(2T\right)\geq\log\left(\frac{1}{s}\right)}{\leq} s+\frac{655360}{3}\delta c_\star s\log^2\left(\frac{1}{s}\right).
\label{eq:null_support_explicit}
\end{align}
Moreover, $\PP(w_{t}\neq w_{t-1})\leq s+\PP(w_{t}\neq0\mid w^{\star}=0),$ because an update can occur either on an active coordinate, whose probability is $s$, or on a null coordinate that is nonzero at time $t$ ($\{w_{t}\neq w_{t-1},w^{\star}=0\}\subseteq \{w_{t}\neq 0,w^{\star}=0\},$ almost surely). From the lower bound in \cref{eq:weak_active_lambda},
\[
c_{\star}s\frac{6\log^{3}\left(\frac{1}{s}\right)}{r^{2}\nu^{2}\varphi(1)\left(\log\left(\frac{1}{s}\right)+\log\left(2T\right)\right)^{2}}\geq\left(\frac{\lambda\sqrt{s}}{(1-2\rho)\nu h(\nu)}\frac{\log\left(\frac{1}{s}\right)}{16\left(\log\left(2T\right)+\log\left(\frac{1}{s}\right)\right)}\right)^{3}
\]
we obtain
\begin{align}
\lambda\sqrt{s}\leq(1-2\rho)h(\nu)\left(\frac{49152\nu}{r^{2}\varphi(1)}\right)^{\frac{1}{3}}c_{\star}^{\frac{1}{3}}\left[s\left(\log\left(\frac{1}{s}\right)+\log(2T)\right)\right]^{\frac{1}{3}}.
\label{eq:lambda_explicit}
\end{align}
Since in the present case $s\left(\log\left(\frac{1}{s}\right)+\log\left(2T\right)\right)<1.$  \cref{eq:lambda_explicit} implies that $\lambda\sqrt{s}\lesssim c_\star^{\frac{1}{3}}$ (up to constants depending on the problem parameters). Thus, by choosing $c_\star$ sufficiently small, $\lambda\sqrt{s}$ is uniformly as small as needed. We henceforth take $s$ sufficiently small that,
\[
\sqrt{s}\left(2+\frac{655360}{3}\delta c_{\star}\log^{2}\left(\frac{1}{s}\right)\right)\leq\min\left\{ 1,\delta\right\} \Phi^{-1}\left(\frac{3}{4}\right)\sqrt{\frac{h(\nu)}{8\delta}}.
\]
Then,
\begin{align}
\label{eq:9} 
\PP(w_{t}\neq w_{t-1})\leq s\left(2+\frac{655360}{3}\delta c_{\star}\log^{2}\left(\frac{1}{s}\right)\right)\leq\lambda\sqrt{s}.
\end{align}
We also proved in \cref{eq:32,eq:33},
\begin{align}
\label{eq:bound1} 
0\leq(1-2\rho)h(\nu)-\theta_t\leq \frac{2(1-2\rho)}{\delta}\PP(w_t\neq w_{t-1})\leq 2(1-2\rho)\lambda\sqrt s,
\end{align}
and therefore since $\PP(w_{t}\neq w_{t-1})\leq\lambda\sqrt{s}$ by \cref{eq:9} so using \cref{eq:normalized_mean_temporal},
\begin{align}
\left|\frac{\E[w^{\star}w_{t}]}{\nu^{2}\sigma_{t}\tau_{t}}-(1-2\rho)h(\nu)\right|\leq\left(\frac{\sqrt{\frac{2}{\pi}}}{\nu}+2(1-2\rho)\right)\lambda\sqrt{s}.
\label{eq:normalized_mean_H0}
\end{align}
Using \cref{eq:normalized_variance_temporal,eq:null_support_explicit}, we obtain,
\begin{equation}
\begin{aligned}
\frac{\sigma_t^2}{(\sigma_t\tau_t)^2}&\leq2s\left(2+\frac{655360}{3}\delta c_{\star}\frac{\log^{3}\left(\frac{1}{s}\right)}{\log\left(\frac{1}{s}\right)+\log(2T)}\right)\left(\lambda^{2}+\frac{1}{\delta}\right)\\&=\left(2+\frac{655360}{3}\delta c_{\star}\frac{\log^{3}\left(\frac{1}{s}\right)}{\log\left(\frac{1}{s}\right)+\log(2T)}\right)\left(2(\lambda\sqrt{s})^{2}+\frac{2s}{\delta}\right)\leq4\lambda\sqrt{s}.
\end{aligned}
\label{eq:bound}
\end{equation}

Here we chose $\lambda\sqrt{s}\left(2+\frac{655360}{3}\delta c_{\star}\frac{\log^{3}\left(\frac{1}{s}\right)}{\log\left(\frac{1}{s}\right)+\log(2T)}\right)\leq1$ and $\frac{s\left(2+\frac{655360}{3}\delta c_{\star}\frac{\log^{3}\left(\frac{1}{s}\right)}{\log\left(\frac{1}{s}\right)+\log(2T)}\right)}{\delta}\leq\lambda\sqrt{s}$ for a sufficiently small $s$. We now prove a uniform upper bound on $\sigma_t\tau_t$. The lower bound follows directly from $\sigma_t\tau_tr_t\geq1$ (see below \cref{eq:22}) and $r_t\leq1$ (see conclusion beneath \cref{eq:34}), and therefore $\sigma_t\tau_t\geq1$. Since $(1-2\rho)h(\nu)=rh(\nu r)$ by \cref{lem:rel} and \[\nu rh(\nu r)=\nu r\int_{\R}\phi''(\nu rz)\varphi(z)\dd z\underset{u=\nu rz}{=}\frac{1}{\sqrt{2\pi}}\int_{\R}\phi''(u)e^{-\frac{u^{2}}{2(\nu r)^{2}}}\dd u<\frac{1}{\sqrt{2\pi}}\int_{\R}\phi''(u)\dd u=\frac{1}{\sqrt{2\pi}},\] we have $1-\nu(1-2\rho)h(\nu)\sqrt{2\pi}>0.$
For simplicity let $Q_0=\max\left\{1,\frac{32\log(2)\left(\frac{\nu(1-2\rho)}{2\sqrt{2\pi}}+\sqrt{\frac{2}{\pi e}}\right)\left(1+\nu(1-2\rho)h(\nu)\sqrt{2\pi}\right)}{\left(1-\nu(1-2\rho)h(\nu)\sqrt{2\pi}\right)\nu^2(1-2\rho)^2h(\nu)^2}\right\}.$
We take $s$ still smaller, if necessary, so that
\begin{align}
&\left(\frac{\sqrt{\frac{2}{\pi}}}{\nu}+2(1-2\rho)\right)\lambda\sqrt{s}\leq\min\left\{\frac{(1-2\rho)h(\nu)}{2},\frac{1-\nu(1-2\rho)h(\nu)\sqrt{2\pi}}{4\nu\sqrt{2\pi}}\right\},
\label{eq:normalized_mean_smallness}
\\&
\lambda\sqrt{s}\leq\min\left\{ \frac{\left(1-\nu(1-2\rho)h(\nu)\sqrt{2\pi}\right)^{2}}{128\pi},\frac{\nu^{2}(1-2\rho)^{2}h(\nu)^{2}}{8\left(\frac{\nu(1-2\rho)}{2\sqrt{2\pi}}+\sqrt{\frac{2}{\pi e}}\right)}\frac{1-\nu(1-2\rho)h(\nu)\sqrt{2\pi}}{1+\nu(1-2\rho)h(\nu)\sqrt{2\pi}}\right\},
\label{eq:normalized_variance_smallness}
\end{align}
These conditions are possible because we may pick $c_\star$ in \cref{eq:lambda_explicit} arbitrary small (depending on the problem parameters). Multiplying the sixth scalar equation by $\sigma_t\tau_t$ and using $\E[2\pi_\rho(\nu Z_1)-1]=0$ gives,
\begin{align}
1=\sigma_t\tau_t\E\left[\frac{2(1-\pi_\rho(\nu Z_1))\phi''\left(\prox_{\gamma_t\phi}(\nu\alpha_tZ_1+\sigma_tZ_2)
\right)}{1+\gamma_t\phi''\left(\prox_{\gamma_t\phi}(\nu\alpha_tZ_1+\sigma_tZ_2)\right)}\right].
\label{eq:sixth_times_q}
\end{align}
Condition on $Z_2=z_2$ and make the change of variables $\nu\alpha_tz_1+\sigma_tz_2=u+\gamma_t\phi'(u)$.
Then $\dd z_1=\frac{1+\gamma_t\phi''(u)}{\nu\alpha_t}\dd u,$ and hence the Jacobian exactly cancels the denominator in \cref{eq:sixth_times_q}. The function $z\mapsto2(1-\pi_\rho(\nu z))\varphi(z)$ is Lipschitz with
constant $\frac{\nu(1-2\rho)}{2\sqrt{2\pi}}+\sqrt{\frac{2}{\pi e}}.$ Indeed,
\begin{align*}  
\left|\frac{\dd}{\dd z}\left[2(1-\pi_\rho(\nu z))\varphi(z)\right]\right|=\left|-2\nu\underbrace{\pi_{\rho}'(\nu z)}_{=\left(1-2\rho\right)\phi''(\nu Z)}\varphi(z)-2(1-\pi_{\rho}(\nu z))z\varphi(z)\right|\leq\frac{\nu(1-2\rho)}{2}\varphi(0)+2\sup_{z\in\mathbb R}|z|\varphi(z),
\end{align*}

and $\varphi(0)=\frac{1}{\sqrt{2\pi}}$ while $\sup_z|z|\varphi(z)=\frac{1}{\sqrt{2\pi e}}$ we focus on $z>0$ due to the symmetry of $|z|\varphi(z).$ Its derivative is given by $\frac{\dd}{\dd z}z\varphi(z)=\frac{1}{\sqrt{2\pi}}e^{-\frac{z^2}{2}}(1-z^2),$ and a stationary point is $z=1$, which is also the maximum. Using this Lipschitz bound after the preceding change of variables gives
\begin{align}
&\left|\sigma_{t}\tau_{t}\E\left[\frac{2(1-\pi_{\rho}(\nu Z_{1}))\phi''\left(\prox_{\gamma_{t}\phi}(\nu\alpha_{t}Z_{1}+\sigma_{t}Z_{2})\right)}{1+\gamma_{t}\phi''\left(\prox_{\gamma_{t}\phi}(\nu\alpha_{t}Z_{1}+\sigma_{t}Z_{2})\right)}\right]-\frac{1}{\sqrt{2\pi\left[\nu^{2}\left(\frac{\alpha_{t}}{\sigma_{t}\tau_{t}}\right)^{2}+\frac{\sigma_{t}^{2}}{(\sigma_{t}\tau_{t})^{2}}\right]}}\right|\notag\\&\stackrel{[1]}{=}\frac{\sigma_{t}\tau_{t}}{\nu\alpha_{t}}\left|\int_{\mathbb{R}^{2}}\Bigg[2\left(1-\pi_{\rho}\left(\frac{u+\gamma_{t}\phi'(u)-\sigma_{t}z_{2}}{\alpha_{t}}\right)\right)\varphi\left(\frac{u+\gamma_{t}\phi'(u)-\sigma_{t}z_{2}}{\nu\alpha_{t}}\right)\right.\notag\\&\hspace{14em}\left.-2\left(1-\pi_{\rho}\left(-\frac{\sigma_{t}z_{2}}{\alpha_{t}}\right)\right)\varphi\left(-\frac{\sigma_{t}z_{2}}{\nu\alpha_{t}}\right)\Bigg]\phi''(u)\varphi(z_{2})\dd u\dd z_{2}\right|\notag\\&\leq\frac{\sigma_{t}\tau_{t}}{\nu\alpha_{t}}\left(\frac{\nu(1-2\rho)}{2\sqrt{2\pi}}+\sqrt{\frac{2}{\pi e}}\right)\int_{\mathbb{R}^{2}}\left|\frac{u+\gamma_{t}\phi'(u)}{\nu\alpha_{t}}\right|\phi''(u)\varphi(z_{2})\dd u\dd z_{2}\notag\\&\leq\frac{\sigma_{t}\tau_{t}}{\nu^{2}\alpha_{t}^{2}}\left(\frac{\nu(1-2\rho)}{2\sqrt{2\pi}}+\sqrt{\frac{2}{\pi e}}\right)\int_{\mathbb{R}^{2}}\left(|u|+\gamma_{t}\phi'(u)\right)\phi''(u)\varphi(z_{2})\dd u\dd z_{2}\notag\end{align}\begin{align}&=\frac{\sigma_{t}\tau_{t}}{\nu^{2}\alpha_{t}^{2}}\left(\frac{\nu(1-2\rho)}{2\sqrt{2\pi}}+\sqrt{\frac{2}{\pi e}}\right)\left[\int_{\mathbb{R}}|u|\phi''(u)\dd u+\gamma_{t}\int_{\mathbb{R}}\phi'(u)\phi''(u)\dd u\right]\notag\\&=\frac{\sigma_{t}\tau_{t}}{\nu^{2}\alpha_{t}^{2}}\left(\frac{\nu(1-2\rho)}{2\sqrt{2\pi}}+\sqrt{\frac{2}{\pi e}}\right)\left(2\log2+\frac{\gamma_{t}}{2}\right)\notag\\&=\frac{\frac{\nu(1-2\rho)}{2\sqrt{2\pi}}+\sqrt{\frac{2}{\pi e}}}{\nu^{2}\left(\frac{\alpha_{t}}{\sigma_{t}\tau_{t}}\right)^{2}}\left[\frac{2\log2}{\sigma_{t}\tau_{t}}+\frac{1}{2}\frac{\gamma_{t}}{\sigma_{t}\tau_{t}}\right].
\label{eq:q_change_variable_error}
\end{align}
where [1] follows by the change of variables $\nu\alpha_t z_1+\sigma_tz_2=u+\gamma_t\phi'(u),$ and from symmetry and $\pi_\rho(-u)+\pi_\rho(u)=1$,
\[
\int_{\mathbb{R}}2\left(1-\pi_{\rho}\left(-\frac{\sigma_{t}z_{2}}{\alpha_{t}}\right)\right)\varphi\left(\frac{\sigma_{t}z_{2}}{\nu\alpha_{t}}\right)\varphi\left(z_{2}\right)\dd z_{2}=\int_{\mathbb{R}}\varphi\left(\frac{\sigma_{t}z_{2}}{\nu\alpha_{t}}\right)\varphi\left(z_{2}\right)\dd z_{2}\]
Now since $\int_\mathbb R\phi''(u)\dd u=1$ and $\int_\mathbb R \varphi\left(\frac{\sigma_tz_2}{\nu\alpha_t}\right)\varphi\left(z_2\right)\dd z_2=\frac{\nu\alpha_t}{\sqrt{2\pi\left(\nu^2\alpha_t^2+\sigma_t^2\right)}}$, this explains the transition. By \cref{eq:normalized_mean_H0,eq:normalized_mean_smallness,eq:normalized_variance_smallness} with the fact that $\E[w^{\star}w_{t}]=\nu^2\alpha_t,$ imply
\begin{align*}
\sqrt{2\pi\left[\nu^{2}\left(\frac{\alpha_{t}}{\sigma_{t}\tau_{t}}\right)^{2}+\frac{\sigma_t^2}{(\sigma_t\tau_t)^2}\right]}&\leq\sqrt{2\pi}\nu\left|\frac{\alpha_{t}}{\sigma_{t}\tau_{t}}\right|+\sqrt{2\pi}\frac{\sigma_t}{\sigma_t\tau_t}\\&\leq\nu(1-2\rho)h(\nu)\sqrt{2\pi}+\left(2+2\nu\sqrt{2\pi}\left(1-2\rho\right)\right)\lambda\sqrt{s}+\sqrt{2\pi}\underbrace{\frac{\sigma_t}{\sigma_t\tau_t}}_{\stackrel{\mathrm{\ref{eq:bound}}}{\leq}2\sqrt{\lambda\sqrt{s}}}\\&\leq\nu(1-2\rho)h(\nu)\sqrt{2\pi}+\frac{1-\nu(1-2\rho)h(\nu)\sqrt{2\pi}}{2}=\frac{1+\nu(1-2\rho)h(\nu)\sqrt{2\pi}}{2}.
\end{align*}
Therefore,
\begin{align}
\frac{1}{\sqrt{2\pi\left[\nu^{2}\left(\frac{\alpha_{t}}{\sigma_{t}\tau_{t}}\right)^{2}+\frac{\sigma_t^2}{(\sigma_t\tau_t)^2}\right]}}\geq\frac{2}{1+\nu(1-2\rho)h(\nu)\sqrt{2\pi}}=1+\frac{1-\nu(1-2\rho)h(\nu)\sqrt{2\pi}}{1+\nu(1-2\rho)h(\nu)\sqrt{2\pi}}
\label{eq:limit_above_one}
\end{align}
Furthermore by \cref{eq:9}, $\frac{\gamma_{t}}{\sigma_{t}\tau_{t}}=\frac{1}{\delta}\PP(w_{t}\neq w_{t-1})\leq\lambda\sqrt{s}.$
Thus, if $\sigma_t\tau_t\geq Q_0$, then the lower bound of $\frac{\alpha_t}{\sigma_t\tau_t}\geq\frac{1}{2}(1-2\rho)h(\nu)$ (from the lower bound established, see \cref{eq:36} and conclusions below it), the estimate $\frac{\gamma_t}{\sigma_t\tau_t}\leq\lambda\sqrt{s}$, the definition of $Q_0$, \cref{eq:normalized_variance_smallness} imply that the absolute error in \cref{eq:q_change_variable_error} is at most $\frac{1}{2}\frac{1-\nu(1-2\rho)h(\nu)\sqrt{2\pi}}{1+\nu(1-2\rho)h(\nu)\sqrt{2\pi}}$.
Together with \cref{eq:limit_above_one}, the right-hand side of \cref{eq:sixth_times_q} would then be at least $1+\frac{1}{2}\frac{1-\nu(1-2\rho)h(\nu)\sqrt{2\pi}}{1+\nu(1-2\rho)h(\nu)\sqrt{2\pi}}>1,$
contradicting \cref{eq:sixth_times_q}. Hence $1\leq\sigma_t\tau_t\leq Q_0.$
From symmetry of $(Z_1,Z_2)$, evenness of $\phi''$, and $\pi_\rho(u)+\pi_\rho(-u)=1$,
\begin{align}
\label{eq:30}
2\E\left[(1-\pi_{\rho}(\nu Z_{1}))\phi''(\nu\alpha_{t}Z_{1}+\sigma_{t}Z_{2})\right]=\E\left[\phi''(\nu\alpha_{t}Z_{1}+\sigma_{t}Z_{2})\right]=h\left(\sqrt{\nu^{2}\alpha_{t}^{2}+\sigma_{t}^{2}}\right).
\end{align}
Therefore the sixth scalar (see \cref{eq:sixth_times_q}) with $\E[2(1-\pi_\rho(\nu Z_1))]=1$ equation gives,
\begin{align*}
&\left|\frac{1}{\sigma_{t}\tau_{t}}-h\left(\sqrt{\nu^{2}\alpha_{t}^{2}+\sigma_{t}^{2}}\right)\right|\leq\E\left[2(1-\pi_{\rho}(\nu Z_{1}))\left|\phi''\left(\prox_{\gamma_{t}\phi}(\nu\alpha_{t}Z_{1}+\sigma_{t}Z_{2})\right)-\phi''(\nu\alpha_{t}Z_{1}+\sigma_{t}Z_{2})\right|\right]\\&\hspace{10em}+\E\left[2(1-\pi_{\rho}(\nu Z_{1}))\phi''\left(\prox_{\gamma_{t}\phi}(\nu\alpha_{t}Z_{1}+\sigma_{t}Z_{2})\right)\frac{\gamma_{t}\phi''\left(\prox_{\gamma_{t}\phi}(\nu\alpha_{t}Z_{1}+\sigma_{t}Z_{2})\right)}{1+\gamma_{t}\phi''\left(\prox_{\gamma_{t}\phi}(\nu\alpha_{t}Z_{1}+\sigma_{t}Z_{2})\right)}\right].
\end{align*}
Since
$\phi''(u)=\phi'(u)(1-\phi'(u))$, the map $p\mapsto p(1-p)$ is $1$-Lipschitz on $[0,1]$, and from the Lipschitzness property we have $|\phi''(u)-\phi''(v)|\leq|\phi'(u)-\phi'(v)|.$
Using the \cref{eq:15},
\begin{align*}
0\leq\phi'(\nu\alpha_{t}Z_{1}+\sigma_{t}Z_{2})-\phi'\left(\prox_{\gamma_{t}\phi}(\nu\alpha_{t}Z_{1}+\sigma_{t}Z_{2})\right)\leq\frac{2\gamma_{t}\phi''\left(\prox_{\gamma_{t}\phi}(\nu\alpha_{t}Z_{1}+\sigma_{t}Z_{2})\right)}{1+\gamma_{t}\phi''\left(\prox_{\gamma_{t}\phi}(\nu\alpha_{t}Z_{1}+\sigma_{t}Z_{2})\right)},
\end{align*}
together with $\phi''\leq\frac{1}{4}$, gives,
\begin{equation}
\label{eq:qtemporal}
\begin{aligned}
\left|\frac{1}{\sigma_{t}\tau_{t}}-h\left(\sqrt{\nu^{2}\alpha_{t}^{2}+\sigma_{t}^{2}}\right)\right|&\leq\frac{9}{4}\E\left[2(1-\pi_{\rho}(\nu Z_{1}))\frac{\gamma_{t}\phi''\left(\prox_{\gamma_{t}\phi}(\nu\alpha_{t}Z_{1}+\sigma_{t}Z_{2})\right)}{1+\gamma_{t}\phi''\left(\prox_{\gamma_{t}\phi}(\nu\alpha_{t}Z_{1}+\sigma_{t}Z_{2})\right)}\right]\\&\stackrel{\mathrm{\ref{eq:32}}}{=}\frac{9}{4}\frac{\gamma_{t}}{\sigma_{t}\tau_{t}}=\frac{9}{4\delta}\PP(w_{t}\neq w_{t-1})\leq\frac{9}{4}\lambda\sqrt{s}.
\end{aligned}
\end{equation}
Next by \cref{eq:normalized_mean_smallness,eq:normalized_mean_H0}, 
$\frac{\alpha_t}{\sigma_t\tau_t}\geq
\frac{1}{2}(1-2\rho)h(\nu)$, so by \cref{eq:normalized_mean_H0,eq:normalized_mean_temporal,eq:bound1,eq:bound},
\begin{align*}\left|\frac{\sqrt{\nu^{2}\alpha_{t}^{2}+\sigma_{t}^{2}}}{\sigma_{t}\tau_{t}}-\nu(1-2\rho)h(\nu)\right|&\leq\left|\sqrt{\nu^{2}\frac{\alpha_{t}^{2}}{(\sigma_{t}\tau_{t})^{2}}+\frac{\sigma_{t}^{2}}{(\sigma_{t}\tau_{t})^{2}}}-\nu\frac{\alpha_{t}}{\sigma_{t}\tau_{t}}\right|+\nu\left|\frac{\alpha_{t}}{\sigma_{t}\tau_{t}}-\theta_{t}\right|+\nu\left|\theta_{t}-(1-2\rho)h(\nu)\right|\\&=\frac{\frac{\sigma_{t}^{2}}{(\sigma_{t}\tau_{t})^{2}}}{\sqrt{\nu^{2}\frac{\alpha_{t}^{2}}{(\sigma_{t}\tau_{t})^{2}}+\frac{\sigma_{t}^{2}}{(\sigma_{t}\tau_{t})^{2}}}+\nu\frac{\alpha_{t}}{\sigma_{t}\tau_{t}}}+\nu\left|\frac{\alpha_{t}}{\sigma_{t}\tau_{t}}-\theta_{t}\right|+\nu\left|\theta_{t}-(1-2\rho)h(\nu)\right|\\&\leq\frac{\frac{\sigma_{t}^{2}}{(\sigma_{t}\tau_{t})^{2}}}{\nu\frac{\alpha_{t}}{\sigma_{t}\tau_{t}}}+\nu\left|\frac{\alpha_{t}}{\sigma_{t}\tau_{t}}-\theta_{t}\right|+\nu\left|\theta_{t}-(1-2\rho)h(\nu)\right|\\&\leq\frac{2}{\nu(1-2\rho)h(\nu)}\frac{\sigma_{t}^{2}}{(\sigma_{t}\tau_{t})^{2}}+\nu\left|\frac{\alpha_{t}}{\sigma_{t}\tau_{t}}-\theta_{t}\right|+\nu\left|\theta_{t}-(1-2\rho)h(\nu)\right|\\&\leq\left[\frac{8}{\nu(1-2\rho)h(\nu)}+\sqrt{\frac{2}{\pi}}+2(1-2\rho)\nu\right]\lambda\sqrt{s}.
\end{align*}
Using $\sigma_t\tau_t\leq Q_0$,
\begin{align}
&\left|\sqrt{\nu^{2}\alpha_{t}^{2}+\sigma_{t}^{2}}-\nu(1-2\rho)h(\nu)\sigma_{t}\tau_{t}\right|\leq Q_{0}\left[\sqrt{\frac{2}{\pi}}+2(1-2\rho)\nu+\frac{8}{\nu(1-2\rho)h(\nu)}\right]\lambda\sqrt{s}.
\label{eq:radius_calibration_error}
\end{align}
The function $h$ is Lipschitz because $|h'(x)|=\left|\E[Z\phi'''(xZ)]\right|\leq\frac{1}{4}\E|Z|=\frac{1}{4}\sqrt{\frac{2}{\pi}}.$
Thus \cref{eq:qtemporal,eq:radius_calibration_error}, and
$\sigma_t\tau_t\leq Q_0$ give,
\begin{equation}
\label{eq:scalar_calibration_error}    
\begin{aligned}
&\left|1-\sigma_{t}\tau_{t}h\left(\nu(1-2\rho)h(\nu)\sigma_{t}\tau_{t}\right)\right|\\&\leq\sigma_{t}\tau_{t}\left|h\left(\sqrt{\nu^{2}\alpha_{t}^{2}+\sigma_{t}^{2}}\right)-h\left(\nu(1-2\rho)h(\nu)\sigma_{t}\tau_{t}\right)\right|+\sigma_{t}\tau_{t}\left|\frac{1}{\sigma_{t}\tau_{t}}-h\left(\sqrt{\nu^{2}\alpha_{t}^{2}+\sigma_{t}^{2}}\right)\right|\\&\leq\frac{\sigma_{t}\tau_{t}}{4}\sqrt{\frac{2}{\pi}}\left|\sqrt{\nu^{2}\alpha_{t}^{2}+\sigma_{t}^{2}}-\nu(1-2\rho)h(\nu)\sigma_{t}\tau_{t}\right|+\sigma_{t}\tau_{t}\left|\frac{1}{\sigma_{t}\tau_{t}}-h\left(\sqrt{\nu^{2}\alpha_{t}^{2}+\sigma_{t}^{2}}\right)\right|\\&\leq\left[\frac{9Q_{0}}{4}+\frac{Q_{0}^{2}}{4}\sqrt{\frac{2}{\pi}}\left(\sqrt{\frac{2}{\pi}}+2(1-2\rho)\nu+\frac{8}{\nu(1-2\rho)h(\nu)}\right)\right]\lambda\sqrt{s}.
\end{aligned}
\end{equation}

Now the function $a\longmapsto ah\left(\nu(1-2\rho)h(\nu)a\right)$
is strictly increasing since,
\[
\frac{\dd}{\dd a}\left[ah\left(\nu(1-2\rho)h(\nu)a\right)\right]=\E\left[Z^{2}\phi''\left(\nu(1-2\rho)h(\nu)aZ\right)\right]>0.
\]
Moreover, because $(1-2\rho)h(\nu)\stackrel{\mathrm{\ref{lem:rel}}}{=}rh(\nu r)$, then $\frac{rh(\nu r)}{(1-2\rho)h(\nu)}=1.$ Notice also that $\frac{r}{(1-2\rho)h(\nu)}=\frac{1}{h(\nu r)}\geq4,$ because $h(\nu r)=\E[\phi''(\nu rZ)]\leq\frac{1}{4}$, while $\sigma_t\tau_t\geq1$. Therefore the interval between $\sigma_t\tau_t$ and $\frac{r}{(1-2\rho)h(\nu)}$ is contained in $\left[1,\max\left\{Q_0,\frac{r}{(1-2\rho)h(\nu)}\right\}\right].$
For every $a$ in this interval by \cref{eq:14},
\[
\frac{\dd}{\dd a}\left[ah\left(\nu(1-2\rho)h(\nu)a\right)\right]=\left.\frac{\dd}{\dd x}\left[xh(\nu x)\right]\right|_{x=(1-2\rho)h(\nu)a}\geq\left(1+\nu(1-2\rho)h(\nu)\max\left\{ Q_{0},\frac{r}{(1-2\rho)h(\nu)}\right\}\right)^{-3}.
\]
The mean-value theorem and
\cref{eq:scalar_calibration_error} therefore give,
\[
\left|\sigma_{t}\tau_{t}-\frac{r}{(1-2\rho)h(\nu)}\right|\leq\frac{{\frac{9Q_{0}}{4}+\frac{Q_{0}^{2}}{4}\sqrt{\frac{2}{\pi}}\left(\sqrt{\frac{2}{\pi}}+2(1-2\rho)\nu+\frac{8}{\nu(1-2\rho)h(\nu)}\right)}}{{\left(1+\nu(1-2\rho)h(\nu)\max\left\{Q_{0},\frac{r}{(1-2\rho)h(\nu)}\right\}\right)^{-3}}}\lambda\sqrt{s}.
\]
Finally by \cref{eq:bound1} and,
\[
\left|\sigma_{t}\tau_{t}\theta_{t}-r\right|\leq\sigma_{t}\tau_{t}\left|\theta_{t}-(1-2\rho)h(\nu)\right|+(1-2\rho)h(\nu)\left|\sigma_{t}\tau_{t}-\frac{r}{(1-2\rho)h(\nu)}\right|,
\]
therefore,
\begin{align}
\left|\sigma_{t}\tau_{t}\theta_{t}-r\right|
\leq\left[2(1-2\rho)Q_{0}
+\frac{(1-2\rho)h(\nu)\left[{\frac{9Q_{0}}{4}+\frac{Q_{0}^{2}}{4}\sqrt{\frac{2}{\pi}}\left(\sqrt{\frac{2}{\pi}}+2(1-2\rho)+\frac{8}{\nu(1-2\rho)h(\nu)}\right)}\right]}{{\left(1+\nu(1-2\rho)h(\nu)\max\left\{ Q_{0},\frac{r}{(1-2\rho)h(\nu)}\right\}\right)^{-3}}}\right]\lambda\sqrt{s}.
\label{eq:temporal_calibration}
\end{align}
We now return to the additive Gaussian event
\cref{eq:additive_gaussian_gap}. Let $J\leq T$ be the last time at which the active coordinate is updated. Then $w_{T}=w_{J}$, and since an update places $w_{J}$ at one of the two endpoints of the temporal interval,
\[
w_{T}-rw^{\star}=\left(\sigma_{J}\tau_{J}\theta_{J}-r\right)w^{\star}+\sigma_{J}\tau_{J}\left(\frac{r_{J}}{\sqrt{\delta}}Z_{J}-\epsilon_{J}\lambda\right),
\]
where $\epsilon_{J}\in\{-1,1\}.$ Hence, by the reverse triangle inequality (twice),
\begin{align*}
|w_{T}-rw^{\star}|&\geq\sigma_{J}\tau_{J}\left(\lambda-\frac{r_{J}}{\sqrt{\delta}}|Z_{J}|\right)-\left|\sigma_{J}\tau_{J}\theta_{J}-r\right||w^{\star}|.
\end{align*}
On the event (\ref{eq:additive_gaussian_gap}), which has probability at least $\frac{1}{2}$, we have simultaneously for every $j\leq T$,
\[
|Z_{j}|\leq\frac{\lambda\sqrt{\delta}}{r_{j}}-\frac{\log\left[\left(\sum_{k=1}^{T}2\Phi\left(-\frac{\lambda\sqrt{\delta}}{r_{k}}\right)\right)^{-1}\right]}{4\sqrt{\log\left(2T\right)+\log\left[\left(\sum_{k=1}^{T}2\Phi\left(-\frac{\lambda\sqrt{\delta}}{r_{k}}\right)\right)^{-1}\right]}}.
\]
Therefore,
\begin{align*}
\lambda-\frac{r_{j}}{\sqrt{\delta}}|Z_{j}|\geq\frac{r_{j}}{4\sqrt{\delta}}\frac{\log\left[\left(\sum_{k=1}^{T}2\Phi\left(-\frac{\lambda\sqrt{\delta}}{r_{k}}\right)\right)^{-1}\right]}{\sqrt{\log\left(2T\right)+\log\left[\left(\sum_{k=1}^{T}2\Phi\left(-\frac{\lambda\sqrt{\delta}}{r_{k}}\right)\right)^{-1}\right]}}\geq\frac{\sqrt{h(\nu)}}{4\sqrt{8\delta}}\frac{\log\left[\left(\sum_{k=1}^{T}2\Phi\left(-\frac{\lambda\sqrt{\delta}}{r_{k}}\right)\right)^{-1}\right]}{\sqrt{\log\left(2T\right)+\log\left[\left(\sum_{k=1}^{T}2\Phi\left(-\frac{\lambda\sqrt{\delta}}{r_{k}}\right)\right)^{-1}\right]}},
\end{align*}
where we used $r_j\geq\sqrt{\frac{h(\nu)}{8}}$ (see \cref{eq:35}). Since $\sigma_j\tau_j\geq1$, it follows that the gap,
\begin{align*}
\sigma_{j}\tau_{j}\left(\lambda-\frac{r_{j}}{\sqrt{\delta}}|Z_{j}|\right)\geq\frac{\sqrt{h(\nu)}}{4\sqrt{8\delta}}\frac{\log\left[\left(\sum_{k=1}^{T}2\Phi\left(-\frac{\lambda\sqrt{\delta}}{r_{k}}\right)\right)^{-1}\right]}{\sqrt{\log\left(2T\right)+\log\left[\left(\sum_{k=1}^{T}2\Phi\left(-\frac{\lambda\sqrt{\delta}}{r_{k}}\right)\right)^{-1}\right]}}.
\end{align*}
By \cref{eq:temporal_calibration}, $\left|\sigma_J\tau_J\theta_J-r\right|\leq K\lambda\sqrt{s}.$
Hence, if we restrict to active coordinates satisfying,
\[
0<|w^{\star}|\leq\frac{\sqrt{h(\nu)}}{16K\lambda\sqrt{8s\delta}}\frac{\log\left[\left(\sum_{k=1}^{T}2\Phi\left(-\frac{\lambda\sqrt{\delta}}{r_{k}}\right)\right)^{-1}\right]}{\sqrt{\log\left(2T\right)+\log\left[\left(\sum_{k=1}^{T}2\Phi\left(-\frac{\lambda\sqrt{\delta}}{r_{k}}\right)\right)^{-1}\right]}},
\]
then,
\[
\left|\sigma_{J}\tau_{J}\theta_{J}-r\right||w^{\star}|\leq\frac{\sqrt{h(\nu)}}{16\sqrt{8\delta}}\frac{\log\left[\left(\sum_{k=1}^{T}2\Phi\left(-\frac{\lambda\sqrt{\delta}}{r_{k}}\right)\right)^{-1}\right]}{\sqrt{\log\left(2T\right)+\log\left[\left(\sum_{k=1}^{T}2\Phi\left(-\frac{\lambda\sqrt{\delta}}{r_{k}}\right)\right)^{-1}\right]}}.
\]
Thus the calibration term consumes at most one quarter of the gap. Combining the estimates gives,
\begin{align*}
|w_{T}-rw^{\star}|&\geq\frac{\sqrt{h(\nu)}}{4\sqrt{8\delta}}\frac{\log\left[\left(\sum_{k=1}^{T}2\Phi\left(-\frac{\lambda\sqrt{\delta}}{r_{k}}\right)\right)^{-1}\right]}{\sqrt{\log\left(2T\right)+\log\left[\left(\sum_{k=1}^{T}2\Phi\left(-\frac{\lambda\sqrt{\delta}}{r_{k}}\right)\right)^{-1}\right]}}-\frac{\sqrt{h(\nu)}}{16\sqrt{8\delta}}\frac{\log\left[\left(\sum_{k=1}^{T}2\Phi\left(-\frac{\lambda\sqrt{\delta}}{r_{k}}\right)\right)^{-1}\right]}{\sqrt{\log\left(2T\right)+\log\left[\left(\sum_{k=1}^{T}2\Phi\left(-\frac{\lambda\sqrt{\delta}}{r_{k}}\right)\right)^{-1}\right]}}\\&=\frac{3\sqrt{h(\nu)}}{16\sqrt{8\delta}}\frac{\log\left[\left(\sum_{k=1}^{T}2\Phi\left(-\frac{\lambda\sqrt{\delta}}{r_{k}}\right)\right)^{-1}\right]}{\sqrt{\log\left(2T\right)+\log\left[\left(\sum_{k=1}^{T}2\Phi\left(-\frac{\lambda\sqrt{\delta}}{r_{k}}\right)\right)^{-1}\right]}}.
\end{align*}
The same bound holds if the coordinate never updates. Indeed, since
$r_j\geq\sqrt{\frac{h(\nu)}{8}}$ (see \cref{eq:35}), we have
\begin{align*}
\log\left[\left(\sum_{k=1}^{T}2\Phi\left(-\frac{\lambda\sqrt{\delta}}{r_{k}}\right)\right)^{-1}\right]&\leq\log\left[\left(2\Phi\left(-\lambda\sqrt{\frac{8\delta}{h(\nu)}}\right)\right)^{-1}\right]\\&\underset{\Phi(-x)\geq\int_{x}^{x+1}\varphi(z)\dd z\geq\varphi(x+1)}{\leq}\log\left[\left(\sqrt{\frac{2}{\pi}}\exp\left(-\frac{1}{2}\left(\lambda\sqrt{\frac{8\delta}{h(\nu)}}+1\right)^{2}\right)\right)^{-1}\right]\\&=\frac{1}{2}\left(\lambda\sqrt{\frac{8\delta}{h(\nu)}}+1\right)^{2}+\frac{1}{2}\log\frac{\pi}{2}\\&\leq\frac{8\lambda^{2}\delta}{h(\nu)}\left[\frac{1}{2}\left(1+\frac{1}{\Phi^{-1}\left(\frac{3}{4}\right)}\right)^{2}+\frac{1}{2\left(\Phi^{-1}\left(\frac{3}{4}\right)\right)^{2}}\log\frac{\pi}{2}\right].
\end{align*}
where in the last inequality we used $\lambda\sqrt{\frac{8\delta}{h(\nu)}}\geq \Phi^{-1}\left(\frac{3}{4}\right)$ (recall we have shown it must hold $\lambda\to \infty$).
Equivalently, since conditionally on being active $w^\star=\frac{\nu}{\sqrt{s}}G$, the preceding lower bound holds for every updated active coordinate satisfying $|G|\leq\frac{\sqrt{h(\nu)}}{16K\lambda\nu\sqrt{8\delta}}\frac{\log\left[\left(\sum_{k=1}^{T}2\Phi\left(-\frac{\sqrt{\delta}}{r_{k}}\right)\right)^{-1}\right]}{\lambda\sqrt{\log(2T)+\log\left[\left(\sum_{k=1}^{T}2\Phi\left(-\frac{\lambda\sqrt{\delta}}{r_{k}}\right)\right)^{-1}\right]}}.$
On the other hand, if the coordinate never updates, then $w_{T}=0$ and $|w_{T}-rw^{\star}|=\frac{r\nu|G|}{\sqrt{s}}.$ We confine $|G|$ to a subinterval so that,
\[\frac{3\sqrt{h(\nu)}\sqrt{s}}{16r\nu\sqrt{8\delta}}\frac{\log\left[\left(\sum_{k=1}^{T}2\Phi\left(-\frac{\lambda\sqrt{\delta}}{r_k}\right)\right)^{-1}\right]}{\sqrt{\log(2T)+\log\left[\left(\sum_{k=1}^{T}2\Phi\left(-\frac{\lambda\sqrt{\delta}}{r_k}\right)\right)^{-1}\right]}}\leq |G|\leq\frac{\sqrt{h(\nu)}}{16K\lambda\nu\sqrt{8\delta}}\frac{\log\left[\left(\sum_{k=1}^{T}2\Phi\left(-\frac{\lambda\sqrt{\delta}}{r_k}\right)\right)^{-1}\right]}{\sqrt{\log(2T)+\log\left[\left(\sum_{k=1}^{T}2\Phi\left(-\frac{\lambda\sqrt{\delta}}{r_k}\right)\right)^{-1}\right]}}.\] 
By the contradicting assumption (\cref{eq:contradiction_temporal}), $\nu^2(\alpha_t-r)^2\leq c_\star s\log\left(\frac{1}{s}\right)\leq\frac{1}{4}r^2\nu^2,$ after choosing $s$ sufficiently small and hence $\alpha_t\leq\frac{3}{2}r.$ Therefore, $\sigma_t\tau_t=\frac{\alpha_t}{\frac{\alpha_t}{\sigma_t\tau_t}}\leq\frac{3r}{(1-2\rho)h(\nu)}=\frac{3}{h(\nu r)}\stackrel{\mathrm{\ref{eq:28}}}{\lesssim}1+\nu,$ which follows from $\frac{\alpha_t}{\sigma_t\tau_t}\geq\frac{1}{2}(1-2\rho)h(\nu)$ as we already established for sufficiently small $s$. With this upper bound we have $K\lesssim\frac{(1+\nu)^6}\nu,$ we decrease in \cref{eq:lambda_explicit} $c_\star$ so that 
$\frac{G_{\min}}{G_{\max}}=\frac{3K\lambda\sqrt{s}}{r}\leq\frac12,$ the interval $G_{\min}\leq|G|\leq G_{\max}$ has one-sided length at least $\frac{G_{\max}}{2},$ namely
$\frac{\sqrt{h(\nu)}}{32K\lambda\nu\sqrt{8\delta}}\frac{\log\left[\left(\sum_{k=1}^{T}2\Phi\left(-\frac{\lambda\sqrt{\delta}}{r_k}\right)\right)^{-1}\right]}{\sqrt{\log(2T)+\log\left[\left(\sum_{k=1}^{T}2\Phi\left(-\frac{\lambda\sqrt{\delta}}{r_k}\right)\right)^{-1}\right]}}.$
Moreover, the preceding upper bound on the logarithm implies that the
upper endpoint of this interval is at most $\frac{1}{16K\nu}\sqrt{\frac{1}{2}\left(1+\frac{1}{\Phi^{-1}\left(\frac{3}{4}\right)}\right)^{2}+\frac{1}{2\left(\Phi^{-1}\left(\frac{3}{4}\right)\right)^{2}}\log\frac{\pi}{2}}.$ Hence the standard Gaussian density throughout this interval is bounded
below by $\varphi\left(\frac{1}{16K\nu}\sqrt{\frac{1}{2}\left(1+\frac{1}{\Phi^{-1}\left(\frac{3}{4}\right)}\right)^{2}+\frac{1}{2\left(\Phi^{-1}\left(\frac{3}{4}\right)\right)^{2}}\log\frac{\pi}{2}}\right).$
Therefore, integrating over these active coordinates,
\begin{align*}
&\nu^{2}(\alpha_{T}-r)^{2}+\sigma_{T}^{2}=\E[(w_{T}-rw^{\star})^{2}]\geq s\left(\frac{3\sqrt{h(\nu)}}{16\sqrt{8\delta}}\right)^{2}\frac{\log^{2}\left[\left(\sum_{k=1}^{T}2\Phi\left(-\frac{\lambda\sqrt{\delta}}{r_{k}}\right)\right)^{-1}\right]}{\log\left(2T\right)+\log\left[\left(\sum_{k=1}^{T}2\Phi\left(-\frac{\lambda\sqrt{\delta}}{r_{k}}\right)\right)^{-1}\right]}\\&\times\frac{\sqrt{h(\nu)}}{16K\lambda\nu\sqrt{8\delta}}\frac{\log\left[\left(\sum_{k=1}^{T}2\Phi\left(-\frac{\lambda\sqrt{\delta}}{r_{k}}\right)\right)^{-1}\right]}{\sqrt{\log\left(2T\right)+\log\left[\left(\sum_{k=1}^{T}2\Phi\left(-\frac{\lambda\sqrt{\delta}}{r_{k}}\right)\right)^{-1}\right]}}\varphi\left(\frac{1}{16K\nu}\sqrt{\frac{1}{2}\left(1+\frac{1}{\Phi^{-1}\left(\frac{3}{4}\right)}\right)^{2}+\frac{1}{2\left(\Phi^{-1}\left(\frac{3}{4}\right)\right)^{2}}\log\frac{\pi}{2}}\right)
\end{align*}
Here we used both signs of $G$.
To eliminate $\lambda$, some $j\leq T$ satisfies $2\Phi\left(-\frac{\lambda\sqrt{\delta}}{r_{j}}\right)\geq\frac{1}{T}\sum_{k=1}^{T}2\Phi\left(-\frac{\lambda\sqrt{\delta}}{r_{k}}\right).$
Therefore, using $2\Phi(-x)\leq e^{-\frac{x^2}{2}}$ and $r_j\leq1$,
\begin{align*}
\frac{1}{T}\sum_{k=1}^{T}2\Phi\left(-\frac{\lambda\sqrt{\delta}}{r_{k}}\right)\leq2\Phi\left(-\frac{\lambda\sqrt{\delta}}{r_{j}}\right)\leq\exp\left(-\frac{\lambda^{2}\delta}{2r_{j}^{2}}\right)\leq\exp\left(-\frac{\lambda^{2}\delta}{2}\right),
\end{align*}
and hence $\lambda\sqrt{\delta}\leq\sqrt{2\left(\log\left(T\right)+\log\left[\left(\sum_{k=1}^{T}2\Phi\left(-\frac{\lambda\sqrt{\delta}}{r_{k}}\right)\right)^{-1}\right]\right)},$
\begin{align*}
&\nu^{2}(\alpha_{T}-r)^{2}+\sigma_{T}^{2}\geq s\left(\frac{3\sqrt{h(\nu)}}{16\sqrt{8\delta}}\right)^{2}\frac{\log^{3}\left[\left(\sum_{k=1}^{T}2\Phi\left(-\frac{\lambda\sqrt{\delta}}{r_{k}}\right)\right)^{-1}\right]}{\left(\log\left(2T\right)+\log\left[\left(\sum_{k=1}^{T}2\Phi\left(-\frac{\lambda\sqrt{\delta}}{r_{k}}\right)\right)^{-1}\right]\right)^{2}}\\&\qquad\times\frac{\sqrt{h(\nu)}}{16K\nu\sqrt{16}}\varphi\left(\frac{1}{16K\nu}\sqrt{\frac{1}{2}\left(1+\frac{1}{\Phi^{-1}\left(\frac{3}{4}\right)}\right)^{2}+\frac{1}{2\left(\Phi^{-1}\left(\frac{3}{4}\right)\right)^{2}}\log\frac{\pi}{2}}\right).
\end{align*}
By \cref{eq:tail_log_lower}, $\log\left[\left(\sum_{k=1}^{T}2\Phi\left(-\frac{\lambda\sqrt{\delta}}{r_{k}}\right)\right)^{-1}\right]\geq\frac{1}{2}\log\left(\frac{1}{s}\right).$
Since $x\longmapsto\frac{x^3}{(\log\left(2T\right)+x)^2}$ is strictly increasing for $x>0$,
\begin{align*}
&\nu^{2}(\alpha_{t}-r)^{2}+\sigma_{t}^{2}
\\&\geq\frac{9h(\nu)^{\frac{3}{2}}}{1048576K\nu\delta}\varphi\left(\frac{1}{16K\nu}\sqrt{\frac{1}{2}\left(1+\frac{1}{\Phi^{-1}\left(\frac{3}{4}\right)}\right)^{2}+\frac{1}{2\left(\Phi^{-1}\left(\frac{3}{4}\right)\right)^{2}}\log\frac{\pi}{2}}\right)s\frac{\log^{3}\left(\frac{1}{s}\right)}{\left(\log\left(2T\right)+\frac{1}{2}\log\left(\frac{1}{s}\right)\right)^{2}}.
\end{align*}
Thus \cref{eq:contradiction_temporal} is contradicted whenever $c_{\star}<\frac{9h(\nu)^{\frac{3}{2}}}{1048576K\nu\delta}\varphi\left(\frac{1}{16K\nu}\sqrt{\frac{1}{2}\left(1+\frac{1}{\Phi^{-1}\left(\frac{3}{4}\right)}\right)^{2}+\frac{1}{2\left(\Phi^{-1}\left(\frac{3}{4}\right)\right)^{2}}\log\frac{\pi}{2}}\right).$
Taking the minimum of this positive constant and the constants obtained
in the preceding cases, the proof yields
\[
\nu^{2}(\alpha_{t}-r)^{2}+\sigma_{t}^{2}\geq c''s\frac{\log^{3}\left(\frac{1}{s}\right)}{\left(\log\left(\frac{1}{s}\right)+\log\left(2T\right)\right)^{2}}
\]
for $c''=\min\left\{\frac{9h(\nu)^{\frac{3}{2}}}{1048576\delta K\nu}\varphi\left(\frac{1}{16K\nu}\sqrt{\frac{1}{2}\left(1+\frac{1}{\Phi^{-1}\left(\frac{3}{4}\right)}\right)^{2}+\frac{1}{2\left(\Phi^{-1}\left(\frac{3}{4}\right)\right)^{2}}\log\frac{\pi}{2}}\right), \frac{r^{2}\varphi\left(1\right)}{384\sqrt{8}\nu\delta^{\frac{3}{2}}(1-2\rho)^{3}}\right\}$. 
Instead of choosing $K$ as we chose, so that $\left|\sigma_J\tau_J\theta_J-r\right|\leq K\lambda\sqrt{s}$, and thereby risking an exponential dependence on the parameters due to the behavior of $\varphi$, to make sure it remains uniformly bounded from below, we set $K=\max\left\{K,\frac{1}{16\nu}\sqrt{\frac{1}{2}\left(1+\frac{1}{\Phi^{-1}\left(\frac{3}{4}\right)}\right)^{2}+\frac{1}{2\left(\Phi^{-1}\left(\frac{3}{4}\right)\right)^{2}}\log\frac{\pi}{2}}\right\}.$
With this choice, the argument of $\varphi$ is below $1$, so it remains uniformly bounded from below by $\varphi(1)$ and the same lower-bound argument applies. Using \cref{eq:28} $h(\nu)\gtrsim\frac{1}{1+\nu}$, and by the relation from \cref{lem:rel} $rh(\nu r)=(1-2\rho)h(\nu)$ we get $r\gtrsim\frac{1-2\rho}{1+\nu}$, $\frac{\dd}{\dd a}(ah(\nu a))\gtrsim(1+\nu a)^{-3}$ by \cref{eq:14}, we obtain $Q_0\lesssim\frac{(1+\nu)^5}{\nu^2(1-2\rho)^2}$ and $K\lesssim\frac{(1+\nu)^{6}}{\nu^8(1-2\rho)^7}$. Consequently, $c''\gtrsim\frac{\nu^7(1-2\rho)^7}{(1+\nu)^{26}(1+\delta)^{\frac{3}{2}}}$ (this choice is also sufficient to make $\frac{3K\lambda\sqrt{s}}{r}\leq \frac{1}{2}$ by $r\gtrsim\frac{1-2\rho}{1+\nu}$).
By \cref{prop:coefficient_error_to_excess_risk},
\[
\mathcal{L}(\mathbf w_T)-\inf_{\vw} \mathcal{L}(\vw)\geq C\frac{\nu^7(1-2\rho)^7}{(1+\nu)^{32}(1+\delta)^{\frac{3}{2}}}\frac{s\log^{3}\left(\frac{1}{s}\right)}{\left(\log\left(\frac{1}{s}\right)+\log\left(2T\right)\right)^{2}}.
\]
for some universal constant $C>0.$
Collecting all the conditions in which $s$ was required to be sufficiently small, it is enough to require, for universal constants $C_1,C_2,C_3,C_4>0$,
\begin{align}
&s\leq C_{1},\qquad s\log^{3}\left(\frac{1}{s}\right)\leq C_{2},\tag{S1}\ \\&s\left(2+\frac{655360}{3}\delta c_{\star}\log^{2}\left(\frac{1}{s}\right)\right)\leq\frac{\delta h(\nu)}{4},\tag{S2}\ \\&\sqrt{s}\left(2+\frac{655360}{3}\delta c_{\star}\log^{2}\left(\frac{1}{s}\right)\right)\leq\min\{1,\delta\}\Phi^{-1}\left(\frac{3}{4}\right)\sqrt{\frac{h(\nu)}{8\delta}},\tag{S3}\ \\&c_{\star}s\log\left(\frac{1}{s}\right)\leq C_{3}r^{2}\nu^{2},\tag{S4}\\&\lambda\sqrt{s}\left(2+\frac{655360}{3}\delta c_{\star}\frac{\log^{3}\left(\frac{1}{s}\right)}{\log\left(\frac{1}{s}\right)+\log\left(2T\right)}\right)\leq1,\tag{S5}\\&
Q\sqrt{\frac{3s}{\nu^2\delta}}\leq\frac{1}{4}(1-2\rho)h(\nu),\tag{S6}
\end{align}
The remaining smallness conditions on $\lambda\sqrt{s}$ in \cref{eq:normalized_mean_smallness,eq:normalized_variance_smallness,eq:normalized_variance_smallness} are satisfied by taking $c_\star$ sufficiently small and therefore impose no additional restriction on $s_0$.
Using \cref{eq:28} $h(\nu)\gtrsim\frac{1}{1+\nu}$ and $c_\star\asymp\frac{\nu^7(1-2\rho)^7}{(1+\nu)^{26}(1+\delta)^{\frac{3}{2}}},$ we have $\delta c_\star\lesssim1$ and $c_\star\lesssim r^2\nu^2$, so (S1) and (S4) follow by taking $s$ smaller than a universal constant. Thus it is sufficient to require
\[
s\log^2\left(\frac{1}{s}\right)\lesssim\frac{\delta}{1+\nu},\qquad s\log^4\left(\frac{1}{s}\right)\lesssim\frac{1}{\delta(1+\nu)}.
\]
For (S6) we need to require $s\leq C_4\nu^2\delta r^2h(\nu r)^4.$
Consequently, for a sufficiently small universal constant $c>0$, it is enough to take
\[
s\leq s_0=c\min\left\{\frac{\frac{\delta}{1+\nu}}{\log_+^2\left(\frac{e(1+\nu)}{\delta}\right)},\frac{\frac{1}{\delta(1+\nu)}}{\log_+^4\left(e\delta(1+\nu)\right)},\frac{\delta\nu^2(1-2\rho)^2}{(1+\nu)^4}\right\},
\]
with the convention that each logarithmic denominator is replaced by a universal constant whenever its argument is bounded.
\end{proof}

\newpage

\subsection{Proof of Lemma~\ref{thm:temporal_l1_large_T_upper}}
\label{prf:temporal_l1_large_T_upper}
\begin{fuller}[\cref{thm:temporal_l1_large_T_upper}]
There exist universal constants $C,\upsilon>0,$ a constant $c_{\nu,\delta}=c(\nu,\delta)>0$ and\linebreak $\lambda=\sqrt{\upsilon\left[\frac{(1+\nu)^7}{
\delta}\left(1+\nu^2+\frac{1}{\delta}\right)+\frac{(1+\nu)^{10}}{\delta^2}\right]+\frac{\varphi(1)}{2(1+e)^2\delta(1+\nu)}\log(2T)}$, such that, for all $T\geq 1$ and $s\in(0,1)$, 
\[
\mathcal{L}(\mathbf w_T)-\inf_{\vw} \mathcal{L}(\vw) \leq 
\frac{C(1+\nu)^{11}}{\delta\log(2T)}+\frac{C(1+\nu)^{30}}{\log^{2}(2T)}\left(1+\nu^{2}+\frac{1}{\delta}\right)^{2}+C(1+\nu)^{8}\exp\left(-c_{\nu,\delta}\sqrt{\frac{ T}{\log (2T)}}\right)
\]
\end{fuller}

\begin{proof}
Let $w_t$ follow the temporal L1 regularization, by \cref{lem:2},
\[
w_{t}=\Pi_{\left[\sigma_{t}\tau_{t}\left(\theta_{t}w^{\star}+\frac{r_{t}}{\sqrt{\delta}}Z_{t}-\lambda\right),\sigma_{t}\tau_{t}\left(\theta_{t}w^{\star}+\frac{r_{t}}{\sqrt{\delta}}Z_{t}+\lambda\right)\right]}(w_{t-1}),
\]
We divide the proof into several steps.
Recall $\frac{\gamma_t}{\sigma_t\tau_t}=\frac{1}{\delta}\PP(w_t\neq w_{t-1}).$
The equation for $\theta_t$ and the proximal relation (\cref{eq:prox_relation_large_T}) give
\[
\theta_{t}=2\E\left[\pi_{\rho}'(\nu Z_{1})\phi'\left(\prox_{\gamma_{t}\phi}(\nu\alpha_{t}Z_{1}+\sigma_{t}Z_{2})\right)\right].
\]
Consequently, $0<\theta_{t}\leq2\E[\pi_{\rho}'(\nu Z_{1})]=2(1-2\rho)h(\nu)\leq\frac{1}{2},$
as well as \cref{eq:qtemporal} gives, \begin{align}
\label{eq:27}\left|\frac{1}{\sigma_{t}\tau_{t}}-h\left(\sqrt{\nu^{2}\alpha_{t}^{2}+\sigma_{t}^{2}}\right)\right|\le C\frac{\gamma_{t}}{\sigma_{t}\tau_{t}}.\end{align} The  fourth scalar equation with the proximal relation (\cref{eq:prox_relation_large_T}) give
\[
r_{t}^{2}=2\E\left[(1-\pi_{\rho}(\nu Z_{1}))\phi'(\prox_{\gamma_{t}\phi}(\nu\alpha_{t}Z_{1}+\sigma_{t}Z_{2}))^{2}\right]\underset{0<\phi'(u)\leq1}{\leq}2\E\left[(1-\pi_{\rho}(\nu Z_{1}))\right]=1.
\]
The sixth scalar equation can be written as
\begin{align}
\frac{1}{\sigma_{t}\tau_{t}}=\E\left[2(1-\pi_{\rho}(\nu Z_{1}))\frac{\phi''(\prox_{\gamma_{t}\phi}(\nu\alpha_{t}Z_{1}+\sigma_{t}Z_{2}))}{1+\gamma_{t}\phi''(\prox_{\gamma_{t}\phi}(\nu\alpha_{t}Z_{1}+\sigma_{t}Z_{2}))}\right].
\label{eq:sixth_large_T}
\end{align}
Since $1-\pi_\rho(\nu Z_1)\leq1$, we obtain
\[
\frac{1}{\sigma_{t}\tau_{t}}\leq2\E\left[\frac{\phi''(\prox_{\gamma_{t}\phi}(\nu\alpha_{t}Z_{1}+\sigma_{t}Z_{2}))}{1+\gamma_{t}\phi''(\prox_{\gamma_{t}\phi}(\nu\alpha_{t}Z_{1}+\sigma_{t}Z_{2}))}\right].
\]
Since $\nu\alpha_tZ_1+\sigma_tZ_2\sim\mathcal N(0,\nu^2\alpha_t^2+\sigma_t^2),$
the change of variables $x=u+\gamma_t\phi'(u),$ $\dd x=(1+\gamma_t\phi''(u))\dd u,$
gives
\[
\E\left[\frac{\phi''(\prox_{\gamma_{t}\phi}(\nu\alpha_{t}Z_{1}+\sigma_{t}Z_{2}))}{1+\gamma_{t}\phi''(\prox_{\gamma_{t}\phi}(\nu\alpha_{t}Z_{1}+\sigma_{t}Z_{2}))}\right]=\int_{\R}\frac{\phi''(u)}{\sqrt{2\pi(\nu^{2}\alpha_{t}^{2}+\sigma_{t}^{2})}}\exp\left(-\frac{(u+\gamma_{t}\phi'(u))^{2}}{2\left(\nu^{2}\alpha_{t}^{2}+\sigma_{t}^{2}\right)}\right)\dd u.
\]
Using $\int_{\R}\phi''(u)\dd u=1,$
we obtain $\frac{1}{\sigma_t\tau_t}\leq\frac{2}{\sqrt{2\pi(\nu^2\alpha_t^2+\sigma_t^2)}},$
and hence
\begin{align}
\frac{\sqrt{\nu^2\alpha_t^2+\sigma_t^2}}{\sigma_t\tau_t}\leq\sqrt{\frac{2}{\pi}}.
\label{eq:R_over_q_temporal}
\end{align}
The inequality is trivial when $\sqrt{\nu^2\alpha_t^2+\sigma_t^2}=0$.
The temporal projection update can be written as
\[
w_{t}=w_{t-1}+\mathcal{S}_{\sigma_{t}\tau_{t}\lambda}\left(\sigma_{t}\tau_{t}\left(\theta_{t}w^{\star}+\frac{r_{t}}{\sqrt{\delta}}Z_{t}\right)-w_{t-1}\right).
\]
by \cref{lem:2}. On the update event $\mathcal U_t=\{w_t\neq w_{t-1}\},$ we therefore have exactly
\begin{align}
\frac{w_{t}}{\sigma_{t}\tau_{t}}=\theta_{t}w^{\star}+\frac{r_{t}}{\sqrt{\delta}}Z_{t}-\lambda\mathrm{sign}\left(\theta_{t}w^{\star}+\frac{r_{t}}{\sqrt{\delta}}Z_{t}-\frac{w_{t-1}}{\sigma_{t}\tau_{t}}\right).
\label{eq:update_endpoint_formula}
\end{align}
Hence, on $\mathcal U_t$ we have $\left|\theta_{t}w^{\star}+\frac{r_{t}}{\sqrt{\delta}}Z_{t}-\frac{w_{t-1}}{\sigma_{t}\tau_{t}}\right|>\lambda.$ If $\left|\theta_tw^{\star}+\frac{r_t}{\sqrt\delta}Z_t\right|\leq\frac{\lambda}{2},$ then $\left|\frac{w_t}{\sigma_t\tau_t}\right|\geq\frac{\lambda}{2}$ an vice versa. Consequently,
\begin{align}
\mathcal{U}_{t}\subseteq\left\{ \left|\theta_{t}w^{\star}+\frac{r_{t}}{\sqrt{\delta}}Z_{t}\right|>\frac{\lambda}{2}\right\} \cup\left\{ \left|\frac{w_{t}}{\sigma_{t}\tau_{t}}\right|>\frac{\lambda}{2}\right\} .
\label{eq:updatnion_bound}
\end{align}
Using $\theta_t\leq\frac{1}{2}$, $r_t\leq1$ (see conclusion beneath \cref{eq:34}), and
\cref{eq:R_over_q_temporal}, we have
\[
\E\left[\left(\theta_{t}w^{\star}+\frac{r_{t}}{\sqrt{\delta}}Z_{t}\right)^{2}\right]\leq\frac{\nu^{2}}{4}+\frac{1}{\delta},
\]
and
\[
\E\left[\left(\frac{w_{t}}{\sigma_{t}\tau_{t}}\right)^{2}\right]=\frac{\nu^{2}\alpha_{t}^{2}+\sigma_{t}^{2}}{(\sigma_{t}\tau_{t})^{2}}\leq\frac{2}{\pi}.
\]
Applying Chebyshev's inequality to
\cref{eq:updatnion_bound} therefore gives $\PP(\mathcal{U}_{t})\leq\frac{4}{\lambda^{2}}\left(\frac{\nu^{2}}{4}+\frac{1}{\delta}+\frac{2}{\pi}\right).$
Consequently,
\begin{align}
\frac{\gamma_{t}}{\sigma_{t}\tau_{t}}=\frac{1}{\delta}\PP(\mathcal{U}_{t})\leq\frac{\frac{4}{\delta}\left(\frac{\nu^{2}}{4}+\frac{1}{\delta}+\frac{2}{\pi}\right)}{\lambda^{2}}.
\label{eq:eta_uniform_lambda}
\end{align}

Applying \cref{eq:26} with $q=\sigma_t\tau_t$, $qH_t=\nu\alpha_tZ_1+\sigma_tZ_2$, $\eta=\frac{\gamma_t}{\sigma_t\tau_t}$, and $R=\sqrt{\nu^2\alpha_t^2+\sigma_t^2}$ gives,
\[
\frac{1}{\sigma_{t}\tau_{t}}\geq\frac{e^{-\frac{1}{2}}}{16\sqrt{2\pi}(1+\nu)\left(1+\sqrt{\nu^{2}\alpha_{t}^{2}+\sigma_{t}^{2}}+\gamma_{t}\right)}.
\]
Thus
\[
\sigma_{t}\tau_{t}\leq16\sqrt{2\pi}e^{\frac{1}{2}}(1+\nu)\left(1+\sqrt{\nu^{2}\alpha_{t}^{2}+\sigma_{t}^{2}}+\gamma_{t}\right).
\]
Using
\cref{eq:eta_uniform_lambda},
\[
\sigma_{t}\tau_{t}\leq16\sqrt{2\pi}e^{\frac{1}{2}}(1+\nu)\left(1+\sqrt{\nu^{2}\alpha_{t}^{2}+\sigma_{t}^{2}}\right)+16\sqrt{2\pi}e^{\frac{1}{2}}(1+\nu)\frac{\frac{4}{\delta}\left(\frac{\nu^{2}}{4}+\frac{1}{\delta}+\frac{2}{\pi}\right)}{\lambda^{2}}\sigma_{t}\tau_{t}.
\]
Hence, if $\lambda^{2}\geq128\sqrt{2\pi}e^{\frac{1}{2}}\frac{1+\nu}{\delta}\left(\frac{\nu^{2}}{4}+\frac{1}{\delta}+\frac{2}{\pi}\right),$
then
\begin{align}
\sigma_t\tau_t\leq32\sqrt{2\pi}e^{\frac{1}{2}}(1+\nu)\left(1+\sqrt{\nu^2\alpha_t^2+\sigma_t^2}\right)
\label{eq:q_linear_R}
\end{align}
uniformly in $t$ and $\rho$. Consequently,
\begin{align}
\gamma_t\leq128\sqrt{2\pi}e^{\frac{1}{2}}\frac{1+\nu}{\delta}\left(\frac{\nu^2}{4}+\frac{1}{\delta}+\frac2\pi\right)\frac{1+\sqrt{\nu^2\alpha_t^2+\sigma_t^2}}{\lambda^2}.
\label{eq:gamma_R_bound}
\end{align}

We next obtain a lower bound on $r_t$. Suppose first that $\sqrt{\nu^2\alpha_t^2+\sigma_t^2}\geq1.$ For $\lambda^2\geq512\sqrt{2\pi}e^{\frac{1}{2}}\frac{(1+\nu)^2}{\delta}\left(\frac{\nu^2}{4}+\frac{1}{\delta}+\frac2\pi\right),$
\cref{eq:gamma_R_bound} gives $\frac{\gamma_t}{\sqrt{\nu^2\alpha_t^2+\sigma_t^2}}\leq\frac{1}{2(1+\nu)}.$
Conditionally on $\nu\alpha_tZ_1+\sigma_tZ_2=x$, we have $Z_{1}\stackrel{d}{=}\frac{\nu\alpha_{t}}{\nu^{2}\alpha_{t}^{2}+\sigma_{t}^{2}}x+\sqrt{1-\frac{\nu^{2}\alpha_{t}^{2}}{\nu^{2}\alpha_{t}^{2}+\sigma_{t}^{2}}}\,G,$ where $G\sim\mathcal N(0,1)$. By symmetry,
\[
\E\left[2\left(1-\pi_{\rho}(\nu Z_{1})\right)\,\middle|\,\nu\alpha_{t}Z_{1}+\sigma_{t}Z_{2}=0\right]=1.
\]
Moreover,
\begin{align}
\label{eq:31}
\left|\frac{\dd}{\dd x}\E\left[2\left(1-\pi_{\rho}(\nu Z_{1})\right)\,\middle|\,\nu\alpha_{t}Z_{1}+\sigma_{t}Z_{2}=x\right]\right|\leq\frac{2\nu|\nu\alpha_{t}|}{\nu^{2}\alpha_{t}^{2}+\sigma_{t}^{2}}\sup_{u\in\mathbb{R}}\pi_{\rho}'(u)\leq\frac{\nu}{2\sqrt{\nu^{2}\alpha_{t}^{2}+\sigma_{t}^{2}}},
\end{align}
where we used $|\nu\alpha_t|\leq\sqrt{\nu^2\alpha_t^2+\sigma_t^2}$ and  $\sup_{u\in\mathbb R}\pi_\rho'(u)\leq\frac{1}{4}$. Therefore, whenever $|x|\leq\frac{\sqrt{\nu^{2}\alpha_{t}^{2}+\sigma_{t}^{2}}}{\nu}\leq\frac{\sqrt{\nu^{2}\alpha_{t}^{2}+\sigma_{t}^{2}}}{1+\nu}$, $\E\left[2\left(1-\pi_{\rho}(\nu Z_{1})\right)\,\middle|\,\nu\alpha_{t}Z_{1}+\sigma_{t}Z_{2}=x\right]\geq\frac{1}{2}.$ Consider the event $\frac{\sqrt{\nu^2\alpha_t^2+\sigma_t^2}}{2(1+\nu)}\leq\nu\alpha_tZ_1+\sigma_t Z_2\leq\frac{\sqrt{\nu^2\alpha_t^2+\sigma_t^2}}{1+\nu}$ this event has probability of $\PP\left(\frac{1}{2(1+\nu)}\leq\frac{\nu\alpha_{t}Z_{1}+\sigma_{t}Z_{2}}{\sqrt{\nu^{2}\alpha_{t}^{2}+\sigma_{t}^{2}}}\leq\frac{1}{1+\nu}\right)=\Phi\left(\frac{1}{1+\nu}\right)-\Phi\left(\frac{1}{2\left(1+\nu\right)}\right).$
Moreover, the proximal relation (\cref{eq:prox_relation_large_T}) on this event yields,
\[
\prox_{\gamma_t\phi}(\nu\alpha_tZ_1+\sigma_tZ_2)\geq\nu\alpha_tZ_1+\sigma_tZ_2-\gamma_t\geq0,
\]
and hence $\phi'\left(\prox_{\gamma_t\phi}(\nu\alpha_tZ_1+\sigma_tZ_2)\right)\geq\frac{1}{2}.$ Therefore,
\begin{align*}
r_t^2&=2\E\left[(1-\pi_\rho(\nu Z_1))\phi'\left(\prox_{\gamma_t\phi}(\nu\alpha_tZ_1+\sigma_tZ_2)\right)^2\right]
\geq\frac{1}{8}\left[\Phi\left(\frac{1}{1+\nu}\right)-\Phi\left(\frac{1}{2(1+\nu)}\right)\right]\geq\frac{\varphi(1)}{16(1+\nu)}.
\end{align*}
Suppose now that $\sqrt{\nu^2\alpha_t^2+\sigma_t^2}<1.$ Under the same condition on $\lambda$,
\cref{eq:gamma_R_bound} gives $\gamma_t\leq1.$
On the event $0\leq\nu\alpha_tZ_1+\sigma_tZ_2\leq\frac{\sqrt{\nu^2\alpha_t^2+\sigma_t^2}}{1+\nu},$ we have $\E\left[2(1-\pi_{\rho}(\nu Z_{1}))\mid\nu\alpha_{t}Z_{1}+\sigma_{t}Z_{2}\right]\geq\frac{1}{2},$ while $\prox_{\gamma_{t}\phi}(\nu\alpha_{t}Z_{1}+\sigma_{t}Z_{2})\geq-\gamma_{t}\geq-1.$
Thus $\phi'\left(\prox_{\gamma_{t}\phi}(\nu\alpha_{t}Z_{1}+\sigma_{t}Z_{2})\right)\geq\phi'(-1)=\frac{1}{1+e}.$ Consequently,
\[
r_{t}^{2}\geq\frac{1}{2(1+e)^{2}}\left[\Phi\left(\frac{1}{1+\nu}\right)-\frac{1}{2}\right]\geq\frac{\varphi(1)}{2(1+e)^{2}(1+\nu)}.
\]
Since
$\frac{\varphi(1)}{2(1+e)^2(1+\nu)}\leq\frac{\varphi(1)}{16(1+\nu)},$ we have uniformly in $t$ and $\rho$, $\sqrt{\frac{\varphi(1)}{2(1+e)^2(1+\nu)}}\leq r_t\leq1.$
We next refine the estimate on $\theta_t$. From symmetry, with $\phi'(x)=1-\phi'(-x)$,
\[
2\E\left[\pi_\rho'(\nu Z_1)\phi'(\nu\alpha_tZ_1+\sigma_tZ_2)\right]=(1-2\rho)h(\nu)=rh(\nu r).
\]
From the proximal relation $\prox_{\gamma_t\phi}(x)=x-\gamma_t\phi'(u)\leq x$ and $\phi'$ is increasing, as a consequence of these identities we have $0\leq rh(\nu r)-\theta_t.$ Moreover,
\[
rh(\nu r)-\theta_{t}=2\E\left[\pi_{\rho}'(\nu Z_{1})\left(\phi'(\nu\alpha_{t}Z_{1}+\sigma_{t}Z_{2})-\phi'\left(\prox_{\gamma_{t}\phi}(\nu\alpha_{t}Z_{1}+\sigma_{t}Z_{2})\right)\right)\right].
\]

If $\sqrt{\nu^2\alpha_t^2+\sigma_t^2}\leq1,$ then, since $\phi'$ is $\frac{1}{4}$-Lipschitz and the proximal relation 
(\cref{eq:prox_relation_large_T}),
\[
0\leq rh(\nu r)-\theta_t\leq\frac{\gamma_t}{2}(1-2\rho)h(\nu)\leq\frac{\gamma_t}{8}.
\]
Using \cref{eq:gamma_R_bound},
\[
rh(\nu r)-\theta_t\leq32\sqrt{2\pi}e^{\frac{1}{2}}\frac{1+\nu}{\delta}\left(\frac{\nu^2}{4}+\frac{1}{\delta}+\frac2\pi\right)\frac{1}{\lambda^2}.
\]
If $\sqrt{\nu^2\alpha_t^2+\sigma_t^2}\geq1,$ then using \cref{eq:29}, $\int_{\R}\left[\phi'(x)-\phi'\left(\prox_{\gamma_{t}\phi}(x)\right)\right]\dd x=\frac{\gamma_{t}}{2}.$
Conditioning on $\nu\alpha_tZ_1+\sigma_tZ_2=x$, using $\pi_\rho'(u)\leq\frac{1}{4}$ and the upper bound on the Gaussian density, gives
\begin{align*}
0\leq rh(\nu r)-\theta_{t}&\leq\frac{1}{2\sqrt{2\pi(\nu^{2}\alpha_{t}^{2}+\sigma_{t}^{2})}}\int_{\R}\left[\phi'(x)-\phi'\left(\prox_{\gamma_{t}\phi}(x)\right)\right]\dd x\\&=\frac{\gamma_{t}}{4\sqrt{2\pi(\nu^{2}\alpha_{t}^{2}+\sigma_{t}^{2})}}\leq32e^{\frac{1}{2}}\frac{1+\nu}{\delta}\left(\frac{\nu^{2}}{4}+\frac{1}{\delta}+\frac{2}{\pi}\right)\frac{1}{\lambda^{2}}.
\end{align*}
Thus, in both cases,
\begin{align}
0\leq rh(\nu r)-\theta_{t}\leq32\sqrt{2\pi}e^{\frac{1}{2}}\frac{1+\nu}{\delta}\left(\frac{\nu^{2}}{4}+\frac{1}{\delta}+\frac{2}{\pi}\right)\frac{1}{\lambda^{2}}.
\label{eq:theta_rhofree}
\end{align}

We first consider bounded $\sqrt{\nu^2\alpha_t^2+\sigma_t^2}$. From \cref{eq:30},
\[
2\E\left[(1-\pi_{\rho}(\nu Z_{1}))\phi''(\nu\alpha_{t}Z_{1}+\sigma_{t}Z_{2})\right]=h\left(\sqrt{\nu^{2}\alpha_{t}^{2}+\sigma_{t}^{2}}\right),
\]
the proximal relation, together with $\sup_u|\phi'''(u)|=\frac{1}{6\sqrt3},$ $\sup_u\phi''(u)=\frac{1}{4},$ gives
\begin{align*}
&\left|\frac{\phi''\left(\prox_{\gamma_{t}\phi}\left(\nu\alpha_{t}Z_{1}+\sigma_{t}Z_{2}\right)\right)}{1+\gamma_{t}\phi''\left(\prox_{\gamma_{t}\phi}\left(\nu\alpha_{t}Z_{1}+\sigma_{t}Z_{2}\right)\right)}-\phi''\left(\nu\alpha_{t}Z_{1}+\sigma_{t}Z_{2}\right)\right|\\&\leq\left|\phi''\left(\prox_{\gamma_{t}\phi}\left(\nu\alpha_{t}Z_{1}+\sigma_{t}Z_{2}\right)\right)-\phi''\left(\nu\alpha_{t}Z_{1}+\sigma_{t}Z_{2}\right)\right|+\\&\hspace{8em}\left|\phi''\left(\prox_{\gamma_{t}\phi}\left(\nu\alpha_{t}Z_{1}+\sigma_{t}Z_{2}\right)\right)-\frac{\phi''\left(\prox_{\gamma_{t}\phi}\left(\nu\alpha_{t}Z_{1}+\sigma_{t}Z_{2}\right)\right)}{1+\gamma_{t}\phi''\left(\prox_{\gamma_{t}\phi}\left(\nu\alpha_{t}Z_{1}+\sigma_{t}Z_{2}\right)\right)}\right|\\&\leq\underbrace{\sup_{u}\phi'''(u)\left|\prox_{\gamma_{t}\phi}\left(\nu\alpha_{t}Z_{1}+\sigma_{t}Z_{2}\right)-\nu\alpha_{t}Z_{1}+\sigma_{t}Z_{2}\right|}_{=\frac{1}{6\sqrt{3}}\gamma_{t}\phi'\left(\prox_{\gamma_{t}\phi}\left(\nu\alpha_{t}Z_{1}+\sigma_{t}Z_{2}\right)\right)}+\underbrace{\frac{\gamma_{t}\phi''\left(\prox_{\gamma_{t}\phi}\left(\nu\alpha_{t}Z_{1}+\sigma_{t}Z_{2}\right)\right)^{2}}{1+\gamma_{t}\phi''\left(\prox_{\gamma_{t}\phi}\left(\nu\alpha_{t}Z_{1}+\sigma_{t}Z_{2}\right)\right)}}_{\leq\gamma_{t}\phi''\left(\prox_{\gamma_{t}\phi}\left(\nu\alpha_{t}Z_{1}+\sigma_{t}Z_{2}\right)\right)^{2}\leq\frac{\gamma_{t}}{16}}
\end{align*}
Thus,
\[
\left|\frac{1}{\sigma_t\tau_t}-h\left(\sqrt{\nu^2\alpha_t^2+\sigma_t^2}\right)\right|\leq\left(\frac{1}{6\sqrt3}+\frac{1}{16}\right)\gamma_t.
\]
Consequently, whenever $\sqrt{\nu^2\alpha_t^2+\sigma_t^2}\leq C(1+\nu)^3$ for a sufficiently large universal constant $C$, we have $h\left(\sqrt{\nu^2\alpha_t^2+\sigma_t^2}\right)\geq\frac{c}{(1+\nu)^3}$ for a universal $c>0$ (see \cref{eq:28}). 
Using \cref{eq:gamma_R_bound}, if $\lambda^2\geq C\frac{(1+\nu)^7}{\delta}\left(1+\nu^2+\frac{1}{\delta}\right),$ then $\frac{1}{\sigma_t\tau_t}\geq\frac{c}{2(1+\nu)^3}$ from \cref{eq:gamma_R_bound}.
Therefore,
\[\left|\sigma_{t}\tau_{t}-\frac{1}{h\left(\sqrt{\nu^{2}\alpha_{t}^{2}+\sigma_{t}^{2}}\right)}\right|=\frac{\sigma_{t}\tau_{t}}{h\left(\sqrt{\nu^{2}\alpha_{t}^{2}+\sigma_{t}^{2}}\right)}\left|\frac{1}{\sigma_{t}\tau_{t}}-h\left(\sqrt{\nu^{2}\alpha_{t}^{2}+\sigma_{t}^{2}}\right)\right|\leq\frac{C(1+\nu)^{10}}{\delta\lambda^{2}}\left(1+\nu^{2}+\frac{1}{\delta}\right)\]
Combining this with \cref{eq:theta_rhofree},
\begin{equation}
\label{eq:center_compact_calibration}
\begin{aligned}
\left|\sigma_{t}\tau_{t}\theta_{t}-\frac{rh(\nu r)}{h\left(\sqrt{\nu^{2}\alpha_{t}^{2}+\sigma_{t}^{2}}\right)}\right|&\leq\sigma_{t}\tau_{t}\left|\theta_{t}-rh(\nu r)\right|+\underbrace{rh(\nu r)}_{=(1-2\rho)h(\nu)\leq\frac{1}{4}}\left|\sigma_{t}\tau_{t}-\frac{1}{h\left(\sqrt{\nu^{2}\alpha_{t}^{2}+\sigma_{t}^{2}}\right)}\right|\\&\leq\frac{C(1+\nu)^{10}}{\delta\lambda^{2}}\left(1+\nu^{2}+\frac{1}{\delta}\right).
\end{aligned}
\end{equation}

The function $h$ is strictly decreasing on $(0,\infty),$ because $\phi''$ is strictly decreasing on $(0,\infty)$. Moreover, $\frac{\dd}{\dd a}\left(ah(a)\right)=\E[Z^2\phi''(aZ)]>0.$ Also, $\frac{\dd^2}{\dd a^2}\left(ah(a)\right)=\E[Z^3\phi'''(aZ)]<0,$ so $ah(a)$ is concave.
For every $a\geq0$, since $\min\left\{1,\frac{1}{a}\right\}\geq\frac{1}{1+a},$ we obtain
\begin{equation}
\label{eq:Gprime_lower_temporal}
\begin{aligned}
\E[Z^{2}\phi''(aZ)]\geq \phi''(1)\E[Z^{2}\mathbf{1}_{{|Z|\leq\min{\left\{1,\frac{1}{a}\right\}}}}]&=2\phi''(1)\int_{0}^{b}z^{2}\varphi(z)\dd z\geq2\varphi(1)\phi''(1)\int_{0}^{b}z^{2}\dd z\\&=2\varphi(1)\phi''(1)\frac{1}{3(1+a)^3}\geq\frac{2\varphi(1)\phi''(1)}{3(1+a)^{3}}.
\end{aligned}
\end{equation}

Now fix any $x\in[0,\nu]$ because $h(u)$ is decreasing while $uh(u)$ is increasing (\cref{lem:mon}). For $R<x$, $R<\frac{xh(x)}{h(R)}<x,$ while for $R>x$, $x<\frac{xh(x)}{h(R)}<R.$
Moreover, because $\frac{xh(x)}{h(R)}$ is between $R$ and $x$,
\[
1-\frac{\left|\frac{xh(x)}{h(R)}-x\right|}{|R-x|}=1-\frac{\left|\frac{xh(x)}{h(R)}-x\right|}{\left|R-\frac{xh(x)}{h(R)}\right|+\left|\frac{xh(x)}{h(R)}-x\right|}=\frac{\left|Rh(R)-xh(x)\right|}{h(R)\left|R-x\right|}.
\]
If $R\leq2(1+\nu),$ the mean-value theorem, $h(R)\leq\frac{1}{4}$, and \cref{eq:Gprime_lower_temporal} give $1-\frac{\left|\frac{xh(x)}{h(R)}-x\right|}{|R-x|}\geq\frac{8\varphi(1)\phi''(1)}{81(1+\nu)^{3}}.$
If instead $R\geq2(1+\nu),$ since $x\leq\nu<R$ and $\frac{xh(x)}{h(R)}$ is between $R$ and $x$, then
\[
\frac{\left|\frac{xh(x)}{h(R)}-x\right|}{R-x}\leq\frac{xh(x)}{Rh(R)}\leq\frac{\nu h(\nu)}{2(1+\nu)h(2(1+\nu))}.
\]
Using again \cref{eq:Gprime_lower_temporal}, since $qh(q)=\int_{\R}\phi''(u)\varphi\left(\frac{u}{q}\right)\dd u\leq \varphi(0)\int_{R}\phi''(u)\dd u=\frac{1}{\sqrt{2\pi}},$ and because $qh(q)$ is increasing we have $2(1+\nu)h(2(1+\nu))-\nu h(\nu)>0$, therefore,
\[
1-\frac{\nu h(\nu)}{2(1+\nu)h(2(1+\nu))}\geq\sqrt{2\pi}\left(2(1+\nu)h(2(1+\nu))-\nu h(\nu)\right)\geq\frac{2\sqrt{2\pi}\varphi(1)\phi''(1)}{81(1+\nu)^{2}}.
\]
Thus, for every $x\in[0,\nu]$ and every $R\geq0$,
\begin{align}
\left|\frac{xh(x)}{h(R)}-x\right|\leq\left(1-\frac{2\sqrt{2\pi}\varphi(1)\phi''(1)}{81(1+\nu)^{3}}\right)|R-x|.
\label{eq:uniform_population_contraction}
\end{align}
Since $\nu r\leq\nu$, taking $x=\nu r$ gives
\begin{align}
\left|\nu\frac{rh(\nu r)}{h(R)}-\nu r\right|\leq\left(1-\frac{2\sqrt{2\pi}\varphi(1)\phi''(1)}{81(1+\nu)^{3}}\right)|R-\nu r|.
\label{eq:population_contraction_explicit}
\end{align}

Combining
\cref{eq:center_compact_calibration,eq:population_contraction_explicit},
whenever $\sqrt{\nu^2\alpha_t^2+\sigma_t^2}\leq C(1+\nu)^3,$
we obtain
\begin{align}
\nu\left|\sigma_{t}\tau_{t}\theta_{t}-r\right|\leq\left(1-\frac{2\sqrt{2\pi}\varphi(1)\phi''(1)}{81(1+\nu)^{3}}\right)\left|\sqrt{\nu^{2}\alpha_{t}^{2}+\sigma_{t}^{2}}-\nu r\right|+\frac{C(1+\nu)^{11}}{\delta\lambda^{2}}\left(1+\nu^{2}+\frac{1}{\delta}\right).
\label{eq:center_compact_explicit}
\end{align}

We next control the large-$\sqrt{\nu^2\alpha_t^2+\sigma_t^2}$ regime. On $\frac{\nu\alpha_{t}Z_{1}+\sigma_{t}Z_{2}}{\sqrt{\nu^{2}\alpha_{t}^{2}+\sigma_{t}^{2}}}=x,$ we have $Z_{1}\stackrel{d}{=}\frac{\nu\alpha_{t}}{\sqrt{\nu^{2}\alpha_{t}^{2}+\sigma_{t}^{2}}}x+\sqrt{1-\frac{\nu^{2}\alpha_{t}^{2}}{\nu^{2}\alpha_{t}^{2}+\sigma_{t}^{2}}}G.$
Let \[\omega_t(x)=\E_{G}\left[2\left(1-\pi_{\rho}\left(\nu\left(\frac{\nu\alpha_{t}}{\sqrt{\nu^{2}\alpha_{t}^{2}+\sigma_{t}^{2}}}x+\sqrt{1-\frac{\nu^{2}\alpha_{t}^{2}}{\nu^{2}\alpha_{t}^{2}+\sigma_{t}^{2}}}G\right)\right)\right)\right],\] from symmetry, $\omega_t(0)=1.$ Moreover, the derivative in $x$ has absolute value at most $\frac{\nu}{2}$ (see \cref{eq:31}). Therefore $\left|\omega_t(x)-1\right|\leq\frac{\nu}{2}|x|.$
Conditioning the sixth equation on $\nu\alpha_tZ_1+\sigma_tZ_2$ and making the exact change of variables $x=u+\gamma_t\phi'(u)$ with $\dd x=(1+\gamma_t\phi''(u))\dd u,$ gives
\begin{align}
&1=\frac{\sigma_{t}\tau_{t}}{\sqrt{2\pi(\nu^{2}\alpha_{t}^{2}+\sigma_{t}^{2})}}\int_{\R}\phi''(u)\omega_t(u+\gamma_t\phi'(u))\exp\left(-\frac{1}{2}\left(\frac{u+\gamma_{t}\phi'(u)}{\sqrt{\nu^{2}\alpha_{t}^{2}+\sigma_{t}^{2}}}\right)^{2}\right)\dd u
\label{eq:large_R_exact_identity}
\end{align}

Using $\int_{\R}|u|\phi''(u)\dd u=2\log2,$ $\int_{\R}\phi'(u)\phi''(u)\dd u=\frac{1}{2},$
we obtain
\begin{align*}
&\left|\int_{\R}\phi''(u)\omega_{t}\left(u+\gamma_{t}\phi'(u)\right)\exp\left(-\frac{1}{2}\left(\frac{u+\gamma_{t}\phi'(u)}{\sqrt{\nu^{2}\alpha_{t}^{2}+\sigma_{t}^{2}}}\right)^{2}\right)\dd u-1\right|\\&=\left|\int_{\R}\phi''(u)\left(\omega_{t}\left(u+\gamma_{t}\phi'(u)\right)\exp\left(-\frac{1}{2}\left(\frac{u+\gamma_{t}\phi'(u)}{\sqrt{\nu^{2}\alpha_{t}^{2}+\sigma_{t}^{2}}}\right)^{2}\right)-1\right)\dd u\right|\\&\leq\left|\int_{\R}\phi''(u)\left(\exp\left(-\frac{1}{2}\left(\frac{u+\gamma_{t}\phi'(u)}{\sqrt{\nu^{2}\alpha_{t}^{2}+\sigma_{t}^{2}}}\right)^{2}\right)\left|\omega_{t}\left(u+\gamma_{t}\phi'(u)\right)-\underbrace{\omega_{t}(0)}_{=1}\right|+\left|\exp\left(-\frac{1}{2}\left(\frac{u+\gamma_{t}\phi'(u)}{\sqrt{\nu^{2}\alpha_{t}^{2}+\sigma_{t}^{2}}}\right)^{2}\right)-1\right|\right)\dd u\right|\\&\underset{\left|1-e^{-\frac{x^{2}}{2}}\right|\leq\left|x\right|}{\leq}\left|\int_{\R}\phi''(u)\left(\frac{\nu}{2\sqrt{\nu^{2}\alpha_{t}^{2}+\sigma_{t}^{2}}}\left|u+\gamma_{t}\phi'(u)\right|+\frac{\left|u+\gamma_{t}\phi'(u)\right|}{\sqrt{\nu^{2}\alpha_{t}^{2}+\sigma_{t}^{2}}}\right)\right|\\&\leq\frac{1+\frac{\nu}{2}}{\sqrt{\nu^{2}\alpha_{t}^{2}+\sigma_{t}^{2}}}\left|\int_{\R}\phi''(u)\left|u+\gamma_{t}\phi'(u)\right|\right|\\&\leq\frac{1+\frac{\nu}{2}}{\sqrt{\nu^{2}\alpha_{t}^{2}+\sigma_{t}^{2}}}\left(2\log2+\frac{\gamma_{t}}{2}\right).
\end{align*}

Recall that
\[
1-\sqrt{2\pi}\nu h(\nu)=\int_{\R}\phi''(u)\left(1-e^{-\frac{u^{2}}{2\nu^{2}}}\right)\dd u.
\]
Using $|u|\geq1$, $1-\sqrt{2\pi}\nu h(\nu)\underset{1-e^{-x}\geq\frac{x}{1+x}}{\geq}\frac{2}{(1+e)(1+2\nu^2)}.$ Since $\sqrt{2\pi}\nu rh(\nu r)\stackrel{\mathrm{\ref{lem:rel}}}{=}\sqrt{2\pi}\nu(1-2\rho)h(\nu)\leq\sqrt{2\pi}\nu h(\nu),$ so,
\begin{align}
1-\sqrt{2\pi}\nu rh(\nu r)\geq\frac{2}{(1+e)(1+2\nu^2)}.
\label{eq:large_R_gap_rhofree}
\end{align}

Hence, whenever $\sqrt{\nu^2\alpha_t^2+\sigma_t^2}\geq C(1+\nu)^3$ and $\lambda^{2}\geq C\frac{(1+\nu)^{7}}{\delta}\left(1+\nu^{2}+\frac{1}{\delta}\right)$ for a sufficiently large universal constant $C$, \cref{eq:gamma_R_bound} gives
$\frac{\gamma_t}{\sqrt{\nu^2\alpha_t^2+\sigma_t^2}}\leq C\frac{1+\nu}{\delta\lambda^2}\left(1+\nu^2+\frac{1}{\delta}\right)\leq\frac{c}{(1+\nu)^6}.$
Consequently, after increasing $C$ if necessary,
\[
\frac{1+\frac{\nu}{2}}{\sqrt{\nu^2\alpha_t^2+\sigma_t^2}}
\left(2\log2+\frac{\gamma_t}{2}\right)
\leq\frac{1}{(1+e)(1+2\nu^2)}.
\]
Therefore, denoting the integral in \cref{eq:large_R_exact_identity} by $I_t$, we have $I_t\geq1-\frac{1}{(1+e)(1+2\nu^2)}.$
Using \cref{eq:large_R_exact_identity}, $\theta_t\leq rh(\nu r)$ (it follows from the fifth equation where $\theta_t\leq2\E[\pi_\rho'(\nu Z_1)\phi'(\nu\alpha_tZ_1+\sigma_tZ_2))]=\E[\pi_\rho'(\nu Z_1)]=(1-2\rho)h(\nu)\stackrel{\mathrm{\ref{lem:rel}}}{=}rh(\nu r)$) and \cref{eq:large_R_gap_rhofree} , we obtain
\begin{align*}
\frac{\nu\sigma_{t}\tau_{t}\theta_{t}}{\sqrt{\nu^{2}\alpha_{t}^{2}+\sigma_{t}^{2}}}=\frac{\sqrt{2\pi}\nu\theta_{t}}{I_{t}}\leq\frac{\sqrt{2\pi}\nu rh(\nu r)}{1-\frac{1}{(1+e)(1+2\nu^{2})}}\leq\frac{1-\frac{2}{(1+e)(1+2\nu^{2})}}{1-\frac{1}{(1+e)(1+2\nu^{2})}}\stackrel{\frac{1}{(1+e)(1+2\nu^{2})}<1}{\leq}1-\frac{1}{2(1+e)(1+2\nu^{2})},
\end{align*}
which gives
\[
\frac{\nu\sigma_t\tau_t\theta_t}{\sqrt{\nu^2\alpha_t^2+\sigma_t^2}}\leq1-\frac{1}{2(1+e)(1+2\nu^2)}.
\]
Since $\nu r\leq\nu$ (\cref{lem:rel}) and $\sqrt{\nu^2\alpha_t^2+\sigma_t^2}\geq C(1+\nu)^3,$
increasing the universal constant $C$ if necessary gives
\begin{align}
\nu\left|\sigma_{t}\tau_{t}\theta_{t}-r\right|\leq\left(1-\frac{1}{2(1+e)(1+2\nu^{2})}\right)\left|\sqrt{\nu^{2}\alpha_{t}^{2}+\sigma_{t}^{2}}-\nu r\right|\leq\left(1-\frac{2\sqrt{2\pi}\varphi(1)\phi''(1)}{81(1+\nu)^{3}}\right)\left|\sqrt{\nu^{2}\alpha_{t}^{2}+\sigma_{t}^{2}}-\nu r\right|.
\label{eq:center_large_explicit}
\end{align}

Combining
\cref{eq:center_compact_explicit,eq:center_large_explicit},
we conclude that, provided
\begin{align}
\lambda^{2}\geq C\left[\frac{(1+\nu)^{7}}{\delta}\left(1+\nu^{2}+\frac{1}{\delta}\right)+\frac{(1+\nu)^{10}}{\delta^{2}}\right]
\label{eq:lambda_lower_rhofree}
\end{align}
for a sufficiently large universal constant $C$, then for every $t$,
\begin{align}
&\nu\left|\sigma_{t}\tau_{t}\theta_{t}-r\right|\leq\left(1-\frac{2\sqrt{2\pi}\varphi(1)\phi''(1)}{81(1+\nu)^{3}}\right)\left|\sqrt{\nu^{2}\alpha_{t}^{2}+\sigma_{t}^{2}}-\nu r\right|
+\frac{C(1+\nu)^{11}}{\delta\lambda^{2}}\left(1+\nu^{2}+\frac{1}{\delta}\right).
\label{eq:center_error_pre_global}
\end{align}

Finally, by the reverse triangle inequality,
\begin{align}
\left|\underbrace{\sqrt{\nu^{2}\alpha_{t}^{2}+\sigma_{t}^{2}}}_{=\left\Vert (\nu\alpha_{t},\sigma_{t})\right\Vert }-\underbrace{\nu r}_{=\left\Vert (\nu r,0)\right\Vert }\right|&\leq\sqrt{\nu^{2}(\alpha_{t}-r)^{2}+\sigma_{t}^{2}}=\sqrt{\E[(w_{t}-rw^{\star})^{2}]}.
\label{eq:11}
\end{align}
Hence,
\begin{align}
&\nu\left|\sigma_{t}\tau_{t}\theta_{t}-r\right|\leq\left(1-\frac{2\sqrt{2\pi}\varphi(1)\phi''(1)}{81(1+\nu)^{3}}\right)\sqrt{\E[(w_{t}-rw^{\star})^{2}]}+\frac{C(1+\nu)^{11}}{\delta\lambda^{2}}\left(1+\nu^{2}+\frac{1}{\delta}\right).
\label{eq:center_error_global}
\end{align}

We now show the following identity. Let $a,b>0$ and fixed $x$, then we have
\begin{align}
\E[\Pi_{[a(Z-b),a(Z+b)]}(x)^2]\leq(1-2\Phi(-b))x^2+C\frac{a^22\Phi(-b)}{b^2}.
\label{eq:centered_projection_moment}
\end{align}
After rescaling it suffices to take $a=1$. The function $g_b(x)=\E[\Pi_{[Z-b,Z+b]}(x)^2]$ is even and, for $x\geq0$,
\[
g_b'(x)=2x\PP(|Z-x|\leq b)\leq 2x(1-2\Phi(-b)),
\]
since a Gaussian interval of fixed length has maximal probability when
centered at zero. Hence $g_b(x)\leq g_b(0)+(1-2\Phi(-b))x^2.$
Moreover,
\begin{align*}
g_{b}(0)&=\E[(|Z|-b)_{+}^{2}]=2\int_{b}^{\infty}(z-b)^{2}\varphi(z)\dd z
=2\varphi(b)\int_{0}^{\infty}y^{2}e^{-by-\frac{y^{2}}{2}}\dd y\leq2\varphi(b)\int_{0}^{\infty}y^{2}e^{-by}\dd y=\frac{4\varphi(b)}{b^{3}},
\end{align*}
and hence, for $b\geq1$, by Mills inequality, $\varphi(b)\leq\left(b+\frac{1}{b}\right)\Phi(-b)\leq2b\Phi(-b).$ This proves \cref{eq:centered_projection_moment}.
Now suppose $X'=\Pi_{[d+a(Z-b),d+a(Z+b)]}(X),$ where $|d|\leq D$. We claim that
\begin{align}
\E[(|X'|-D)_{+}^{2}\mid X]\leq(1-2\Phi(-b))(|X|-D)_{+}^{2}+C\frac{a^{2}2\Phi(-b)}{b^{2}}.
\label{eq:moving_band_projection}
\end{align}
To prove the claim, first assume $X>D$.
Note that
\[
X'=d+\Pi_{[a(Z-b),a(Z+b)]}(X-d).
\]
Since
$\Pi_{[a(Z-b),a(Z+b)]}(\cdot)$
is increasing and $1$-Lipschitz,
\begin{align*}
(X'-D)_{+}&=\Big(\Pi_{[a(Z-b),a(Z+b)]}(X-d)-\Pi_{[a(Z-b),a(Z+b)]}(X-D)
+\Pi_{[a(Z-b),a(Z+b)]}(X-D)-(D-d)\Big)_{+}\\&\leq\left(\Pi_{[a(Z-b),a(Z+b)]}(X-D)\right)_{+}.
\end{align*}
Likewise,
\[
(-D-X')_+\leq\left(-\Pi_{[a(Z-b),a(Z+b)]}(X-D)\right)_+,
\]
and hence
\[
(|X'|-D)_+\leq\left|\Pi_{[a(Z-b),a(Z+b)]}(X-D)\right|.
\]
The case $X<-D$ follows by reflection.

Finally, suppose $|X|\leq D$. Since $-D-d\leq X-d\leq D-d,$ monotonicity and the $1$-Lipschitz property give $(X'-D)_+\leq\left(\Pi_{[a(Z-b),a(Z+b)]}(0)\right)_+,$ and $(-D-X')_+\leq\left(-\Pi_{[a(Z-b),a(Z+b)]}(0)\right)_+.$
Thus $(|X'|-D)_+\leq\left|\Pi_{[a(Z-b),a(Z+b)]}(0)\right|.$
Applying \cref{eq:centered_projection_moment} proves \cref{eq:moving_band_projection}.

Fix integers $u<t$. By \cref{eq:center_error_global},
\begin{align}
&\nu\max_{u<j\leq t}|\sigma_{j}\tau_{j}\theta_{j}-r|\leq\left(1-\frac{2\sqrt{2\pi}\varphi(1)\phi''(1)}{81(1+\nu)^{3}}\right)\max_{u\leq j\leq t}\sqrt{\E[(w_{j}-rw^{\star})^{2}]}+\frac{C(1+\nu)^{11}}{\delta\lambda^{2}}\left(1+\nu^{2}+\frac{1}{\delta}\right).
\label{eq:block_center_bound}
\end{align}
The scalar proximal update is written as,
\[
w_{j}-rw^{\star}=\Pi_{\left[\sigma_{j}\tau_{j}\left(\theta_{j}w^{\star}+\frac{r_{j}}{\sqrt{\delta}}Z_{j}\right)-\lambda\sigma_{j}\tau_{j}-rw^{\star},\sigma_{j}\tau_{j}\left(\theta_{j}w^{\star}+\frac{r_{j}}{\sqrt{\delta}}Z_{j}\right)+\lambda\sigma_{j}\tau_{j}-rw^{\star}\right]}\left(w_{j-1}-rw^{\star}\right)
\]
Each projection center belongs to
$\left[-|w|\max_{u<j\leq t}|\sigma_j\tau_j\theta_j-r|,|w|\max_{u<j\leq t}|\sigma_j\tau_j\theta_j-r|\right].$
By $\sqrt{\frac{\varphi(1)}{2(1+e)^2(1+\nu)}}\leq r_j$,
\[
2\Phi\left(-\frac{\lambda\sqrt{\delta}}{r_{j}}\right)\geq2\Phi\left(-\lambda\sqrt{\frac{2(1+e)^{2}\delta(1+\nu)}{\varphi(1)}}\right).
\]
Using \cref{eq:11},
\[
\sqrt{\nu^2\alpha_j^2+\sigma_j^2}\leq\nu r+\sqrt{\E[(w_j-rw^{\star})^2]},
\]
and therefore from \cref{eq:q_linear_R} and $r\leq 1$,
\begin{align*}
\sigma_{j}\tau_{j}&\leq32\sqrt{2\pi}e^{\frac{1}{2}}(1+\nu)\left(1+\nu+\sqrt{\E[(w_{j}-rw^{\star})^{2}]}\right)\leq32\sqrt{2\pi}e^{\frac{1}{2}}(1+\nu)^{2}\left(1+\max_{u\leq k\leq t}\sqrt{\E[(w_{k}-rw^{\star})^{2}]}\right).
\end{align*}
Consequently using $r_j\leq 1$,
\begin{align}
\frac{\left(\frac{\sigma_{j}\tau_{j}r_{j}}{\sqrt{\delta}}\right)^{2}}{\left(\frac{\lambda\sqrt{\delta}}{r_{j}}\right)^{2}}=\frac{(\sigma_{j}\tau_{j})^{2}r_{j}^{4}}{\delta^{2}\lambda^{2}}\leq\frac{C(1+\nu)^{4}}{\delta^{2}\lambda^{2}}\left(1+\max_{u\leq k\leq t}\sqrt{\E[(w_{k}-rw^{\star})^{2}]}\right)^{2}.
\label{eq:block_noise_scale}
\end{align}

Applying \cref{eq:moving_band_projection} successively over the block, and using the telescopic identity
\[
\sum_{j=u+1}^{t}2\Phi\left(-\frac{\lambda\sqrt{\delta}}{r_{j}}\right)\prod_{k=j+1}^{t}\left(1-2\Phi\left(-\frac{\lambda\sqrt{\delta}}{r_{k}}\right)\right)=1-\prod_{j=u+1}^{t}\left(1-2\Phi\left(-\frac{\lambda\sqrt{\delta}}{r_{j}}\right)\right)\leq1,
\]
with $d=(\sigma_j\tau_j\theta_j-r)w^{\star},$ $D=|w^{\star}|\max_{u<k\leq t}|\sigma_k\tau_k\theta_k-r|,$ $X=w_{j-1}-rw^{\star},$ $X'=w_j-rw^{\star},$
gives
\begin{align*}
\E\left[\left(|w_{t}-rw^{\star}|-|w^{\star}|\max_{u<k\leq t}|\sigma_{k}\tau_{k}\theta_{k}-r|\right)^{2}_+\right]\leq&\exp\left(-2\Phi\left(-\lambda\sqrt{\frac{2(1+e)^{2}\delta(1+\nu)}{\varphi(1)}}\right)(t-u)\right)\E[(w_{u}-rw^{\star})^{2}]\\&+\frac{C(1+\nu)^{4}}{\delta^{2}\lambda^{2}}\left(1+\max_{u\leq j\leq t}\sqrt{\E[(w_{j}-rw^{\star})^{2}]}\right)^{2}.
\end{align*}
From Minkowski inequality, $|w_{t}-rw^{\star}|\leq\left(|w_{t}-rw^{\star}|-|w^{\star}|\max_{u<k\leq t}|\sigma_{k}\tau_{k}\theta_{k}-r|\right)_{+}+|w^{\star}|\max_{u<k\leq t}|\sigma_{k}\tau_{k}\theta_{k}-r|$,
\begin{align*}
\sqrt{\E[(w_{t}-rw^{\star})^{2}]}\leq\sqrt{\E\left[\left(|w_{t}-rw^{\star}|-|w^{\star}|\max_{u<k\leq t}|\sigma_{k}\tau_{k}\theta_{k}-r|\right)^{2}_+\right]}+\nu\max_{u<k\leq t}|\sigma_{k}\tau_{k}\theta_{k}-r|.
\end{align*}
Combining with \cref{eq:block_center_bound} yields
\begin{equation}
\label{eq:block_main_inequality}
\begin{aligned}
\sqrt{\E[(w_{t}-rw^{\star})^{2}]}&\leq\exp\left(-\Phi\left(-\lambda\sqrt{\frac{2(1+e)^{2}\delta(1+\nu)}{\varphi(1)}}\right)(t-u)\right)\sqrt{\E[(w_{u}-rw^{\star})^{2}]}\\&\quad+\left(1-\frac{2\sqrt{2\pi}\varphi(1)\phi''(1)}{81(1+\nu)^{3}}+\frac{C(1+\nu)^{2}}{\delta\lambda}\right)\max_{u\leq j\leq t}\sqrt{\E[(w_{j}-rw^{\star})^{2}]}\\&\quad+\frac{C(1+\nu)^{2}}{\delta\lambda}+\frac{C(1+\nu)^{11}}{\delta\lambda^{2}}\left(1+\nu^{2}+\frac{1}{\delta}\right).
\end{aligned}
\end{equation}
By \cref{eq:lambda_lower_rhofree}, after increasing the universal constant $C$ if necessary, $\frac{C(1+\nu)^{2}}{\delta\lambda}\leq\frac{\sqrt{2\pi}\varphi(1)\phi''(1)}{324(1+\nu)^{3}}.$
Partition time into consecutive blocks of length $\left\lceil \frac{1}{\Phi\left(-\lambda\sqrt{\frac{2(1+e)^{2}\delta(1+\nu)}{\varphi(1)}}\right)}\log\left(\frac{324(1+\nu)^{3}}{\sqrt{2\pi}\varphi(1)\phi''(1)}\right)\right\rceil.$
Then the exponential factor across a full block is at most $\frac{\sqrt{2\pi}\varphi(1)\phi''(1)}{324(1+\nu)^3}.$
Applying \cref{eq:block_main_inequality} to a point attaining the maximum on each block gives two cases. If the maximum on the current block does not exceed the maximum on the previous block, then
\begin{align*}
&\max_{\text{curr. block}}\!\!\sqrt{\E[(w_{t}-rw^{\star})^{2}]}\leq\left(1-\frac{c}{(1+\nu)^{3}}\right)\max_{\text{prev. block}}
\!\!\!\sqrt{\E[(w_{t}-rw^{\star})^{2}]}
+\frac{C(1+\nu)^{2}}{\delta\lambda}+\frac{C(1+\nu)^{11}}{\delta\lambda^{2}}\left(1+\nu^{2}+\tfrac{1}{\delta}\right).
\end{align*}
If the maximum on the current block exceeds the maximum on the
previous block, moving the current-block maximum appearing on the right-hand side of
\cref{eq:block_main_inequality}
to the left gives
\begin{align*}
&\max_{\text{curr. block}}\sqrt{\E[(w_{t}-rw^{\star})^{2}]}\leq\frac{1}{6}\max_{\text{prev. block}}\sqrt{\E[(w_{t}-rw^{\star})^{2}]}
+C(1+\nu)^{3}\left[\frac{(1+\nu)^{2}}{\delta\lambda}+\frac{(1+\nu)^{11}}{\delta\lambda^{2}}\left(1+\nu^{2}+\frac{1}{\delta}\right)\right].
\end{align*}
Consequently, after subtracting $\frac{C(1+\nu)^{5}}{\delta\lambda}+\frac{C(1+\nu)^{14}}{\delta\lambda^{2}}\left(1+\nu^{2}+\frac{1}{\delta}\right),$
the maxima over consecutive blocks contract by a factor at most $1-\frac{c}{(1+\nu)^3}>\frac{1}{6}.$ The first block is bounded by $C(1+\nu)^4$ under \cref{eq:lambda_lower_rhofree}. Iterating the block inequality therefore gives
\begin{align}
&\sqrt{\E[(w_{t}-rw^{\star})^{2}]}\leq C(1+\nu)^{4}\exp\left(-\frac{ct\Phi\left(-\lambda\sqrt{\frac{2(1+e)^{2}\delta(1+\nu)}{\varphi(1)}}\right)}{(1+\nu)^{3}\log(C(1+\nu))}\right)+\frac{C(1+\nu)^{5}}{\delta\lambda}+\frac{C(1+\nu)^{14}}{\delta\lambda^{2}}\left(1+\nu^{2}+\frac{1}{\delta}\right).
\label{eq:eT_fixed_lambda}
\end{align}
Squaring and adjusting universal constants gives
\begin{align}
\E[(w_t-rw^{\star})^2]&\leq C(1+\nu)^8\exp\left(-\frac{ct\Phi\left(-\lambda\sqrt{\frac{2(1+e)^2\delta(1+\nu)}{\varphi(1)}}\right)}{(1+\nu)^3\log(C(1+\nu))}\right)+\frac{C(1+\nu)^{10}}{\delta^2\lambda^2}+\frac{C(1+\nu)^{28}}{\delta^2\lambda^4}\left(1+\nu^2+\frac{1}{\delta}\right)^2.
\label{eq:large_T_fixed_lambda_bound}
\end{align}

Choose $\lambda^2=C\left[\frac{(1+\nu)^7}{
\delta}\left(1+\nu^2+\frac{1}{\delta}\right)+\frac{(1+\nu)^{10}}{\delta^2}\right]+\frac{\varphi(1)}{2(1+e)^2\delta(1+\nu)}\log(2T)$ at all tasks, 
where $C$ is a sufficiently large universal constant.
Then \cref{eq:lambda_lower_rhofree} holds for every $T\geq1$, and $\lambda^{-2}\leq\frac{2(1+e)^2\delta(1+\nu)}{\varphi(1)\log(2T)}.$ Consequently, $\frac{C(1+\nu)^{10}}{\delta^{2}\lambda^{2}}\leq\frac{C(1+\nu)^{11}}{\delta\log(2T)},$ and
\[
\frac{C(1+\nu)^{28}}{\delta^{2}\lambda^{4}}\left(1+\nu^{2}+\frac{1}{\delta}\right)^{2}\leq\frac{C(1+\nu)^{30}}{\log^{2}(2T)}\left(1+\nu^{2}+\frac{1}{\delta}\right)^{2}.
\]
Moreover,
\[
\left(\lambda\sqrt{\frac{2(1+e)^{2}\delta(1+\nu)}{\varphi(1)}}\right)^{2}=\log(2T)+C\frac{2(1+e)^{2}}{\varphi(1)}\left[(1+\nu)^{8}\left(1+\nu^{2}+\frac{1}{\delta}\right)+\frac{(1+\nu)^{11}}{\delta}\right].
\]
By the Gaussian Mills lower bound,
\[
\Phi\left(-\lambda\sqrt{\frac{2(1+e)^{2}\delta(1+\nu)}{\varphi(1)}}\right)\geq\frac{c\exp\left(-C\left[(1+\nu)^{8}\left(1+\nu^{2}+\frac{1}{\delta}\right)+\frac{(1+\nu)^{11}}{\delta}\right]\right)}{\sqrt{\log(2T)+C\left[(1+\nu)^{8}\left(1+\nu^{2}+\frac{1}{\delta}\right)+\frac{(1+\nu)^{11}}{\delta}\right]}}T^{-\frac{1}{2}}.
\]
Substituting this into
\cref{eq:large_T_fixed_lambda_bound}
gives, for every $T\geq1$,
\begin{align*}
&\E[(w_{T}-rw^{\star})^{2}]\leq C(1+\nu)^{8}\exp\Bigg(-\frac{c\exp\left(-C\left[(1+\nu)^{8}\left(1+\nu^{2}+\frac{1}{\delta}\right)+\frac{(1+\nu)^{11}}{\delta}\right]\right)\sqrt{T}}{(1+\nu)^{3}\log(C(1+\nu))\sqrt{\log(2T)+C\left[(1+\nu)^{8}\left(1+\nu^{2}+\frac{1}{\delta}\right)+\frac{(1+\nu)^{11}}{\delta}\right]}}\Bigg)\\&\hspace{12em}+\frac{C(1+\nu)^{11}}{\delta\log(2T)}+\frac{C(1+\nu)^{30}}{\log^{2}(2T)}\left(1+\nu^{2}+\frac{1}{\delta}\right)^{2},
\end{align*}
Finally, \cref{prop:coefficient_error_to_excess_risk} closes the proof.
\end{proof}

\newpage

\subsection{Proof for Lemma~\ref{thm:finite-L1-bound}}
\label{prf:finite-L1-bound}
\begin{fuller}[\cref{thm:finite-L1-bound}]
There exist universal constants $c,C>0$, such that the following holds. 
If 
$$s\left(\log\left(\frac{1}{s}\right)+\log\left(2T\right)\right)\le c\frac{\delta}{(1+\nu)^{18}}\,$$ 
and we choose the same regularization parameter at every task as $\lambda=\frac{2}{\sqrt{\delta}}\sqrt{\log\left(\frac{1}{s}\right)+\log\left(2T\right)},$ then
\[
\mathcal{L}(\mathbf w_T)-\inf_{\vw} \mathcal{L}(\vw)\le C\frac{s(1+\nu)^{12}}{\delta}\left(\log\left(\frac{1}{s}\right)+\log(2T)\right).
\]
\end{fuller}
\begin{proof}
The temporal L1 regularization, by \cref{lem:2}, follows
\[
w_{t}=\Pi_{\left[\sigma_{t}\tau_{t}\left(\theta_{t}w^{\star}
+\frac{r_{t}}{\sqrt{\delta}}Z_{t}-\lambda\right),
\sigma_{t}\tau_{t}\left(\theta_{t}w^{\star}
+\frac{r_{t}}{\sqrt{\delta}}Z_{t}+\lambda\right)\right]}(w_{t-1}).
\]

We first control the probability of an update. On $\{w^{\star}=0\}$, as long as the coordinate has not moved from zero, an update at task $j$ requires $\left|\frac{r_j}{\sqrt{\delta}}Z_j\right|>\lambda.$ From the fourth scalar equation (see \cref{eq:rt_identity} and derivations above it),
\[
r_{t}^{2}=2\E\left[(1-\pi_{\rho}(\nu Z_{1}))\phi'\left(\prox_{\gamma_{t}\phi}(\nu\alpha_{t}Z_{1}+\sigma_{t}Z_{2})\right)^{2}\right].
\]
Since $0<\phi'(u)<1$ and $2\mathbb E[1-\pi_\rho(\nu Z_1)]=1$, we have $0<r_t\leq1$ (see conclusion beneath \cref{eq:34}). Then almost surely (unless one of the endpoints in the projection becomes exactly zero),
\[
\PP\left(w_{t}\neq0\mid w^{\star}=0\right)=\PP\left(\exists j:\left|\frac{r_{j}}{\sqrt{\delta}}Z_{j}\right|>\lambda\right)\leq2T\Phi\left(-\lambda\sqrt{\delta}\right)\leq Te^{-\frac{\delta\lambda^{2}}{2}}\underset{\lambda^{2}=\frac{4}{\delta}\left(\log\left(\frac{1}{s}\right)+\log\left(2T\right)\right)}{=}\frac{s^{2}}{4T}.
\]
Consequently, because an update can occur either on an active
coordinate, whose probability is $s$, or on a null coordinate that is nonzero at time $t$, as $\{w_{t}\neq w_{t-1},w^{\star}=0\}\subseteq\{w_{t}\neq0,w^{\star}=0\}$ almost surely, $\PP(w_{t}\neq w_{t-1})\leq s+\frac{s^{2}}{4T}\leq2s.$
Using $\frac{\gamma_t}{\sigma_t\tau_t}=\frac{1}{\delta}\PP(w_t\neq w_{t-1})$ we obtain uniformly for $t\leq T$, $\frac{\gamma_t}{\sigma_t\tau_t}\leq\frac{2s}{\delta}.$
We next control $\theta_t$. From \cref{eq:24},
\begin{align*}
2\E\left[\pi_{\rho}'(\nu Z_{1})\phi'\left(\nu\alpha_{t}Z_{1}+\sigma_{t}Z_{2}\right)\right]=(1-2\rho)h(\nu)\stackrel{\mathrm{\ref{lem:rel}}}{=}rh(\nu r).
\end{align*}
Therefore using the equation for $\theta_t$ (see \cref{eq:22}),
\[
0\leq rh(\nu r)-\theta_{t}=2\E\Big[\pi_{\rho}'(\nu Z_{1})\Big(\phi'(\nu\alpha_{t}Z_{1}+\sigma_{t}Z_{2})-\phi'\left(\prox_{\gamma_{t}\phi}(\nu\alpha_{t}Z_{1}+\sigma_{t}Z_{2})\right)\Big)\Big].
\]

We will also use the sixth scalar equation. By \cref{eq:25}
\[
\frac{1}{\sigma_{t}\tau_{t}}=2\E\left[\left(1-\pi_{\rho}(\nu Z_{1})\right)\frac{\phi''\left(\prox_{\gamma_{t}\phi}\left(\nu\alpha_{t}Z_{1}+\sigma_{t}Z_{2}\right)\right)}{1+\gamma_{t}\phi''\left(\prox_{\gamma_{t}\phi}\left(\nu\alpha_{t}Z_{1}+\sigma_{t}Z_{2}\right)\right)}\right].\]
The change of variables $u=\prox_{\gamma_t\phi}(x)$ which gives $x=u+\gamma_t\phi'(u)$ and hence $\dd x=(1+\gamma_t\phi''(u))\dd u,$ so,
\[
\frac{1}{\sigma_{t}\tau_{t}}=\frac{2}{\sqrt{2\pi\left(\nu^{2}\alpha_{t}^{2}+\sigma_{t}^{2}\right)}}\int_{\R}\phi''(u)\exp\left(-\frac{\left(u+\gamma_{t}\phi'(u)\right)^{2}}{2\left(\nu^{2}\alpha_{t}^{2}+\sigma_{t}^{2}\right)}\right)\dd u\leq\frac{2}{\sqrt{2\pi\left(\nu^{2}\alpha_{t}^{2}+\sigma_{t}^{2}\right)}}\underbrace{\int_{\R}\phi''(u)\dd u}_{=1}.
\]
Note that we integrate over the random variable $\nu\alpha_tZ_1+\sigma_t Z_2\sim \mathcal N\left(0,\sqrt{\nu^2\alpha_t^2+\sigma_t^2}\right)$ therefore $\frac{\sqrt{\nu^{2}\alpha_{t}^{2}+\sigma_{t}^{2}}}{\sigma_{t}\tau_{t}}\leq\sqrt{\frac{2}{\pi}}.$ Applying \cref{eq:26} with $q=\sigma_t\tau_t$, $qH_t=\nu\alpha_tZ_1+\sigma_tZ_2$, $\eta=\frac{\gamma_t}{\sigma_t\tau_t}$, and $R=\sqrt{\nu^2\alpha_t^2+\sigma_t^2}$ gives,

\begin{align}
\frac{1}{\sigma_t\tau_t}\geq\frac{c}{(1+\nu)\left(1+\sqrt{\nu^2\alpha_t^2+\sigma_t^2}+\gamma_t\right)}
\label{eq:q_lower_small_s}
\end{align}
for a universal constant $c>0$.
If $\sqrt{\nu^2\alpha_t^2+\sigma_t^2}\leq1$, combining \cref{eq:q_lower_small_s} with $\frac{\gamma_t}{\sigma_t\tau_t}\leq\frac{2s}{\delta}$, we obtain $\sigma_t\tau_t\leq C(1+\nu)$, provided $\frac{s}{\delta}\leq\frac{c}{1+\nu}$. Since $\phi'$ is $\frac{1}{4}$-Lipschitz, $|\pi_{\rho}'(u)|=|(1-2\rho)\sigma(u)(1-\sigma(u))|\leq \frac{1}{4}$ and 
\[
\left|\nu\alpha_tZ_1+\sigma_tZ_2-\prox_{\gamma_t\phi}(\nu\alpha_tZ_1+\sigma_tZ_2)\right|=\left|\gamma_t\phi'\left(\prox_{\gamma_t\phi}(\nu\alpha_tZ_1+\sigma_tZ_2)\right)\right|\leq\gamma_t.
\]
therefore,
\[
rh(\nu r)-\theta_t\leq\frac{\gamma_t}{8}\leq C(1+\nu)\frac{s}{\delta}.
\]
If $\sqrt{\nu^2\alpha_t^2+\sigma_t^2}\geq1$, then by
\cref{eq:q_lower_small_s} $\frac{\gamma_{t}}{\sqrt{\nu^{2}\alpha_{t}^{2}+\sigma_{t}^{2}}}=\frac{\gamma_{t}}{\sigma_{t}\tau_{t}}\frac{\sigma_{t}\tau_{t}}{\sqrt{\nu^{2}\alpha_{t}^{2}+\sigma_{t}^{2}}}\underset{\frac{\gamma_{t}}{\sigma_{t}\tau_{t}}\leq\frac{2s}{\delta}}{\leq}\frac{2s}{\delta}\frac{\sigma_{t}\tau_{t}}{\sqrt{\nu^{2}\alpha_{t}^{2}+\sigma_{t}^{2}}}$, again provided $\frac{s}{\delta}\leq\frac{c}{1+\nu}$ (we take such sufficiently small $s$) so $\frac{\sigma_t\tau_t}{\sqrt{\nu^2\alpha_t^2+\sigma_t^2}}\leq C(1+\nu)$. Moreover let $f(x)=\phi'(x)$ and $g(x)=\phi'(\prox_{\gamma_t\phi}(x)).$ Since $f^{-1}(y)=(\phi'^{-1}(y))$, while the proximal identity $x=u+\gamma_t\phi'(u)$ gives $g^{-1}(y)=(\phi')^{-1}(y)+\gamma_ty,$ we obtain
\begin{align}
\label{eq:29} 
\int_{\R}\left[\phi'(x)-\phi'\left(\prox_{\gamma_{t}\phi}(x)\right)\right]\dd x=\int_{0}^{1}\left[g^{-1}(y)-f^{-1}(y)\right]\dd y=\int_{0}^{1}\gamma_{t}y\dd y=\frac{\gamma_{t}}{2}.
\end{align}

Conditioning on $\nu\alpha_tZ_1+\sigma_tZ_2=x$ and using $\pi_\rho'(u)\leq\frac{1}{4}$ therefore gives
\[
rh(\nu r)-\theta_t\leq C\frac{\gamma_t}{\sqrt{\nu^2\alpha_t^2+\sigma_t^2}}\leq C(1+\nu)\frac{s}{\delta}.
\]
Thus, uniformly in $t\leq T$ and uniformly in $\rho$,
\begin{align}
\left|\theta_{t}-rh(\nu r)\right|\leq C(1+\nu)\frac{s}{\delta}.
\label{eq:thetamall_s_rhofree}
\end{align}

Also, from \cref{eq:27} and $\frac{\gamma_t}{\sigma_t\tau_t}\leq\frac{2s}{\delta}$,
\begin{align}
\left|\frac{1}{\sigma_{t}\tau_{t}}-h\left(\sqrt{\nu^{2}\alpha_{t}^{2}+\sigma_{t}^{2}}\right)\right|\leq C\frac{s}{\delta}.
\label{eq:qtemporal_small_s}
\end{align}

Since the projection output belongs to its defining interval, $\left|\frac{w_{t}}{\sigma_{t}\tau_{t}}-\theta_{t}w^{\star}-\frac{r_{t}}{\sqrt{\delta}}Z_{t}\right|\leq\lambda.$
Multiplying by $|w^{\star}|$ and taking expectations, using
$\E|w^{\star}|=\nu\sqrt{\frac{2s}{\pi}},$ and $\E[w^{\star} w_t]=\nu^2\alpha_t$, yields
\[
\left|\frac{\alpha_{t}}{\sigma_{t}\tau_{t}}-\theta_{t}\right|\leq\frac{\lambda}{\nu}\sqrt{\frac{2s}{\pi}}\leq\frac{C}{\nu\sqrt{\delta}}\sqrt{s\left(\log\left(\frac{1}{s}\right)+\log(2T)\right)}.
\]
Combining this with \cref{eq:thetamall_s_rhofree},
\begin{align}
\left|\frac{\alpha_{t}}{\sigma_{t}\tau_{t}}-rh(\nu r)\right|\leq\frac{C}{\nu\sqrt{\delta}}\sqrt{s\left(\log\left(\frac{1}{s}\right)+\log(2T)\right)}+C(1+\nu)\frac{s}{\delta}.
\label{eq:alpha_normalized_small_s}
\end{align}

Moreover, by \cref{eq:projection_variance_temporal},
\[
\frac{\nu^{2}\alpha_{t}^{2}+\sigma_{t}^{2}}{\left(\sigma_{t}\tau_{t}\right)^{2}}-\nu^{2}\left(\frac{\alpha_{t}}{\sigma_{t}\tau_{t}}\right)^{2}=\frac{\sigma_{t}^{2}}{(\sigma_{t}\tau_{t})^{2}}\leq\E\left[\left(\frac{w_{t}}{\sigma_{t}\tau_{t}}-\theta_{t}w^{\star}\right)^{2}\right].
\]
On $\{w^{\star}\neq0\}$ because $r_t\leq 1$ and the projection output belongs to its defining interval,
\[
\E\left[\left(\frac{w_{t}}{\sigma_{t}\tau_{t}}-\theta_{t}w^{\star}\right)^{2}\mathbf{1}_{\{w^{\star}\neq0\}}\right]\leq s\E\left[\left(\lambda+\frac{|Z_{t}|}{\sqrt{\delta}}\right)^{2}\right]\leq C\frac{s}{\delta}\left(\log\left(\frac{1}{s}\right)+\log(2T)\right).
\]
On $\{w^{\star}=0\}$ the coordinate is zero except on an event of probability at most $\frac{s^2}{4T}$; using the same tail bound as above shows that the null contribution is smaller than the same right-hand side. Hence
\begin{align}
\frac{\sigma_{t}^{2}}{(\sigma_{t}\tau_{t})^{2}}\leq C\frac{s}{\delta}\left(\log\left(\frac{1}{s}\right)+\log(2T)\right).
\label{eq:12}
\end{align}

Consequently, by
\cref{eq:alpha_normalized_small_s,eq:12}, and using $\frac{s}{\delta}\left(\log\left(\frac{1}{s}\right)+\log(2T)\right)\leq\frac{c}{(1+\nu)^{18}},$ and the reverse triangle inequality \[\left|\underbrace{\sqrt{\frac{\nu^{2}\alpha_{t}^{2}+\sigma_{t}^{2}}{\sigma_{t}^{2}\tau_{t}^{2}}}}_{=\left\Vert \left(\frac{\nu\alpha_{t}}{\sigma_{t}\tau_{t}},\frac{\sigma_t}{\sigma_t\tau_t}\right)\right\Vert }-\underbrace{\nu rh(\nu r)}_{=\left\Vert \left(\nu rh(\nu r),0\right)\right\Vert }\right|\leq\sqrt{\left(\frac{\nu\alpha_{t}}{\sigma_{t}\tau_{t}}-\nu rh\left(\nu r\right)\right)^{2}+\frac{\sigma_t^2}{(\sigma_t\tau_t)^2}}\leq O\left(\sqrt{\frac{s}{\delta}\left(\log\left(\frac{1}{s}\right)+\log(2T)\right)}\right)\] where the constant in $O(\cdot)$ is universal, then we have
\begin{align}
\frac{\sqrt{\nu^{2}\alpha_{t}^{2}+\sigma_{t}^{2}}}{\sigma_{t}\tau_{t}}=\nu rh(\nu r)+O\left(\sqrt{\frac{s}{\delta}\left(\log\left(\frac{1}{s}\right)+\log(2T)\right)}\right),
\label{eq:13}
\end{align}
and $\sigma_t\tau_t\leq C(1+\nu)^3$ uniformly for $t\leq T$. Indeed, $\frac{1}{\sqrt{2\pi}}-\nu h(\nu)=\int_{\R}\phi''(u)\left(1-e^{-\frac{u^{2}}{2\nu^{2}}}\right)\dd u\geq\frac{c}{(1+\nu)^{2}}.$ By the sixth scalar equation, $\frac{\sqrt{\nu^{2}\alpha_{t}^{2}+\sigma_{t}^{2}}}{\sigma_{t}\tau_{t}}=\frac{1}{\sqrt{2\pi}}\int_{\R}F_{t}\left(x\right)e^{-\frac{x^{2}}{2\left(\nu^{2}\alpha_{t}^{2}+\sigma_{t}^{2}\right)}}\frac{\dd}{\dd x}\phi'\left(\prox_{\gamma_{t}\phi}\left(x\right)\right)\dd x$, and so
\begin{align*}
&\left|\frac{\sqrt{\nu^{2}\alpha_{t}^{2}+\sigma_{t}^{2}}}{\sigma_{t}\tau_{t}}-\frac{1}{\sqrt{2\pi}}\right|\leq\frac{1}{\sqrt{2\pi}}\int_{\R}\left|F_{t}\left(x\right)-1\right|\frac{\dd}{\dd x}\phi'\left(\prox_{\gamma_{t}\phi}\left(x\right)\right)\\&\qquad+\frac{1}{\sqrt{2\pi}}\int_{\R}F_{t}\left(x\right)\frac{\dd}{\dd x}\phi'\left(\prox_{\gamma_{t}\phi}\left(x\right)\right)\left|e^{-\frac{x^{2}}{2\left(\nu^{2}\alpha_{t}^{2}+\sigma_{t}^{2}\right)}}-1\right|\dd x\leq C\frac{\left(1+\nu\right)\left(1+\gamma_{t}\right)}{\sqrt{\nu^{2}\alpha_{t}^{2}+\sigma_{t}^{2}}},
\end{align*}
where $F_t(x)=\E[1-2\pi_\rho(\nu Z_1)\mid \nu \alpha_t Z_1+\sigma_t Z_2=x]$ and $\int_{\R}\frac{\dd}{\dd x}\phi'\left(\prox_{\gamma_{t}\phi}\left(x\right)\right)\dd x=1$ together with the conditional estimate above and $|e^{-u}-1|\leq |u|$ with \[\int_{\R}\left|x\right|\frac{\dd}{\dd x}\phi'\left(\prox_{\gamma_{t}\phi}\left(x\right)\right)\dd x=\int_{\R}\left|u+\gamma_{t}\phi'(u)\right|\phi''(u)\dd u\leq\underbrace{\int_{\R}|u|\phi''(u)\dd u}_{=2\int_{0}^{\infty}\frac{1}{1+e^{-u}}\dd u=2\log2}+\gamma_{t}\underbrace{\int_{\R}\phi'(u)\phi''(u)\dd u}_{=\frac{1}{2}\int_{\R}\frac{\dd}{\dd u}\left(\phi'(u)^{2}\right)\dd u=\frac{1}{2}},\] because $\frac{\gamma_t}{\sigma_t\tau_t}\leq \frac{2s}{\delta},$ $\sigma_t\tau_t\leq C(1+\nu)$ thus $\frac{\left(1+\nu\right)\left(1+\gamma_{t}\right)}{\sqrt{\nu^{2}\alpha_{t}^{2}+\sigma_{t}^{2}}}\leq C\frac{1+\nu}{\sqrt{\nu^{2}\alpha_{t}^{2}+\sigma_{t}^{2}}}+C\frac{\left(1+\nu\right)^{2}}{\sqrt{\nu^{2}\alpha_{t}^{2}+\sigma_{t}^{2}}}\frac{s}{\delta}$ and if $\sqrt{\nu^2\alpha_t^2+\sigma_t^2}\geq C(1+\nu)^3,$
\[
\frac{\sqrt{\nu^{2}\alpha_{t}^{2}+\sigma_{t}^{2}}}{\sigma_{t}\tau_{t}}=\frac{1}{\sqrt{2\pi}}+O\left(\frac{1+\nu}{\sqrt{\nu^{2}\alpha_{t}^{2}+\sigma_{t}^{2}}}+\frac{1}{1+\nu}\frac{s}{\delta}\right).
\]
For a sufficiently large universal $C$ and sufficiently small
$\frac{s}{\delta}(\log\left(\frac{1}{s}\right)+\log(2T))$, this contradicts
\cref{eq:13}, since
\[
\nu rh(\nu r)=\nu(1-2\rho)h(\nu)\leq\nu h(\nu)\leq\frac{1}{\sqrt{2\pi}}-\frac{c}{(1+\nu)^{2}}.
\]
Hence $\sqrt{\nu^2\alpha_t^2+\sigma_t^2}\leq C(1+\nu)^3.$ Using \cref{eq:qtemporal_small_s} and $h(x)\geq\frac{c}{1+x}$ (see \cref{eq:28}) we conclude that $\sigma_t\tau_t\leq C(1+\nu)^3$ uniformly for $t\leq T,$ after choosing $\frac{s}{\delta}$ sufficiently small. It follows that
\[
\left|\sqrt{\nu^{2}\alpha_{t}^{2}+\sigma_{t}^{2}}-\nu rh(\nu r)\sigma_{t}\tau_{t}\right|\leq C\frac{(1+\nu)^{3}}{\sqrt{\delta}}\sqrt{s\left(\log\left(\frac{1}{s}\right)+\log(2T)\right)}.
\]
Using \cref{eq:qtemporal_small_s} and since $h$ is Lipschitz as $\sup_u|\phi'''(u)|=\frac{1}{6\sqrt{3}}$, we may
write
\begin{align}
\left|\sigma_{t}\tau_{t}h\left(\nu rh(\nu r)\sigma_{t}\tau_{t}\right)-1\right|\leq C\frac{(1+\nu)^{6}}{\sqrt{\delta}}\sqrt{s\left(\log\left(\frac{1}{s}\right)+\log(2T)\right)}.
\label{eq:F_crude_small_s}
\end{align}

The function $x\longmapsto xh\left(\nu rh(\nu r)x\right)$ is strictly increasing, and by \cref{eq:14}
\begin{align}
\frac{\dd}{\dd x}\left[xh\left(\nu rh(\nu r)x\right)\right]=\E\left[Z^{2}\phi''\left(\nu rh(\nu r)xZ\right)\right]\geq\frac{c}{(1+\nu rh(\nu r)x)^3}.
\label{eq:F_derivative_small_s}
\end{align}
and it equals one at $x=\frac{1}{h(\nu r)}.$ On $\frac{1}{2h(\nu r)}\leq x\leq\frac{2}{h(\nu r)}$ we have $\nu rh(\nu r)x\leq 2\nu r\leq 2\nu$ (because $r\leq 1$ see \cref{lem:rel}).
Integrating \cref{eq:F_derivative_small_s} from $\frac{1}{h(\nu r)}$ to either $\frac{1}{2h(\nu r)}$ or $\frac{2}{h(\nu r)}$ shows that
\[
\left|xh\left(\nu rh(\nu r)x\right)-1\right|\geq\frac{c}{(1+\nu)^3}
\]
at both endpoints. Therefore
\cref{eq:F_crude_small_s}, together with $\frac{s}{\delta}\left(\log\left(\frac{1}{s}\right)+\log(2T)\right)\leq\frac{c}{(1+\nu)^{18}},$
implies the inequality $\frac{1}{2h(\nu r)}\leq\sigma_t\tau_t\leq\frac2{h(\nu r)}.$
Since $\nu r\leq\nu$ and $h(\nu)\geq\frac{c}{1+\nu}$ (see \cref{eq:28}), we have in particular
\begin{align}
\sigma_t\tau_t\leq C(1+\nu).
\label{eq:q_sharp_small_s}
\end{align}

Returning to \cref{eq:13,eq:qtemporal_small_s}, now using the sharper
bound \cref{eq:q_sharp_small_s}, gives
\begin{align}
\left|\sigma_{t}\tau_{t}h\left(\nu rh(\nu r)\sigma_{t}\tau_{t}\right)-1\right|\leq C\frac{(1+\nu)^{2}}{\sqrt{\delta}}\sqrt{s\left(\log\left(\frac{1}{s}\right)+\log(2T)\right)}.
\label{eq:F_sharp_small_s}
\end{align}
Applying \cref{eq:F_derivative_small_s} and the mean-value theorem,
\[
\left|\sigma_{t}\tau_{t}-\frac{1}{h(\nu r)}\right|\leq C\frac{(1+\nu)^{5}}{\sqrt{\delta}}\sqrt{s\left(\log\left(\frac{1}{s}\right)+\log(2T)\right)}.
\]
Finally, using
\[
\sigma_{t}\tau_{t}\theta_{t}-r=\sigma_{t}\tau_{t}\left(\theta_{t}-rh(\nu r)\right)+rh(\nu r)\left(\sigma_{t}\tau_{t}-\frac{1}{h(\nu r)}\right),
\]
together with
\cref{eq:thetamall_s_rhofree,eq:q_sharp_small_s}, we obtain
\begin{align}
\left|\sigma_{t}\tau_{t}\theta_{t}-r\right|\leq C\frac{(1+\nu)^{5}}{\sqrt{\delta}}\sqrt{s\left(\log\left(\frac{1}{s}\right)+\log(2T)\right)}.
\label{eq:center_small_s_explicit}
\end{align}

Now, we split into cases. If the coordinate $w_0$ never updates, then $\left|\sigma_{t}\tau_{t}\theta_{t}w^{\star}+\sigma_{t}\tau_{t}\frac{r_{t}}{\sqrt{\delta}}Z_{t}\right|\leq\sigma_{t}\tau_{t}\lambda,$ for all $t$. Thus $w_T=w_0=0$,
\begin{align*}
|w_{T}-rw^{\star}|=r|w^{\star}|&\leq|r-\sigma_{1}\tau_{1}\theta_{1}||w^{\star}|+\sigma_{1}\tau_{1}|\theta_{1}w^{\star}|\\&\leq C\frac{(1+\nu)^{5}}{\sqrt{\delta}}\sqrt{s\left(\log\left(\frac{1}{s}\right)+\log(2T)\right)}|w^{\star}|+C(1+\nu)\left(\lambda+\frac{\max_{j\leq T}|Z_{j}|}{\sqrt{\delta}}\right).
\end{align*}
Otherwise, let $J\leq T$ be the last update. Since an update has been made, $w_J=w_t$ has to be one of the endpoints in the interval, so $w_{T}=w_{J}=\sigma_{J}\tau_{J}\left(\theta_{J}w^{\star}+\frac{r_{J}}{\sqrt{\delta}}Z_{J}\pm\lambda\right),$
and then
\begin{align*}
|w_{T}-rw^{\star}|&\leq|\sigma_{J}\tau_{J}\theta_{J}-r||w^{\star}|+\sigma_{J}\tau_{J}\frac{r_{J}}{\sqrt{\delta}}|Z_{J}|+\sigma_{J}\tau_{J}\lambda\\&\leq C\frac{(1+\nu)^{5}}{\sqrt{\delta}}\sqrt{s\left(\log\left(\frac{1}{s}\right)+\log(2T)\right)}|w^{\star}|+C(1+\nu)\left(\lambda+\frac{\max_{j\leq T}|Z_{j}|}{\sqrt{\delta}}\right).
\end{align*}
Because $\PP\left(\max_{j\leq T}|Z_j|>u\right)\leq2Te^{-\frac{u^{2}}{2}},$
we have $\E[\max_{j\leq T}Z_j^2]\leq2\log(2T)+2.$
On the event $\{w^{\star}\neq0\}$, recall that
$\PP(w^{\star}\neq0)=s$ and
$\E[(w^{\star})^2]=\nu^2$. Hence
\begin{align}
\E\left[(w_{T}-rw^{\star})^{2}\mathbf{1}_{\{w^{\star}\neq0\}}\right]\leq C\frac{(1+\nu)^{12}}{\delta}s\left(\log\left(\frac{1}{s}\right)+\log(2T)\right).
\label{eq:active_final_small_s}
\end{align}

Otherwise, if $\{w^{\star}=0\}$, we distinguish between coordinates that were already nonzero at time $T-1$ and newly activated coordinates. If $w_{T-1}\neq0$, then the event $\{w^{\star}=0,w_{T-1}\neq0\}$ depends only on the previous Gaussian variables and is therefore independent of the fresh $Z_t$. By $r_t\leq 1$, $\left|\frac{w_t}{\sigma_t\tau_t}\right|\leq\lambda+\frac{|Z_t|}{\sqrt{\delta}},$
and hence
\begin{align*}
&\E\left[\left(\frac{w_{T}}{\sigma_{T}\tau_{T}}\right)^{2}\mathbf{1}_{\{w^{\star}=0,w_{t-1}\neq0\}}\right]\leq2(1-s)\PP(w_{T-1}\neq0\mid w^{\star}=0)\left(\lambda^{2}+\frac{1}{\delta}\right)\leq C\frac{s^{2}}{T}\left(\lambda^{2}+\frac{1}{\delta}\right),
\end{align*}
where we used
\[
\PP(w_{T-1}\neq0\mid w^{\star}=0)\leq2T\Phi(-\lambda\sqrt{\delta})\leq\frac{s^2}{4T}.
\]
If instead $w_{T-1}=0$, then on a null coordinate the temporal
soft-thresholding update gives exactly the equality $\frac{w_{T}}{\sigma_{T}\tau_{t}}=\mathrm{sign}(Z_{T})\left(\frac{r_{T}}{\sqrt{\delta}}|Z_{T}|-\lambda\right)_{+}.$ Therefore
\[
\E\left[\left(\frac{w_{T}}{\sigma_{T}\tau_{T}}\right)^{2}\mathbf{1}_{\{w^{\star}=0,w_{T-1}=0\}}\right]\leq\frac{2}{\delta}\Phi(-\lambda\sqrt{\delta})\leq C\frac{s^{2}}{\delta T^{2}},
\]
where we used $r_t\leq1$ (see conclusion beneath \cref{eq:34}) and \[\int_{x}^{\infty}(z-x)^{2}\varphi(z)\dd z\leq\int_{x}^{\infty}z(z-x)\varphi(z)\dd z=\int_{x}^{\infty}z^{2}\varphi(z)\dd z-\int_{x}^{\infty}zx\varphi(z)\dd z=(x\varphi(x)+\Phi(-x))-x\varphi(-x)=\Phi(-x).\]
Combining the two cases and using
\cref{eq:q_sharp_small_s},
\[
\E\left[w_{T}^{2}\mathbf{1}_{\{w^{\star}=0\}}\right]\leq C\frac{(1+\nu)^{2}}{\delta}s\left(\log\left(\frac{1}{s}\right)+\log(2T)\right),
\]
where we used $s\leq1$ and $T\geq1$.
Combining this with
\cref{eq:active_final_small_s}, we obtain
\[
\E[(w_{T}-rw^{\star})^{2}]\leq C\frac{s(1+\nu)^{12}}{\delta}\left(\log\left(\frac{1}{s}\right)+\log(2T)\right).
\]
The constant $C$ is universal and the bound is uniform over
$\rho\in[0,\frac{1}{2})$.
Finally, \cref{prop:coefficient_error_to_excess_risk} closes the proof.
\end{proof}
\subsection{Proof of lower bound for temporal L1 when $T=1$}
\label{prf:single_L1}
\begin{theorem}
\label{thm:single_L1}
There exist universal constants $c,C>0$ such that, for all $s\leq c\frac{\delta\nu^2(1-2\rho)^2}{(1+\nu)^2\log\left(\frac{1+\nu}{\delta\nu^2(1-2\rho)^2}\right)}$ and $\lambda>0$,
\[
\mathcal{L}(\mathbf w_1)-\inf_{\vw} \mathcal{L}(\vw)\geq \frac{C}{(1+\nu)^6}\frac{s}{\delta}\log\left(\frac{1}{s}\right).
\]
\end{theorem}

\begin{proof}
Since $w_0=0$, the first temporal L1 iterate by \cref{lem:2} follows,
\[
w_1=\sigma_1\tau_1\mathcal S_\lambda\left(\theta_1w^{\star}+\frac{r_1}{\sqrt\delta}Z\right).
\]
Recall that $\E[(w_1-rw^{\star})^2]=\nu^2(\alpha_1-r)^2+\sigma_1^2$ by \cref{eq:7}.
We argue by contradiction. Suppose that, for a sufficiently small universal constant $c_\star>0$,
\[
\nu^2(\alpha_1-r)^2+\sigma_1^2<c_\star\frac{s}{\delta}\log\left(\frac{1}{s}\right).
\]
Since $r>0$ (see \cref{lem:rel}) for every fixed $\rho<\frac{1}{2}$ and $s\log\left(\frac{1}{s}\right)\to0$, by decreasing $s_0$ if necessary, $\frac r2\leq\alpha_1\leq\frac{3r}{2}.$
We first note that $(\sigma_1\tau_1)r_1\geq1$. Indeed, the sixth scalar equation with $\E[1-2\pi_\rho(\nu Z_1)]=0$ gives,
\[
\frac{1}{\sigma_1\tau_1}=\E\left[\frac{2(1-\pi_\rho(\nu Z_1))\phi''(\prox_{\gamma_1\phi}(\nu\alpha_1Z_1+\sigma_1Z_2))}{1+\gamma_1\phi''(\prox_{\gamma_1\phi}(\nu\alpha_1Z_1+\sigma_1Z_2))}\right].
\]
Since $\phi''(u)\leq\phi'(u)$ and $\phi''(u)\geq 0$,
\[
\frac{1}{\sigma_1\tau_1}\leq\E\left[2(1-\pi_\rho(\nu Z_1))\phi'(\prox_{\gamma_1\phi}(\nu\alpha_1Z_1+\sigma_1Z_2))\right].
\]
By Cauchy--Schwarz and $2\E[1-\pi_\rho(\nu Z_1)]=1$ and the proximal optimality condition (\cref{eq:prox_relation_large_T}), using the fourth scalar equation, gives,
\[
\frac{1}{\sigma_1\tau_1}\leq\sqrt{2\E\left[(1-\pi_\rho(\nu Z_1))\phi'(\prox_{\gamma_1\phi}(\nu\alpha_1Z_1+\sigma_1Z_2))^2\right]}=r_1.
\]
Hence $(\sigma_1\tau_1)r_1\geq1.$ We next show that the threshold must be large on null coordinates. Let $Z\sim \mathcal N(0,1)$ be independent of $w^{\star}$, $\E[Z(w_1-rw^{\star})]=\E[Zw_1].$ Stein's Lemma gives
\[
\E[Zw_1]=\frac{\sigma_1\tau_1r_1}{\sqrt\delta}\PP\left(\left|\theta_1w^{\star}+\frac{r_1}{\sqrt\delta}Z\right|>\lambda\right).
\]
Restricting to $\{w^{\star}=0\}$ and using $(\sigma_1\tau_1)r_1\geq1$,
\[
\E[Zw_1]\geq\frac{2(1-s)}{\sqrt\delta}\Phi\left(-\frac{\lambda\sqrt\delta}{r_1}\right).
\]
Therefore, by Cauchy--Schwarz,
\[
\nu^{2}(\alpha_{1}-r)^{2}+\sigma_{1}^{2}=\E[(w_{1}-rw^{\star})^{2}]\geq\E[Z(w_{1}-rw^{\star})]^{2}\geq\frac{4(1-s)^{2}}{\delta}\Phi\left(-\frac{\lambda\sqrt{\delta}}{r_{1}}\right)^{2}.
\]
Thus, for $s\leq\frac{1}{2}$, $\E[(w_1-rw^{\star})^2]\geq\frac{1}{\delta}\Phi\left(-\frac{\lambda\sqrt\delta}{r_1}\right)^2.$
If $\frac{\lambda^2\delta}{r_1^2}\leq\frac{1}{2}\log\left(\frac{1}{s}\right),$ then the lower Mills bound $\Phi(-x)\geq c\frac{e^{-\frac{x^{2}}{2}}}{1+x}$ gives $\Phi\left(-\frac{\lambda\sqrt\delta}{r_1}\right)^2\geq c\frac{\sqrt{s}}{1+\log\left(\frac{1}{s}\right)}.$ Since $\frac{\sqrt{s}}{1+\log\left(\frac{1}{s}\right)}\gg s\log\left(\frac{1}{s}\right),$ this contradicts the assumed upper bound for a sufficiently small $s$. Consequently, $\frac{\lambda^2\delta}{r_1^2}\geq\frac{1}{2}\log\left(\frac{1}{s}\right).$

We now use the active coordinates. Conditionally on $\{w^{\star}\neq0\}$, write $w^{\star}=\frac{\nu}{\sqrt{s}}G$, with $G\sim \mathcal N(0,1)$ independent of $Z$. Then
\[
\theta_1w^{\star}+\frac{r_1}{\sqrt\delta}Z=\frac{\theta_1\nu}{\sqrt{s}}G+\frac{r_1}{\sqrt\delta}Z.
\]
Let $p=2\Phi\left(-\frac{\lambda}{\sqrt{\frac{\theta_1^2\nu^2}{s}+\frac{r_1^2}{\delta}}}\right).$
Stein's Lemma gives the exact identity $\alpha_1=(\sigma_1\tau_1)\theta_1p.$
Indeed,
\begin{align*}
\nu^{2}\alpha_{1}
&=\E[w^{\star}w_{1}]=s\frac{\nu}{\sqrt{s}}(\sigma_{1}\tau_{1})\E\left[G\mathcal{S}_{\lambda}\left(\frac{\theta_{1}\nu}{\sqrt{s}}G+\frac{r_{1}}{\sqrt{\delta}}Z\right)\right]=s\frac{\nu}{\sqrt{s}}(\sigma_{1}\tau_{1})\E\left[\mathcal{S}_{\lambda}'\left(\frac{\theta_{1}\nu}{\sqrt{s}}G+\frac{r_{1}}{\sqrt{\delta}}Z\right)\right]\\&=s\frac{\nu}{\sqrt{s}}(\sigma_{1}\tau_{1})\E\left[\mathbf{1}_{\left\{ \left|\frac{\theta_{1}\nu}{\sqrt{s}}G+\frac{r_{1}}{\sqrt{\delta}}Z\right|>\lambda\right\} }\right]
=(\sigma_{1}\tau_{1})\theta_{1}\nu^{2}p.
\end{align*}

We next lower bound the residual on active coordinates. Put $y=\frac{\lambda}{\sqrt{\frac{\theta_1^2\nu^2}{s}+\frac{r_1^2}{\delta}}},$
and let $U\sim \mathcal N(0,1)$. Since
\[
\mathcal S_\lambda\left(\frac{\theta_1\nu}{\sqrt{s}}G+\frac{r_1}{\sqrt\delta}Z\right)\stackrel{d}{=}\sqrt{\frac{\theta_1^2\nu^2}{s}+\frac{r_1^2}{\delta}}\mathcal S_y(U),
\]
we obtain
\begin{align*}
\sigma_{1}^{2}=\E[(w_{1}-\alpha_{1}w^{\star})^{2}]&\geq s\E\left[\left(\sigma_{1}\tau_{1}\mathcal{S}_{\lambda}\left(\frac{\theta_{1}\nu}{\sqrt{s}}G+\frac{r_{1}}{\sqrt{\delta}}Z\right)-\frac{\nu\alpha_{1}}{\sqrt{s}}G\right)^{2}\mid w^{\star}\neq0\right]
\geq\alpha_{1}^{2}\nu^{2}\left(\frac{\E[\mathcal{S}_{y}(U)^{2}]}{p^{2}}-1\right).
\end{align*}
Here we used $p=\PP(|U|>y)=\E[U\mathcal S_y(U)]$ by Stein's Lemma, and
\[
(\sigma_1\tau_1)^2\left(\frac{\theta_1^2\nu^2}{s}+\frac{r_1^2}{\delta}\right)\geq(\sigma_1\tau_1)^2\frac{\theta_1^2\nu^2}{s}=\frac{\alpha_1^2\nu^2}{sp^2},
\]
from $\alpha_1=(\sigma_1\tau_1)\theta_1p$ by the first scalar equation. For $y\geq1$,
\[
\frac{\E[\mathcal S_y(U)^2]}{p^2}-1=\frac{\E[(\mathcal S_y(U)-pU)^2]}{p^2}\geq\E[U^2\mathbf1_{\{|U|\leq1\}}]>0.
\]
Hence $y\geq1$ would imply $\sigma_1^2\geq c\alpha_1^2\nu^2\geq c r^2\nu^2,$
contradicting $\sigma_1^2\to0$. For $0\leq y\leq1$,
\[
\E[\mathcal S_y(U)^2]-p^2=\E[(\mathcal S_y(U)-pU)^2].
\]
Moreover, writing $1-p=\PP(|U|\leq y)=2\int_{0}^{y}\varphi(u)\dd u\leq2\varphi(0)y\leq\sqrt{\frac{2}{\pi}}y$, by Mill's ratio bound. On the event $1\leq U\leq\frac{9}{8}$,
\[
|\mathcal S_y(U)-pU|=|U-y-pU|=|y-(1-p)U|\geq\left(1-\frac98\sqrt{\frac2\pi}\right)y.
\]
Therefore there exists a universal constant $c>0$ such that $\frac{\E[\mathcal S_y(U)^2]}{p^2}-1\geq cy^2,$ and hence $\sigma_1^2\geq c\alpha_1^2\nu^2y^2\geq cr^2\nu^2y^2,$ where we use $s$ sufficiently small so that $\alpha_1\geq\frac{r}{2}.$
It remains to lower bound $y$. Using $\alpha_1=(\sigma_1\tau_1)\theta_1p$,
\[
(\sigma_1\tau_1)\frac{\theta_1\nu}{\sqrt{s}}=\frac{\alpha_1\nu}{p\sqrt{s}}\leq C\frac{r\nu}{\sqrt{s}},
\]
where we used $\alpha_1\leq\frac{3r}{2}$ for a sufficiently small $s$ and $p\geq2\Phi(-1)$. Also, $(\sigma_1\tau_1)\frac{r_1}{\sqrt\delta}\geq\frac{1}{\sqrt\delta}.$
Since $u\mapsto\frac{u^2}{a^2+u^2}$ is increasing for $u>0$,
\begin{align*}
y^{2}=\frac{(\sigma_{1}\tau_{1})^{2}\lambda^{2}}{(\sigma_{1}\tau_{1})^{2}\frac{\theta_{1}^{2}\nu^{2}}{s}+(\sigma_{1}\tau_{1})^{2}\frac{r_{1}^{2}}{\delta}}=\frac{\lambda^{2}\delta}{r_{1}^{2}}\frac{\left((\sigma_{1}\tau_{1})\frac{r_{1}}{\sqrt{\delta}}\right)^{2}}{(\sigma_{1}\tau_{1})^{2}\frac{\theta_{1}^{2}\nu^{2}}{s}+\left((\sigma_{1}\tau_{1})\frac{r_{1}}{\sqrt{\delta}}\right)^{2}}\geq\frac{\frac{1}{\delta}\frac{\lambda^{2}\delta}{r_{1}^{2}}}{C\frac{r^{2}\nu^{2}}{s}+\frac{1}{\delta}}\geq c\frac{s}{\delta r^{2}\nu^{2}}\frac{\lambda^{2}\delta}{r_{1}^{2}},
\end{align*}
where in the last step we decreased $s_0$ so that $s\leq\delta r^2\nu^2$. Using the threshold lower bound obtained above, $y^2\geq c\frac{s}{\delta r^2\nu^2}\log\left(\frac{1}{s}\right).$ Consequently,
\[
\nu^2(\alpha_1-r)^2+\sigma_1^2\geq\sigma_1^2\geq cr^2\nu^2y^2\geq\frac{c}{\delta}s\log\left(\frac{1}{s}\right),
\]
contradicting the assumption for $c_\star>0$ sufficiently small.

For some universal constants $C_1,C_2,C_3>0,$ a sufficiently small $s<s_0$ would need to make
\begin{align*}
&\frac{r}{2}\leq\alpha_1\leq\frac{3r}{2}\implies s\log\left(\frac{1}{s}\right)\leq C_1\delta\nu^2r^2\\&
s\leq C_2\delta r^2\nu^2.
\end{align*}
By \cref{lem:rel}, $rh(\nu r)=(1-2\rho)h(\nu)$ therefore since $h(\nu r)\leq h(0)=\frac{1}{4}$ so $r\geq 4(1-2\rho)h(\nu)$, by solving $s_0\log\left(\frac{1}{s_0}\right)\asymp \delta\nu^2(1-2\rho)^2h(\nu)^2$ we get, that there exists $C>0$ so that, $s_0\leq C\frac{\delta\nu^2(1-2\rho)^2}{(1+\nu)^2\log_+\left(\frac{1+\nu}{\delta\nu^2(1-2\rho)^2}\right)},$ as $h(\nu)=\frac{1}{\nu}\int_\R \phi''(u)\varphi\left(\frac{u}{\nu}\right)\dd u\leq \min\left\{\frac{1}{4},\frac{1}{\sqrt{2\pi}\nu}\right\}\leq C_3\frac{1}{1+\nu}.$ Combining with \cref{prop:coefficient_error_to_excess_risk} we get
\[
\mathcal{L}_1-\inf_{\vw} \mathcal{L}(\vw)\geq C\frac{1}{(1+\nu)^6}\frac{s}{\delta}\log\left(\frac{1}{s}\right),
\]
for some universal constant $C>0$. 
\end{proof}
\newpage
\subsection{Proof of \cref{prop:deep-equivalence}}
\label{prf:equ}
\begin{recall}[\cref{prop:deep-equivalence}]
Denote by $\mathbf{w}(\boldsymbol{\theta})$ the linear predictor induced by a {depth-$h$} DLN parameterized by $\boldsymbol{\theta}$. Then, for any predictor $\mathbf w\in\mathbb R^d$ and weighting function $c(\cdot,\cdot)>0$,
\begin{align*}
\sum_{i=1}^{d}c_{i}|w_{i}|^{p/h}=\min_{\substack{\mathbf{w}_{1},\ldots,\mathbf{w}_{h}\in\mathbb{R}^{d}:\\
F(\mathbf{w}_{1},\ldots,\mathbf{w}_{h})=\mathbf{w}
}
}\frac{1}{h}\sum_{i=1}^{d}c_{i}\sum_{\ell=1}^{h}|(\mathbf{w}_{\ell})_{i}|^{p}.
\end{align*}
\end{recall}

\begin{proof}
Since $w_i=\prod_{\ell=1}^{h}(\mathbf w_\ell)_i$, the constraint and the objective separate over the coordinates. For every $i$, $|w_i|^p=\prod_{\ell=1}^{h}|(\mathbf w_\ell)_i|^p$. Applying the arithmetic--geometric mean inequality to the $h$ nonnegative quantities $|(\mathbf w_1)_i|^p,\ldots,|(\mathbf w_h)_i|^p$ and multiplying by $c_{i}$ gives
\begin{align*}
\frac{c_{i}}{h}\sum_{\ell=1}^{h}|(\mathbf{w}_{\ell})_{i}|^{p}\geq c_{i}\left(\prod_{\ell=1}^{h}|(\mathbf{w}_{\ell})_{i}|^{p}\right)^{1/h}=c_{i}|w_{i}|^{p/h}.
\end{align*}
When $w_i\neq 0$, equality is attained by choosing $(\mathbf w_1)_i=\sign(w_i)|w_i|^{1/h}$ and $(\mathbf w_\ell)_i=|w_i|^{1/h}$ for every $\ell\in\{2,\ldots,h\}$. Indeed, these choices satisfy $\prod_{\ell=1}^{h}(\mathbf w_\ell)_i=w_i$, and all $h$ quantities in the arithmetic--geometric mean inequality equal $|w_i|^{p/h}$. Because $c_{i}$ multiplies all $h$ terms associated with coordinate $i$, it does not alter the equality condition. If $w_i=0$, equality is attained by taking $(\mathbf w_\ell)_i=0$ for every $\ell$. Summing over $i$ proves the result.
\end{proof}

\newpage

\subsection{Information-Theoretic Lower Bound}
\label{sec:lowerproof}
\begin{theorem}[Information-theoretic lower bound]
\label{thm:dense_info_lower}
Assume $\mathbf x\sim\mathcal N(0,\frac{1}{d}\mathbf I_d)$ and, conditionally on $\mathbf x$,
\[
\PP(y=1\mid\mathbf x)=\rho+(1-2\rho)\frac{e^{\mathbf x^\top\teacher}}{1+e^{\mathbf x^\top\teacher}},
\]
where $\rho\in[0,\frac{1}{2}),$ and $\frac{\Vert\teacher\Vert_2^2}{d}=\nu^2,$ with $\nu>0.$ Let $r>0$ be the solution of $r=\frac{(1-2\rho)\E[\phi''(\nu Z)]}{\E[\phi''(\nu r Z)]},$ where $Z\sim\mathcal N(0,1).$ Then, there exists $c_{\rho,\nu}>0$ (depending on $\rho,\nu$) such that
\[
\inf_{\widehat{\mathbf{w}}}\sup_{\Vert\teacher\Vert_2=\nu \sqrt d}= \nu\E\left[\mathcal{L}(\widehat{\mathbf{w}})-\mathcal{L}(r\teacher)\right]\ge c_{\rho,\nu}\min\left\{ 1,\frac{1}{\delta T}\right\} .
\]
\end{theorem}

\begin{proof}
We allow our estimator $\widehat{\mathbf w}$ to depend jointly on all $nT$ samples. Thus any lower bound applies to every sequential learner. Recall that
\[
\mathcal{L}(\mathbf{w})=\E\left[\phi(\mathbf{x}^{\top}\mathbf{w})-\left(\rho+(1-2\rho)\phi'(\mathbf{x}^{\top}\teacher)\right)\mathbf{x}^{\top}\mathbf{w}\right].
\]
Since $\E[\mathbf x]=0,$ the $\rho$-term vanishes. By Stein's Lemma, $\E[\phi'(\mathbf x^\top\teacher)\mathbf x]=\frac{h(\nu)}{d}\teacher.$ Therefore,
\[
\mathcal{L}(\mathbf{w})=\E[\phi(\mathbf{x}^{\top}\mathbf{w})]-\frac{(1-2\rho)h(\nu)}{d}\teacher^{\top}\mathbf{w}.
\]

For fixed $\Vert\mathbf w\Vert_2,$ the first term depends only on $\Vert\mathbf w\Vert_2,$ because $\mathbf x^\top\mathbf w\sim\mathcal N(0,\frac{\Vert\mathbf w\Vert_2^2}{d}).$ The second term is minimized by maximizing $\teacher^{\top}\mathbf w.$ By Cauchy--Schwarz, this happens when $\mathbf w$ is a nonnegative multiple of $\teacher.$ Thus the minimizer has the form $\mathbf w=a\teacher,$ with $a\ge0.$ For such we have $\mathbf w,$ $\mathbf x^\top(a\teacher)\sim\mathcal N(0,a^2\nu^2),$ and
\[
\frac{\dd}{\dd a}\mathcal{L}(a\teacher)=\nu^{2}\left[ah(\nu a)-(1-2\rho)h(\nu)\right].
\]
Moreover, $\frac{\dd}{\dd q}(qh(q))=\E[Z^2\phi''(qZ)]>0.$ Hence the unique minimizer is $r\teacher,$ where $rh(\nu r)=(1-2\rho)h(\nu).$ The Hessian is $\nabla^2\mathcal{L}(\mathbf w)=\E[\phi''(\mathbf x^\top\mathbf w)\mathbf x\mathbf x^\top].$ Thus, for every $\mathbf u\in\mathbb R^d,$
\[
\mathbf{u}^{\top}\nabla^{2}\mathcal{L}(\mathbf{w})\mathbf{u}=\E[\phi''(\mathbf{x}^{\top}\mathbf{w})(\mathbf{x}^{\top}\mathbf{u})^{2}].
\]
By rotational invariance, we take $\mathbf w=\|\mathbf w\|_2\mathbf e_1.$ Write $\mathbf x=d^{-\frac{1}{2}}(Z_1,\dots,Z_d)$ and $\mathbf u=u_\parallel\mathbf e_1+\mathbf u_\perp,$ 
\[
\E\left[\phi''(\mathbf{x}^{\top}\mathbf{w})(\mathbf{x}^{\top}\mathbf{u})^{2}\right]=\frac{1}{d}\left(u_{\parallel}^{2}\E\left[\phi''\left(\frac{\|\mathbf w\|_2}{\sqrt d}Z\right)Z^{2}\right]+\|\mathbf{u}_{\perp}\|_{2}^{2}\E\left[\phi''\left(\frac{\|\mathbf w\|_2}{\sqrt d}Z\right)\right]\right).
\]
Let $c=\min_{|a|\leq \nu}\{\min\{\E[\phi''(aZ)Z^2],\E[\phi''(aZ)]\}\}>0,$ then $\E[\phi''(\mathbf{x}^{\top}\mathbf{w})(\mathbf{x}^{\top}\mathbf{u})^{2}]\ge\frac{c}{d}\|\mathbf{u}\|_{2}^{2},$ 
On every set where $\frac{\Vert\mathbf w\Vert_2}{\sqrt d}$ is bounded by a constant depending only on $\nu,$ Gaussian rotational invariance and $\phi''>0$ imply $\nabla^2\mathcal{L}(\mathbf w)\succeq c_\nu d^{-1}\mathbf I_d.$ Therefore, by Taylor's theorem, if $\Vert\mathbf w-r\teacher\Vert_2\le c\sqrt d,$
\[
\mathcal{L}(\mathbf{w})-\mathcal{L}(r\teacher)\ge c_{\rho,\nu}\frac{\Vert\mathbf{w}-r\teacher\Vert_{2}^{2}}{d}.
\]
Moreover the risk is continuous, has the unique minimizer $r\teacher,$ and is coercive, $\E[\phi(\x^\top\mathbf w)]\geq \E[(\x^\top\mathbf w)_+]=\frac{\Vert\mathbf w\Vert}{\sqrt{2\pi}}$ and $-(1-2\rho)h(\nu)\mathbf{\teacher^{\top}\mathbf{w}}\geq-(1-2\rho)\nu h(\nu)\Vert\mathbf{w}\Vert$. Therefore,
\[
\mathcal{L}(\sqrt{d}\mathbf{w})=\left[\frac{1}{\sqrt{2\pi}}-\left(1-2\rho\right)\nu h(\nu)\right]\Vert\mathbf{w}\Vert\xrightarrow{\Vert\mathbf{w}\Vert\to\infty}\infty
\]
since $\frac{1}{\sqrt{2\pi}}-\nu h(\nu)=\int_{\R}\phi''(u)\left(1-e^{-\frac{u^{2}}{2\nu^{2}}}\right)\dd u\geq\frac{c}{(1+\nu)^{2}}.$
Hence, after decreasing $c_{\rho,\nu}$ if necessary, for all $\mathbf w\in\mathbb R^d,$
\begin{equation}
\mathcal{L}(\mathbf{w})-\mathcal{L}(r\teacher)\ge c_{\rho,\nu}\min\left\{ 1,\frac{\Vert\mathbf{w}-r\teacher\Vert_{2}^{2}}{d}\right\} .
\label{eq:global_risk_parameter_lower}
\end{equation}

We now use Assouad's Lemma. Assouad's Lemma provides a methodological way to give an uniform lower bound of the objective. Set $b^2=c_0\frac{d}{nT\vee d},$ where $c_0>0$ is sufficiently small depending only on $\nu.$ For every $\mathbf v\in\{-1,+1\}^{d-1},$ define
\[
\mathbf{w}_{\mathbf{v}}=\sqrt{\nu^{2}d-(d-1)b^{2}}\mathbf{e}_{1}+b\sum_{j=1}^{d-1}v_{j}\mathbf{e}_{j+1}.
\]
Then $\Vert\mathbf w_{\mathbf v}\Vert_2^2=\nu^2d.$
For two parameters $\mathbf v,\mathbf v'$ differing only in coordinate $j,$ $\Vert\mathbf w_{\mathbf v}-\mathbf w_{\mathbf v'}\Vert_2^2=4b^2.$ For the ordinary logistic model, its KL divergence is given by,
\[
D_{{\rm KL}}\left(\mathrm{Ber}(\phi'(u))\Vert\mathrm{Ber}(\phi'(v))\right)=\phi(v)-\phi(u)-\phi'(u)(v-u)\le\frac{1}{8}(u-v)^{2},
\]
because $\phi''\leq\frac{1}{4},$ and Taylor's Theorem. The noisy model with parameter $\rho$ is obtained by passing the logistic label through an independent binary symmetric channel with flip probability $\rho.$ The data-processing inequality for KL says that applying the same stochastic channel to two distributions cannot increase their KL divergence,
\[
D_{{\rm KL}}\left(\mathrm{Ber}(\rho+(1-2\rho)\phi'(u))\Vert\mathrm{Ber}(\rho+(1-2\rho)\phi'(v))\right)\le\frac{1}{8}(u-v)^{2}.
\]
Thus, for one observation,
\[
D_{{\rm KL}}(P_{\mathbf{v}}\Vert P_{\mathbf{v}'})\le\frac{1}{8}\E\left[(\mathbf{x}^{\top}(\mathbf{w}_{\mathbf{v}}-\mathbf{w}_{\mathbf{v}'}))^{2}\right]=\frac{1}{8d}\Vert\mathbf{w}_{\mathbf{v}}-\mathbf{w}_{\mathbf{v}'}\Vert_{2}^{2}=\frac{b^{2}}{2d}=\frac{c_{0}}{2(nT\vee d)}.
\]
For $nT$ observations, $D_{\rm KL}(P_{\mathbf v}^{\otimes nT}\Vert P_{\mathbf v'}^{\otimes nT})\le\frac{c_0}{2}.$ Choosing $c_0$ sufficiently small, Assouad's lemma implies that for every estimator $\widehat{\mathbf v}\in\{-1,+1\}^{d-1},$
\[
\sup_{\mathbf{v}\in\{-1,+1\}^{d-1}}\E_{\mathbf{v}}[d_{{\rm H}}(\widehat{\mathbf{v}},\mathbf{v})]\ge c(d-1).
\]
where $d_H(\mathbf v,\mathbf v')=\sum_{j=1}^{d-1}\mathbf{1}\{\hat{v}_{j}\neq v_{j}\}$ is the Hamming distance.
Given an arbitrary estimator $\widehat{\mathbf w},$ define its induced sign estimator by $\widehat v_j=\mathrm{sign}(\widehat w_{j+1}),$ with an arbitrary convention when $\widehat w_{j+1}=0.$ If $\widehat v_j\neq v_j,$ then $|\widehat w_{j+1}-rbv_j|\ge rb.$ Therefore
\[
\Vert\widehat{\mathbf w}-r\mathbf w_{\mathbf v}\Vert_2^2\ge r^2b^2d_{\rm H}(\widehat{\mathbf v},\mathbf v).
\]
Using \eqref{eq:global_risk_parameter_lower} and choosing $c_0$ smaller if necessary so that $r^2c_0\le1,$
\[
\mathcal{L}_{\mathbf{w}_{\mathbf{v}}}(\widehat{\mathbf{w}})-\mathcal{L}_{\mathbf{w}_{\mathbf{v}}}(r\mathbf{w}_{\mathbf{v}})\ge c_{\rho,\nu}\frac{r^{2}b^{2}}{d}d_{{\rm H}}(\widehat{\mathbf{v}},\mathbf{v}).
\]
Taking expectations and applying Assouad's lemma gives
\begin{align*}
\inf_{\widehat{\mathbf{w}}}\sup_{\Vert\teacher\Vert_{2}=\nu\sqrt{d}}\E\left[\mathcal{L}(\widehat{\mathbf{w}})-\mathcal{L}(r\teacher)\right]&
\ge c_{\rho,\nu}\frac{b^{2}}{d}\inf_{\widehat{\mathbf{v}}}\sup_{\mathbf{v}\in\{-1,+1\}^{d-1}}\E_{\mathbf{v}}[d_{{\rm H}}(\widehat{\mathbf{v}},\mathbf{v})]
\\&
\ge c_{\rho,\nu}\frac{b^{2}}{d}(d-1)\ge c_{\rho,\nu}\frac{d}{nT\vee d}=c_{\rho,\nu}\min\left\{ 1,\frac{d}{nT}\right\} .
\end{align*}

Finally, since $\frac{n}{d}=\delta\in(0,\infty),$ and $\frac{d}{nT}=\frac{1}{T(n/d)}=\frac{1}{\delta T}.$ Therefore,
\[
\inf_{\widehat{\mathbf{w}}}\sup_{\Vert\teacher\Vert_{2}=\nu\sqrt{d}}\E\left[\mathcal{L}(\widehat{\mathbf{w}})-\mathcal{L}(r\teacher)\right]\ge c_{\rho,\nu}\min\left\{ 1,\frac{1}{\delta T}\right\} .
\]
\end{proof}

\end{document}

%% file: 00_packages.tex
\usepackage{capt-of}
\usepackage{booktabs}
\usepackage{makecell}
\usepackage{cancel}
\usepackage{pifont}
\usepackage{enumitem}
\usepackage{wrapfig}
\usepackage{xspace}
\usepackage{float}
\usepackage{comment}
\usepackage{amsmath,amssymb,amsthm,mathtools,bm}
\usepackage{hyperref}
\usepackage{url}
\usepackage{amsmath,amssymb,amsfonts}
\usepackage{tikz}
\usepackage{pgfplots}
\usetikzlibrary{pgfplots.groupplots}
\pgfplotsset{compat=1.18}
\usepackage{caption}
\usepackage{booktabs}
\usepackage{multirow}
\usepackage{tabularx}
\usepackage{graphicx}
\usepackage{subcaption}
\usepackage{microtype}
\usepackage{subcaption}
\usepackage{hyperref}
\usepackage{amssymb}
\usepackage{mathtools}
\usepackage{amsthm}
\usepackage{aliascnt}
\usepackage{cancel}
\usepackage[capitalize,noabbrev]{cleveref}
\usepackage{algorithm,algorithmic}
\usepackage{graphicx}
\usepackage{subcaption}
\usepackage{cleveref}
\usepackage{xcolor}

\usepackage{dblfloatfix}

%% file: 00_custom.tex
\usepackage{aliascnt}

\newtheorem{theorem}{Theorem}
\newtheorem{assumption}{Assumption}

\newaliascnt{definition}{theorem}

\aliascntresetthe{definition}

\newaliascnt{proposition}{theorem}
\newtheorem{proposition}[proposition]{Proposition}
\aliascntresetthe{proposition}

\newaliascnt{corollary}{theorem}

\aliascntresetthe{corollary}

\newaliascnt{reduction}{theorem}

\aliascntresetthe{reduction}

\newaliascnt{example}{theorem}
\newtheorem{example}[example]{Example}
\aliascntresetthe{example}

\newaliascnt{lemma}{theorem}
\newtheorem{lemma}[lemma]{Lemma}
\aliascntresetthe{lemma}

\newaliascnt{remark}{theorem}
\newtheorem{remark}[remark]{Remark}
\aliascntresetthe{remark}

\newaliascnt{simplifiedtheorem}{theorem}
\newtheorem{simplifiedtheorem}[simplifiedtheorem]{Simplified Theorem}
\aliascntresetthe{simplifiedtheorem}
\newaliascnt{simplifiedlemma}{lemma}
\newtheorem{simplifiedlemma}[simplifiedlemma]{Simplified Lemma}
\aliascntresetthe{simplifiedlemma}

\crefname{assumption}{Assumption}{Assumptions}

\crefname{appendix}{Appendix}{Appendices}
\Crefname{appendix}{Appendix}{Appendices}

\crefname{definition}{Definition}{Definitions}
\Crefname{definition}{Definition}{Definitions}

\crefname{proposition}{Proposition}{Propositions}
\Crefname{proposition}{Proposition}{Propositions}

\crefname{corollary}{Corollary}{Corollaries}
\Crefname{corollary}{Corollary}{Corollaries}

\crefname{lemma}{Lemma}{Lemmas}
\Crefname{lemma}{Lemma}{Lemmas}

\crefname{equation}{Eq.}{Eqs.}
\Crefname{equation}{Eq.}{Eqs.}

\crefname{remark}{Remark}{Remarks}
\Crefname{remark}{Remark}{Remarks}

\crefname{simplifiedtheorem}{Theorem}{Theorems}
\Crefname{simplifiedtheorem}{Theorem}{Theorems}

\crefname{simplifiedlemma}{Lemma}{Lemmas}
\Crefname{simplifiedlemma}{Lemma}{Lemmas}

\newcommand{\vect}[1]{\mathbf{#1}}

\newcommand{\0}{\mat{0}}
\newcommand{\teacher}{\vw_{\star}}
\newcommand{\mat}[1]{\mathbf{#1}}
\newcommand{\vw}{\vect{w}}

\newcommand{\X}{\mat{X}}

\newcommand{\x}{\vect{x}}

\newcommand{\y}{\vect{y}}

\newcommand{\I}{\mat{I}}

\newcommand{\tparagraph}[1]{\vspace{-8pt}\paragraph{#1}}

\newcommand{\dd}{\mathop{}\!\mathrm{d}}

\newcommand{\lessorder}[2]{%
  \mathchoice
    {\mathord{\mathop{\bigO}\limits_{#1}}}
    {\mathord{\bigO_{#1}}}
    {\mathord{\bigO_{#1}}}
    {\mathord{\bigO_{#1}}}%
  \prn{#2}%
}

\newcommand{\gtrorder}[2]{%
  \mathchoice
    {\mathord{\mathop{\Omega}\limits_{#1}}}
    {\mathord{\Omega_{#1}}}
    {\mathord{\Omega_{#1}}}
    {\mathord{\Omega_{#1}}}%
  \prn{#2}%
}

\newcommand{\sameorder}[2]{%
  \mathchoice
    {\mathord{\mathop{\Theta}\limits_{#1}}}
    {\mathord{\Theta_{#1}}}
    {\mathord{\Theta_{#1}}}
    {\mathord{\Theta_{#1}}}%
  \prn{#2}%
}
\newcommand{\hfrac}[2]{{#1}/{#2}}
\newcommand{\bigO}[0]{\mathcal{O}}

\newcommand{\prn}[1]{\left({#1}\right)}

\newcommand{\bigprn}[1]{\big({#1}\big)}

\newcommand{\biggprn}[1]{\bigg({#1}\bigg)}

\newcommand{\norm}[1]{\left\Vert{#1}\right\Vert}

\DeclareMathOperator*{\argmin}{argmin}
\newcommand{\R}{\mathbb{R}}
\newcommand{\E}{\mathbb{E}}

\newcommand{\sign}{\mathrm{sign}}

\newcommand{\prox}{\mathrm{Prox}}

\newcommand{\Var}{\mathrm{Var}}

\newcommand{\PP}{\mathbb{P}}

\newcommand{\Ber}{\mathrm{Ber}}
\newcommand{\iidsim}{\stackrel{\mathrm{iid}}{\sim}}

\newcommand{\ie}{\textit{i.e.,}\xspace}
\newcommand{\eg}{\textit{e.g.,}\xspace}
\newcommand{\iid}{i.i.d.\@\xspace}

\crefname{appendix}{App.}{Apps.}
\Crefname{appendix}{App.}{Apps.}

\makeatletter
\newenvironment{recall}[1][\proofname]{\par
\normalfont \topsep6\p@\@plus6\p@\relax
\trivlist
\item\relax
{\bfseries
Recall #1}%
{\bfseries\@addpunct{.}}\hspace\labelsep\ignorespaces
}
\makeatother

\makeatletter
\newenvironment{fuller}[1][\proofname]{\par
\normalfont \topsep6\p@\@plus6\p@\relax
\trivlist
\item\relax
{\bfseries
Full #1}%
{\bfseries\@addpunct{.}}\hspace\labelsep\ignorespaces
}
\makeatother

\usepackage{titletoc} 


%% file: ICLR/math_commands.tex
\usepackage{amsmath,amsfonts,bm}

\def\eqref#1{equation~\ref{#1}}

\def\1{\bm{1}}

\def\mM{{\mathbf{M}}}

\DeclareMathAlphabet{\mathsfit}{\encodingdefault}{\sfdefault}{m}{sl}
\SetMathAlphabet{\mathsfit}{bold}{\encodingdefault}{\sfdefault}{bx}{n}



%% file: 02_penalty_figure.tex
\begin{figure}[h!]
\centering

\captionsetup[subfigure]{justification=raggedright,singlelinecheck=false}

\begin{subfigure}[t]{0.49\columnwidth}
\centering
\begin{tikzpicture}
\begin{axis}[
    scale only axis,
    width=0.74\linewidth,
    height=0.52\linewidth,
    xmin=-1.2,
    xmax=1.2,
    ymin=-0.03,
    ymax=1.2,
    xtick={-1,0,1},
    ytick={0,0.5,1},
    xlabel={$w_{t,i}$},
    ylabel={Penalty},
    ylabel style={
        at={(axis description cs:-0.15,0.5)},
        anchor=south,
    },
    ytick={0,1},
    extra y ticks={0.5},
    extra y tick labels={},
    extra y tick style={
        tick style={draw=none},
        grid=major,
    },
    grid=major,
    grid style={black!12},
    axis x line*=bottom,
    axis y line*=left,
    tick label style={font=\scriptsize},
    label style={font=\small},
    legend style={
        at={(axis description cs:0.5,1.01)},
        anchor=south,
        draw=none,
        fill=none,
        font=\small,
        legend columns=2,
        inner sep=0pt,
        /tikz/every even column/.append style={column sep=0.12cm},
    },
]

\addplot[
    swgreen,
    line width=1.2pt,
    domain=-1.5:1.5,
    samples=200,
]
{abs(x)};
\addlegendentry{SW-L1}

\addplot[
    lonegray,
    line width=2.2pt,
    dash pattern=on 4pt off 3pt,
    domain=-1.5:1.5,
    samples=200,
]
{abs(x)};
\addlegendentry{L1}

\end{axis}
\end{tikzpicture}

\vspace{-5pt}
\caption{
\textbf{Previously inactive coordinate ($w_{t-1,i}=0$).}
SW-L1 and L1 coincide, so activation is equally penalized in either
direction.
}
\label{fig:sw-l1-inactive}
\end{subfigure}
\hfill
%
\begin{subfigure}[t]{0.47\columnwidth}
\centering
\begin{tikzpicture}
\begin{axis}[
    scale only axis,
    width=0.82\linewidth,
    height=0.52\linewidth,
    xmin=-1.2,
    xmax=1.2,
    ymin=-0.03,
    ymax=1.2,
    xtick={-1,0,1},
    ytick={0,0.5,1},
    xlabel={$w_{t,i}$},
    ytick={0,1},
    extra y ticks={0.5},
    extra y tick labels={},
    extra y tick style={
        tick style={draw=none},
        grid=major,
    },
    grid=major,
    grid style={black!12},
    axis x line*=bottom,
    axis y line*=left,
    tick label style={font=\scriptsize},
    label style={font=\small},
    legend style={
        at={(axis description cs:0.5,1.01)},
        anchor=south,
        draw=none,
        fill=none,
        font=\small,
        legend columns=2,
        inner sep=0pt,
        /tikz/every even column/.append style={column sep=0.12cm},
    },
]

\path[
    fill=flipred,
    fill opacity=0.07,
    draw=none,
]
(axis cs:-1.5,-0.03) rectangle (axis cs:0,1.58);

\path[
    fill=swgreen,
    fill opacity=0.07,
    draw=none,
]
(axis cs:0,-0.03) rectangle (axis cs:1.5,1.58);

\addplot[
    swgreen,
    line width=1.2pt,
    domain=-1.5:1.5,
    samples=200,
]
{x < 0 ? -x : x/2};
\addlegendentry{SW-L1}

\addplot[
    lonegray,
    line width=2.2pt,
    dash pattern=on 4pt off 3pt,
    domain=-1.5:1.5,
    samples=200,
]
{abs(x)};
\addlegendentry{L1}

\end{axis}
\end{tikzpicture}

\vspace{-5pt}
\caption{
\textbf{Previously positive coordinate ($w_{t-1,i}>0$).}
SW-L1 penalizes same-sign updates less than sign flips.
Here $w_{t-1,i}=\epsilon=1$.
}
\label{fig:sw-l1-positive}
\end{subfigure}

\vspace{-0.1cm}
\caption{
\textbf{Coordinate-wise comparison of L1 and our sign-weighted L1
regularization.}
}
\label{fig:sw-l1-penalty}
\end{figure}

%% file: 03_main_figure.tex
\begin{figure*}[t]
\centering


\begin{subfigure}{0.3\textwidth}
    \centering
    \includegraphics[width=\linewidth]{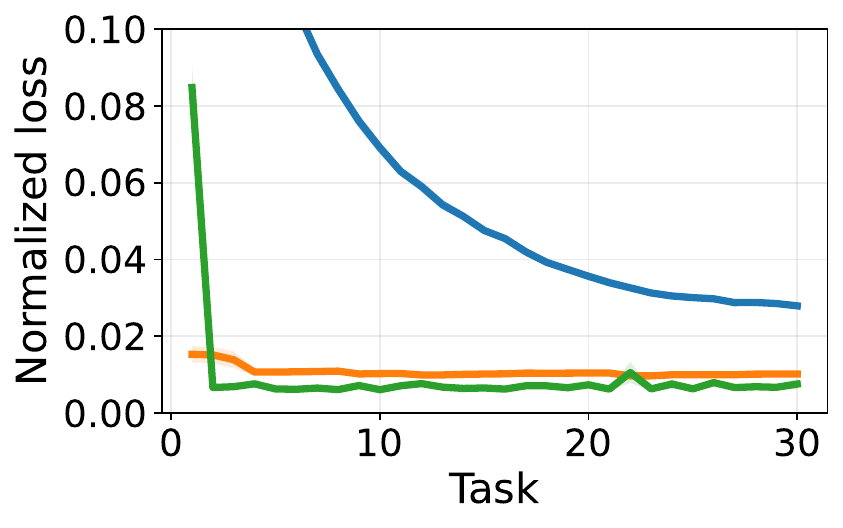}
    \vspace{-.6cm}
    \caption{\textbf{High sparsity} ($s=10^{-3}$)}
\end{subfigure}
\hfill
\begin{subfigure}{0.3\textwidth}
    \centering
    \includegraphics[width=\linewidth]{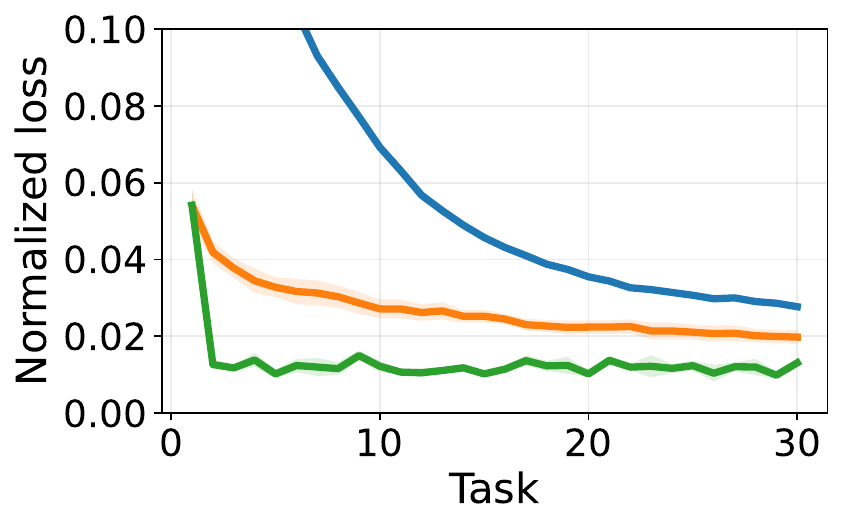}
    \vspace{-.6cm}
    \caption{\textbf{Medium sparsity} ($s=10^{-2}$)}
\end{subfigure}
\hfill
\begin{subfigure}{0.3\textwidth}
    \centering
    \includegraphics[width=\linewidth]{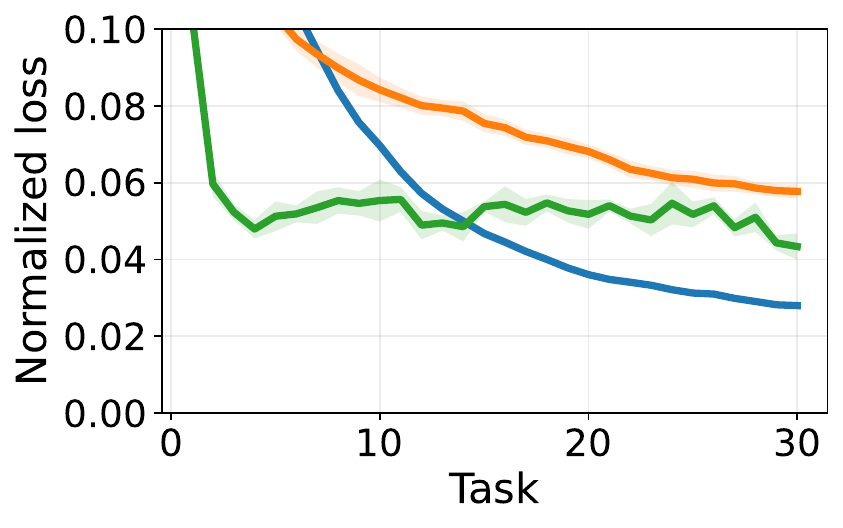}
    \vspace{-.6cm}
    \caption{\textbf{Low sparsity} ($s=10^{-1}$)}
\end{subfigure}

\vspace{0.3em}
{\small(Setting I) \textbf{Linear classifier for Gaussian synthetic tasks.}\par}

\vspace{0.6em}


\begin{subfigure}{0.3\textwidth}
    \centering
    \includegraphics[width=\linewidth]{Figures/excess_linear_s0p001_k1.pdf}
    \vspace{-.6cm}
    \caption{\textbf{High sparsity} ($s=10^{-3}$)}
\end{subfigure}
\hfill
\begin{subfigure}{0.3\textwidth}
    \centering
    \includegraphics[width=\linewidth]{Figures/excess_linear_s0p01_k8.pdf}
    \vspace{-.6cm}
    \caption{\textbf{Medium sparsity} ($s=10^{-2}$)}
\end{subfigure}
\hfill
\begin{subfigure}{0.3\textwidth}
    \centering
    \includegraphics[width=\linewidth]{Figures/excess_linear_s0p1_k78.pdf}
    \vspace{-.6cm}
    \caption{\textbf{Low sparsity} ($s=10^{-1}$)}
\end{subfigure}

\vspace{0.3em}
{\small(Setting II) \textbf{Linear classifier for MNIST.}\par}

\vspace{0.6em}


\begin{subfigure}{0.3\textwidth}
    \centering
    \includegraphics[width=\linewidth]{Figures/excess_relu_s0p001_k2.pdf}
    \vspace{-.6cm}
    \caption{\textbf{High sparsity} ($s=10^{-3}$)}
\end{subfigure}
\hfill
\begin{subfigure}{0.3\textwidth}
    \centering
    \includegraphics[width=\linewidth]{Figures/excess_relu_s0p01_k20.pdf}
    \vspace{-.6cm}
    \caption{\textbf{Medium sparsity} ($s=10^{-2}$)}
\end{subfigure}
\hfill
\begin{subfigure}{0.3\textwidth}
    \centering
    \includegraphics[width=\linewidth]{Figures/excess_relu_s0p1_k200.pdf}
    \vspace{-.6cm}
    \caption{\textbf{Low sparsity} ($s=10^{-1}$)}
\end{subfigure}

\vspace{0.3em}
{\small(Setting III) \textbf{Two-layer ReLU for MNIST.}\par}

\caption{\textbf{Effect of teacher sparsity across models and datasets.}
sign weighted L1 performs best at high teacher sparsity and remains competitive at moderate sparsity, while L2 becomes preferable as sparsity decreases.}

\vspace{-.5em}

\label{fig:sparsity_all}
\end{figure*}